\documentclass[10pt,twocolumn,letterpaper]{article}

\usepackage[pagenumbers]{cvpr} 

\renewcommand{\paragraph}[1]{\vspace{.2em}\noindent\textbf{#1.}}

\usepackage{siunitx}

\usepackage{graphicx}
\usepackage{amsmath,amssymb}
\usepackage{booktabs}
\usepackage{multirow}
\usepackage{xcolor}
\usepackage{xspace}
\usepackage{tikz}
\usetikzlibrary{arrows.meta,positioning,fit,backgrounds,calc,shapes.geometric,shapes.arrows,shadows}
\usepackage{pifont}
\usepackage{enumitem}
\usepackage{etoolbox}
\usepackage{silence}
\makeatletter
\robustify\@latex@warning@no@line
\makeatother
\usepackage{authblk}

\newcommand{\method}{Keystone\xspace}  

\newcommand{\Sim}{\mathrm{Sim}(3)}
\newcommand{\vK}{\mathbf{K}}

\newcommand{\vR}{\mathbf{R}}
\newcommand{\vS}{\mathbf{S}}
\newcommand{\vX}{\mathbf{X}}

\newcommand{\vc}{\mathbf{c}}
\newcommand{\vd}{\mathbf{d}}
\newcommand{\vt}{\mathbf{t}}
\newcommand{\vx}{\mathbf{x}}

\newcommand{\pred}[1]{\hat{#1}}

\newcommand{\fade}{\textcolor[gray]{0.4}}
\definecolor{oursrow}{RGB}{230,230,230}
\definecolor{keystonerow}{RGB}{230,230,230}

\usepackage{colortbl}


\definecolor{pastel_yellow}{RGB}{255, 243, 176} 
\definecolor{pastel_blue}{RGB}{190, 225, 245}   
\definecolor{pastel_pink}{RGB}{244, 210, 194}   
\newrobustcmd{\first}{\cellcolor{pastel_yellow}}
\newrobustcmd{\second}{\cellcolor{pastel_blue}}
\newrobustcmd{\third}{\cellcolor{pastel_pink}}

\definecolor{rankA}{HTML}{FCEFB0}
\definecolor{rankB}{HTML}{CFE3F5}
\definecolor{rankC}{HTML}{F6D3C6}

\definecolor{cvprblue}{rgb}{0.21,0.49,0.74}
\usepackage[pagebackref,breaklinks,colorlinks,allcolors=cvprblue]{hyperref}

\def\confName{CVPR}
\def\confYear{2027}

\title{Keystone: Proper Refinement Matters More Than Accurate SfM Initialization} 

\author{
\vspace{-1.4em}
Sithu Aung$^1$, \quad 
Viktor Kocur$^2$, \quad 
Yaqing Ding$^3$, \quad 
Torsten Sattler$^4$, \quad 
Zuzana Kukelova$^1$ \\
$^1$VRG, FEE, Czech Technical University in Prague \quad
$^2$FMPH, Comenius University in Bratislava \quad
$^3$Southeast University \quad
$^4$CIIRC, Czech Technical University in Prague \\
\href{https://keystone-refine.github.io/}{https://keystone-refine.github.io/}
}

\begin{document}
\twocolumn[{%
\renewcommand\twocolumn[1][]{#1}%
\maketitle
\vspace{-2.4em}
\input{fig/teaser}
}]

\begin{abstract}
Feed-forward reconstruction networks, \eg, VGGT, predict camera poses and dense geometry for a set of images in a single forward pass. They are highly efficient, making them an attractive alternative to classical Structure-from-Motion (SfM) approaches based on local feature matches. 
However, their camera pose estimates are clearly less accurate than those obtained by feature-based SfM. 
Recent work has therefore proposed increasingly elaborate pipelines that combine feed-forward predictions with classical SfM. 
In this paper, we explore a much simpler alternative.
We take initial camera poses from any method, such as an off-the-shelf reconstruction network, and iteratively refine them with local feature matches through multiple rounds of triangulation and bundle adjustment.
Scenes too large for one forward pass are predicted in two interleaved partitions of overlapping chunks, placed in a common frame by averaging similarity transformations estimated on their shared images, and then refined as a whole.
Despite its simplicity, this approach, which we call \textbf{Keystone}, substantially improves feed-forward, classical, and hybrid initializations alike and brings them to nearly the same accuracy.
In particular, the noisy poses of feed-forward networks are good enough: starting from them, Keystone outperforms both classical SfM and (significantly more complex) recent hybrid pipelines on standard benchmarks, including challenging settings involving lens distortion and long video sequences.
Given its simplicity and strong results, Keystone represents a solid baseline for future research on feed-forward methods for accurate 3D reconstruction and practical SfM systems.
\end{abstract}

\section{Introduction}
\label{sec:intro}
Structure-from-Motion (SfM) estimates camera poses and sparse 3D structure from a collection of images and is a fundamental component of novel view synthesis, visual localization, and 3D reconstruction.
Classical incremental~\cite{colmap, snavely2006photo, wu2013towards} and global~\cite{glomap, theia-manual, moulon2016openmvg} pipelines rely on local feature correspondences~\cite{sift, superpoint, sarlin2020superglue, lindenberger2023lightglue} and bundle adjustment (BA)~\cite{zhan2026bundle} to reach sub-degree pose accuracy when sufficient image overlap is available.
However, they can be slow on large collections~\cite{wu2013towards, snavely2008skeletal} and are prone to failure when the view graph is sparse~\cite{mpsfm} or contains incorrect image pairs and correspondences~\cite{dfsfm, sferrazza2025match}, particularly in scenes with repeated structures~\cite{wilson2014robust, doppelgangers}.

Feed-forward (FF) reconstruction networks~\cite{vggt, pi3, keetha2026mapanything, lin2025depthanything3, wang2026vggtomega, burzio2026dejaview, jin2026zipmap} take a different approach and predict camera poses and dense geometry for all input images in a single forward pass.
Their main appeal is efficiency, and under certain conditions they are also more robust to visual ambiguities~\cite{dust3r, mast3rsfm, doppelgangers++}.
However, they have two limitations.
First, their camera poses remain substantially less accurate than those of feature-based SfM~\cite{meza2026benchmarking}.
Second, the memory requirements of a single forward pass limits how many images can be processed together~\cite{shen2026fastvggt, wang2026flashvggt, elflein2026vggt3}, so larger scenes must be split into chunks.
One line of work addresses this limitation within the networks themselves, exploiting the sparsity of cross-view interactions through token merging~\cite{shen2026fastvggt}, sparse or compressed global attention~\cite{wang2026block, sun2026avggt, wang2026flashvggt, ren2026speed3r}, or linear-time alternatives to global attention~\cite{elflein2026vggt3}.
These methods avoid aligning chunks, but they approximate full global attention, often require retraining, and still grow in cost with the number of images; they thus reduce rather than remove the need for chunking and leave the accuracy of the predicted poses unchanged.

Other works instead operate on the predictions of unmodified networks, but address the two limitations separately.
For accuracy, BA on image correspondences substantially improves FF poses~\cite{vggt, chen2026ggpt, pan2026gluemap}; most directly, Meza~\etal~\cite{meza2026benchmarking} refine FF poses by triangulation and BA on local feature matches and find that this offers a favorable trade-off between efficiency and quality for novel view synthesis.
For scale, chunk-wise methods~\cite{vggtslam, deng2025vggtlong, pilongcode2025, murai2024_mast3rslam, cheng2026merg3r} align independently predicted chunks along a sequence or a tree and correct the accumulated drift with loop closures.
What remains open is which design choices of such a refinement matter for pose accuracy, whether it helps only FF networks or any initialization, and how to apply it to scenes beyond a single forward pass without chaining chunks.

We address these questions with \textbf{\method}, a simple refinement that builds on Meza~\etal.
\method takes the cameras of any method, whether an FF network, a classical SfM pipeline, or a hybrid one, and refines them over several rounds of triangulation and BA on a fixed set of local feature matches.
Since each round re-triangulates the matches at the updated cameras, it recovers matches that the previous estimates rejected as inconsistent, which significantly improves the pose accuracy.
As a result, initializations that differ widely in accuracy reach nearly the same accuracy after \method (\cref{fig:teaser}): \emph{the refinement matters more than the initialization}, and an off-the-shelf FF network becomes the cheapest starting point.
For scenes beyond a single forward pass, we predict the images in two interleaved partitions of chunks and place all chunks jointly from their shared images, so that they can be refined exactly like a small scene.

In summary, our contributions are threefold:
\begin{itemize}
  \item \textbf{A refinement for any initialization.} We identify which design choices of iterative triangulation and BA matter for pose accuracy, and show that the resulting refinement improves FF, classical SfM, and hybrid initializations alike, bringing them to nearly the same level of accuracy.
  \item \textbf{Scaling to large scenes.} Two interleaved partitions and global similarity averaging extend the refinement of FF methods to scenes beyond a single forward pass, without per-chunk refinement or merging.
  \item \textbf{An extensive evaluation.} We evaluate several FF networks and baselines on nine datasets, covering lens distortion, video sequences, and large collections, and establish \method as a strong baseline for future research.
\end{itemize}

\section{Related Work}
\label{sec:related_work}
\noindent \textbf{Classical SfM.} 
Incremental methods~\cite{snavely2006photo, agarwal2011building, colmap} register images one at a time and repeatedly refine the growing model with BA~\cite{triggs1999bundle}, which makes them accurate but slow for large image collections.
Global methods~\cite{theia-manual,moulon2016openmvg,glomap} estimate all camera rotations and positions jointly from a view graph, and divide-and-conquer approaches~\cite{bhowmick2014divide, Zhu_2018_CVPR} reconstruct partitions separately and merge them using similarity transformations.
All these pipelines rely on local feature correspondences, whose quality has steadily improved through learned detectors and matchers~\cite{superpoint, sarlin2020superglue,lindenberger2023lightglue,nordstrom2026loma,lindenberger2026flexglue}, dense matchers~\cite{romav1, romav2, zhang2025ufm}, and keypoint refinement~\cite{pixsfm, dfsfm, densesfm}.
Despite their different strategies, these methods are carefully engineered to obtain reliable initial poses that BA then refines.
We show that, with proper refinement, much coarser initializations such as FF predictions can reach a similar accuracy, and that the same refinement also improves the poses of classical pipelines.

\input{fig/overview}

\noindent \textbf{Feed-forward reconstruction.} 
DUSt3R~\cite{dust3r} demonstrated that neural networks can directly regress point maps from image pairs, and MASt3R~\cite{mast3r} added local features for matching.
Extending pairwise predictions to larger scenes, however, requires a global alignment, which motivated multi-view architectures.
VGGT~\cite{vggt} predicts camera poses, depth maps, point maps, and tracks for multiple images in one forward pass.
$\pi^3$~\cite{pi3} removes the reference view, while Pow3R~\cite{jang2025pow3r}, MapAnything~\cite{keetha2026mapanything}, and DepthAnything3~\cite{lin2025depthanything3} accept additional input conditions.
More recent works scale these models or make them recurrent: VGGT-$\Omega$~\cite{wang2026vggtomega} scales data and model size, and DVLT~\cite{burzio2026dejaview} iterates weight-tied transformer blocks, which Solv3R~\cite{kwak2026solv3r} turns into a multi-view solver that can continue from any current state.
To reduce the cost of global attention over many views, several works sparsify or approximate it~\cite{shen2026fastvggt, wang2026block, sun2026avggt, wang2026flashvggt, ren2026speed3r}, while streaming approaches~\cite{wang2025spann3r, cabon2025must3r, wang2025cut3r, wu2026point3r, zhuo2026streaming, lan2026stream3r, yuan2026infinitevggt} process the images sequentially with a persistent scene state or memory.
Test-time training methods~\cite{chen2026ttt3r, jin2026zipmap, elflein2026vggt3} instead compress the scene into a fixed-size state updated at test time, so that the cost grows linearly with the number of views.
Despite their efficiency and robustness to visual ambiguities, these networks still produce less accurate camera poses than feature-based SfM~\cite{meza2026benchmarking}.
We treat them as fast initializations: \method refines the predictions of any of them, to an accuracy close to that of the other initializations it refines.

\noindent \textbf{Feed-forward SLAM.} 
Instead of extending the network's context, another line of work predicts smaller chunks of images and aligns them afterwards.
VGGT-Long~\cite{deng2025vggtlong} aligns overlapping chunks with similarity transformations estimated from shared frames and removes drift with loop closures in a pose graph, a strategy also applied to $\pi^3$ and DepthAnything3~\cite{pilongcode2025, lin2025depthanything3}.
VGGT-SLAM~\cite{vggtslam} aligns sub-maps with 15-DoF projective transformations, which VGGT-SLAM 2.0~\cite{maggio2026vggt} revisits with a new factor graph formulation.
MASt3R-SLAM~\cite{murai2024_mast3rslam} builds dense SLAM on two-view MASt3R predictions.
Since \method only requires initial cameras, any of these methods can initialize it.
For our own FF initialization, we keep the chunking simple: the refinement corrects the remaining misalignment, so two interleaved partitions and a similarity averaging only need to place the chunks roughly.

\noindent \textbf{Hybrid reconstruction.} 
Several works combine learned geometry with optimization against image correspondences.
Some retain the classical SfM pipeline and add learned geometric priors to improve its robustness.
MP-SfM~\cite{mpsfm} introduces monocular depth and surface normal priors into incremental SfM, enabling registration even with very limited overlap, and DGSfM~\cite{aung2026dgsfm} uses monocular depth to turn scale-ambiguous epipolar constraints into scale-aware ones for view-graph filtering and the initialization of global positioning.
VidMap~\cite{pataki2026vidmap} extends global SfM to videos, combining temporal tracks, loop closures, and metric depth.
These methods improve the constraints and the camera initialization but largely preserve the structure of classical SfM.
Others start from FF predictions and show that the predicted cameras improve substantially through optimization against image correspondences.
VGGSfM~\cite{vggsfm} makes the SfM pipeline differentiable and performs BA on predicted tracks, and VGGT~\cite{vggt} refines its predictions by BA on its own tracks.
MASt3R-SfM~\cite{mast3rsfm} optimizes MASt3R~\cite{mast3r} predictions globally, and GGPT~\cite{chen2026ggpt} estimates camera poses from dense matches in a lightweight SfM stage to ground the network's point maps.
MERG3R~\cite{cheng2026merg3r} merges overlapping subsets of unordered collections through confidence-weighted BA.
GLUEMAP~\cite{pan2026gluemap} incorporates FF predictions into global SfM before a final BA, and reports a large improvement over FF poses refined by BA alone.
Closest to our work, Meza~\etal~\cite{meza2026benchmarking} benchmark camera pose estimation for novel view synthesis and show that refining FF predictions with triangulation and BA offers a favorable trade-off between efficiency and accuracy.
We build on this observation and study which design choices of the refinement matter for pose accuracy.
The resulting refinement is complementary to these pipelines: it improves the poses of these hybrid methods as well, and brings plain FF predictions, without relative poses or averaging, to a similar accuracy.

\section{Method}
\label{sec:method}
\noindent \textbf{Problem setup.}
Given a set of images $\{I_i\}_{i=1}^{N}$, we estimate for each image a world-to-camera rotation $\vR_i \in SO(3)$, a camera center $\vc_i \in \mathbb{R}^3$, and intrinsic parameters $\vK_i$, which include radial distortion coefficients $\mathbf{k}_i \in \mathbb{R}^{2}$ for images with lens distortion.
\method starts from initial estimates of these cameras, which may come from classical SfM, a hybrid pipeline, or an FF network, and treats their source as a black box (\cref{fig:overview}).
It then matches local features between the images and refines all cameras over $R$ rounds of triangulation and bundle adjustment (\cref{sec:refine}).

\noindent \textbf{Feed-forward initialization.}
An FF network $f$ takes a set $\mathcal{V}$ of at most $M$ images and predicts, for each image $i \in \mathcal{V}$, a rotation $\pred{\vR}_i$, a center $\pred{\vc}_i$, and intrinsics $\pred{\vK}_i$, together with a depth map and a confidence map.
In most networks, $\pred{\vK}_i$ describes a pinhole camera without distortion, often with the principal point at the image center.
The predictions of one forward pass share a coordinate frame that is defined up to a similarity transformation $\vS_\mathcal{V} = (s_\mathcal{V}, \vR_\mathcal{V}, \vt_\mathcal{V})$, which acts on a 3D point as $\vS_\mathcal{V}(\vX) = s_\mathcal{V}\vR_\mathcal{V}\vX + \vt_\mathcal{V}$.
If $\vS_\mathcal{V}$ maps this frame to the world, the world camera of image $i$ is
\vspace{-0.3em}
\begin{equation}
    \vc_i = s_\mathcal{V}\vR_\mathcal{V}\pred{\vc}_i + \vt_\mathcal{V}, \qquad \vR_i = \pred{\vR}_i\vR_\mathcal{V}^{\top}.
    \label{eq:sim3}
\vspace{-0.3em}
\end{equation}
Scenes with $N \le M$ images are predicted in a single forward pass, whose frame we take as the world.
Larger scenes are predicted in chunks, and we estimate one similarity per chunk to bring them into a common frame (\cref{sec:chunks}).

\subsection{Iterative Feature-based Refinement}
\label{sec:refine}
Initial cameras are typically close to the correct solution but not precise, especially those predicted by FF networks, which are considerably noisier than the poses of SfM or hybrid pipelines.
Following Meza~\etal~\cite{meza2026benchmarking}, we refine them over several rounds of triangulation and bundle adjustment, with track completion, merging, and filtering.

\noindent \textbf{Correspondences.} 
We match local features on a set of image pairs $\mathcal{P}$, which contains all pairs in small unordered collections, each frame with its next $w$ frames and its top-$k$ retrieval neighbors in video sequences, and each image with its top-$k$ retrieval neighbors in large collections.
Each pair $(i, i') \in \mathcal{P}$ is verified by LO-RANSAC~\cite{chum2003locally} with COLMAP's E/F/H model selection, using the initial intrinsics $\vK_i$ and $\vK_{i'}$, and we keep its inlier matches $\mathcal{M}_{ii'}$ if there are at least 15.

\noindent \textbf{Refinement rounds and inner loops.} 
The inlier matches $\mathcal{M}_{ii'}$ link keypoints across images into feature tracks $\mathcal{T}_j = \{(i, \vx_{ij})\}$, each ideally observing a single 3D point $\vX_j$.
We write $\pi_i(\vX) = \pi\big(\vK_i, \vR_i(\vX - \vc_i)\big)$ for the projection of a point into image $i$, including the distortion $\mathbf{k}_i$, and $e_{ij} = \|\pi_i(\vX_j) - \vx_{ij}\|$ for its reprojection error.
The refinement runs $R$ \emph{outer rounds}, each of which first triangulates every track at the current cameras.
Since the reprojection error of an inaccurate camera can be large, an observation $\vx_{ij}$ is accepted if its viewing ray $\vd_{ij} = \vR_i^\top \pi^{-1}(\vK_i, \vx_{ij})$ points close to the triangulated point,
\vspace{-0.3em}
\begin{equation}
    \angle\big(\vd_{ij},\, \vX_j - \vc_i\big) < \tau_a \enspace ,
    \label{eq:tri_angle}
\vspace{-0.3em}
\end{equation}
with a loose $\tau_a = 2^\circ$, so that tracks form even from inaccurate cameras.
With noisy initial cameras, some observations of a track still disagree with the others and are left out, and the rest may even split into several 3D points.
To recover them and fuse the split points as the poses improve, an \emph{inner loop} follows COLMAP's iterative global refinement and repeats four steps.
First, \emph{bundle adjustment} optimizes all cameras and points,
\vspace{-0.3em}
\begin{equation}
    \min_{\{\vR_i, \vc_i, \vK_i\},\, \{\vX_j\}} \; \sum_{j} \sum_{(i, \vx_{ij}) \in \mathcal{T}_j} \rho\big( e_{ij}^2 \big) \enspace ,
    \label{eq:ba}
\vspace{-0.3em}
\end{equation}
where $\rho$ is the soft-$\ell_1$ loss.
Second, \emph{track completion} adds every left-out observation of $\mathcal{T}_j$ with $e_{ij} < \tau_r$.
Third, \emph{track merging} fuses two points of the same track if the merged point has $e_{ij} < \tau_r$ in all their observations.
Fourth, \emph{outlier filtering} removes observations with $e_{ij} \geq \tau_r$ and points whose largest triangulation angle $\max_{i,i'} \angle(\vX_j - \vc_i, \vX_j - \vc_{i'})$ is below $\tau_\theta$.
We set $\tau_r = 4$\,px and $\tau_\theta = 1.5^\circ$.
The inner loop stops after five iterations, or earlier once fewer than 0.1\% of the observations change.

\noindent \textbf{Camera model.}
The initial intrinsics are often inconsistent across images; FF networks in particular predict $\pred{\vK}_i$ independently for each image and forward pass, and such freely varying focal lengths leave BA poorly constrained.
When all images of a scene come from one camera, we therefore share a single camera among them, initialize it with the median of the initial intrinsics, and refine its focal length and principal point (\cref{tab:intrinsics}).
Otherwise, every image keeps its own camera with a fixed principal point, since a free principal point per image can absorb rotation errors.
For strongly distorted images, we add the radial distortion $1 + k_1 r^2 + k_2 r^4$ with $\mathbf{k}_i = (k_1, k_2)$ and refine it together with the focal length, starting from $\mathbf{0}$ for pinhole initializations such as FF networks.
Known intrinsics, including their distortion, remain fixed.

\begin{figure}[t]
\centering
\definecolor{ink}{HTML}{1A1A1A}
\definecolor{muted}{HTML}{6A6A6A}
\definecolor{cA}{HTML}{3E7CB1}
\definecolor{cB}{HTML}{D9643A}
\resizebox{0.9\columnwidth}{!}{%
\begin{tikzpicture}[
    font=\sffamily\small, text=ink,
    cam/.pic={\fill[ink!28] (0,0) -- (#1-24:0.21) -- (#1+24:0.21) -- cycle; \fill[ink!65] (0,0) circle (0.04);},
    camh/.pic={\fill[ink!28] (0,0) -- (#1-24:0.21) -- (#1+24:0.21) -- cycle; \fill[ink] (0,0) circle (0.055); \draw[ink, line width=0.6pt] (0,0) circle (0.11);},
    ra/.style={fill=cA!10, draw=cA, line width=0.9pt, dash pattern=on 3pt off 1.8pt, rounded corners=5pt},
    rb/.style={draw=cB, line width=1.0pt, rounded corners=7pt},
    la/.style={text=cA!80!black, font=\sffamily\small\bfseries},
    lbb/.style={text=cB!85!black, font=\sffamily\small\bfseries},
    na/.style={circle, fill=cA!10, draw=cA, line width=0.9pt, dash pattern=on 2.5pt off 1.5pt, minimum size=0.46cm, inner sep=0pt, font=\sffamily\small\bfseries, text=cA!80!black},
    nb/.style={circle, fill=cB!6, draw=cB, line width=0.9pt, minimum size=0.46cm, inner sep=0pt, font=\sffamily\small\bfseries, text=cB!85!black},
    ed/.style={ink!55, line width=0.7pt},
    title/.style={font=\sffamily\small\bfseries, text=ink, anchor=south},
    sub/.style={font=\sffamily\footnotesize, text=ink, align=center, anchor=north},
    arr/.style={-{Latex[length=2mm,width=1.6mm]}, line width=0.8pt, ink!60},
]
\begin{scope}
\node[title] at (1.62,2.85) {Interleaved partitions $\mathcal{A}$ and $\mathcal{B}$};
\draw[ra] (-0.36,-0.40) rectangle (1.50,0.98);
\draw[ra] (-0.36,1.12) rectangle (1.50,2.24);
\draw[ra] (1.72,-0.40) rectangle (3.58,0.98);
\draw[ra] (1.72,1.12) rectangle (3.58,2.24);
\draw[rb] (-0.56,-0.62) rectangle (0.86,2.46);
\draw[rb] (1.02,-0.62) rectangle (2.40,2.46);
\draw[rb] (2.56,-0.62) rectangle (3.80,2.46);
\pic at (-0.02,-0.02) {cam=333};
\pic at (-0.01,0.58) {cam=211};
\pic at (-0.10,1.32) {cam=227};
\pic at (0.09,1.81) {cam=109};
\pic at (0.49,0.06) {cam=250};
\pic at (0.47,0.73) {cam=347};
\pic at (0.67,1.36) {cam=57};
\pic at (0.46,1.95) {cam=21};
\pic at (1.14,-0.12) {camh=11};
\pic at (1.23,0.56) {camh=303};
\pic at (1.25,1.36) {cam=215};
\pic at (1.29,1.93) {cam=100};
\pic at (2.17,0.12) {cam=302};
\pic at (2.08,0.52) {cam=83};
\pic at (1.94,1.18) {cam=96};
\pic at (1.98,2.05) {cam=139};
\pic at (2.78,0.07) {cam=0};
\pic at (2.74,0.68) {cam=120};
\pic at (2.78,1.29) {cam=26};
\pic at (2.67,2.03) {cam=97};
\pic at (3.12,-0.09) {cam=347};
\pic at (3.33,0.46) {cam=89};
\pic at (3.12,1.18) {cam=60};
\pic at (3.15,1.96) {cam=161};
\node[la] at (-0.9,0.0) {$A_1$};
\node[la] at (4.15,0.0) {$A_2$};
\node[la] at (-0.9,1.9) {$A_3$};
\node[la] at (4.15,1.9) {$A_4$};
\node[lbb] at (0.15,2.68) {$B_1$};
\node[lbb] at (1.71,2.68) {$B_2$};
\node[lbb] at (3.18,2.68) {$B_3$};
\draw[ink, line width=0.5pt] (1.38,0.42) -- (1.95,-0.82);
\node[font=\sffamily\footnotesize, text=ink, anchor=north, inner sep=1pt] at (1.95,-0.84) {shared images of $A_1$ and $B_2$};
\node[sub] at (1.62,-1.2) {every image lies in one chunk of $\mathcal{A}$ (blue)\\and one chunk of $\mathcal{B}$ (orange);\\each chunk of $\mathcal{B}$ spans several chunks of $\mathcal{A}$};
\end{scope}
\draw[arr] (4.65,0.95) -- (5.15,0.95);
\begin{scope}[shift={(6.0,0)}]
\node[title] at (1.1,2.85) {Chunk graph};
\node[na] (A1) at (0,0.1) {$A_1$};
\node[na] (A2) at (2.2,0.1) {$A_2$};
\node[na] (A3) at (0,1.8) {$A_3$};
\node[na] (A4) at (2.2,1.8) {$A_4$};
\node[nb] (B1) at (-0.35,0.95) {$B_1$};
\node[nb] (B2) at (1.1,0.95) {$B_2$};
\node[nb] (B3) at (2.55,0.95) {$B_3$};
\foreach \a/\b in {A1/B1, A3/B1, A3/B2, A4/B2, A2/B3, A4/B3, A2/B2} {\draw[ed] (\a) -- (\b);}
\draw[ink, line width=1.1pt] (A1) -- (B2);
\node[font=\sffamily\footnotesize, text=ink, anchor=north west, inner sep=1pt] at (0.45,0.5) {$\mathbf{S}_{u\to v}$};
\node[sub] at (1.1,-1.2) {one $\mathrm{Sim}(3)$ per pair of\\chunks that share images;\\robustly averaged};
\end{scope}
\end{tikzpicture}}
\vspace{-0.4em}
\caption{\textbf{Interleaved partitions.} Partition $\mathcal{B}$ cuts the same retrieval graph as $\mathcal{A}$ with the edges inside $\mathcal{A}$ chunks down-weighted, so that each of its chunks spans several chunks of $\mathcal{A}$. Chunks that share images are related by similarity transforms, which are jointly averaged to map every chunk into a common frame.}
\label{fig:partitions}
\vspace{-1.4em}
\end{figure}
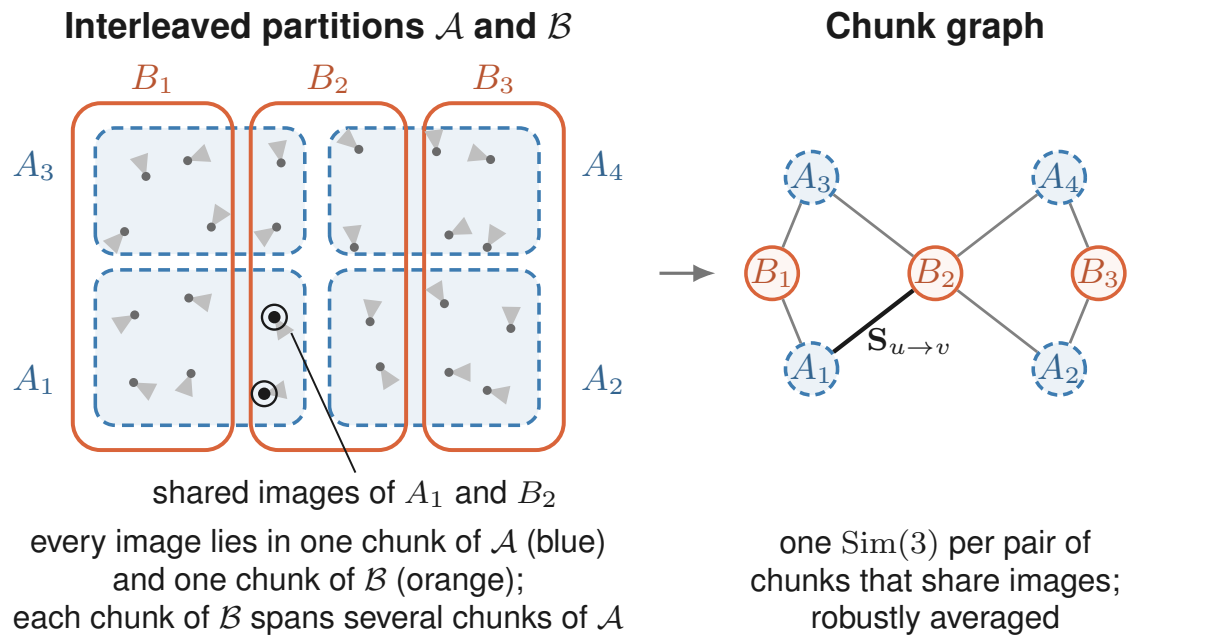

\begin{table*}[t]
\centering
\caption{\textbf{Camera pose accuracy on unordered collections.} Absolute recall (\%) after $\Sim$ alignment, and relative pose AUC (\%). 
``+ \method'' refines the cameras of the row above. 
$^\dagger$: with the ground-truth intrinsics.
\colorbox{pastel_yellow}{Best}, \colorbox{pastel_blue}{second}, and \colorbox{pastel_pink}{third} results are highlighted.
}
\label{tab:main}
\setlength{\tabcolsep}{2.6pt}
\renewcommand{\arraystretch}{0.9}
\resizebox{\linewidth}{!}{
\begin{tabular}{l cccccc ccccccccc ccccccccc}
\toprule
& \multicolumn{6}{c}{ETH3D} & \multicolumn{9}{c}{IMC2021} & \multicolumn{9}{c}{SMERF} \\
\cmidrule(lr){2-7}\cmidrule(lr){8-16}\cmidrule(lr){17-25}
& \multicolumn{3}{c}{Abs. recall} & \multicolumn{3}{c}{AUC} & \multicolumn{3}{c}{5 images (900)} & \multicolumn{3}{c}{10 images (450)} & \multicolumn{3}{c}{25 images (225)} & \multicolumn{3}{c}{Low overlap} & \multicolumn{3}{c}{Medium overlap} & \multicolumn{3}{c}{High overlap} \\
\cmidrule(lr){2-4}\cmidrule(lr){5-7}\cmidrule(lr){8-10}\cmidrule(lr){11-13}\cmidrule(lr){14-16}\cmidrule(lr){17-19}\cmidrule(lr){20-22}\cmidrule(lr){23-25}
Method & 5cm & 10cm & 25cm & @1$^\circ$ & @3$^\circ$ & @5$^\circ$ & @3$^\circ$ & @5$^\circ$ & @10$^\circ$ & @3$^\circ$ & @5$^\circ$ & @10$^\circ$ & @3$^\circ$ & @5$^\circ$ & @10$^\circ$ & @1$^\circ$ & @5$^\circ$ & @20$^\circ$ & @1$^\circ$ & @5$^\circ$ & @20$^\circ$ & @1$^\circ$ & @5$^\circ$ & @20$^\circ$\\
\midrule
COLMAP & 89.0 & 91.5 & 92.0 & 54.3 & 75.4 & 80.7 & 43.3 & 55.3 & 68.6 & 54.9 & 66.2 & 77.8 & 64.7 & 74.7 & 84.7 & 24.2 & 39.7 & 46.0 & 42.9 & 61.7 & 68.0 & 61.1 & 77.2 & 81.6 \\
\rowcolor{keystonerow} \quad \textbf{+ \method} & 90.4 & 90.9 & 91.2 & 72.8 & 82.7 & 85.4 & 45.6 & 57.6 & 70.4 & 56.8 & 67.7 & 78.8 & \third{65.8} & 75.4 & 85.1 & 33.7 & 44.5 & 48.1 & 53.9 & 65.9 & 69.5 & 69.1 & 79.5 & 82.3 \\
GLOMAP & 75.1 & 77.4 & 78.6 & 46.8 & 62.9 & 67.1 & 44.5 & 56.4 & 69.6 & 57.0 & 68.1 & 79.4 & 63.3 & 73.2 & 83.4 & 7.4 & 19.2 & 32.3 & 24.1 & 41.3 & 52.3 & 41.1 & 56.5 & 65.0 \\
\rowcolor{keystonerow} \quad \textbf{+ \method} & 78.2 & 78.4 & 79.4 & 60.7 & 69.3 & 71.6 & 45.8 & 58.0 & 71.1 & 56.9 & 68.1 & 79.4 & 65.6 & 75.2 & 85.0 & 27.9 & 37.8 & 43.9 & 46.4 & 57.5 & 61.4 & 57.6 & 66.8 & 70.3 \\
DGSfM & 93.5 & 97.4 & \third{98.4} & 53.9 & 79.4 & 85.9 & 41.4 & 55.0 & 70.8 & 46.6 & 59.5 & 74.1 & 47.2 & 59.5 & 73.6 & 11.8 & 46.4 & 74.4 & 21.0 & 50.4 & 75.3 & 33.2 & 62.0 & 81.3 \\
\rowcolor{keystonerow} \quad \textbf{+ \method} & \third{98.1} & \third{98.1} & \third{98.4} & 79.4 & 90.6 & 93.2 & 49.1 & 62.2 & 76.4 & 59.2 & 70.7 & \third{82.2} & 64.8 & 74.5 & 84.4 & 58.8 & 83.1 & 91.4 & 69.0 & 88.5 & 94.5 & 79.3 & 93.6 & \third{97.7} \\
GGPT & 94.4 & 96.2 & 96.6 & 62.6 & 82.2 & 87.0 & 41.4 & 55.9 & 72.5 & 50.9 & 64.1 & 77.9 & 55.4 & 67.3 & 79.8 & 42.2 & 61.3 & 68.8 & 51.2 & 67.2 & 74.9 & 27.2 & 34.0 & 36.7 \\
\rowcolor{keystonerow} \quad \textbf{+ \method} & 96.1 & 96.5 & 96.8 & 78.4 & 89.0 & 91.5 & \third{49.5} & \third{62.7} & \second{76.8} & \second{59.4} & \second{71.0} & \first{82.6} & \first{66.3} & \first{76.0} & \first{85.7} & 48.8 & 69.4 & 76.1 & 62.6 & 79.1 & 82.9 & 32.4 & 39.4 & 41.0 \\
GLUEMAP & 90.4 & 97.5 & \second{99.1} & 42.1 & 73.3 & 82.1 & 43.1 & 57.0 & 72.7 & 53.6 & 66.2 & 79.3 & 62.1 & 72.9 & 83.5 & 12.4 & 69.4 & 91.5 & 24.8 & 77.8 & 94.3 & 47.8 & 88.3 & 97.1 \\
\rowcolor{keystonerow} \quad \textbf{+ \method} & \first{99.6} & \first{99.7} & \first{100.0} & \second{80.0} & \first{92.0} & \first{95.0} & \third{49.5} & 62.6 & \third{76.7} & 59.1 & 70.6 & 82.1 & \third{65.8} & 75.4 & 85.0 & \first{65.7} & \first{91.6} & \first{97.5} & \first{74.8} & \first{94.8} & \first{98.7} & \first{82.3} & \first{96.4} & \first{99.1} \\
Solv3R & 75.2 & 91.0 & 97.5 & 22.0 & 58.6 & 72.5 & 24.9 & 40.1 & 60.3 & 27.8 & 43.4 & 63.0 & 31.7 & 47.2 & 66.0 & 1.3 & 35.5 & 76.7 & 2.3 & 45.4 & 83.0 & 4.1 & 54.0 & 87.1 \\
\rowcolor{keystonerow} \quad \textbf{+ \method} & \second{98.2} & \second{99.3} & \first{100.0} & \third{79.9} & \second{91.7} & \second{94.7} & 47.7 & 61.4 & 76.2 & 57.2 & 69.5 & 81.8 & 63.4 & 73.9 & 84.5 & \second{63.8} & \second{89.9} & \second{96.0} & \second{73.9} & \second{94.4} & \second{98.5} & \second{82.0} & \first{96.4} & \first{99.1} \\
Pi3X & 53.6 & 75.5 & 87.7 & 15.8 & 44.3 & 57.1 & 41.0 & 55.4 & 71.7 & 42.1 & 56.6 & 72.4 & 43.8 & 58.0 & 73.5 & 0.5 & 23.8 & 64.5 & 0.7 & 27.6 & 69.1 & 0.9 & 31.3 & 72.7 \\
\rowcolor{keystonerow} \quad \textbf{+ \method} & 94.7 & 97.2 & \second{99.1} & 78.5 & 89.7 & 92.7 & \second{49.6} & \second{62.8} & \first{77.1} & \third{59.3} & \third{70.8} & \second{82.4} & \third{65.8} & \third{75.6} & \third{85.3} & 59.4 & 82.6 & 89.2 & 68.6 & 87.5 & 92.7 & 80.0 & \third{93.7} & 96.7 \\
DVLT & 66.3 & 84.9 & 93.5 & 17.1 & 50.9 & 65.6 & 27.7 & 42.8 & 62.0 & 31.9 & 46.9 & 65.1 & 35.4 & 50.2 & 67.8 & 0.7 & 31.9 & 74.6 & 1.2 & 38.2 & 79.2 & 1.9 & 43.4 & 82.4 \\
\rowcolor{keystonerow} \quad \textbf{+ \method} & 95.3 & 96.9 & \first{100.0} & 79.4 & 90.2 & 93.3 & 47.5 & 61.2 & 75.9 & 57.1 & 69.2 & 81.4 & 62.9 & 73.5 & 84.2 & \third{60.8} & \third{87.0} & \third{94.1} & \third{72.9} & \third{93.3} & 97.5 & \third{80.9} & \second{95.4} & \second{98.5} \\
VGGT-$\Omega$ & 72.6 & 89.6 & 95.7 & 21.4 & 54.9 & 68.7 & 34.3 & 49.0 & 66.9 & 35.0 & 50.0 & 67.5 & 35.4 & 50.5 & 68.1 & 1.2 & 33.3 & 74.2 & 1.7 & 38.5 & 78.9 & 2.1 & 41.3 & 79.1 \\
\rowcolor{keystonerow} \quad \textbf{+ \method} & 98.0 & \first{99.7} & \first{100.0} & \first{80.1} & \third{91.6} & \third{94.6} & \first{49.8} & \first{63.0} & \first{77.1} & \first{59.5} & \first{71.1} & \first{82.6} & \second{66.1} & \second{75.8} & \second{85.5} & 60.6 & 86.9 & \third{94.1} & 71.4 & 92.5 & \third{97.6} & 79.5 & 93.0 & 96.1 \\
\midrule \midrule
GLUEMAP$^\dagger$ & \third{88.6} & \third{97.0} & \third{97.7} & \third{43.2} & \third{73.8} & \third{81.9} & \third{71.4} & \second{81.1} & \third{89.7} & \third{78.2} & \third{85.6} & \third{92.0} & \third{83.2} & \third{88.8} & \third{93.5} & \third{12.5} & \third{68.8} & \second{91.1} & \third{25.1} & \third{77.8} & \second{94.3} & \third{47.5} & \third{88.2} & \third{97.0} \\
\rowcolor{keystonerow} \quad \textbf{+ \method} & \first{99.3} & \first{99.7} & \first{100.0} & \first{81.3} & \first{92.3} & \first{95.1} & \second{77.1} & \first{84.9} & \second{91.7} & \second{81.8} & \second{88.1} & \second{93.5} & \second{85.6} & \second{90.5} & \second{94.7} & \first{65.1} & \first{91.1} & \first{97.2} & \first{75.1} & \first{94.8} & \first{98.7} & \first{82.9} & \first{96.5} & \first{99.1} \\
Pi3X$^\dagger$ & 60.7 & 76.9 & 86.3 & 21.9 & 50.7 & 61.6 & 54.8 & \third{68.2} & 81.3 & 54.9 & 68.2 & 81.2 & 56.6 & 69.4 & 81.9 & 0.5 & 22.4 & 63.6 & 0.7 & 26.9 & 68.6 & 1.0 & 30.2 & 70.3 \\
\rowcolor{keystonerow} \quad \textbf{+ \method} & \second{97.1} & \second{99.0} & \second{99.2} & \second{81.0} & \second{91.2} & \second{93.9} & \first{77.2} & \first{84.9} & \first{91.8} & \first{82.0} & \first{88.4} & \first{93.7} & \first{85.8} & \first{90.7} & \first{94.9} & \second{59.2} & \second{82.1} & \third{89.0} & \second{70.0} & \second{88.3} & \third{93.1} & \second{80.8} & \second{94.1} & \second{97.1} \\
 
\bottomrule
\end{tabular}}
\vspace{-1.0em}
\end{table*}

\subsection{Scaling to Large Scenes}
\label{sec:chunks}
\label{sec:align}
Initializations from SfM or hybrid pipelines cover the whole scene, but a single FF forward pass is limited by GPU memory, and its run time grows quadratically with the number of views (\cref{fig:chunks}, right), so larger scenes have to be predicted in chunks.
A chain~\cite{deng2025vggtlong,lin2025depthanything3,cheng2026merg3r} or tree of chunks rests on one link per chunk, which can misplace a whole chunk, whereas the stars~\cite{pan2026gluemap} link each image to many others but need one forward pass per image (\cref{fig:chunk_layouts}).
We instead link every chunk to several others through two interleaved partitions, at one forward pass per chunk, and place all chunks jointly.
Since the refinement corrects the remaining misalignment, as it does for any initialization, the chunks only need to be placed roughly, and they should be as large as the GPU allows to give each prediction more context.

\noindent \textbf{Interleaved partitions.} 
We divide the images into two partitions of disjoint chunks with at most $M$ images each, $\mathcal{A}=\{A_u\}_{u=1}^{a}$ 
and $\mathcal{B}=\{B_v\}_{v=1}^{b}$, so that every image is predicted twice, once in its chunk of $\mathcal{A}$ and once in its chunk of $\mathcal{B}$.
The two partitions are built so that every chunk $B_v$ shares images with several chunks $A_u$, \ie, $A_u \cap B_v \neq \emptyset$, and the shared images connect all chunks (\cref{fig:partitions}).
For video sequences, consecutive frames are co-visible, so $\mathcal{A}$ cuts the sequence into contiguous chunks of $M$ frames, and $\mathcal{B}$ shifts the same cuts by $M/2$ frames.
For unordered collections, we build a graph that connects every image to its $k$ nearest neighbors, weighted by the descriptor similarity $w_{ii'}$, and obtain $\mathcal{A}$ by spectral clustering~\cite{ng2001spectral,damle2019simple}, as commonly used to partition large SfM problems~\cite{bhowmick2014divide,Zhu_2018_CVPR,chen2020graph}.
Clusters larger than $M$ are split in half at the median of their Fiedler vector~\cite{shi2000normalized}, and very small clusters are merged into their best-connected neighbor.
For $\mathcal{B}$, we cluster the same graph again, with the weights of edges inside the chunks of $\mathcal{A}$ scaled by a small $\varepsilon= 0.005$:
\vspace{-0.3em}
\begin{equation}
    w'_{ii'} = \begin{cases} \varepsilon\, w_{ii'} & \text{if $i, i'$ are in the same chunk of $\mathcal{A}$} \enspace , \\ w_{ii'} & \text{otherwise} \enspace , \end{cases}
\vspace{-0.3em}
\end{equation}
so that cutting through the chunks of $\mathcal{A}$ becomes cheap and the images that $\mathcal{A}$ separates are grouped together.

\noindent \textbf{Chunk similarities.}
For every pair of chunks $A_u$ and $B_v$ that share at least three images, we fit a similarity $\vS_{A_u \to B_v}$ to their shared cameras with the Umeyama method~\cite{umeyama1991least} inside RANSAC~\cite{fischler1981random}.
A shared camera is an inlier if its mapped rotation lies within $10^\circ$ of its prediction in $B_v$ and its mapped center lies within $10\%$ of the spread of the shared cameras, which makes the test independent of the scale of each prediction.
We keep a pair if at least three of its $n$ shared cameras are inliers and weight it by the squared inlier fraction $\lambda_{uv} = (n_\text{in}/n)^2$, so that chunks of visually similar but different places, whose predictions disagree on the shared views, are not linked.

\noindent \textbf{Similarity averaging.}
Rather than composing these similarities along a chain or tree, which accumulates drift, we estimate one similarity $\vS_{A_u}$ or $\vS_{B_v}$ per chunk jointly, like the global alignment of DUSt3R~\cite{dust3r} and the motion averaging of GLUEMAP~\cite{pan2026gluemap}, but over a few hundred chunks instead of thousands of image pairs.
We fix a chunk of the largest connected component as the reference and minimize
\vspace{-0.3em}
\begin{equation}
    \sum_{(u,v)} \lambda_{uv}\, \rho\big( \| \mathrm{Log}( \vS_{A_u \to B_v}^{-1}\, \vS_{B_v}^{-1} \vS_{A_u} ) \|^2 \big) \enspace ,
\vspace{-0.3em}
\end{equation}
where $\mathrm{Log}$ maps a similarity to its 7-dim. tangent vector and $\rho$ is the Cauchy loss.
Intuitively, we seek one similarity per chunk that agrees with all kept relative similarities, \ie, $\vS_{A_u}\approx\vS_{B_v}\,\vS_{A_u \to B_v}$.
We solve it with IRLS~\cite{chatterjee2013efficient}, dropping edges with large residuals so that a few wrong links do not distort the solution.
The initial camera of every image then follows from \cref{eq:sim3} with the similarity of its chunk in $\mathcal{A}$, or in $\mathcal{B}$ if the former is not connected to the reference, and enters the refinement like any other initialization.


\begin{table}[t]
\centering
\caption{\textbf{Camera pose accuracy on images with lens distortion.} 
\method estimates the radial distortion, or keeps the ground-truth distortion fixed in the case of calibrated cameras (denoted as $^\dagger$).
}
\label{tab:distorted}
\vspace{-0.3em}
\setlength{\tabcolsep}{2.8pt}
\renewcommand{\arraystretch}{0.9}
\resizebox{\linewidth}{!}{
\begin{tabular}{l cccccc cccc}
\toprule
& \multicolumn{6}{c}{ETH3D} & \multicolumn{4}{c}{Rotunda} \\
\cmidrule(lr){2-7}\cmidrule(lr){8-11}
Method & 5cm & 10cm & 25cm & @1$^\circ$ & @3$^\circ$ & @5$^\circ$ & @1$^\circ$ & @3$^\circ$ & @5$^\circ$ & @30$^\circ$ \\
\midrule

COLMAP & 91.2 & 93.3 & 93.8 & 55.8 & 77.4 & 82.8 & 50.4 & 81.4 & 88.5 & \third{98.0} \\
\rowcolor{oursrow} \quad \textbf{+ \method} & 93.5 & 93.9 & 94.2 & 70.9 & 84.2 & 87.3 & 54.7 & 82.9 & 89.4 & \second{98.2} \\
GLOMAP & 78.9 & 79.7 & 80.0 & 46.4 & 63.1 & 67.3 & 9.6 & 30.5 & 40.0 & 61.1 \\
\rowcolor{oursrow} \quad \textbf{+ \method} & 80.7 & 80.7 & 81.0 & 58.9 & 69.4 & 72.0 & 21.2 & 44.4 & 51.5 & 65.8 \\
DGSfM & 36.9 & 52.0 & 62.1 & 7.7 & 28.6 & 42.6 & 0.0 & 0.4 & 1.9 & 43.3 \\
\rowcolor{oursrow} \quad \textbf{+ \method} & 94.1 & 94.9 & 96.3 & \second{75.5} & 87.5 & 90.5 & 33.0 & 53.1 & 58.2 & 66.5 \\
GGPT & 30.6 & 45.5 & 56.0 & 5.4 & 22.9 & 36.8 & 0.0 & 0.7 & 4.0 & 71.4 \\
\rowcolor{oursrow} \quad \textbf{+ \method} & 90.1 & 92.1 & 94.4 & 71.3 & 83.7 & 87.3 & \third{55.3} & \second{83.3} & \third{89.6} & \second{98.2} \\
GLUEMAP & 84.6 & 91.5 & 97.1 & 41.1 & 71.8 & 80.5 & 45.5 & 79.7 & 87.7 & 97.9 \\
\rowcolor{oursrow} \quad \textbf{+ \method} & \first{97.4} & \second{98.2} & 98.9 & \first{75.9} & \first{89.8} & \first{93.1} & \first{55.7} & \second{83.3} & \second{89.7} & \second{98.2} \\
Solv3R & 42.1 & 64.4 & 79.9 & 6.7 & 33.6 & 51.2 & 0.1 & 6.2 & 23.8 & 85.2 \\
\rowcolor{oursrow} \quad \textbf{+ \method} & 94.0 & \first{98.9} & \second{99.6} & 75.2 & \second{88.9} & \second{92.8} & 20.5 & 64.2 & 78.2 & 96.3 \\
Pi3X & 32.1 & 53.9 & 76.1 & 3.9 & 26.1 & 41.9 & 0.2 & 6.4 & 24.0 & 84.8 \\
\rowcolor{oursrow} \quad \textbf{+ \method} & 92.3 & \third{97.3} & 98.0 & 73.1 & 86.7 & 90.3 & 54.5 & \third{83.0} & 89.5 & \second{98.2} \\
DVLT & 40.8 & 59.6 & 75.6 & 5.2 & 29.8 & 47.3 & 0.8 & 10.6 & 27.0 & 83.0 \\
\rowcolor{oursrow} \quad \textbf{+ \method} & \third{94.9} & 96.0 & \third{99.2} & 75.1 & \third{88.6} & 92.3 & 53.6 & 82.6 & 89.2 & \second{98.2} \\
VGGT-$\Omega$ & 41.2 & 68.5 & 84.8 & 5.9 & 33.4 & 51.3 & 0.0 & 1.9 & 8.4 & 75.5 \\
\rowcolor{oursrow} \quad \textbf{+ \method} & \second{95.4} & 97.2 & \first{99.7} & \third{75.3} & \second{88.9} & \third{92.6} & \second{55.6} & \first{83.4} & \first{89.8} & \first{98.3} \\
\midrule 
GLUEMAP$^\dagger$ & \third{87.0} & \third{92.9} & \second{96.3} & \third{39.7} & \third{71.8} & \third{80.1} & \second{65.2} & \second{87.8} & \second{92.6} & \second{98.7} \\
\rowcolor{oursrow} \quad \textbf{+ \method} & \first{97.7} & \first{98.1} & \first{98.8} & \first{77.2} & \first{89.7} & \first{92.8} & \third{63.3} & \third{87.1} & \third{92.2} & \second{98.7} \\
Pi3X$^\dagger$ & 51.7 & 72.2 & \third{84.9} & 9.9 & 39.4 & 53.5 & 0.1 & 4.2 & 16.7 & \third{81.9} \\
\rowcolor{oursrow} \quad \textbf{+ \method} & \second{95.5} & \second{97.5} & \first{98.8} & \second{75.9} & \second{88.9} & \second{92.2} & \first{65.5} & \first{88.0} & \first{92.8} & \first{98.8} \\
 
\bottomrule
\end{tabular}}
\vspace{-1.6em}
\end{table}

\section{Experiments}
\label{sec:experiments}
\noindent \textbf{Datasets.} 
For \textit{unordered collections}, we use ETH3D~\cite{eth3dsfm}, comprising 25 scenes with 7--110 images each; IMC2021~\cite{imc2021}, using 900, 450, and 225 bags of 5, 10, and 25 internet photos, respectively; and SMERF~\cite{duckworth2024smerf}, whose 100 collections of 29--265 images sample four ZipNeRF~\cite{barron2023zipnerf} videos at different overlap levels~\cite{mpsfm}. 
All three use undistorted images.
For \textit{lens distortion}, we use the original distorted ETH3D images and the 62 GoPro images of the Rotunda scene~\cite{kocur2026minimal}.
For \textit{video sequences}, we use 55 ETH3D SLAM sequences~\cite{schops2019bad} with 17--885 frames each and the nine Aachen v1.1 sequences~\cite{Zhang2020ARXIV} with 17--1{,}044 frames each.
For \textit{large collections}, we use the HGE and LIN scenes of LaMAR~\cite{sarlin2022lamar}, with 7{,}553 and 9{,}319 images, and the 4{,}324 database images of Aachen~\cite{Sattler2018CVPR, Sattler2012BMVC}.

\noindent \textbf{Baselines.} 
We compare against a wide range of baselines: \textit{Classical SfM} (COLMAP~\cite{colmap}, GLOMAP~\cite{glomap} with LoMa~\cite{nordstrom2026loma} matches), \textit{SfM with learned priors} (DGSfM~\cite{aung2026dgsfm}, VidMap~\cite{pataki2026vidmap}),  \textit{hybrid} methods combining FF predictions with classical methods (GGPT~\cite{chen2026ggpt}, GLUEMAP~\cite{pan2026gluemap}, MERG3R~\cite{cheng2026merg3r}), \textit{scalable FF networks} (Solv3R~\cite{kwak2026solv3r},  CUT3R~\cite{wang2025cut3r}), \textit{FF-based SLAM approaches} (VGGT-Long~\cite{deng2025vggtlong}), and \textit{FF networks} (Pi3X~\cite{pi3}, DVLT~\cite{burzio2026dejaview}, VGGT-$\Omega$~\cite{wang2026vggtomega}). 
We apply \method on top of each baseline.

\noindent \textbf{Implementation details.} 
FF networks predict each scene of \textit{unordered collections} and \textit{lens distortion} in a single forward pass, while \textit{video sequences} and \textit{large collections} are split into chunks of at most $M = 60$ views unless stated otherwise.
GLUEMAP, MERG3R and VGGT-Long use their own chunking layout; other baselines use none.
For \method, we use LoMa-B~\cite{nordstrom2026loma} matches with 4{,}096 keypoints, exhaustively for small collections, within a window of $w = 10$ frames plus the top-$20$ MegaLoc~\cite{Berton_2025_MegaLoc} neighbors for \textit{video sequences}, and with the top-$60$ neighbors for \textit{large collections}.
Intrinsics are shared per scene on ETH3D (undistorted, distorted, SLAM), SMERF, Aachen sequences and LaMAR.
The refinement uses $R=5$ rounds with 5 inner loops unless stated otherwise. 
We use a pinhole model by default, a simple pinhole model on IMC2021, where every image has its own camera, and the radial model on distorted images.
We use COLMAP's global bundle adjustment with the soft-$\ell_1$ loss, at most 100 iterations.

\noindent \textbf{Hardware.}
Feed-forward inference, local feature matching, and retrieval run on a A100 GPU.
Similarity averaging and \method refinement run on a CPU node with 8 threads and 64\,GB of memory.

\noindent \textbf{Metrics.} 
We report the AUC of the relative pose error, defined as the maximum of rotation and translation-direction errors, over all image pairs~\cite{aung2026dgsfm}, with thresholds set per dataset.
For datasets with metric ground truth, we also report the absolute recall: percentage of all images whose camera position and orientation lie within given thresholds after RANSAC-based $\Sim$ alignment to the ground truth.

\begin{table}[t]
\centering
\caption{\textbf{Camera pose accuracy on video sequences.} 
$^\dagger$: calibrated. $^\ddagger$: our chunking (\cref{sec:chunks}). $^\S$: the method's own chunking. 
}
\label{tab:sequences}
\vspace{-0.3em}
\setlength{\tabcolsep}{2.8pt}
\renewcommand{\arraystretch}{0.9}
\resizebox{\linewidth}{!}{
\begin{tabular}{l cccccc cccccc}
\toprule
& \multicolumn{6}{c}{ETH3D SLAM} & \multicolumn{6}{c}{Aachen Seq.} \\
\cmidrule(lr){2-7}\cmidrule(lr){8-13}
Method & 5cm & 10cm & 25cm & @1$^\circ$ & @3$^\circ$ & @5$^\circ$ & 0.5m & 1m & 2m & @3$^\circ$ & @10$^\circ$ & @30$^\circ$\\
\midrule
COLMAP & 50.7 & 58.2 & 60.2 & 0.9 & 17.1 & 36.8 & \second{80.3} & \second{83.6} & \third{85.9} & \first{63.3} & 77.3 & 82.9 \\
\rowcolor{keystonerow} \quad \textbf{+ \method} & 90.0 & 91.6 & 92.1 & \first{42.9} & \second{70.2} & 78.2 & \first{83.1} & \first{84.0} & 84.1 & \first{63.3} & 77.2 & 83.0 \\
GLOMAP & 50.9 & 59.2 & 61.6 & 0.9 & 15.4 & 33.8 & 65.6 & 67.1 & 67.9 & 45.6 & 55.5 & 61.0 \\
\rowcolor{keystonerow} \quad \textbf{+ \method} & \second{94.2} & \second{96.2} & \second{96.3} & \third{40.7} & \third{69.2} & \second{78.4} & 69.2 & 69.6 & 69.6 & 47.6 & 57.9 & 62.7 \\
DGSfM & 13.4 & 16.6 & 17.4 & 0.2 & 3.7 & 9.4 & 40.0 & 50.0 & 55.0 & 28.7 & 59.5 & 82.5 \\
\rowcolor{keystonerow} \quad \textbf{+ \method} & 38.2 & 39.1 & 39.7 & 12.8 & 24.8 & 29.6 & 72.6 & 76.5 & 78.4 & 56.7 & \first{80.0} & \first{92.3} \\
VidMap & 52.1 & 54.5 & 55.0 & 4.4 & 17.7 & 29.6 & 47.1 & 61.2 & 75.0 & 18.9 & 55.0 & 82.7 \\
\rowcolor{keystonerow} \quad \textbf{+ \method} & 89.5 & 90.5 & 90.7 & 39.3 & 65.0 & 73.2 & 72.3 & \third{82.9} & \first{88.7} & 43.1 & \third{78.8} & \second{92.2} \\
CUT3R & 2.5 & 5.0 & 6.6 & 0.0 & 0.3 & 1.1 & 9.4 & 13.6 & 21.1 & 1.4 & 14.3 & 32.3 \\
\rowcolor{keystonerow} \quad \textbf{+ \method} & 53.6 & 55.8 & 56.4 & 16.7 & 33.8 & 40.9 & 55.6 & 57.5 & 58.8 & 35.0 & 45.1 & 52.8 \\
Solv3R & 30.9 & 37.6 & 39.2 & 0.1 & 5.4 & 16.9 & 29.5 & 39.3 & 49.4 & 12.8 & 31.5 & 49.7 \\
\rowcolor{keystonerow} \quad \textbf{+ \method} & 88.7 & 89.7 & 90.7 & 34.8 & 63.0 & 72.0 & 57.2 & 58.9 & 60.9 & 36.6 & 48.2 & 61.3 \\
VGGT-Long$^\S$ & 18.8 & 25.3 & 29.0 & 0.1 & 4.2 & 13.1 & 32.4 & 37.7 & 50.6 & 14.7 & 39.7 & 62.6 \\
\rowcolor{keystonerow} \quad \textbf{+ \method} & 81.8 & 84.0 & 84.4 & 31.6 & 57.3 & 65.4 & 58.5 & 62.5 & 65.5 & 40.1 & 55.3 & 70.0 \\
MERG3R$^\S$ & 30.9 & 37.1 & 38.8 & 0.3 & 7.5 & 19.9 & 51.5 & 59.6 & 65.0 & 30.0 & 53.0 & 72.9 \\
\rowcolor{keystonerow} \quad \textbf{+ \method} & 88.0 & 89.4 & 90.2 & 35.3 & 63.4 & 72.4 & 73.7 & 77.3 & 79.1 & 54.5 & 70.6 & 82.2 \\
GLUEMAP$^\S$ & 41.8 & 47.4 & 48.9 & 1.1 & 14.1 & 32.1 & 73.1 & 77.3 & 77.9 & 57.6 & 75.3 & 83.4 \\
\rowcolor{keystonerow} \quad \textbf{+ \method} & \first{95.8} & \first{96.8} & \first{97.3} & \second{42.3} & \first{71.7} & \first{80.7} & 76.7 & 79.1 & 80.9 & \third{59.9} & 76.7 & 84.0 \\
Pi3X$^\ddagger$ & 28.6 & 36.0 & 37.8 & 0.2 & 6.5 & 18.7 & 53.7 & 62.1 & 73.0 & 38.8 & 67.4 & 83.1 \\
\rowcolor{keystonerow} \quad \textbf{+ \method} & 92.2 & 93.8 & 94.3 & 38.7 & 67.8 & 76.6 & \third{77.9} & 81.1 & \second{86.6} & \second{61.0} & \second{79.7} & 88.7 \\
DVLT$^\ddagger$ & 31.8 & 38.7 & 41.2 & 0.3 & 7.8 & 21.7 & 53.9 & 58.9 & 65.3 & 29.8 & 62.7 & 84.3 \\
\rowcolor{keystonerow} \quad \textbf{+ \method} & 93.5 & \third{95.1} & \third{95.8} & 38.6 & 68.9 & \third{78.3} & 76.4 & 80.8 & 84.7 & 57.3 & 78.6 & \third{90.6} \\
VGGT-$\Omega$$^\ddagger$ & 41.7 & 52.2 & 54.6 & 0.4 & 10.3 & 25.5 & 57.4 & 69.9 & 75.9 & 36.8 & 66.3 & 86.2 \\
\rowcolor{keystonerow} \quad \textbf{+ \method} & \third{93.7} & 95.0 & 95.4 & 40.0 & 69.0 & 78.1 & 72.4 & 76.9 & 80.5 & 56.2 & \first{80.0} & \second{92.2} \\
\midrule \midrule
Pi3X$^{\dagger\ddagger}$ & \second{24.2} & \second{28.7} & \second{32.8} & \second{0.2} & \second{5.6} & \second{17.0} & \second{61.5} & \second{67.2} & \second{73.9} & \second{43.9} & \second{73.1} & \second{89.2} \\
\rowcolor{keystonerow} \quad \textbf{+ \method} & \first{91.1} & \first{92.2} & \first{93.2} & \first{49.5} & \first{72.4} & \first{79.1} & \first{77.4} & \first{81.7} & \first{85.1} & \first{62.6} & \first{82.9} & \first{92.9} \\
 
\bottomrule
\end{tabular}}
\vspace{-1em}
\end{table}

\begin{table}[t]
\centering
\caption{\textbf{Camera pose accuracy on large image collections.} $^\ddagger$: our chunking (\cref{sec:chunks}). $^\S$: the method's own chunking. 
}
\label{tab:large}
\vspace{-0.3em}
\setlength{\tabcolsep}{2.6pt}
\renewcommand{\arraystretch}{0.9}
\resizebox{\linewidth}{!}{
\begin{tabular}{l ccccc ccccc ccccc}
\toprule
& \multicolumn{5}{c}{LaMAR HGE (7,553)} & \multicolumn{5}{c}{LaMAR LIN (9,319)} & \multicolumn{5}{c}{Aachen DB (4,324)}\\
\cmidrule(lr){2-6}\cmidrule(lr){7-11}\cmidrule(lr){12-16}
\cmidrule(lr){2-3}\cmidrule(lr){4-6}\cmidrule(lr){7-8}\cmidrule(lr){9-11}\cmidrule(lr){12-13}\cmidrule(lr){14-16}
Method & 1m & 10m & @3$^\circ$ & @10$^\circ$ & @30$^\circ$ & 1m & 10m & @3$^\circ$ & @10$^\circ$ & @30$^\circ$ & 1m & 2m & @3$^\circ$ & @10$^\circ$ & @30$^\circ$\\
\midrule
COLMAP & 13.2 & 14.7 & 3.2 & 6.1 & 13.4 & 73.6 & 88.7 & 40.4 & 67.4 & 77.4 & 29.0 & 29.0 & 23.8 & 28.2 & 30.1 \\
\rowcolor{keystonerow} \quad \textbf{+ \method} & 13.3 & 14.7 & 3.5 & 6.2 & 13.4 & \third{78.7} & 89.1 & \second{46.6} & \third{69.9} & 78.3 & 28.9 & 28.9 & 23.6 & 28.2 & 30.1 \\
GLOMAP & 0.0 & 0.0 & 1.4 & 3.7 & 8.3 & 18.7 & 81.4 & 18.1 & 54.1 & 72.4 & 56.4 & 57.0 & 31.3 & 37.2 & 40.3 \\
\rowcolor{keystonerow} \quad \textbf{+ \method} & 0.4 & 11.8 & 1.5 & 3.7 & 8.4 & 24.7 & 84.6 & 23.4 & 55.9 & 73.1 & \third{56.7} & 57.1 & 32.2 & 37.5 & 40.4 \\
GLUEMAP$^\S$ & \second{61.4} & \first{98.1} & \second{37.1} & \second{77.4} & \second{92.0} & \second{78.9} & \second{94.0} & \third{44.6} & \second{74.8} & \second{85.5} & \first{58.9} & \second{64.4} & \first{51.4} & \first{67.3} & \first{79.6} \\
\rowcolor{keystonerow} \quad \textbf{+ \method} & \first{66.4} & \first{98.1} & \first{40.6} & \first{79.0} & \first{92.5} & \first{84.6} & \first{94.1} & \first{49.3} & \first{77.2} & \first{86.0} & \first{58.9} & \first{64.7} & \second{50.8} & \second{67.1} & \second{79.5} \\
Pi3X$^\ddagger$ & 38.8 & \third{81.8} & 13.3 & 48.6 & 70.3 & 5.9 & 43.5 & 4.0 & 24.4 & 53.5 & 23.5 & 41.7 & 14.6 & 40.3 & 53.8 \\
\rowcolor{keystonerow} \quad \textbf{+ \method} & \third{42.8} & \second{86.5} & \third{24.3} & \third{56.6} & \third{73.4} & 52.1 & 88.0 & 26.3 & 60.7 & 76.4 & \second{57.1} & \third{60.5} & \third{41.9} & \third{54.3} & \third{58.1} \\
DVLT$^\ddagger$ & 1.0 & 11.2 & 0.4 & 6.7 & 33.2 & 1.0 & 15.8 & 0.7 & 9.5 & 42.4 & 18.0 & 29.6 & 6.3 & 28.2 & 46.2 \\
\rowcolor{keystonerow} \quad \textbf{+ \method} & 23.8 & 48.5 & 6.3 & 22.3 & 47.3 & 31.7 & 83.9 & 15.2 & 50.9 & 72.4 & 28.9 & 35.7 & 13.4 & 32.9 & 48.0 \\
VGGT-$\Omega$$^\ddagger$ & 8.2 & 17.0 & 1.9 & 8.3 & 17.2 & 11.9 & 45.7 & 3.3 & 24.5 & 59.9 & 6.5 & 8.3 & 4.9 & 22.6 & 37.3 \\
\rowcolor{keystonerow} \quad \textbf{+ \method} & 15.3 & 21.5 & 4.9 & 18.8 & 35.0 & 61.3 & \third{90.5} & 30.5 & 65.3 & \third{79.0} & 48.7 & 55.8 & 37.1 & 50.6 & 57.4 \\
 
\bottomrule
\end{tabular}}
\vspace{-1.6em}
\end{table}
\begin{figure*}[t]
\centering
\includegraphics[width=0.85\linewidth]{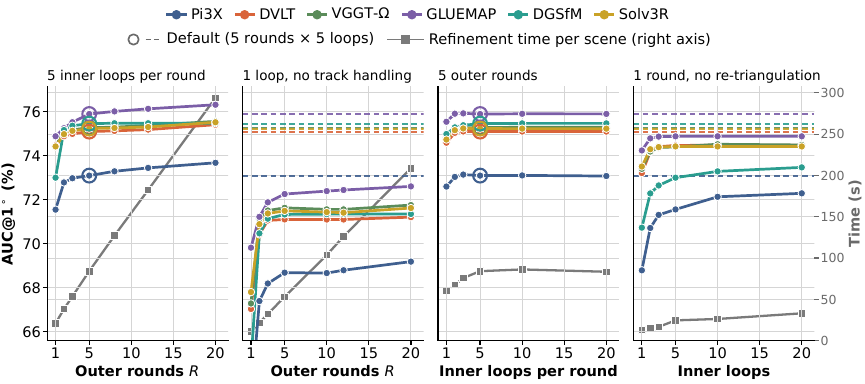}
\vspace{-0.2em}
\caption{\textbf{Refinement loop on ETH3D distorted.} AUC and refinement time per scene (median over initializations) against outer rounds $R$ (left two) and inner loops per round (right two), each with and without the remaining steps of the loop as in plot titles.
}
\label{fig:refinement_loop}
\vspace{-1.2em}
\end{figure*}

\subsection{Evaluation}
\label{sec:benchmarks}

\noindent \textbf{Unordered collections.} 
\Cref{tab:main} reports the results on ETH3D, IMC2021, and SMERF.
On their own, the three FF networks are substantially less accurate than SfM on ETH3D.
After refinement with \method, all three achieve the highest AUC@1$^\circ$, about four times their initial accuracy.
\method also improves the cameras from every other initialization, and the refined cameras reach nearly the same accuracy whether they start from an FF network, a hybrid method, or SfM method with learned priors.
\Cref{fig:teaser} shows the same trend for four more FF networks, DA3~\cite{lin2025depthanything3}, MapAnything~\cite{keetha2026mapanything}, ZipMap~\cite{jin2026zipmap}, and FastVGGT~\cite{shen2026fastvggt}.
Across the seven FF networks, the spread in AUC@3$^\circ$ shrinks from 44 to 11 points after refinement.
\method thus turns noisy but roughly correct poses into highly accurate ones, so \emph{the initialization matters far less than its refinement.}
The gain is largest on SMERF, where weak overlap leaves the FF networks with almost no precise cameras; with \method, they surpass every baseline on its own, and GLUEMAP and Solv3R refined by \method perform best.
On IMC2021, \method is again the most accurate on small bags, though its margin over COLMAP shrinks as the bags grow, since the incremental pipeline benefits more from additional views.
Each internet photo from the IMC2021 dataset comes from a different camera, so its focal length must be estimated separately from the other images.
Using the ground-truth intrinsics increases AUC@3$^\circ$ by more than 20 points, compared with the small gain observed for ETH3D (where intrinsics are shared between images).

\noindent \textbf{Images with lens distortion.} 
\Cref{tab:distorted} reports the results on the distorted ETH3D images and on Rotunda. 
FF networks do not model distortion. 
Especially on the strongly distorted Rotunda images, they recover almost no camera pose precisely. 
\method nonetheless recovers from these initializations, refining the radial distortion together with the other camera parameters. 
On Rotunda, every FF network refined by \method matches COLMAP (which uses the same camera model to model distortions) refined by \method. 
Visualizations in \cref{fig:distorted_map} show that the initial FF poses are inaccurate rather than completely wrong.  
Wrong pose estimates cannot be repaired: on Rotunda, some cameras of DGSfM and GLOMAP remain inaccurate after refinement. (\cf \cref{fig:distorted_map_b} for visualizations).

\noindent \textbf{Video sequences.} 
\Cref{tab:sequences} reports the results on the ETH3D SLAM and Aachen sequences.
All FF networks run on 60-frame chunks, 
except Solv3R and CUT3R, which runs in its streaming mode.
On their own, these predictions are coarse: the chunks are linked only through the frames they share, so small errors accumulate along each sequence. 
\method removes this drift, since retrieval matches between distant frames join the same tracks and BA corrects all chunks together. 
COLMAP and GLOMAP perform poorly on these sequences on their own, but become as accurate as the refined FF networks with \method. 
Thus, \emph{the refinement again matters more than the initialization.}
On the Aachen sequences, COLMAP is the most precise method at the tightest thresholds, and \method does not improve it further.
The refined FF networks come close to it at 3$^\circ$ and 0.5\,m, and are clearly ahead at 10$^\circ$ and 30$^\circ$, due to a few badly misplaced cameras of COLMAP.
\method also corrects the drift of VidMap and DGSfM on Aachen. 
On ETH3D SLAM, DGSfM reconstructs only 22 of the 55 sequences, which explains its poor performance. 

\noindent \textbf{Large image collections.} 
\Cref{tab:large} reports results on LaMAR and Aachen database, which exceed the memory of a single FF call (\cref{fig:chunks}, right). 
We thus chunk them for the FF networks, and \method again clearly improves their accuracy.
The classical pipelines vary across scenes: COLMAP is among the most precise methods on LIN but reconstructs only a small part of HGE and Aachen, and GLOMAP fails on HGE. 
\method also improves the classical and hybrid pipelines on LaMAR, mostly at tight thresholds, and GLUEMAP with \method is the most accurate on HGE and LIN.
On Aachen, it changes little, since the remaining errors there are lost or misplaced parts of the scene rather than imprecise cameras (\cref{sec:failure}).
After GLUEMAP, Pi3X with \method is the most accurate on HGE and Aachen. 
The gap likely lies in placing the chunks rather than predicting them: GLUEMAP predicts a chunk per image, whereas we use far fewer, less overlapping chunks (\cf \cref{fig:chunk_layouts}).
\emph{On large collections, the refinement thus depends on chunks being placed roughly right.}

\subsection{Ablation Studies}
\noindent \textbf{Refinement loop.} 
\Cref{fig:refinement_loop} varies the outer and inner loops of the refinement on ETH3D distorted for six initializations.
The second round gives the largest gain, as it re-triangulates at the cameras improved by the first, and beyond $R=5$, the accuracy gains little while processing time keeps growing (left).
Without the inner loop, the accuracy stays clearly below the default even after 20 rounds, as the correspondences of inaccurate initial cameras never form clean, long tracks (second).
More than five inner loops bring no benefit, as the optimization typically converges within five (third).
Repeated re-triangulation is also necessary: with a single triangulation, the cameras get stuck in local minima, however many inner loops are run (right).
\emph{Track handling is thus the main ingredient and re-triangulation the second}; we therefore use 5 rounds of 5 inner loops.

\begin{figure}[t]
\centering
\includegraphics[width=\linewidth]{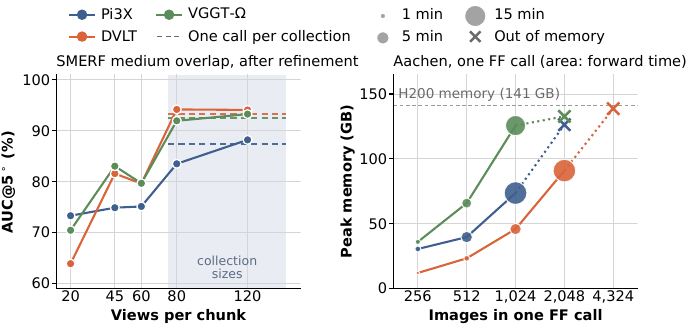}
\vspace{-1.2em}
\caption{\textbf{Chunk size and the cost of a single call.} Left: \method on the SMERF medium overlap collections (75 to 142 images) for different chunk sizes. Right: peak memory and forward time of a single FF call on the first $N$ Aachen images.}
\label{fig:chunks}
\vspace{-0.2em}
\end{figure}

\begin{figure}[t]
\centering
{\scriptsize\setlength{\tabcolsep}{1pt}
\renewcommand{\arraystretch}{0.8}
\begin{tabular*}{0.95\linewidth}{@{\extracolsep{\fill}} l cc rrrr @{}}
\toprule
Layout & FF & + \method & Chunks & FF Calls & Time & Links \\
\midrule
\input{tab/rows/chunks_layout_rows.tex} 
\bottomrule
\vspace{0.1em}
\end{tabular*}}
\includegraphics[width=\linewidth]{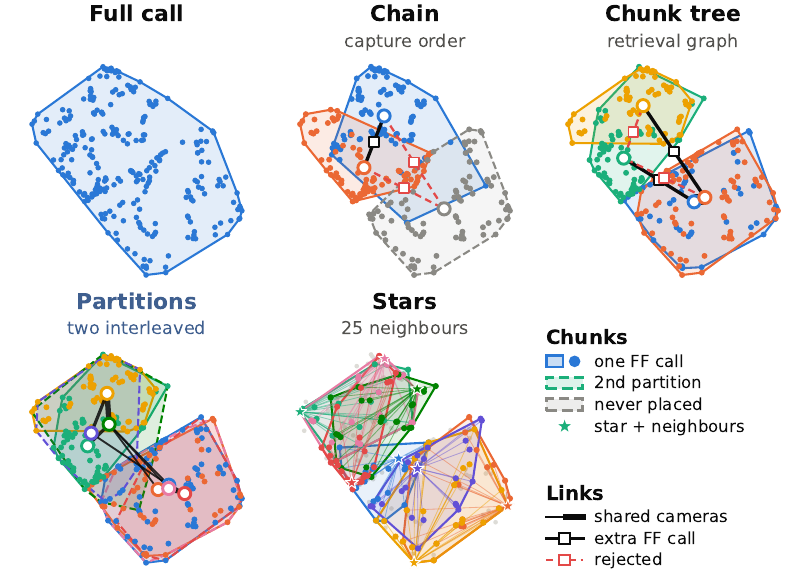}
\vspace{-1.2em}
\caption{\textbf{Chunk layouts on SMERF high-overlap} (London, 265 images, Pi3X). Top: AUC@3$^\circ$, calls and accepted links. Bottom: chunks and links seen from above (8 of 265 stars shown). The chain and tree reject most links and lose whole chunks; the partitions keep all and nearly match the full call at twice its runtime.}
\label{fig:chunk_layouts}
\vspace{-1.6em}
\end{figure}

\noindent \textbf{Chunk size and runtime.} 
\Cref{fig:chunks} (left) runs the partition chunking on the 20 medium-overlap SMERF collections with 20 to 120 images per chunk, although each fits in a single call.
The accuracy after refinement grows with the chunk size and reaches that of a single FF call once a chunk holds most of the collection.
Smaller chunks give each prediction less context and share fewer cameras for the alignment, so their initial cameras are less accurate, and the refinement cannot fully recover them.
Chunks should thus be as large as memory allows.
A single call, however, scales poorly (\cref{fig:chunks}, right): its memory and runtime grow quickly with the image count, so on a 141\,GB H200, Pi3X and VGGT-$\Omega$ run out of memory at 2{,}048 Aachen images, and even DVLT cannot process all 4{,}324.
\emph{Chunking is thus a necessity for large collections, not a choice}, and the partitions of \cref{sec:chunks} make it affordable.

\noindent \textbf{Chunk layouts.} 
\Cref{fig:chunk_layouts} compares the partitions chunking with prior chunk layouts on one SMERF collection.
The chain~\cite{deng2025vggtlong,pilongcode2025,lin2025depthanything3,cheng2026merg3r} links each chunk only to the next, in capture order or, for MERG3R~\cite{cheng2026merg3r}, by visual similarity.
The chunk tree~\cite{Zhu_2018_CVPR,chen2020graph} groups images on the retrieval graph and links each chunk to a single neighbor.
Both rest on one link per chunk, so a rejected link loses or misplaces a whole chunk, which refinement cannot recover.
The stars~\cite{pan2026gluemap} predict every image with its 25 nearest neighbors, giving redundant links at one FF call per image and 18 times the full call's runtime.
The interleaved partitions instead link every chunk to several others through shared cameras, keep all links, and nearly match the full call at twice its runtime.

\begin{table}[t]
\centering
\setlength{\tabcolsep}{2.8pt}
\renewcommand{\arraystretch}{0.9}
\caption{\textbf{Intrinsics refinement on ETH3D.} Abs. recall at 5cm and AUC@1$^\circ$/@3$^\circ$ per network after one round; pp: principal point.}
\label{tab:intrinsics}
\vspace{-0.4em}
\resizebox{\linewidth}{!}{%
\begin{tabular}{l cc cc cc}
\toprule
Model & \multicolumn{2}{c}{Pi3X} & \multicolumn{2}{c}{DVLT} & \multicolumn{2}{c}{VGGT-$\Omega$} \\
\midrule
\fade{Feed-forward} & 53.6 & 15.8\,/\,44.3 & 66.3 & 17.1\,/\,50.9 & 72.6 & 21.4\,/\,54.9 \\
Per image & 90.4 & \underline{57.2}\,/\,\underline{79.8} & 76.1 & 32.9\,/\,65.7 & 95.2 & 55.3\,/\,80.6 \\
\quad + pp. & 83.3 & 55.6\,/\,75.5 & \underline{87.3} & \underline{56.7}\,/\,\underline{76.6} & 87.8 & \underline{57.6}\,/\,79.4 \\
Shared & \underline{91.3} & 53.6\,/\,78.0 & 84.8 & 41.3\,/\,71.7 & \underline{95.4} & 55.2\,/\,\underline{80.8} \\
\quad + pp. & \textbf{93.7} & \textbf{76.3}\,/\,\textbf{88.1} & \textbf{95.3} & \textbf{78.9}\,/\,\textbf{90.0} & \textbf{98.0} & \textbf{80.0}\,/\,\textbf{91.6} \\
\bottomrule
\end{tabular}}
\vspace{-0.6em}
\end{table}

\noindent \textbf{Camera model.} 
\Cref{tab:intrinsics} varies the refined camera model on ETH3D. 
Sharing intrinsics between images taken with the same camera (``Shared''), combined with refining the principal point (``+pp''), significantly improves \method compared to optimizing individual intrinsics per image (``Per image''). 
Note that BA can absorb a rotation error into a shifted principal point~\cite{hartley2003multiple}, which can easily happen if intrinsic are not shared. 
\emph{The accuracy at tight thresholds is thus limited by the intrinsics rather than the poses.}

\begin{table}[t]
\centering
\caption{\textbf{Ablation of initial 3D points on ETH3D distorted.} AUC@1$^\circ$\,/\,@3$^\circ$ of \method with five rounds and five inner loops. Default triangulates the matches at current cameras in every round.}
\label{tab:loop_variants}
\vspace{-0.3em}
\resizebox{\linewidth}{!}{%
\begin{tabular}{l ccc}
\toprule
Variant & Pi3X & DVLT & VGGT-$\Omega$ \\
\midrule
\textbf{Default: triangulation, full inner loop} & \textbf{73.1\,/\,86.7} & \textbf{75.1\,/\,88.6} & \textbf{75.3\,/\,88.9} \\
Points from FF depth, lifted once & 69.6\,/\,84.8 & 72.6\,/\,87.7 & 73.0\,/\,88.0 \\
Points from FF depth, re-lifted every round & 61.6\,/\,77.9 & 61.4\,/\,78.9 & 69.8\,/\,86.2 \\
\bottomrule
\end{tabular}}
\vspace{-1.0em}
\end{table}

\begin{figure*}[t]
\centering
\includegraphics[width=0.95\linewidth]{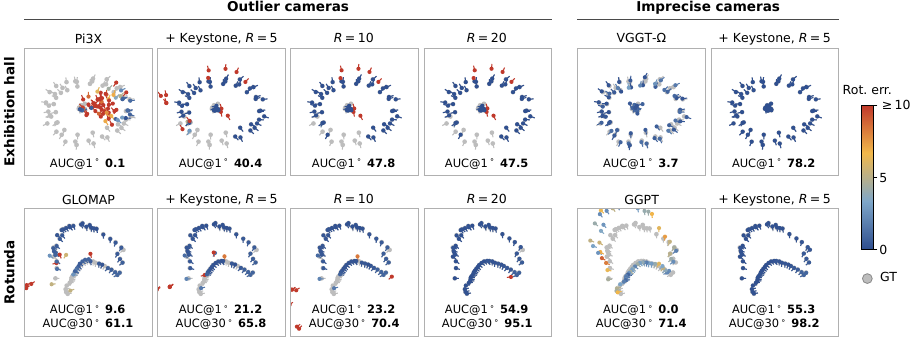}
\caption{\textbf{Outlier vs.\ imprecise initial cameras.} Top-down camera layouts against the ground truth (grey), colored by rotation error, before and after \method with $R$ rounds. \method corrects imprecise cameras within a few rounds, but outlier cameras only slowly (Rotunda) or not at all (exhibition hall).}
\label{fig:failure_cameras}
\vspace{-1.4em}
\end{figure*}

\noindent \textbf{3D points from FF depth.}
\Cref{tab:loop_variants} replaces the triangulation with 3D points lifted from the FF depth maps along the matches, either once before the first round or in every round.
Both are worse than triangulation, by 2 to 4 AUC@1$^\circ$ points when lifted once and by up to 14 points when lifted every round.
The predicted depth maps do not agree exactly with each other or with the refined cameras, so the lifted points inherit the errors of the forward pass.
Lifted once, these errors are only partly corrected by the later rounds; lifted every round, they are restored each time, and the cameras stay close to the FF prediction.
Triangulation instead places the points at the current cameras, so they improve together.
\emph{\method thus uses the FF predictions only for the initial cameras}, not their dense geometry.

\subsection{Limitations}
\label{sec:failure}

\noindent \textbf{Poor initialization.} 
\method needs only roughly correct initial cameras: it corrects imprecise cameras, but not outliers (\cref{fig:failure_cameras}).
On ETH3D's exhibition hall, VGGT-$\Omega$ starts below 4 AUC@1$^\circ$ with every camera degrees off, and \method brings it to 78.
Pi3X starts from a nearly random layout: \method recovers most of its cameras, but in this symmetric hall a fifth stay turned the wrong way, and the accuracy stays below 50 even with more rounds.
On Rotunda, GLOMAP places a few cameras far from the rest; \method pulls them in gradually and needs 20 rounds to match GGPT, which is imprecise everywhere but corrected after 5.
The large collections show the same limit: BA cannot recover cameras placed in the wrong part of the scene.
\method's tolerance is thus wide, but not unlimited.

\noindent \textbf{Poor chunk linking.} 
On large collections, FF networks usually predict each chunk accurately, but chunks can be misaligned.
Under perceptual aliasing, such as the repeated corridors of HGE, the shared cameras of two chunks agree on a wrong similarity, placing a whole chunk in the wrong part of the scene, beyond the refinement's reach.
GLUEMAP~\cite{pan2026gluemap} and DGSfM~\cite{aung2026dgsfm} filter such pairs with Doppelgangers++~\cite{doppelgangers++}, improving their robustness; a similar filter could be applied to chunk links.
Weakly connected parts, such as the eastern districts of the Aachen database, are lost or displaced by all methods (\cf \cref{fig:large_scenes_map} for visualizations).

\section{Conclusion}
We studied a simple refinement of initial cameras with local feature correspondences, repeated triangulation, and bundle adjustment with track updates, and analyzed which of its design choices matter for pose accuracy. 
\method significantly improves the cameras of a wide range of methods, from classical SfM and hybrid pipelines to feed-forward networks, across diverse settings, and brings them to nearly the same accuracy.
In particular, the noisy cameras of feed-forward networks are accurate enough to reach state-of-the-art results after refinement. 

\section*{Acknowledgements}
This work was supported by the Czech Science Foundation (GACR) JUNIOR STAR, Grant No.~22-23183M. The access to the computational infrastructure of the OP VVV funded project CZ.02.1.01/0.0/0.0/16\_019/0000765 ``Research Center for Informatics'' is also gratefully acknowledged. V.K. was funded by the EU NextGenerationEU through the Recovery and Resilience Plan for Slovakia under the project No. 09I02-03-V01-00029.

\begin{figure*}[p]
\centering
\includegraphics[width=\textwidth,height=0.92\textheight,keepaspectratio]{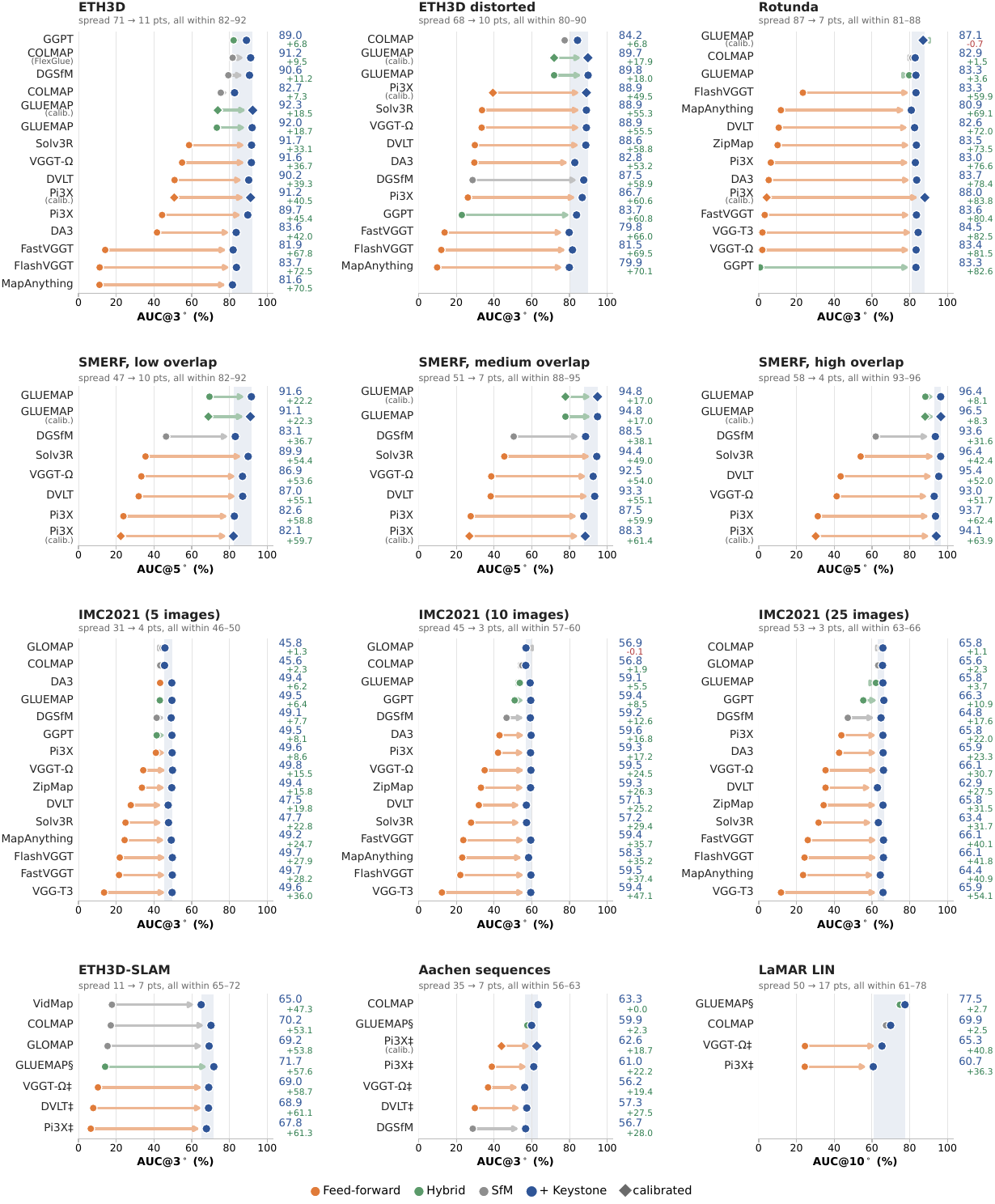}
\vspace{-0.4em}
\caption{\textbf{Keystone converges from any start.} Accuracy of every method before (coloured) and after Keystone (blue), sorted by the starting accuracy; failed methods (unregistered images) are omitted. The blue band is the range of the refined methods, the subtitle the spread before and after, and the right column the refined value and gain. $^\ddagger$: our chunking; $^\S$: the method's own chunking. Initializations that differ by 11 to 87 points end within a few points of each other on every dataset but LaMAR LIN.}
\label{fig:gain_all}
\end{figure*}

\clearpage

\clearpage
\maketitlesupplementary
This supplementary material first analyses how \method converges from different initializations (\cref{sec:convergence}) and presents further ablation studies (\cref{sec:more_ablation}). 
It then shows qualitative results (\cref{sec:qualitative}), and gives additional details on the experiments (\cref{sec:experiment_details}).

\section{Convergence Across Initializations}
\label{sec:convergence}
\Cref{fig:gain_all} shows how much the final accuracy depends on the initialization. 
On every dataset, initializations that differ by tens of points end within a few points of each other after \method (blue band), even on Rotunda, where some start close to zero. 
The ranking before refinement says little about the ranking after: methods that start far apart end side by side. 
The weakest feed-forward initializations gain the most and the SfM ones the least, since SfM already uses the same matches and bundle adjustment. 
The width of the band depends on the dataset, not on the initialization. 
On SMERF, it shrinks with increasing the overlap.
On IMC2021, all methods, SfM included, reach the same ceiling.
On the video sequences, the feed-forward initializations drift along the chain of chunks; the retrieval pairs connect distant frames and remove this drift. 
On the large LaMAR LIN collection, a chunk linked at the wrong place causes a large, consistent error that refinement cannot correct (see \cref{fig:large_scenes_map}). 
Apart from this, the final accuracy is set by \method rather than by the initialization, so the initialization can be chosen for speed and robustness (placing every camera roughly right) rather than precision.

\section{More Ablation Studies}
\label{sec:more_ablation}

\begin{figure}[t]
\centering
\includegraphics[width=\linewidth]{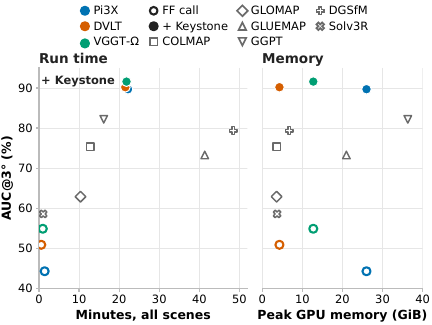}
\caption{\textbf{Accuracy against run time and memory on ETH3D.} AUC@3$^\circ$ against run time summed over all scenes (left) and peak GPU memory (right), every method on one NVIDIA H200 node with 16 CPU cores. Hollow circles: the feed-forward call alone; filled circles: the same network with \method; grey markers: prior work, each with its full pipeline.}
\label{fig:cost_eth3d}
\end{figure}

\begin{figure}[t]
\centering
\includegraphics[width=\columnwidth]{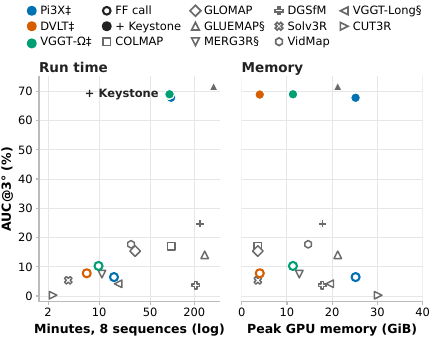}
\caption{\textbf{Accuracy against run time and memory on ETH3D SLAM.} AUC@3$^\circ$ against run time summed over eight sequences (left, logarithmic) and peak GPU memory (right), every method on one NVIDIA H200 node with 16 CPU cores. Hollow circles: the FF call on our chunks, including retrieval and alignment; filled circles: with \method; grey markers: prior work, each with its full pipeline. $^\ddagger$: our chunking. $^\S$: the method's own chunking.}
\label{fig:cost_eth3d_slam2}
\vspace{-1.0em}
\end{figure}

\noindent \textbf{Run time and memory.}
\Cref{fig:cost_eth3d,fig:cost_eth3d_slam2} relate the accuracy to the cost of each pipeline on ETH3D and ETH3D SLAM. 
On ETH3D, an FF call takes about a minute for all scenes, and \method adds about 20 minutes for matching and refinement (\cref{tab:runtime_refinement}). 
This is slower than COLMAP (13 min.) and GLOMAP (10 min.) but faster than GLUEMAP (41 min.) and DGSfM (49 min.), and 7--9 points AUC@3$^\circ$ more accurate than GGPT, the most accurate baseline. 
On the ETH3D SLAM sequences, FF networks with \method take about as long as COLMAP and less than half the time of DGSfM and GLUEMAP, mostly in global bundle adjustment (\cref{tab:runtime_refinement_slam2}), and are about four times as accurate as any baseline on its own. 
\method also improves GLUEMAP, which then becomes the most accurate pipeline, slightly ahead of the FF networks with \method but about four times slower. 
DGSfM with \method remains far behind, since DGSfM registers no images on five of the eight sequences, and \method can only refine cameras that the initialization provides.
The refinement runs on the CPU, so \method adds no GPU memory beyond that of the FF call and the matcher (4 GiB). 
On both datasets, Pi3X needs about twice the memory of VGGT-$\Omega$, the reverse of Aachen (\cref{fig:chunks}, right). 
The peak memory has two parts: the transformer, which grows with the number of images, and the heads that predict the dense outputs, which process images in batches. Pi3X runs its heads on 64 images at once and VGGT-$\Omega$ on 8, so on small scenes and chunks, the heads set the peak and Pi3X needs more memory. 
On hundreds of images the transformer dominates, and there Pi3X is lighter.

\begin{table}[t]
\centering
\setlength{\tabcolsep}{3pt}
\caption{\textbf{Run time of the steps of \method on ETH3D.} Minutes summed over all scenes; five refinement rounds each and share of the total time, from matching to the refined model, on one NVIDIA H200 node with 16 CPU cores; all steps after matching run on the CPU. $^\star$: computed once and shared by the three networks.}
\label{tab:runtime_refinement}
\resizebox{\linewidth}{!}{%
\begin{tabular}{l rr rr rr}
\toprule
& \multicolumn{2}{c}{Pi3X} & \multicolumn{2}{c}{DVLT} & \multicolumn{2}{c}{VGGT-$\Omega$} \\
\cmidrule(lr){2-3}\cmidrule(lr){4-5}\cmidrule(lr){6-7}
Step & min & share & min & share & min & share \\
\midrule
Matching (LoMa-B)$^\star$ & 3.61 & 17\% & 3.61 & 17\% & 3.61 & 17\% \\
Two-view verification & 1.94 & 9\% & 1.95 & 9\% & 1.95 & 9\% \\
Triangulation & 0.21 & 1\% & 0.20 & 1\% & 0.20 & 1\% \\
Inner loop & 15.02 & 72\% & 15.21 & 73\% & 15.13 & 72\% \\
\quad Global bundle adjustment & 14.24 & 68\% & 14.35 & 68\% & 14.28 & 68\% \\
\quad Track completion, merging, filtering & 0.40 & 2\% & 0.44 & 2\% & 0.45 & 2\% \\
\midrule
Total & \textbf{20.79} & 100\% & \textbf{20.97} & 100\% & \textbf{20.88} & 100\% \\
\bottomrule
\end{tabular}}
\vspace{-1.0em}
\end{table}

\begin{table}[t]
\centering
\setlength{\tabcolsep}{3pt}
\caption{\textbf{Run time of the steps of \method on ETH3D SLAM.} Minutes summed over eight of the sequences (2,584 images); five refinement rounds each and share of the total, from retrieval to the refined model, on one NVIDIA H200 node with 16 CPU cores; all steps after matching run on the CPU.}
\label{tab:runtime_refinement_slam2}
\resizebox{\linewidth}{!}{%
\begin{tabular}{l rr rr rr}
\toprule
& \multicolumn{2}{c}{Pi3X} & \multicolumn{2}{c}{DVLT} & \multicolumn{2}{c}{VGGT-$\Omega$} \\
\cmidrule(lr){2-3}\cmidrule(lr){4-5}\cmidrule(lr){6-7}
Step & min & share & min & share & min & share \\
\midrule
Retrieval (MegaLoc)$^\star$ & 1.69 & 2\% & 1.69 & 2\% & 1.69 & 2\% \\
Matching (LoMa-B)$^\star$ & 12.59 & 16\% & 12.59 & 16\% & 12.59 & 16\% \\
Two-view verification & 5.26 & 7\% & 5.25 & 7\% & 5.25 & 7\% \\
Triangulation & 1.55 & 2\% & 1.60 & 2\% & 1.58 & 2\% \\
Inner loop & 55.28 & 72\% & 57.69 & 73\% & 56.13 & 73\% \\
\quad Global bundle adjustment & 42.78 & 56\% & 50.61 & 64\% & 49.10 & 64\% \\
\quad Track completion, merging, filtering & 5.18 & 7\% & 5.34 & 7\% & 5.33 & 7\% \\
\midrule
Total & \textbf{76.37} & 100\% & \textbf{78.82} & 100\% & \textbf{77.24} & 100\% \\
\bottomrule
\end{tabular}}
\vspace{-1.2em}
\end{table}

\noindent \textbf{Run time of each step.}
\Cref{tab:runtime_refinement,tab:runtime_refinement_slam2} split the run time of \method into its steps on ETH3D and ETH3D SLAM.
On ETH3D, most of the time (about two thirds) goes to global bundle adjustment. 
Matching and verification take most of the rest, and the other steps take almost none.
ETH3D SLAM shows the same split, with three times as many images and about four times the run time. 
There, track completion, merging and filtering take a larger share (7\%), as the long sequences produce longer tracks.
On both datasets, the total is nearly the same for all three networks, since the cost depends mainly on the number of images and matches.
Retrieval and matching are computed once and shared by all initializations. 
We run bundle adjustment on the CPU, which keeps the GPU free for the forward passes and lets refinements run on CPU nodes in parallel; GPU-accelerated bundle adjustment could reduce the run time further.

\begin{table}[t]
\centering
\caption{\textbf{Ablation of spatial verification on ETH3D distorted.} AUC@1$^\circ$\,/\,@3$^\circ$ of \method. Matches of each pair are verified by the given model with RANSAC before triangulation; without verification every match enters the triangulation.}
\label{tab:verification}
\vspace{-0.3em}
\setlength{\tabcolsep}{2.8pt}
\renewcommand{\arraystretch}{0.9}
\resizebox{\linewidth}{!}{%
\begin{tabular}{l ccc}
\toprule
Variant & Pi3X & DVLT & VGGT-$\Omega$ \\
\midrule
\textbf{E/F/H with model selection (COLMAP)} & \textbf{73.1\,/\,86.7} & \textbf{75.1\,/\,88.6} & \textbf{75.3\,/\,88.9} \\
Essential matrix & 72.8\,/\,86.7 & 75.0\,/\,88.6 & 75.3\,/\,89.0 \\
Fundamental matrix & 73.3\,/\,87.0 & 75.1\,/\,88.7 & 75.3\,/\,88.9 \\
None (all matches) & 73.4\,/\,86.8 & 75.6\,/\,88.8 & 75.8\,/\,89.1 \\
\bottomrule
\end{tabular}}
\vspace{-0.6em}
\end{table}

\noindent \textbf{Spatial verification.}
\Cref{tab:verification} varies the two-view verification of the matches before triangulation on ETH3D distorted.
All variants lie within 0.5 points of each other on average.
Without verification, all matches enter the tracks, and triangulation uses only those that agree with the current cameras within $\tau_a$. 
The rest are not discarded: as the cameras improve over the rounds, track completion and merging can add them back.
Verification instead removes matches permanently before refinement, based on the two-view geometry alone. 
Since it does not depend on the initial cameras, it rejects wrong matches that happen to fit inaccurate cameras, but it also loses correct matches when the two-view geometry of a pair is poorly estimated.
We keep COLMAP's verification as the default to protect against inaccurate initial cameras.

\begin{table}[t]
\centering
\caption{\textbf{LoMa vs. FlexGlue on ETH3D.} COLMAP and GLOMAP built on LoMa-B~\cite{nordstrom2026loma} matches or on FlexGlue-S~\cite{lindenberger2026flexglue} with DaD~\cite{edstedt2025dad} features, alone and with \method started from their cameras and refined on either set of matches. Abs. recall at 5\,cm (R) and AUC@1$^\circ$ / @3$^\circ$; bold: the better refinement per row.}
\label{tab:matchers}
\setlength{\tabcolsep}{3pt}
\resizebox{\linewidth}{!}{%
\begin{tabular}{ll ccc ccc ccc}
\toprule
& & & & & \multicolumn{6}{c}{+ \method, refined on} \\
\cmidrule(lr){6-11}
& SfM & \multicolumn{3}{c}{SfM alone} & \multicolumn{3}{c}{LoMa-B} & \multicolumn{3}{c}{FlexGlue} \\
\cmidrule(lr){3-5}\cmidrule(lr){6-8}\cmidrule(lr){9-11}
Method & matches & R & @1$^\circ$ & @3$^\circ$ & R & @1$^\circ$ & @3$^\circ$ & R & @1$^\circ$ & @3$^\circ$ \\
\midrule
COLMAP & LoMa-B & 89.0 & 54.3 & 75.4 & 90.4 & \textbf{72.8} & 82.7 & \textbf{90.9} & \textbf{72.8} & \textbf{82.9} \\
 & FlexGlue & 94.6 & 58.1 & 81.7 & \textbf{98.6} & \textbf{80.1} & \textbf{91.2} & \textbf{98.6} & 79.1 & 90.8 \\
\midrule
GLOMAP & LoMa-B & 75.1 & 46.8 & 62.9 & 78.2 & 60.7 & 69.3 & \textbf{80.0} & \textbf{61.9} & \textbf{71.2} \\
 & FlexGlue & 97.5 & 61.0 & 84.5 & \textbf{99.7} & \textbf{82.5} & \textbf{93.5} & \textbf{99.7} & 81.3 & 93.0 \\
\bottomrule
\end{tabular}}
\vspace{-1.0em}
\end{table}

\noindent \textbf{Matcher for initialization and refinement.}
Unlike LoMa-B~\cite{nordstrom2026loma}, FlexGlue-S~\cite{lindenberger2026flexglue} disambiguates repeated structures during matching.
\Cref{tab:matchers} builds COLMAP and GLOMAP on either matcher and refines each with \method on either matcher.
On their own, both pipelines are clearly less accurate with LoMa-B, especially GLOMAP.
The largest failure is the exhibition hall, whose nearly identical repeated structures yield wrong LoMa-B matches that are geometrically consistent and pass verification, so both pipelines fold the hall onto itself while registering every image (see \cref{fig:distorted_map_b}).
\method cannot undo such a fold, as it moves cameras only locally, and refining on FlexGlue matches does not help: from the LoMa-B initializations, both matchers give nearly the same accuracy.
From the FlexGlue initializations, \method reaches over 98\% recall with either matcher, and LoMa-B is even slightly more precise.
In our experience, some feed-forward networks disambiguate repeated structures well: on exhibition hall, \method from DVLT, VGGT-$\Omega$ or Solv3R reaches the same accuracy as from the disambiguating SfM methods (DGSfM and GLUEMAP), while other networks, such as Pi3X, still collapse.
\emph{Disambiguation thus matters for building the initialization, not for refining it}: once the cameras are roughly correct, triangulation at the current cameras rejects the wrong matches.

\begin{figure}[t]
\centering
\includegraphics[width=0.95\linewidth]{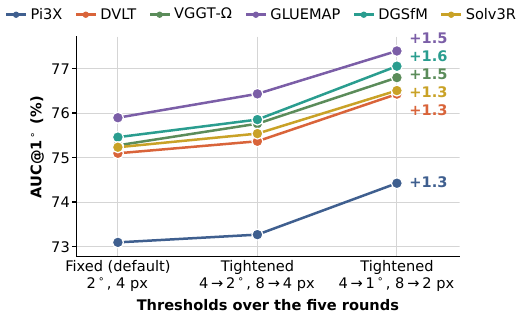}
\vspace{-0.2em}
\caption{\textbf{Threshold schedules on ETH3D distorted.} AUC@1$^\circ$ of \method after five rounds with fixed or tightening triangulation and filtering thresholds. Loose early rounds alone change little; tightening further gains up to 1.6 points for every initialization.}
\label{fig:thresholds}
\vspace{-0.4em}
\end{figure}

\noindent \textbf{Refinement thresholds.}
\Cref{fig:thresholds} varies the triangulation and filtering thresholds over the five rounds on ETH3D distorted.
Starting loose and tightening to the default ($4^\circ \to 2^\circ$, $8 \to 4$\,px) changes little, as the default is already loose enough to form tracks at roughly correct initial cameras.
Tightening further in the last rounds ($4^\circ \to 1^\circ$, $8 \to 2$\,px) gains 1.3 to 1.6 AUC@1$^\circ$ points for every initialization, since once the cameras are accurate, it removes slightly wrong observations that would otherwise bias BA.
We keep the fixed default for simplicity, as tight thresholds discard more observations and may hurt noisier initializations or sparser scenes; tightening over the rounds offers further precision where the initialization allows.

\begin{figure}[t]
\centering
\includegraphics[width=\linewidth]{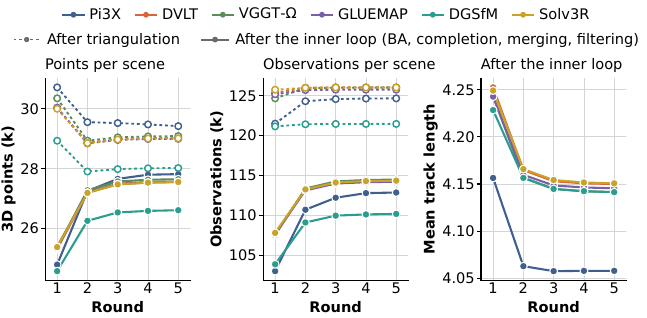}
\vspace{-1.4em}
\caption{\textbf{Tracks over the rounds on ETH3D distorted.} 3D points and observations per scene after the triangulation of each round (open markers) and after its inner loop (filled), and the mean track length after the loop, averaged over the scenes. The first round triangulates the most points but keeps only 81\% of them; later rounds triangulate at better cameras, keep 92--94\%, and the structure is flat after round 3.}
\label{fig:tracks}
\vspace{-1.2em}
\end{figure}

\noindent \textbf{Tracks over the rounds.}
\Cref{fig:tracks} follows the 3D points and observations through the refinement rounds on ETH3D distorted.
In the first round, triangulation at the initial cameras forms the most points, but the inner loop removes about a fifth of them, as many are triangulated from inconsistent observations or split from the same track.
Later rounds triangulate at the improved cameras: fewer points are formed, but more of them survive the inner loop, so both the points and the observations kept after the loop grow until round 3 and then stay flat.
Since the kept points grow slightly faster than the observations, the mean track length drops slightly; the added points are mostly short tracks, which the first round rejects at the noisy initial cameras.
\emph{The structure thus stabilizes within three rounds}, consistent with the small gains of further rounds in \cref{fig:refinement_loop}.

\begin{figure}[t]
\centering
\includegraphics[width=0.75\linewidth]{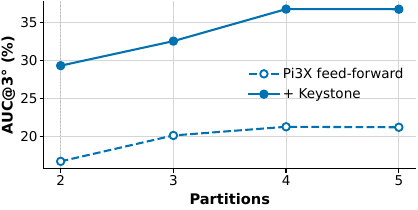}
\par
\vspace{-0.4em}
\caption{\textbf{Number of partitions on LaMAR LIN.} Default with two partitions; each further partition adds another retrieval grouping. AUC@3$^\circ$ of the aligned FF chunks (hollow) and after \method (filled). More partitions link more chunks, and improve the accuracy up to 4 partitions.}
\label{fig:session_partitions}
\vspace{-1.4em}
\end{figure}

\noindent \textbf{Number of partitions.}
\Cref{fig:session_partitions} varies the number of partitions on LaMAR LIN. 
Each further partition groups the images differently, so every chunk overlaps more chunks from the other partitions. 
The averaging then has more links to choose from, and a wrong link is more likely to be outvoted. 
Both the feed-forward and the refined cameras improve up to four partitions and then level off. 
However, each partition adds a feed-forward pass over all images, so the cost grows linearly with the number of partitions; we therefore keep two by default.


 

\section{Qualitative Results}
\label{sec:qualitative}
\Cref{fig:distorted_map,fig:distorted_map_b,fig:imc_map,fig:smerf_map,fig:eth3d_slam2_map,fig:aachen_seq_map,fig:large_scenes_map} show the cameras of each initialization before and after \method.
They share one pattern: cameras that are imprecise but roughly in place converge to nearly the same solution from any initialization, whereas grossly misplaced cameras remain wrong.
The datasets differ in which errors occur and in what limits the final accuracy.

\noindent \textbf{Images with lens distortion.}
COLMAP and GLUEMAP start close to the ground truth, whereas GGPT and the FF networks place many cameras noticeably off; all of them reach 92 to 96 AUC@3$^\circ$ from 6 to 87 before (\cref{fig:distorted_map}).
The errors that remain are gross ones (\cref{fig:distorted_map_b}): GLOMAP folds ``exhibition hall'' and places a few cameras of ``playground'' and ``Rotunda'' far away, Pi3X starts from a nearly random layout of ``exhibition hall'', and DGSfM collapses a group of ``Rotunda'' cameras onto one place, which \method spreads out only partly.

\noindent \textbf{Internet photos (IMC2021).}
Here, the converged cameras even share the same displacements from the ground truth for every initialization (\eg Milan Cathedral), at only 40 to 57 AUC@3$^\circ$ (\cref{fig:imc_map}).
This common error stems from the per-image focal lengths, which the few matches of a 25-image bag constrain poorly: with the ground-truth intrinsics, the displacements vanish, and Pi3X with \method reaches 79 to 85.
For the same reason, an already accurate initialization can lose slightly, as GGPT on London Bridge.

\noindent \textbf{SMERF collections.}
Even at low overlap, every initialization reaches above 90 AUC@5$^\circ$ after \method, within about 2 points of each other (\cref{fig:smerf_map}).
\method also corrects larger displacements, such as a whole group of cameras that DGSfM shifts on ``Alameda'', since the group keeps its relative layout and enough matches to the rest of the scene.
Weak overlap thus limits the initialization rather than the refinement. 
Once the cameras are roughly in place, the matches suffice to reach the same precise solution from any of them.

\noindent \textbf{Video sequences.}
The initializations mostly fail by drift: the FF networks and MERG3R bend or offset parts of the trajectory, DGSfM drifts at the ends of ``nexus6'' and ``nexus8'', and GLOMAP collapses a segment of ``mannequin face 1'' (\cref{fig:eth3d_slam2_map,fig:aachen_seq_map}).
Since the frames are matched not only within a temporal window but also with retrieval neighbors that link revisited places, the global BA corrects such drift over long distances, even for CUT3R, which drifts most.
It stops short only where the trajectory loses its shape: VidMap scatters the frames of ``planar 3'', MERG3R folds them, and CUT3R distorts the long loop of ``nexus1''.

\noindent \textbf{Large image collections.}
Here, the gross errors come from missing or misplaced parts of the scene (\cref{fig:large_scenes_map}).
COLMAP reconstructs only part of HGE and the center of Aachen, and GLUEMAP misplaces the eastern district of Aachen; \method refines the cameras they place, but cannot add or move these parts.
The FF networks reconstruct every chunk well but offset or rotate some relative to the others, duplicating streets on LIN and corridors on HGE.
\method pulls moderately offset chunks back into place, which substantially improves Pi3X and VGGT-$\Omega$ on LIN and Aachen, but chunks linked to the wrong part of the scene remain misplaced, such as the corridors that VGGT-$\Omega$ folds into the center of HGE.

\emph{Across all settings, the accuracy after \method is thus bounded by gross errors of the initialization and, for internet photos, by the intrinsics, not by its imprecision.}

\section{Experimental Details}
\label{sec:experiment_details}

\subsection{Dataset details}

\noindent \textbf{ETH3D.}
We use the 25 high-resolution multi-view scenes~\cite{eth3dsfm} with their undistorted DSLR images, different from DGSfM, that uses only 22 scenes.
Each image is resized to a longest side of 1{,}600 pixels and center-cropped to a multiple of 16 pixels, and the ground-truth intrinsics are adjusted accordingly, following~\cite{aung2026dgsfm, pixsfm}.
All images of a scene come from one camera.

\noindent \textbf{ETH3D distorted.}
We take the original distorted DSLR images of the same scenes and resize and crop them in the same way, to $1{,}600 \times 1{,}056$ pixels.
ETH3D calibrates them with a thin-prism fisheye model, where as we use the radial model in \method.

\noindent \textbf{ETH3D SLAM.}
We use the 55 sequences of VidMap~\cite{pataki2026vidmap}, which are the training split of ETH3D SLAM~\cite{schops2019bad} without its six dark sequences.
To keep the sequences tractable, we keep a frame whenever its ground-truth camera has moved at least 5\,cm or turned at least $5^\circ$ since the last kept frame, which leaves 13{,}831 of the 45{,}872 frames with ground truth.
Every kept frame is evaluated, and the images are not brightened.
Each sequence has one camera.

\noindent \textbf{IMC2021.}
We use the bags of 5, 10, and 25 images with 100 of each per scene, sampled from the 100-image sets of nine phototourism scenes~\cite{imc2021}.
Every image is treated as taken with its own camera.

\noindent \textbf{SMERF.}
We use the image collections of MP-SfM~\cite{mpsfm} on the four ZipNeRF~\cite{barron2023zipnerf} scenes Alameda, Berlin, London, and NYC, with five each of low, medium, and high overlap per scene.
The images are ZipNeRF's undistorted images at half resolution, $2{,}000 \times 1{,}140$ pixels for Berlin and about $1{,}393 \times 793$ pixels for the other three scenes.
Each collection is reconstructed independently and evaluated against its subset of the scene ground truth.

\noindent \textbf{Rotunda.}
The ground truth of the Rotunda scene of RD-Bench~\cite{kocur2026minimal} is reconstructed by COLMAP from LoMa matches on all 157 images from phones and a GoPro, with a radial fisheye camera model.
We evaluate only on the 62 GoPro images, captured in two modes at $3{,}000 \times 2{,}250$ and $4{,}000 \times 3{,}000$ pixels, at half resolution.
Their ground-truth cameras and poses are taken from the full reconstruction, with intrinsics adjusted accordingly.

\noindent \textbf{Aachen.}
The Aachen v1.1 dataset~\cite{Zhang2020ARXIV, Sattler2012BMVC, Sattler2018CVPR} holds 6{,}697 database and sequence images.
We undistort all images to a pinhole model and thus, 4{,}324 database photos form the large collection.
The remaining nine sequences (gopro3 and nexus4 sequences 1--8, 2{,}369 frames in total) are evaluated separately, using one shared camera per sequence.

\noindent \textbf{LaMAR.}
We follow the preparation of GLUEMAP~\cite{pan2026gluemap} and use all iOS frames of the map sessions of HGE and LIN, with ground-truth poses and intrinsics exported by the official LaMAR code~\cite{sarlin2022lamar}.

\subsection{Baseline Details}

\noindent \textbf{Common protocol.}
We run every baseline with its official code and default settings unless stated otherwise, on the original images of each processed dataset.
Unregistered images and failed scenes count as failures in all metrics.
To refine a baseline with \method, we convert their predictions into initial cameras, with the intrinsics of its calibration in original pixels.
Images it leaves unregistered remain unregistered.
Focal lengths below $0.1$ or above $10$ times the image diagonal are replaced by the median of the scene.

\noindent \textbf{COLMAP \& GLOMAP.}
Both use the same LoMa-B matches and pairs with the default settings unless otherwise mentioned specifically.
We use the camera models of \method and share one camera per scene on ETH3D (undistorted and distorted), ETH3D SLAM, and the Aachen sequences, and one camera per image elsewhere.

\noindent \textbf{DGSfM.}
We use MoGe-2~\cite{wang2025moge2} depth, Doppelgangers++~\cite{doppelgangers++} pair filtering with a threshold of $0.8$ with LoMa-B matches as well.
We use a simple pinhole camera without distortion, shared per scene except on IMC2021 and Rotunda.

\noindent \textbf{VidMap.}
We run its uncalibrated base configuration, with intrinsics predicted by Depth Anything 3~\cite{lin2025depthanything3}, on the video sequences.
VidMap estimates poses only for keyframes, so we interpolate the other frames between their neighboring keyframes in frame order and evaluate them.

\noindent \textbf{GGPT.}
We use only its SfM stage up to bundle adjustment, with VGGT point maps and RoMa~v2~\cite{romav2} dense matches on all pairs, and a shared camera except on IMC2021 and Rotunda.
GGPT is not run on the video sequences and large collections, which exceed its memory.

\noindent \textbf{GLUEMAP.}
We use its benchmark configuration: SALAD~\cite{izquierdo2024optimal} retrieval with 100 neighbors (30 sequential neighbors for videos), Doppelgangers++~\cite{doppelgangers++} filtering with a threshold of $0.8$, Pi3X star predictions with 25 neighbors, global mapping, and its final augmented BA.
Intrinsics are shared on the same datasets as for COLMAP and on SMERF, and estimated per camera otherwise.
Videos and LaMAR use its sequential pairing.
For the calibrated variant, we keep the ground-truth intrinsics, including their distortion, fixed throughout.

\noindent \textbf{MERG3R.}
We run its video mode with Pi3X, subsets of 100 images with an overlap of 5, weighted iterative alignment, and global bundle adjustment.

\noindent \textbf{Solv3R.}
Unordered collections use its offline mode with 16 calls.
Videos and large collections use its default streaming setting, with 8 anchor frames, 6 recent frames, a memory of 64 frames sampled every 16th frame, and 8 calls per new frame.

\noindent \textbf{CUT3R.}
CUT3R processes the frames one at a time in capture order with its recurrent state, in a single pass without revisiting earlier frames, we use the online mode as in its relative-pose evaluation.

\noindent \textbf{VGGT-Long.}
We use chunks of 60 frames with an overlap of 30 and SALAD loop closure, at VGGT~\cite{vggt}'s input width of 518 pixels.


\ifdefined\distW\else\newlength{\distW}\fi\setlength{\distW}{1.08in}
\begin{figure*}[p]
\def\mappanel#1{{\setlength{\fboxsep}{0pt}\setlength{\fboxrule}{0.3pt}\fcolorbox{black!20}{white}{\includegraphics[width=\dimexpr\distW-0.6pt\relax]{fig/distorted_map/#1}}}}
\def\mapauc#1#2{\mappanel{#1}\llap{\raisebox{1.2pt}{\setlength{\fboxsep}{0.8pt}\colorbox{white}{\tiny AUC@3$^\circ$ #2}}\hspace{1.2pt}}}
\def\mapaucbf#1#2{\mappanel{#1}\llap{\raisebox{1.2pt}{\setlength{\fboxsep}{0.8pt}\colorbox{white}{\tiny\bfseries AUC@3$^\circ$ #2}}\hspace{1.2pt}}}
\def\maprow#1{\rotatebox{90}{\scriptsize #1}}
\def\mapname#1#2{\raisebox{0pt}[0pt][0pt]{\parbox{\distW}{\centering\includegraphics[width=0.8\distW]{fig/distorted_map/#1}\\[1pt]\footnotesize\bfseries #2}}}  
\centering
\setlength{\tabcolsep}{0.8pt}
\renewcommand{\arraystretch}{0.7}
\begin{tabular}{@{}>{\centering\arraybackslash}m{0.9em} >{\centering\arraybackslash}m{\distW}@{\hspace{4pt}} *{5}{>{\centering\arraybackslash}m{\distW}} @{}}
 & \small\bfseries GT & \small\bfseries COLMAP & \small\bfseries GGPT & \small\bfseries GLUEMAP & \small\bfseries Pi3X & \small\bfseries DVLT \\[2pt]

\maprow{Initialization} & \mappanel{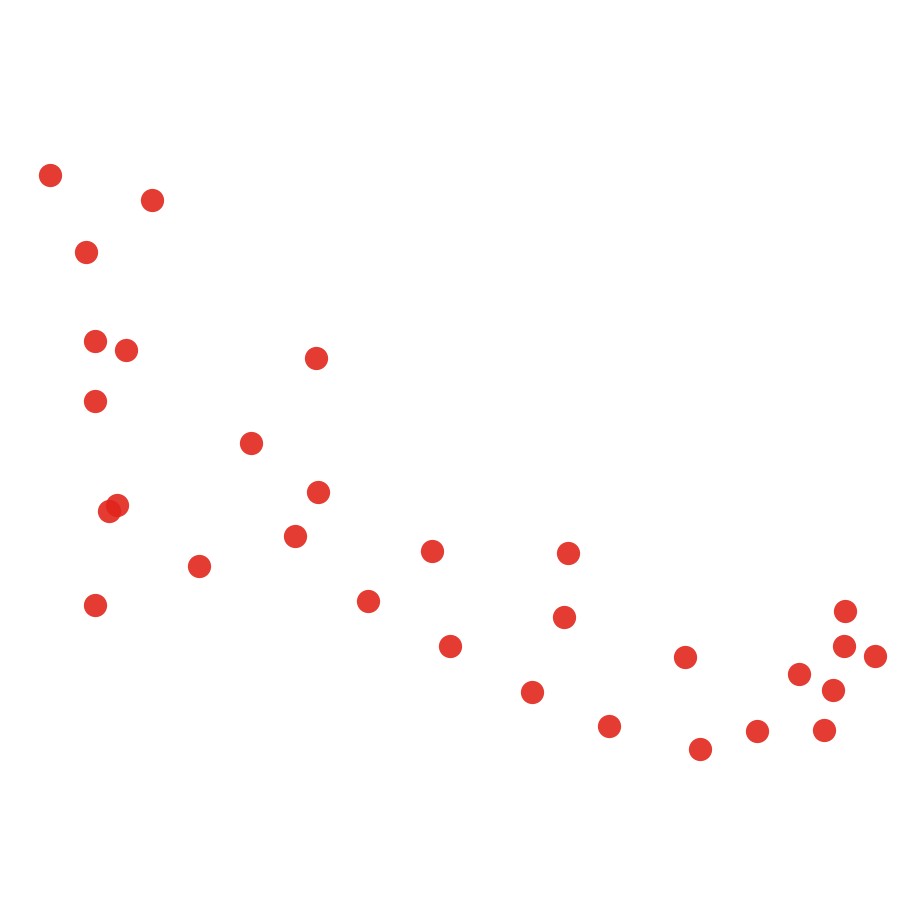} & \mapauc{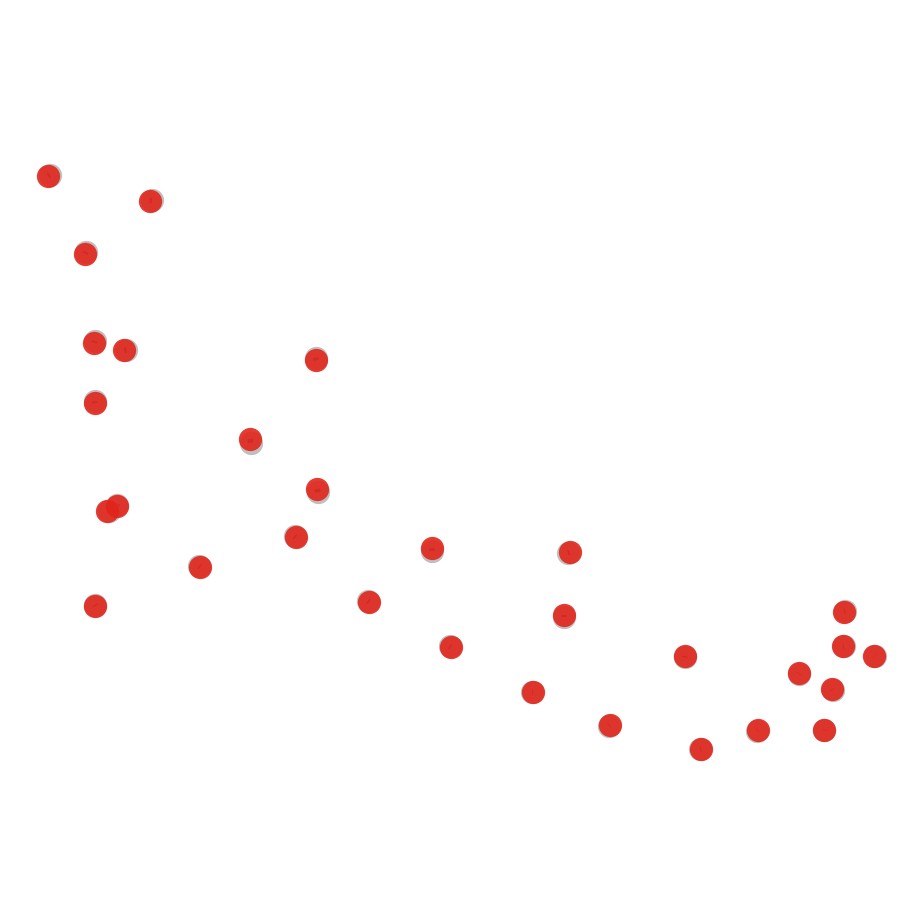}{86.7} & \mapauc{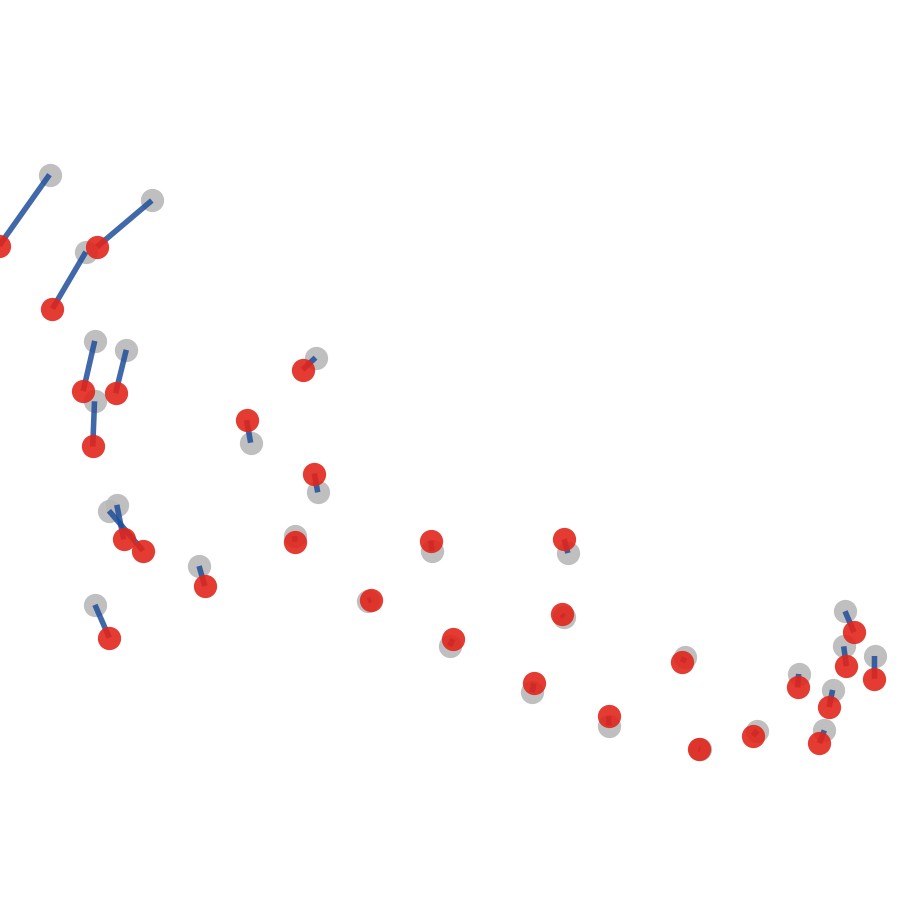}{7.1} & \mapauc{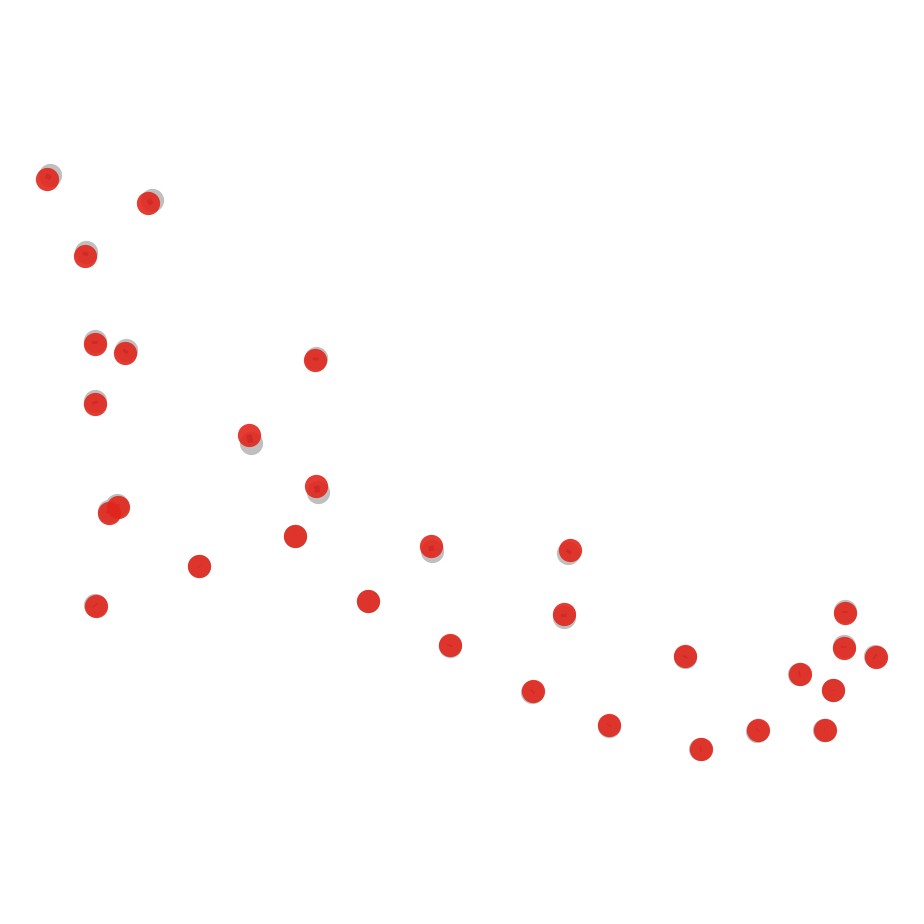}{74.5} & \mapauc{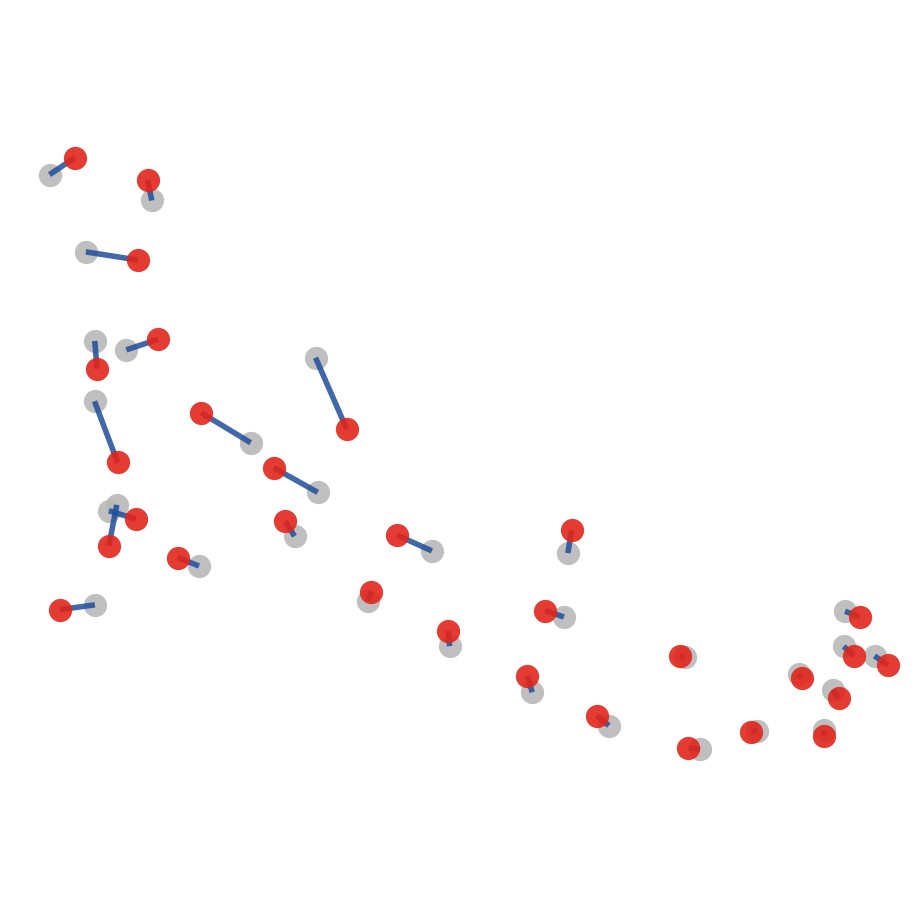}{8.3} & \mapauc{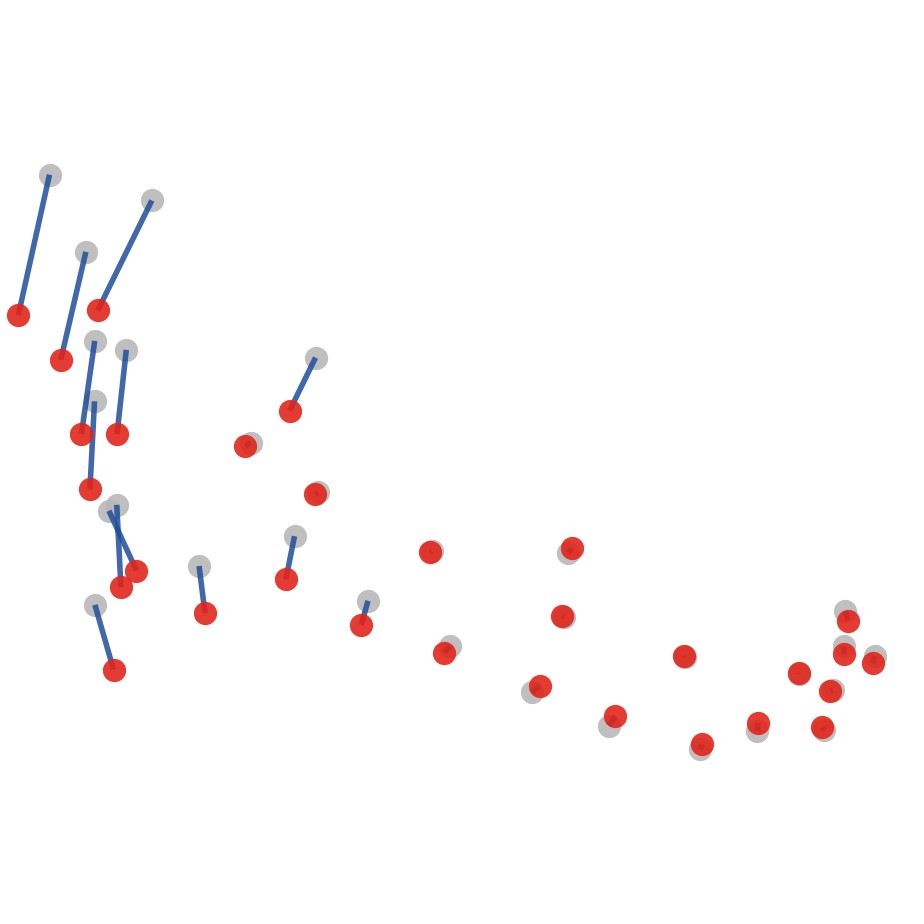}{5.7} \\[0pt]
\maprow{\textbf{+ \method}} & \mapname{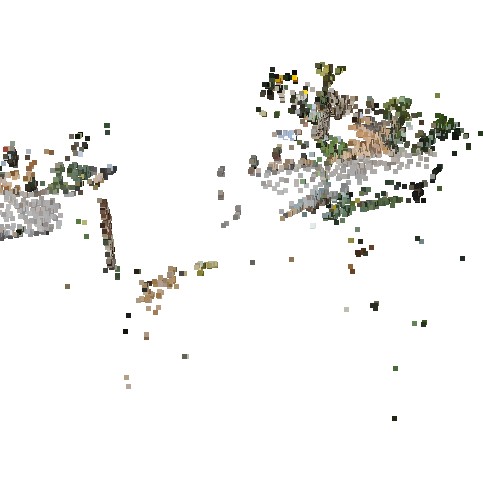}{botanical garden} & \mapaucbf{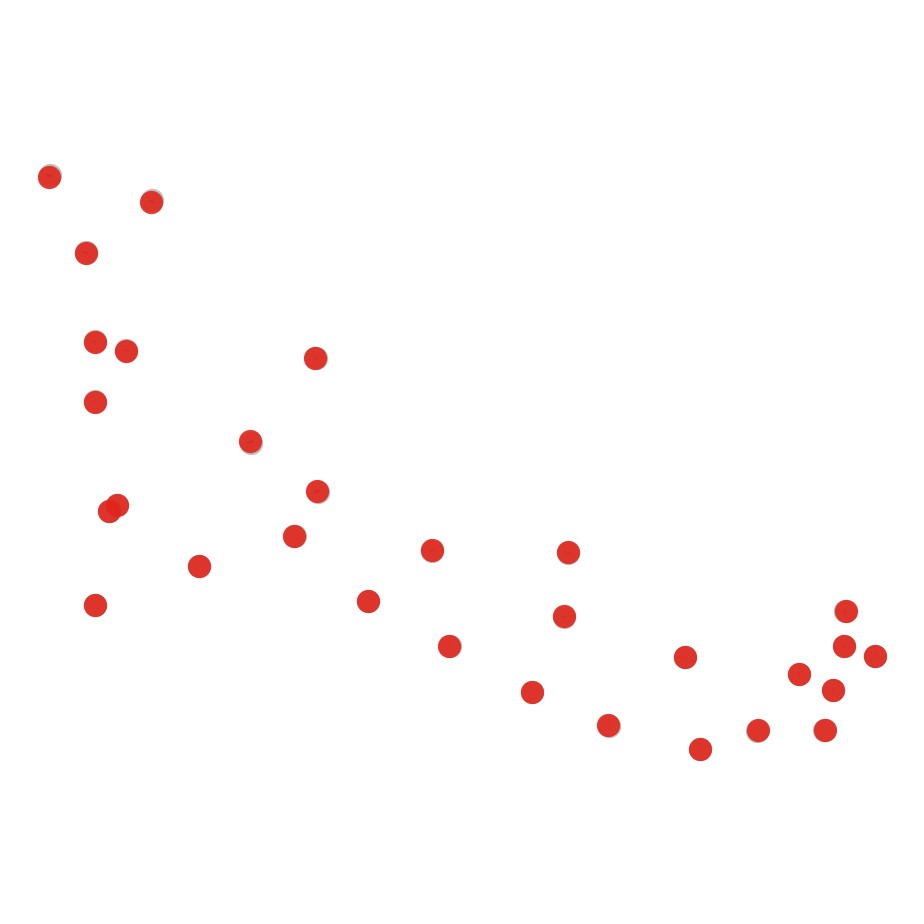}{93.0} & \mapaucbf{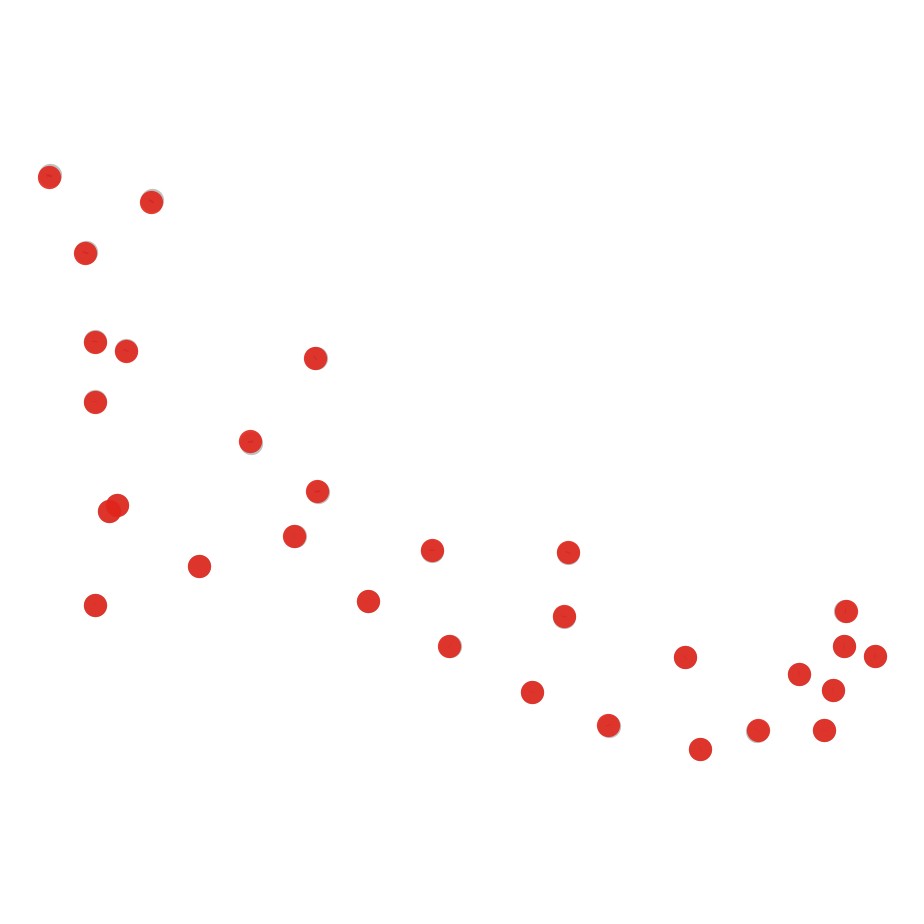}{92.5} & \mapaucbf{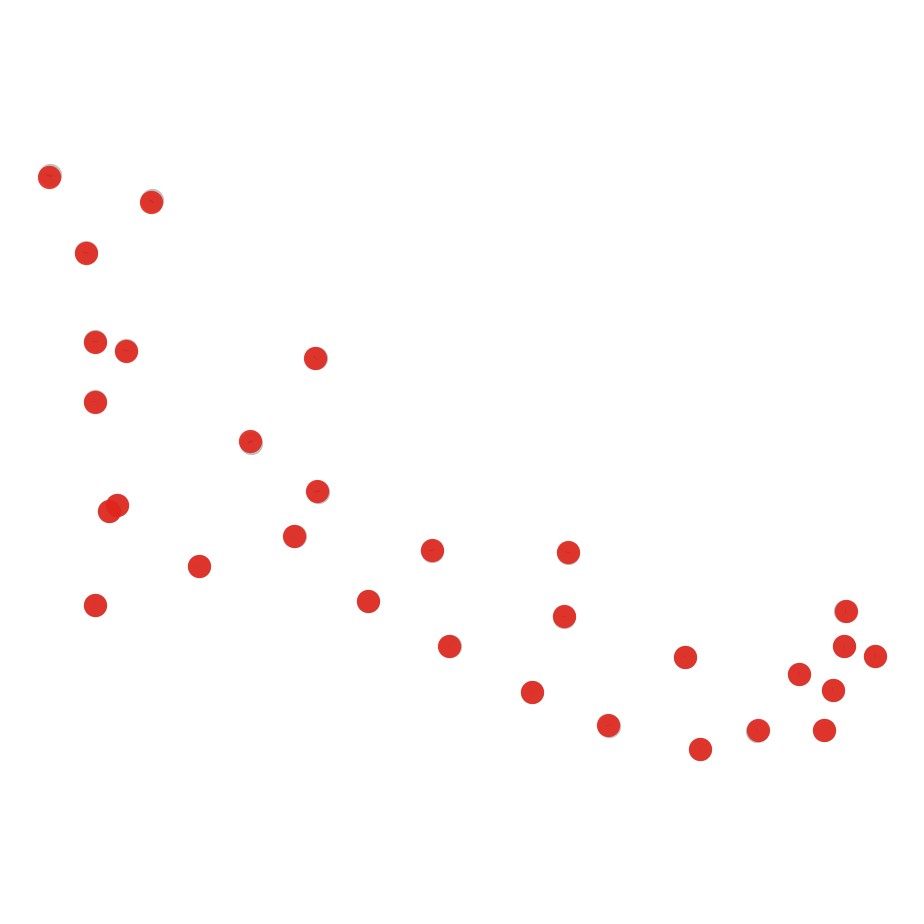}{92.9} & \mapaucbf{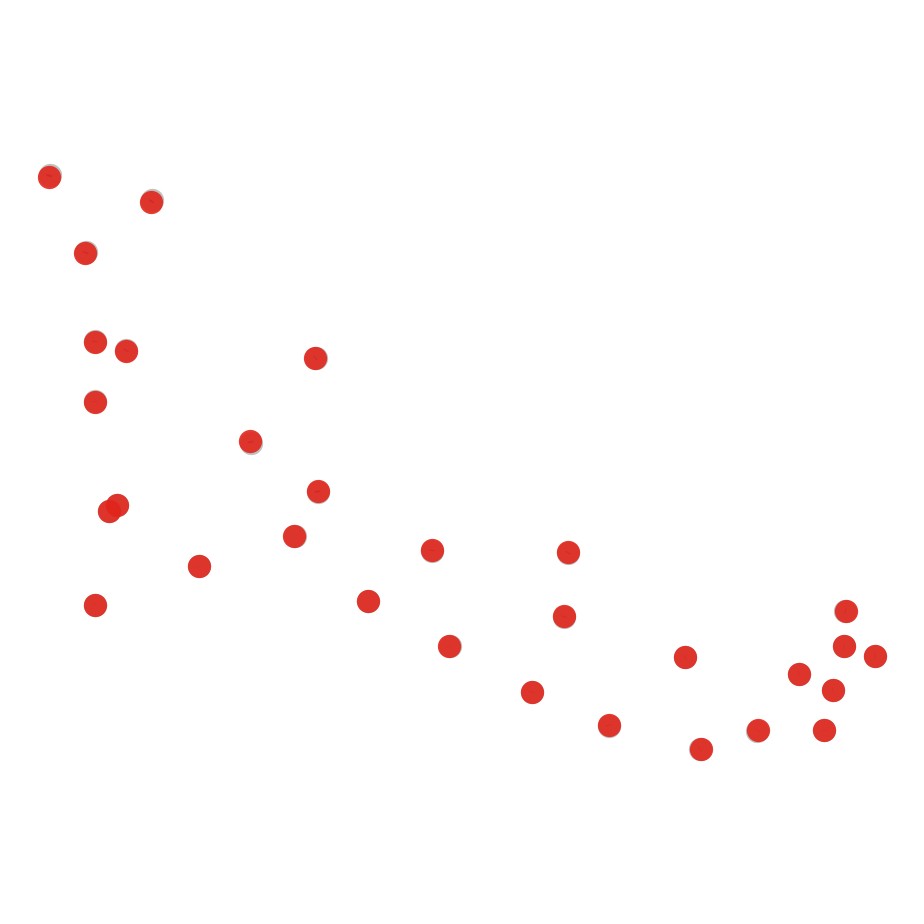}{93.4} & \mapaucbf{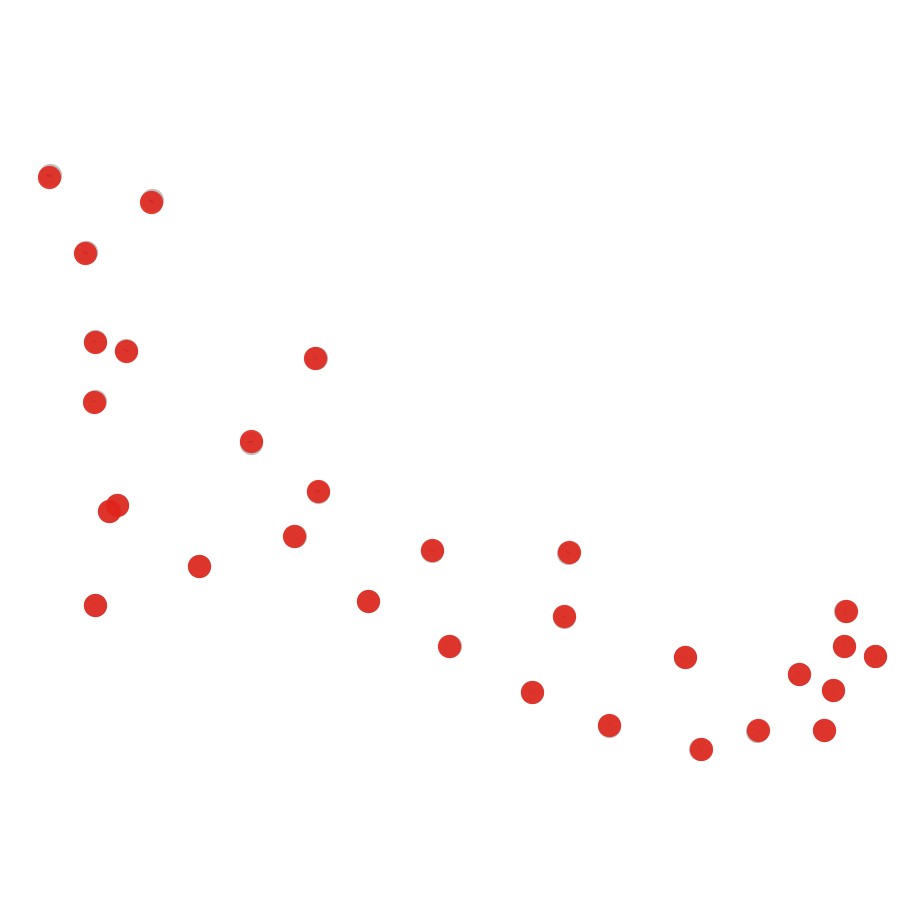}{92.4} \\[4pt]
\maprow{Initialization} & \mappanel{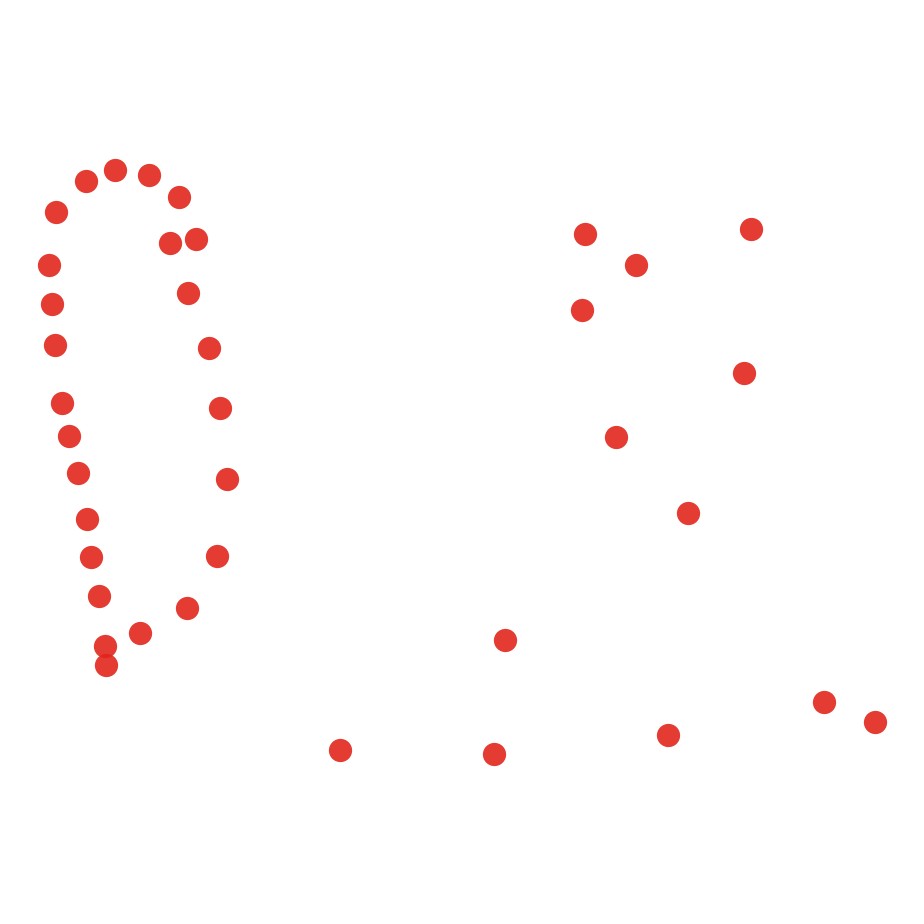} & \mapauc{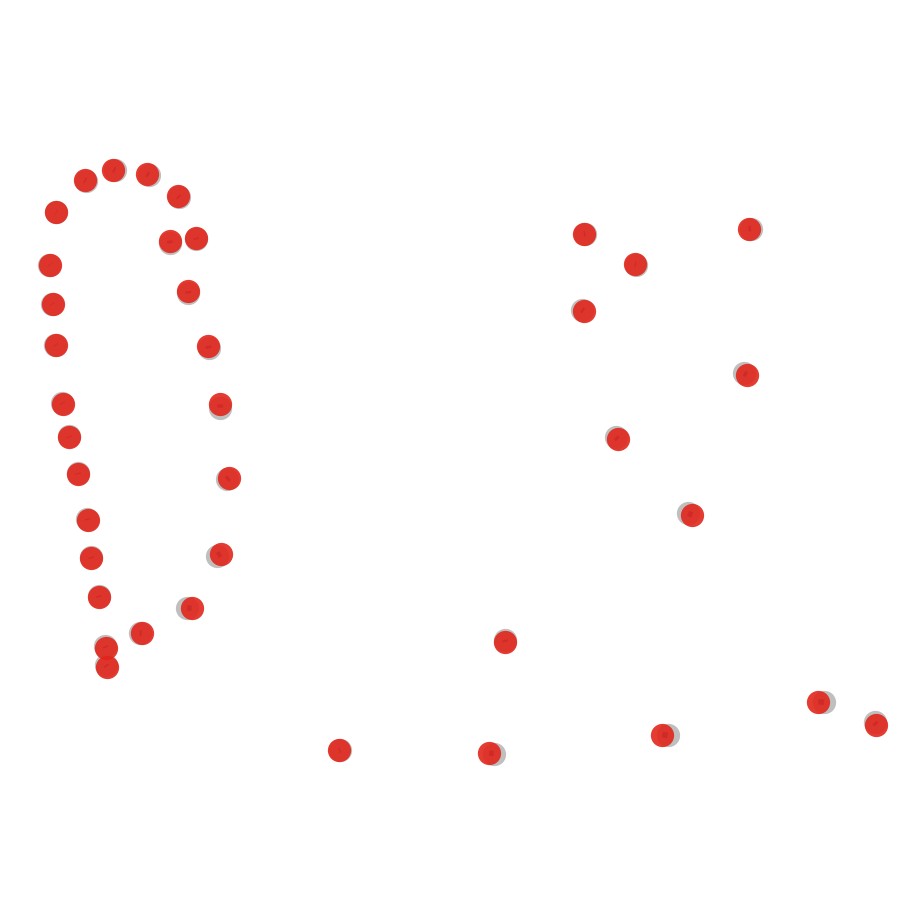}{84.9} & \mapauc{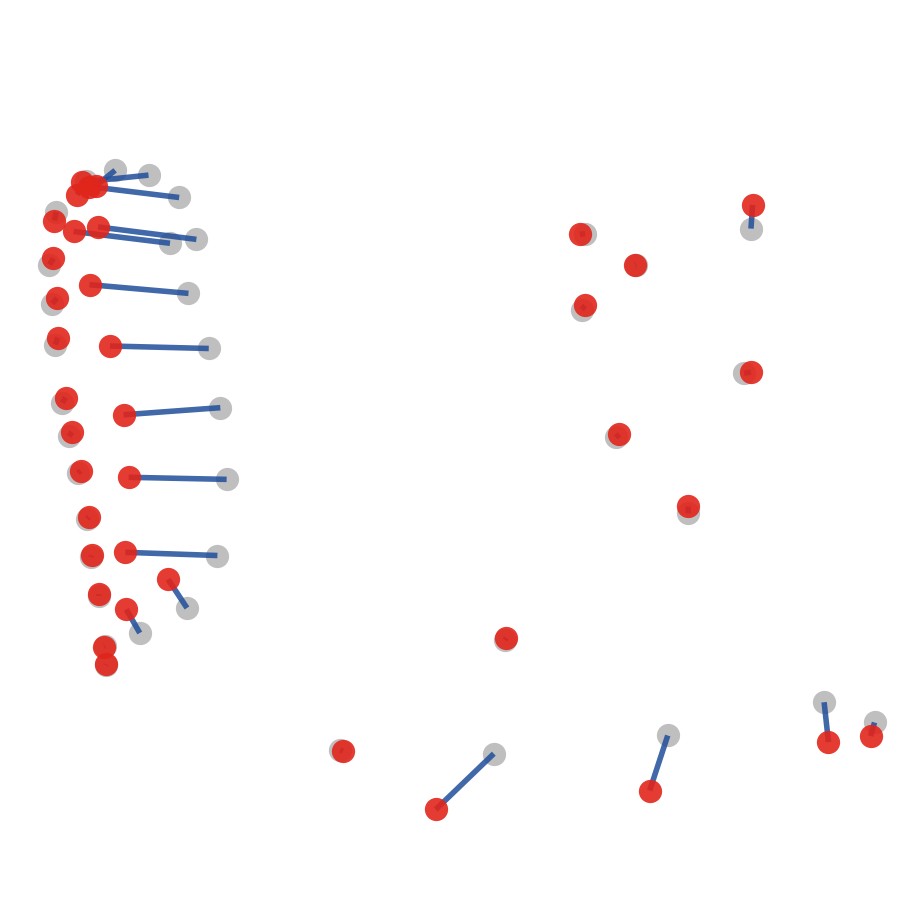}{11.0} & \mapauc{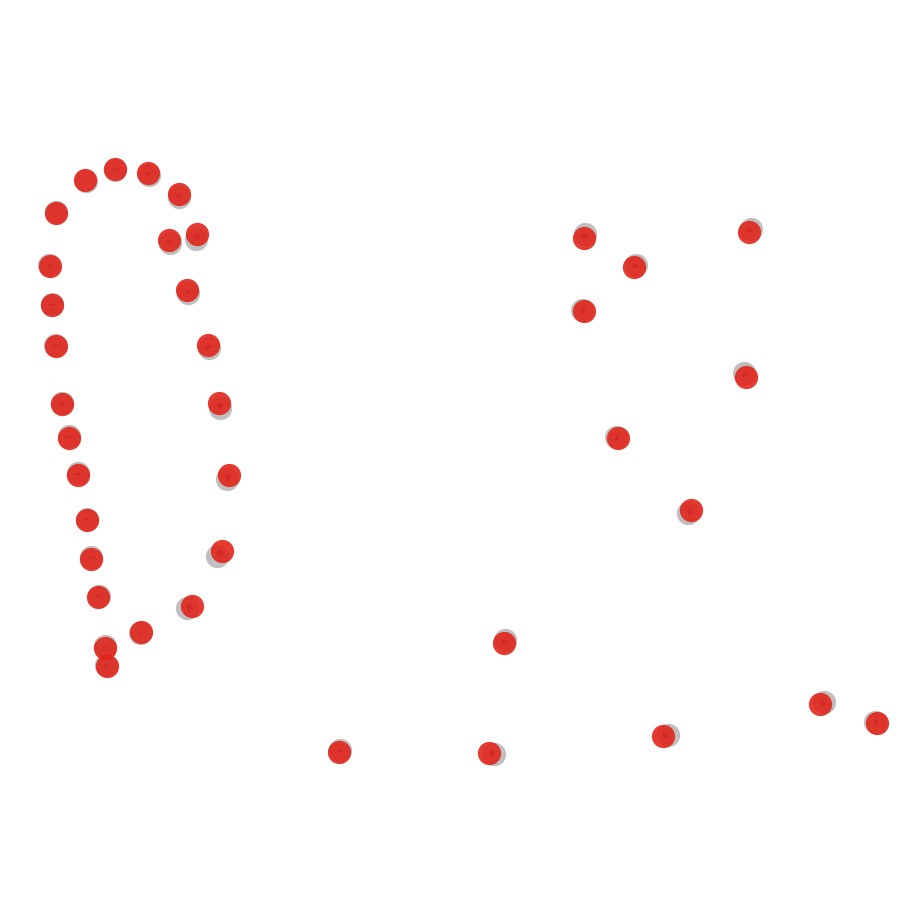}{76.4} & \mapauc{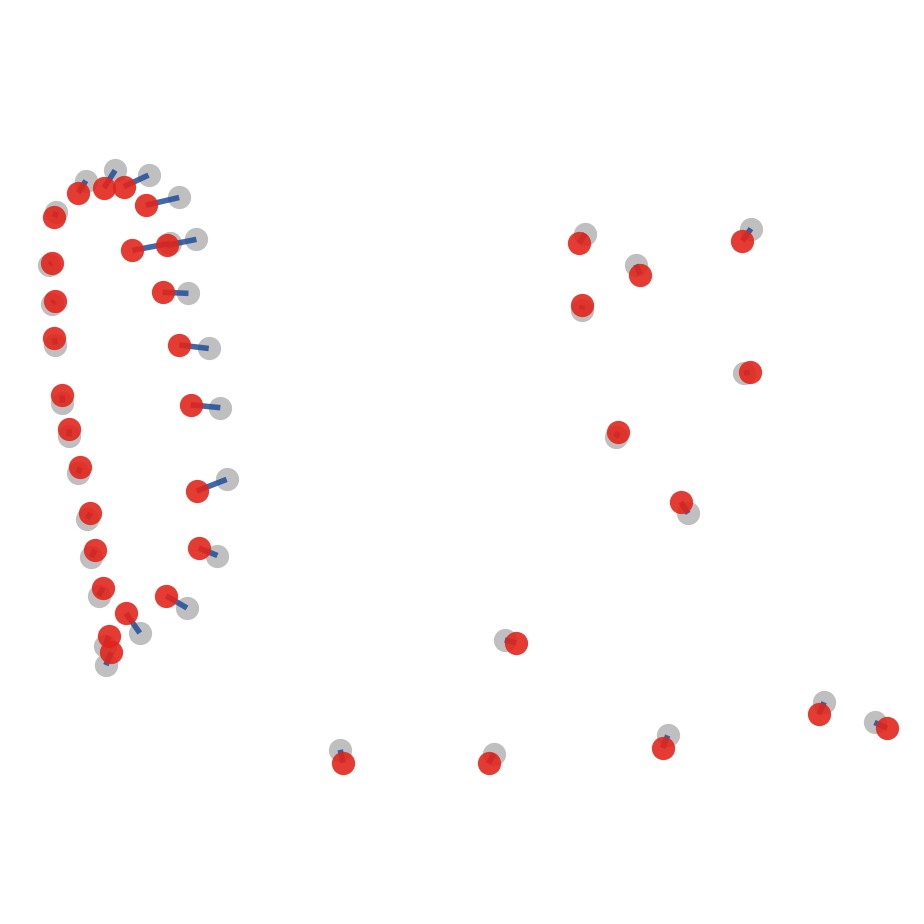}{30.2} & \mapauc{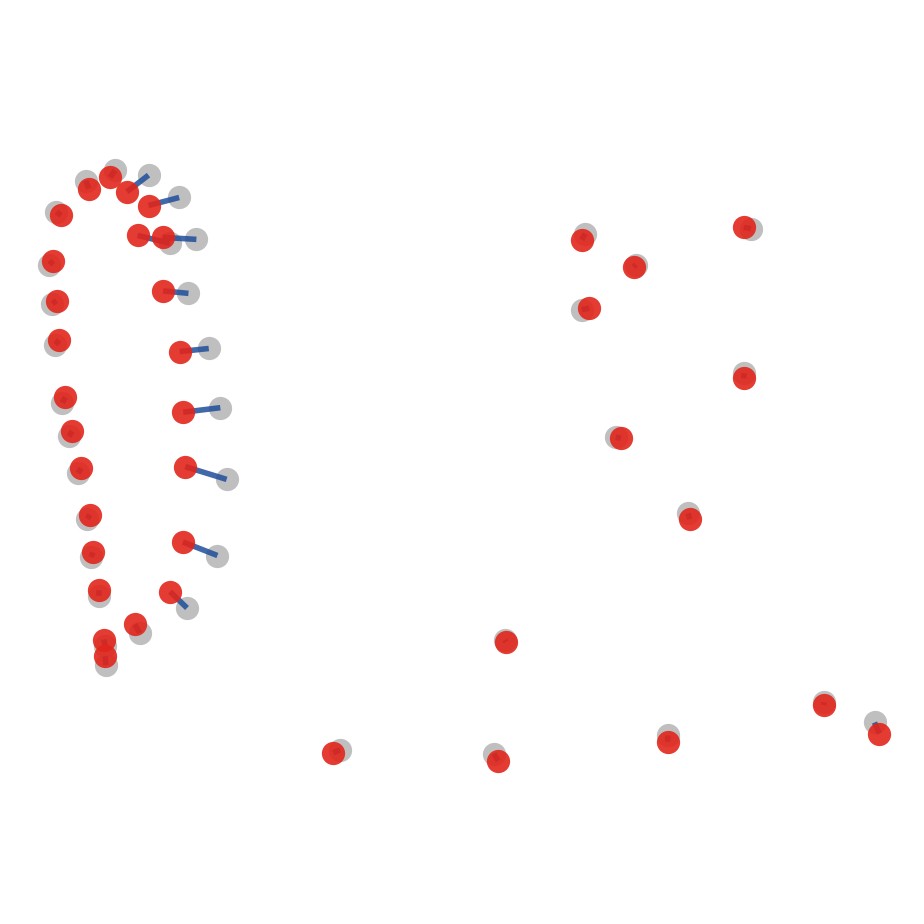}{25.6} \\[0pt]
\maprow{\textbf{+ \method}} & \mapname{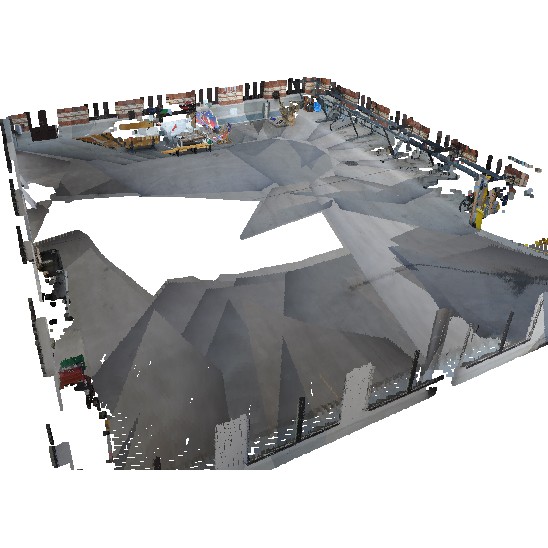}{courtyard} & \mapaucbf{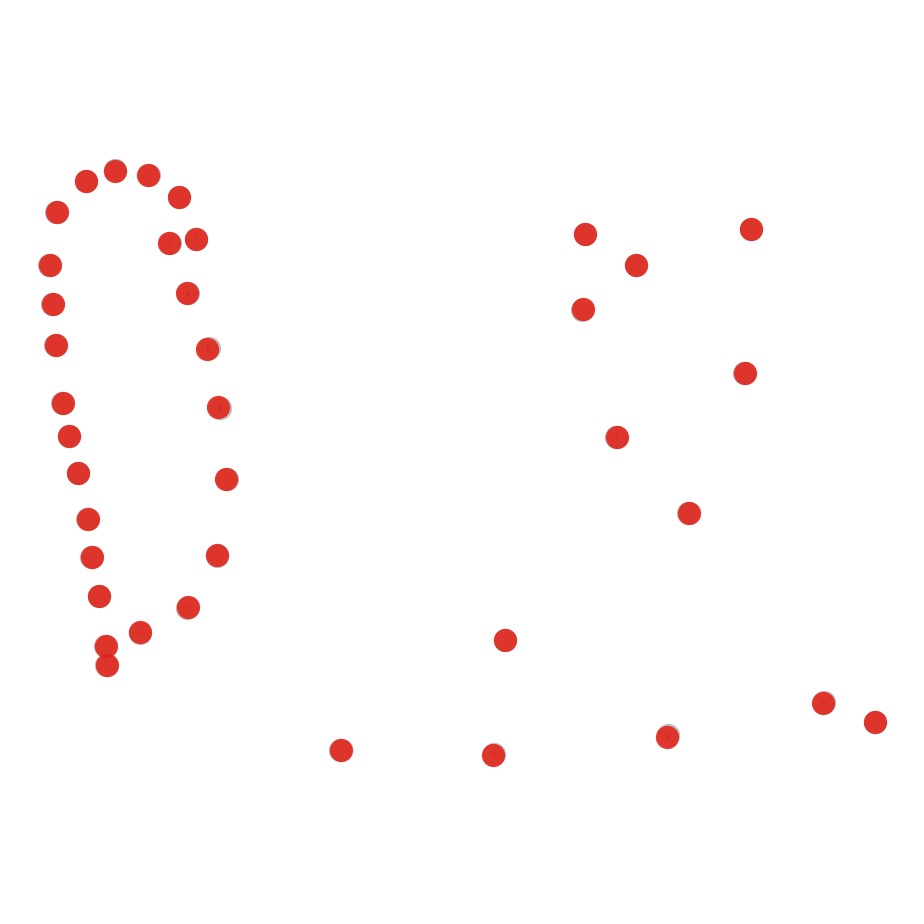}{93.0} & \mapaucbf{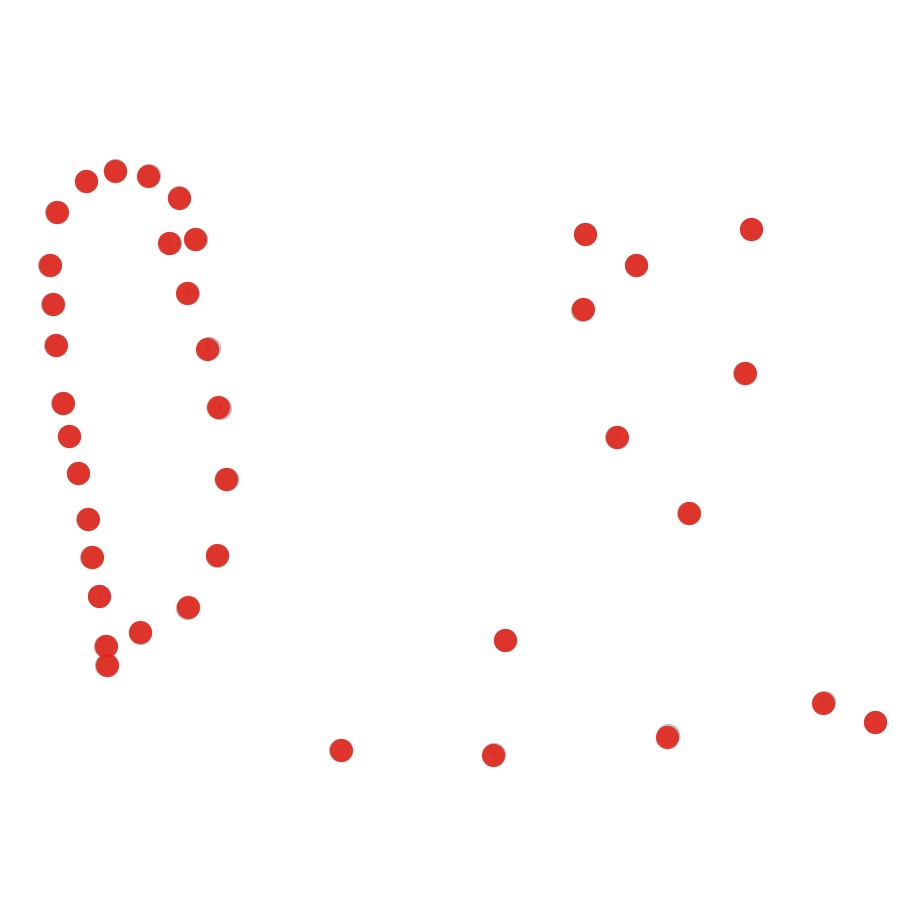}{92.9} & \mapaucbf{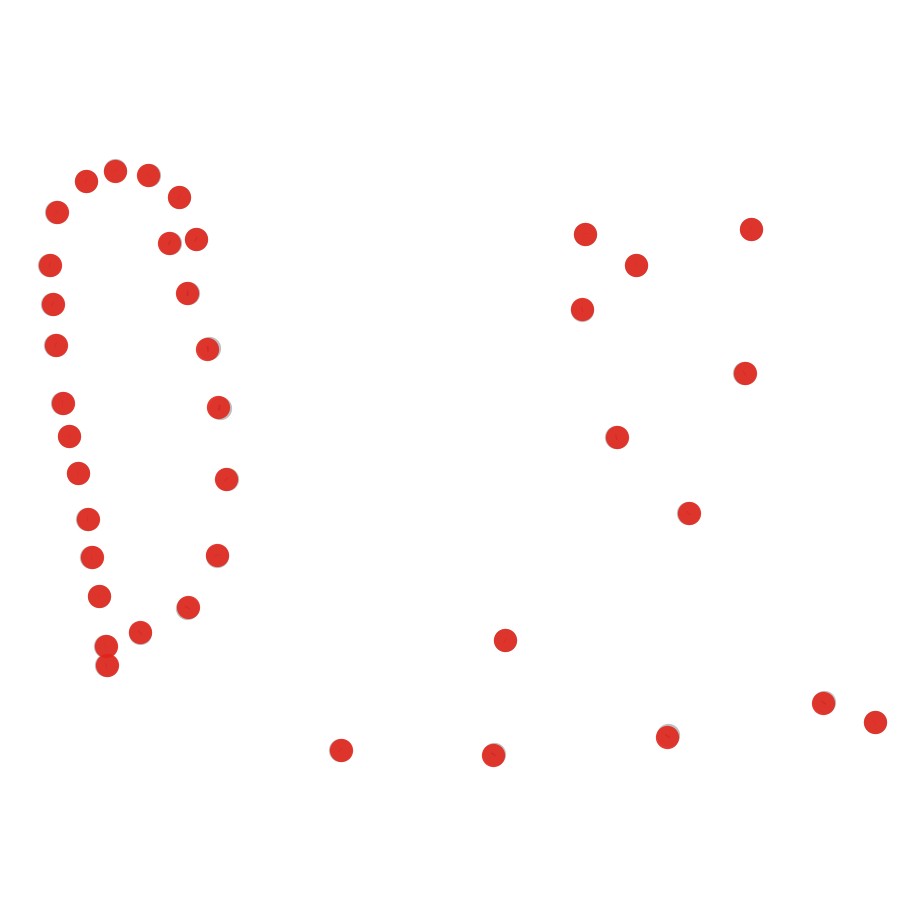}{93.2} & \mapaucbf{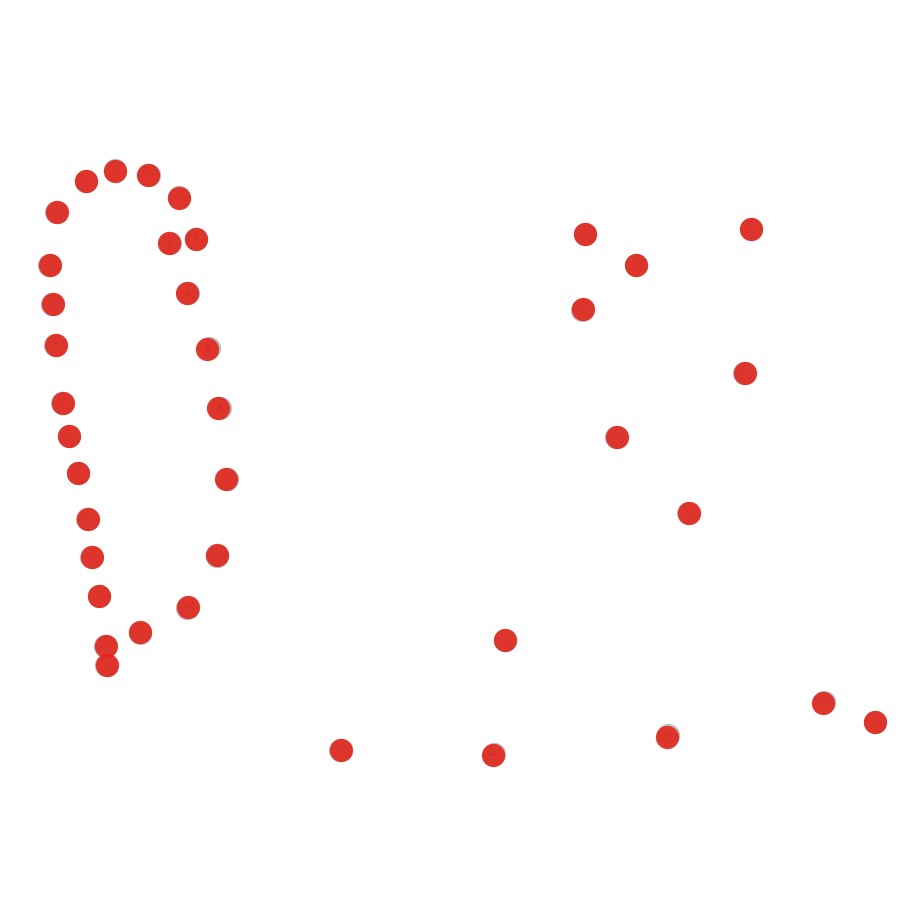}{93.3} & \mapaucbf{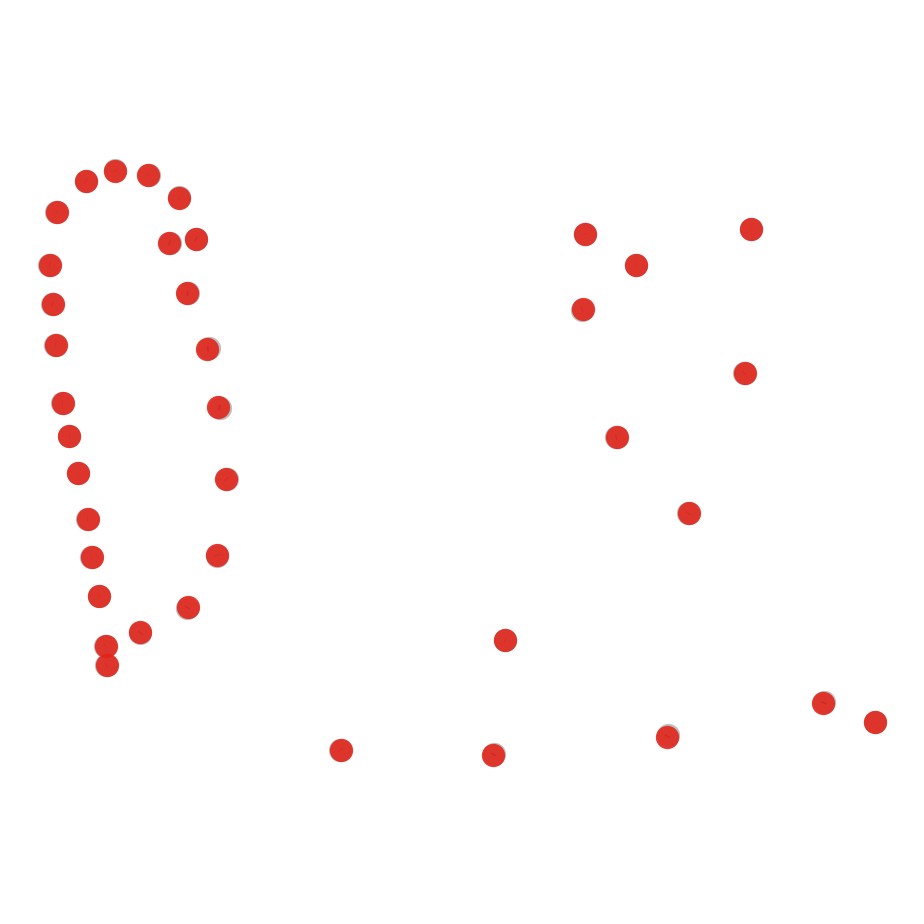}{93.0} \\[4pt]
\maprow{Initialization} & \mappanel{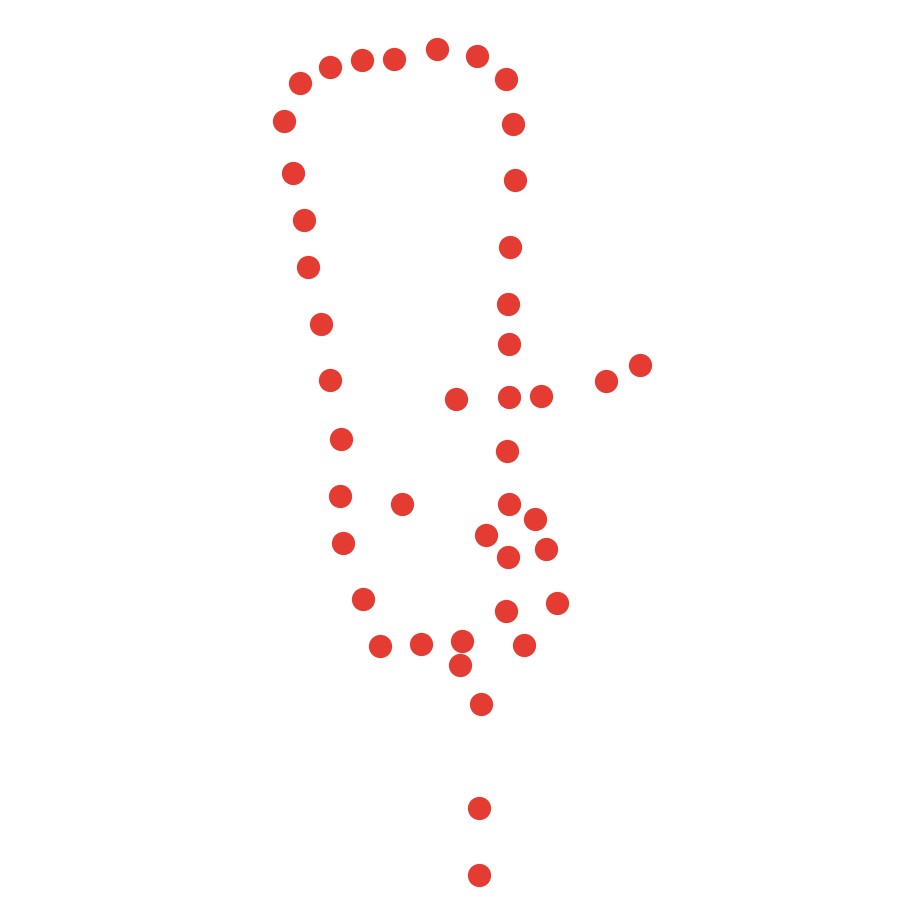} & \mapauc{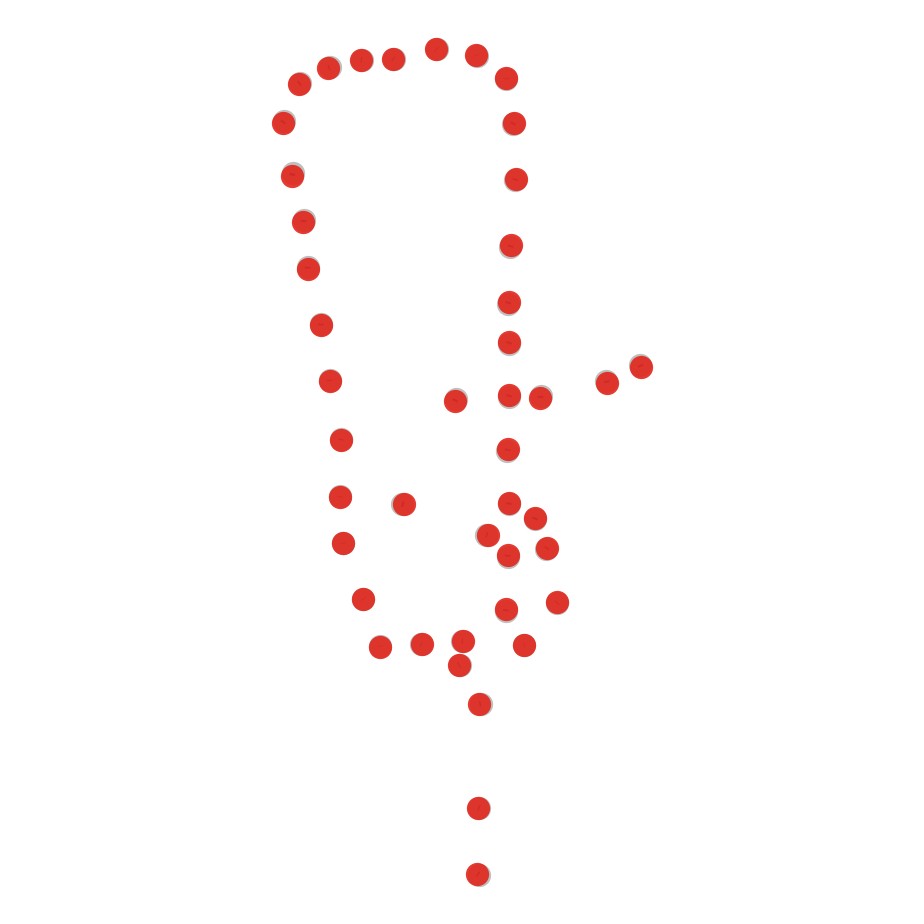}{87.1} & \mapauc{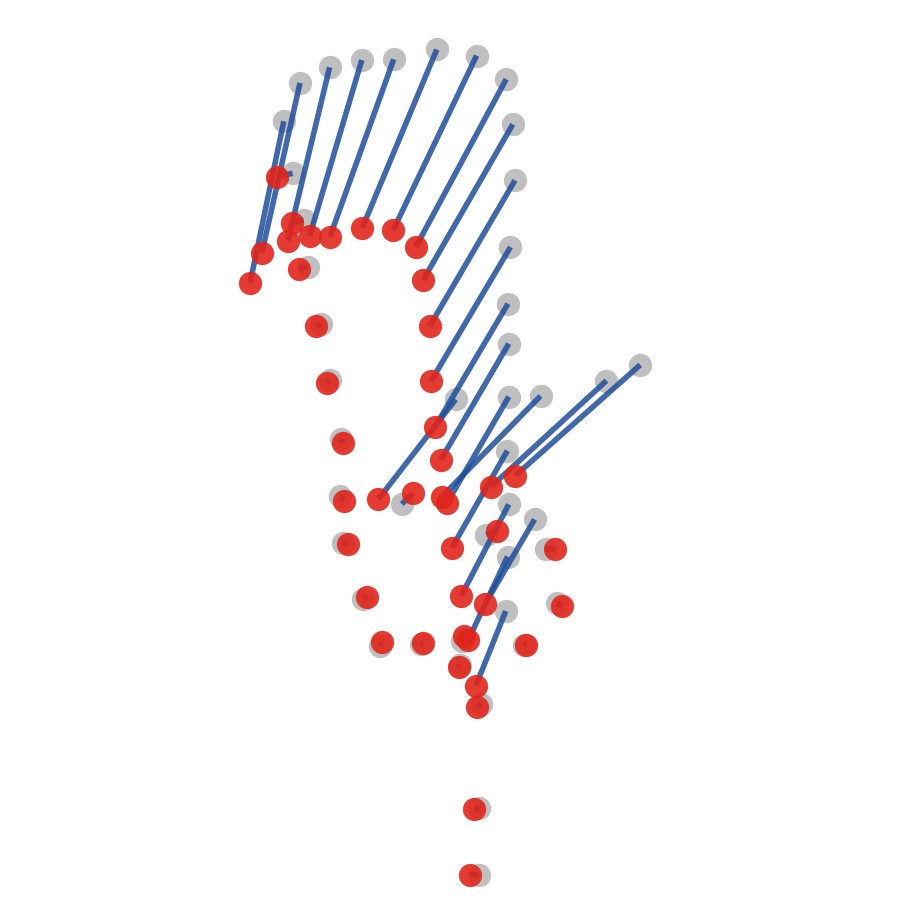}{5.6} & \mapauc{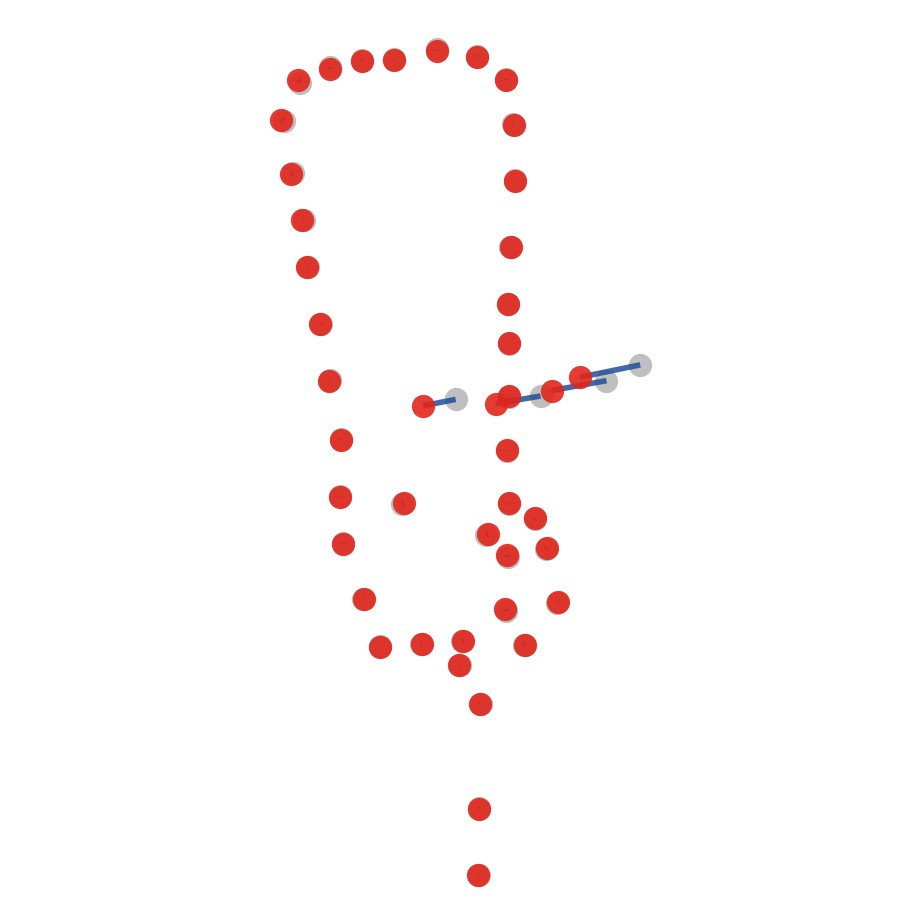}{69.8} & \mapauc{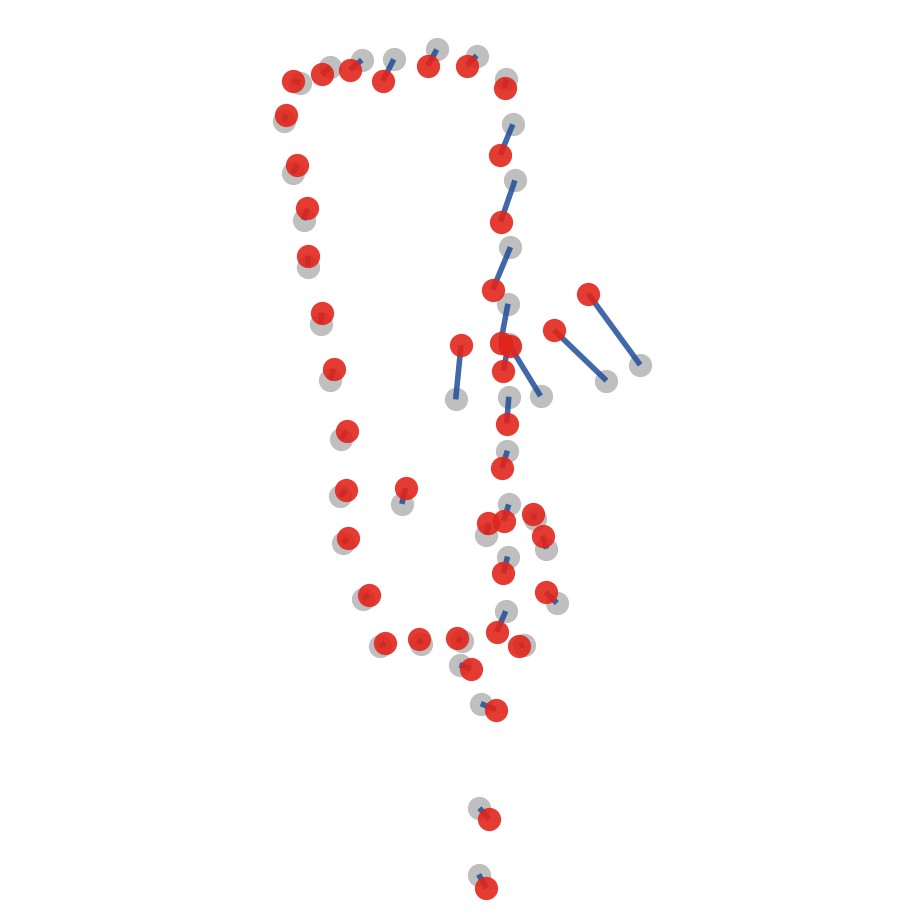}{10.6} & \mapauc{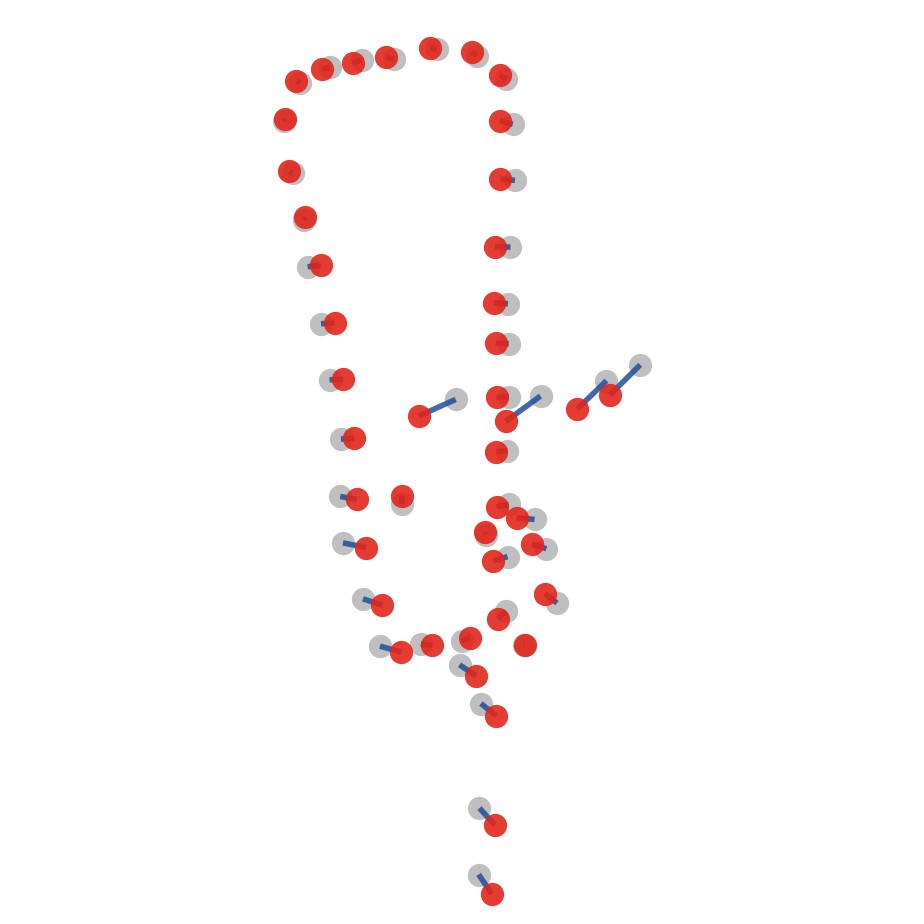}{11.1} \\[0pt]
\maprow{\textbf{+ \method}} & \mapname{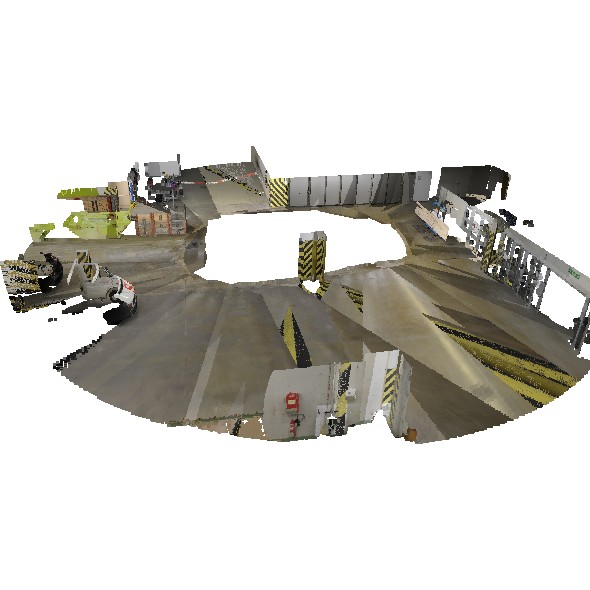}{delivery area} & \mapaucbf{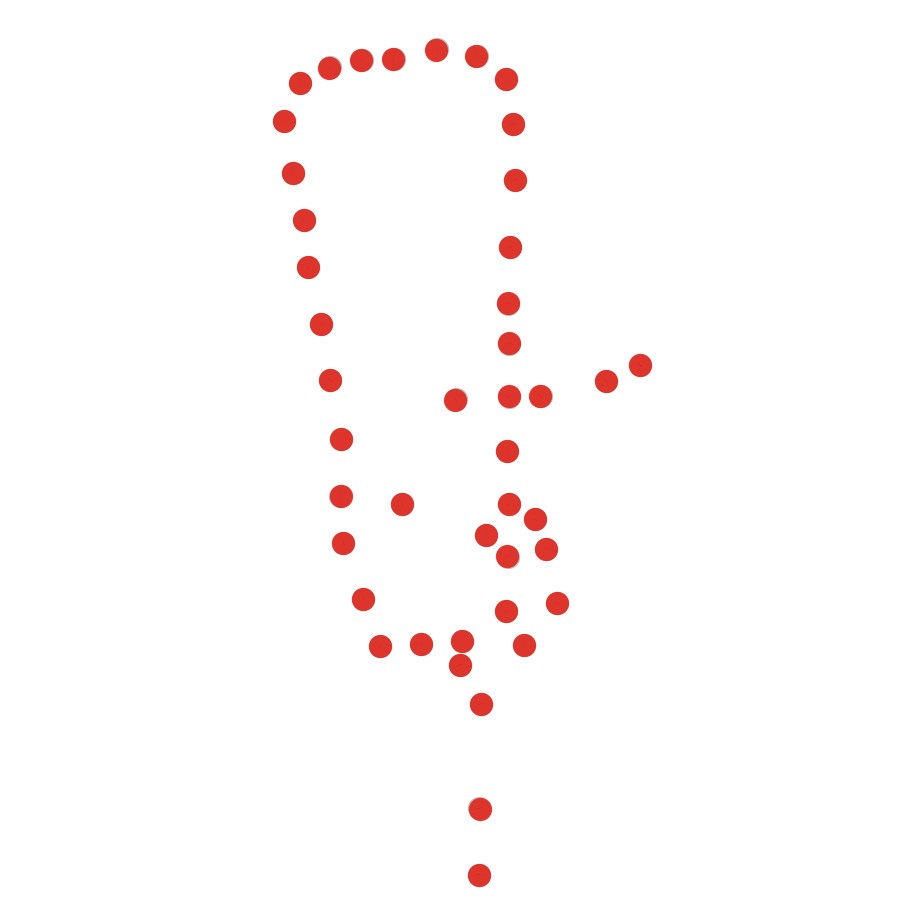}{96.2} & \mapaucbf{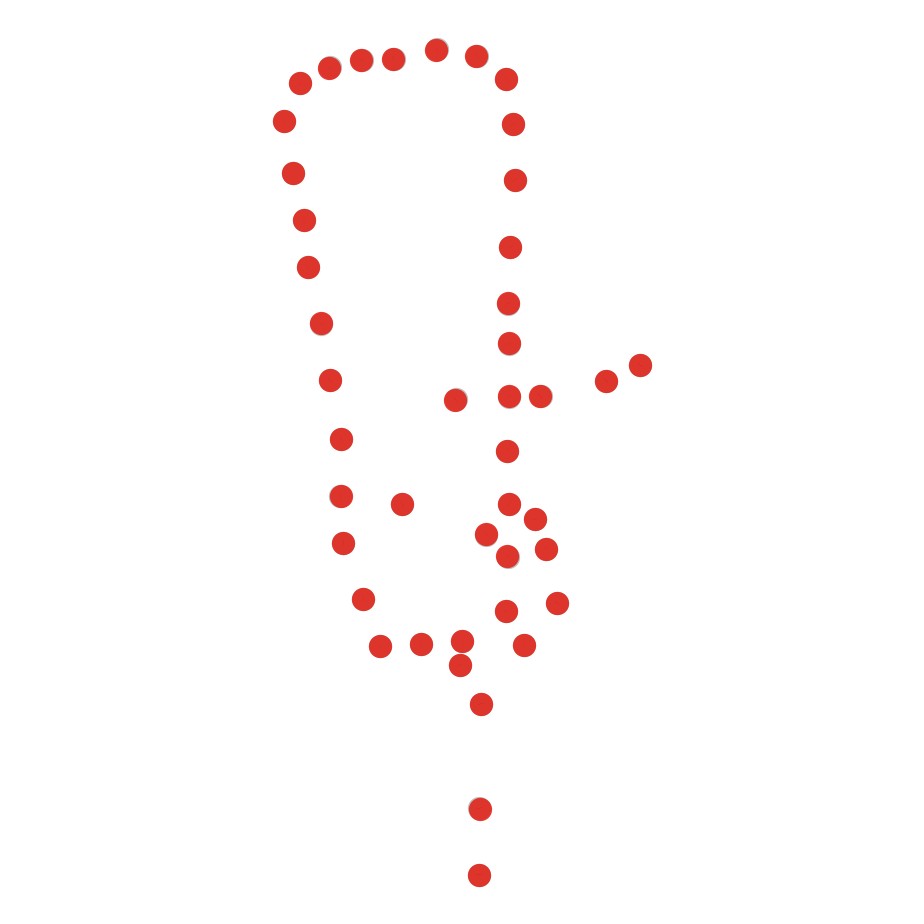}{96.4} & \mapaucbf{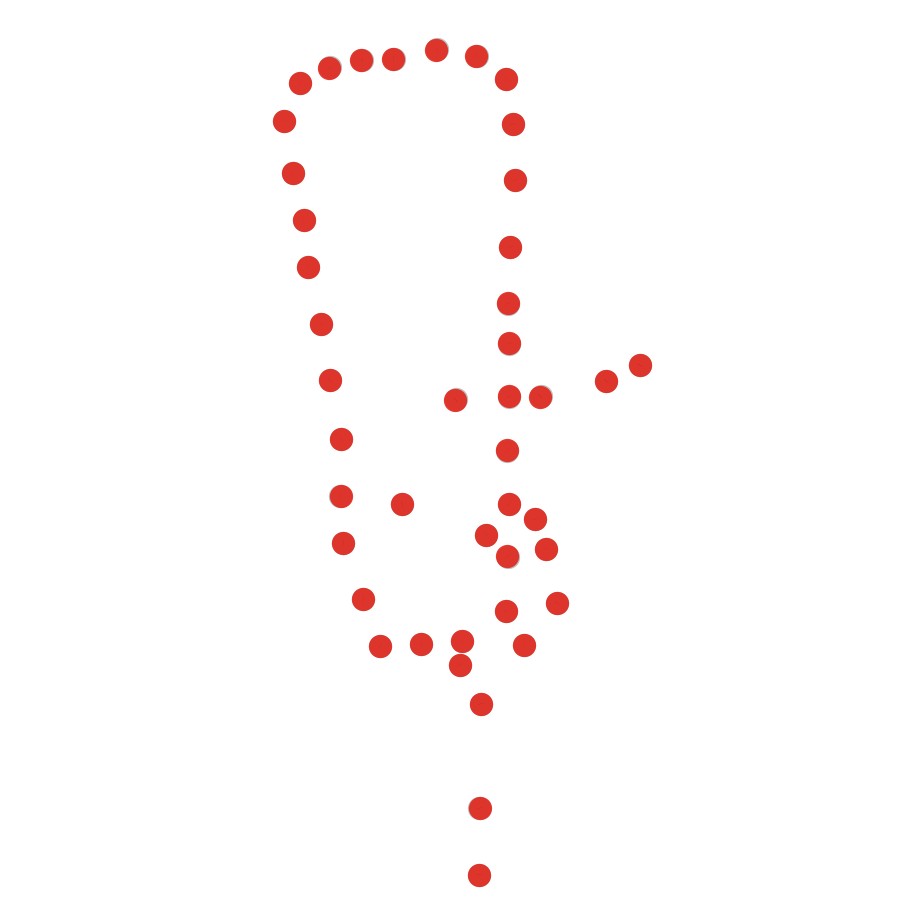}{96.2} & \mapaucbf{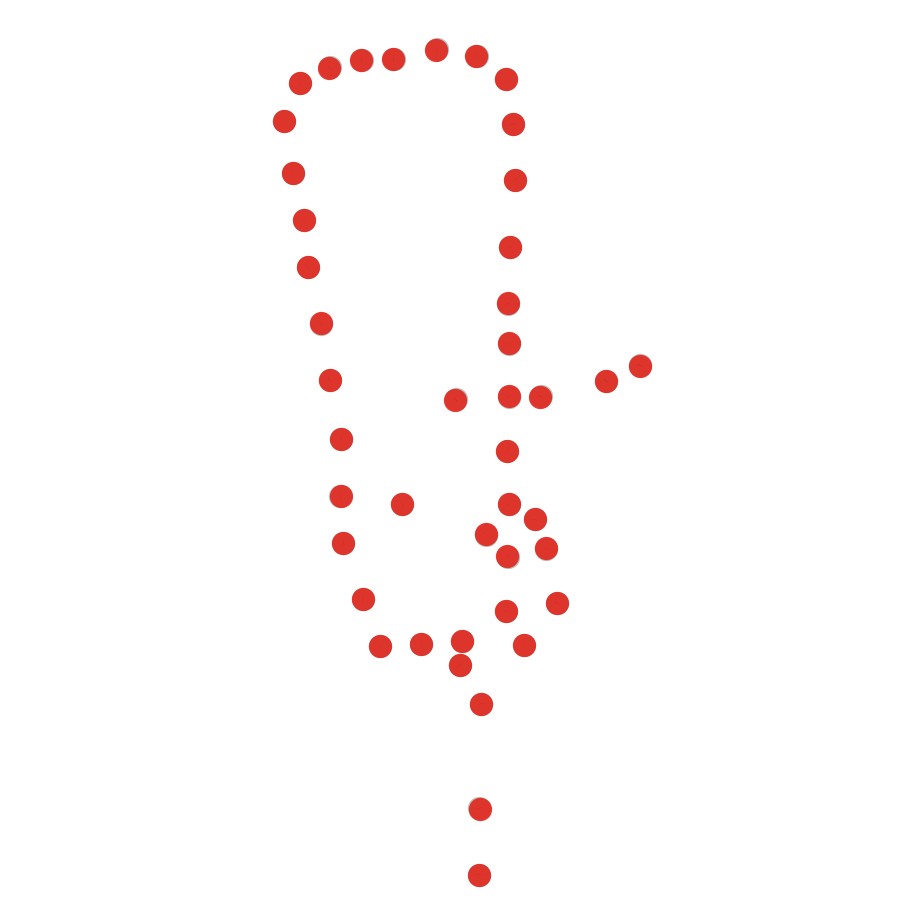}{96.1} & \mapaucbf{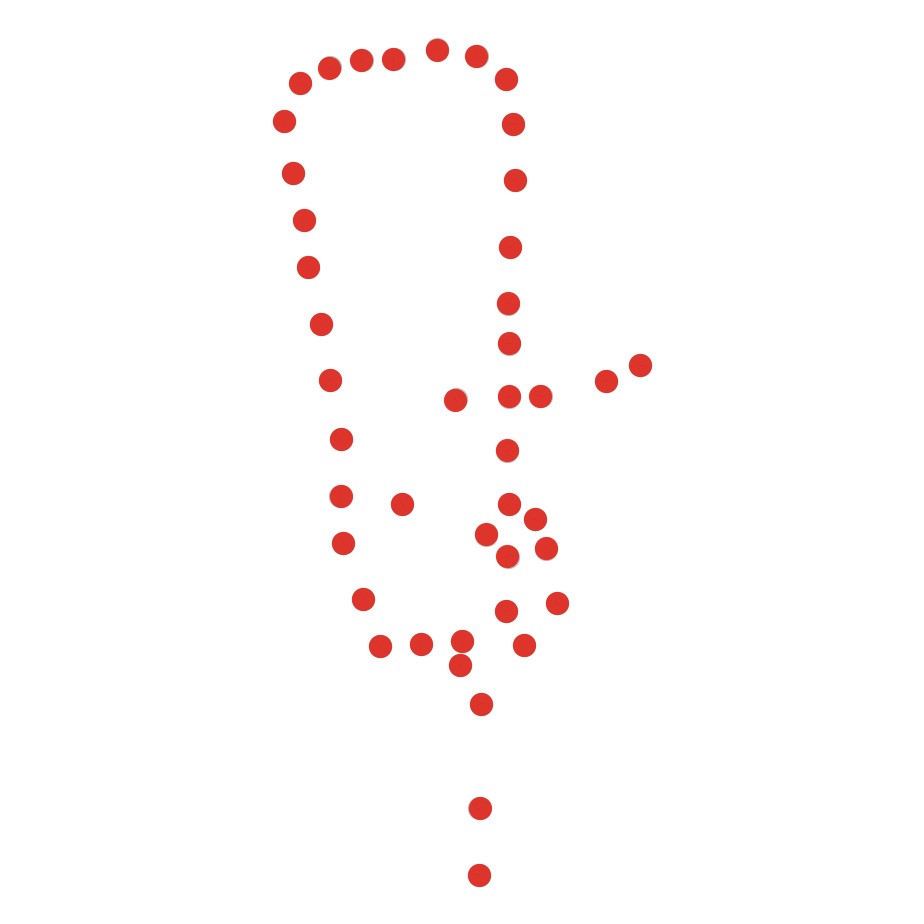}{96.3} \\[4pt]
\end{tabular}
\caption{Camera positions on ETH3D scenes with lens distortion: the Initialization cameras of each method (upper row of each scene) and the same cameras after \method (lower row), aligned to the ground truth; each estimated camera (red) is joined to its ground-truth position (grey) by a blue line, so longer lines are larger position errors. More scenes in \cref{fig:distorted_map_b}.}
\label{fig:distorted_map}
\end{figure*}

\ifdefined\distW\else\newlength{\distW}\fi\setlength{\distW}{1.08in}
\begin{figure*}[p]
\def\mappanel#1{{\setlength{\fboxsep}{0pt}\setlength{\fboxrule}{0.3pt}\fcolorbox{black!20}{white}{\includegraphics[width=\dimexpr\distW-0.6pt\relax]{fig/distorted_map/#1}}}}
\def\mapauc#1#2{\mappanel{#1}\llap{\raisebox{1.2pt}{\setlength{\fboxsep}{0.8pt}\colorbox{white}{\tiny AUC@3$^\circ$ #2}}\hspace{1.2pt}}}
\def\mapaucbf#1#2{\mappanel{#1}\llap{\raisebox{1.2pt}{\setlength{\fboxsep}{0.8pt}\colorbox{white}{\tiny\bfseries AUC@3$^\circ$ #2}}\hspace{1.2pt}}}
\def\maprow#1{\rotatebox{90}{\scriptsize #1}}
\def\mapname#1#2{\raisebox{0pt}[0pt][0pt]{\parbox{\distW}{\centering\includegraphics[width=0.8\distW]{fig/distorted_map/#1}\\[1pt]\footnotesize\bfseries #2}}}  
\centering
\setlength{\tabcolsep}{0.8pt}
\renewcommand{\arraystretch}{0.7}
\begin{tabular}{@{}>{\centering\arraybackslash}m{0.9em} >{\centering\arraybackslash}m{\distW}@{\hspace{4pt}} *{5}{>{\centering\arraybackslash}m{\distW}} @{}}
 & \small\bfseries GT & \small\bfseries GLOMAP & \small\bfseries DGSfM & \small\bfseries GLUEMAP & \small\bfseries Pi3X & \small\bfseries DVLT \\[2pt]

\maprow{Initialization} & \mappanel{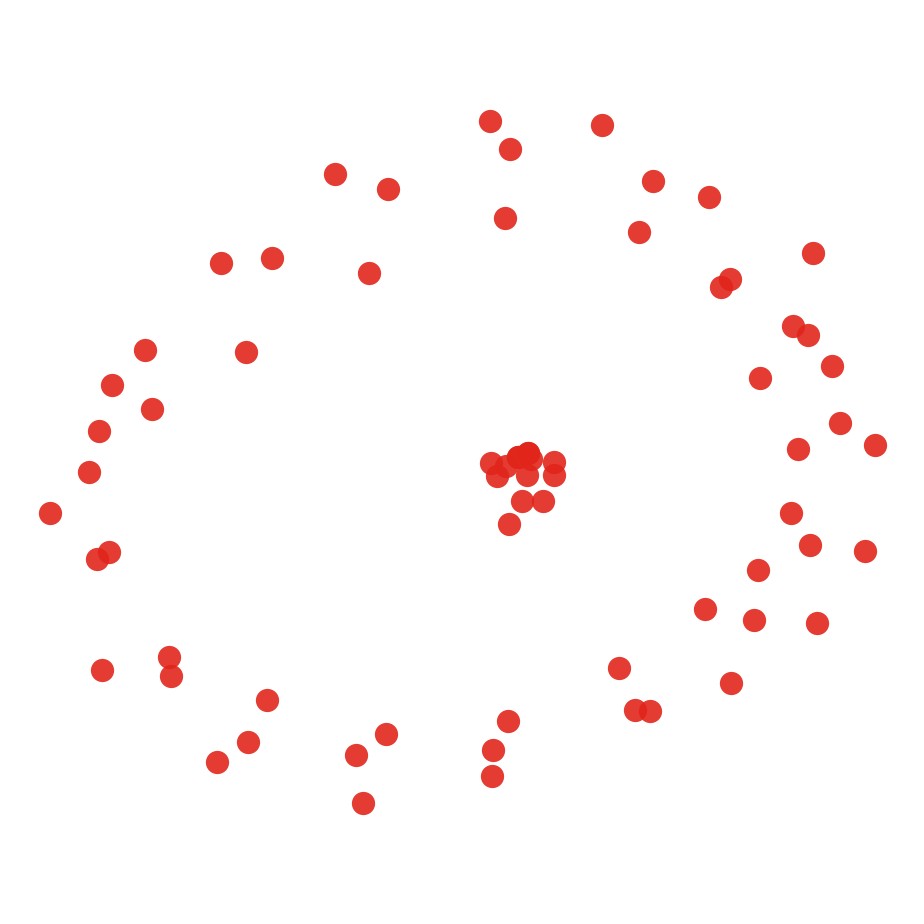} & \mapauc{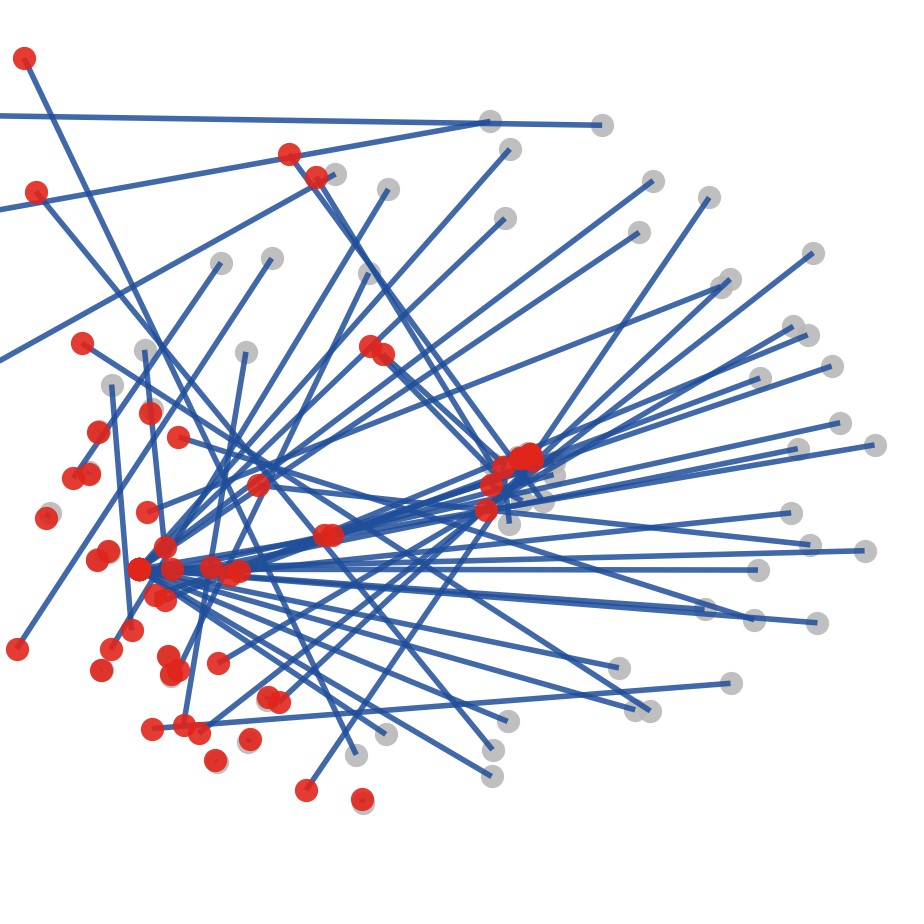}{6.6} & \mapauc{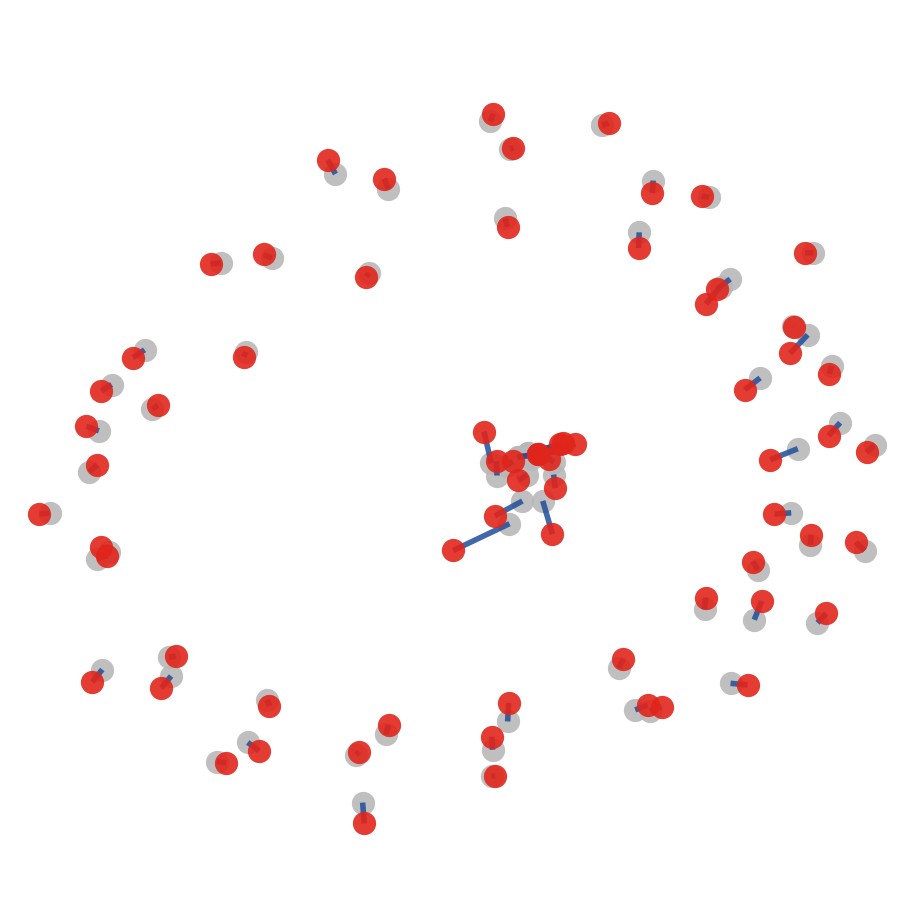}{25.1} & \mapauc{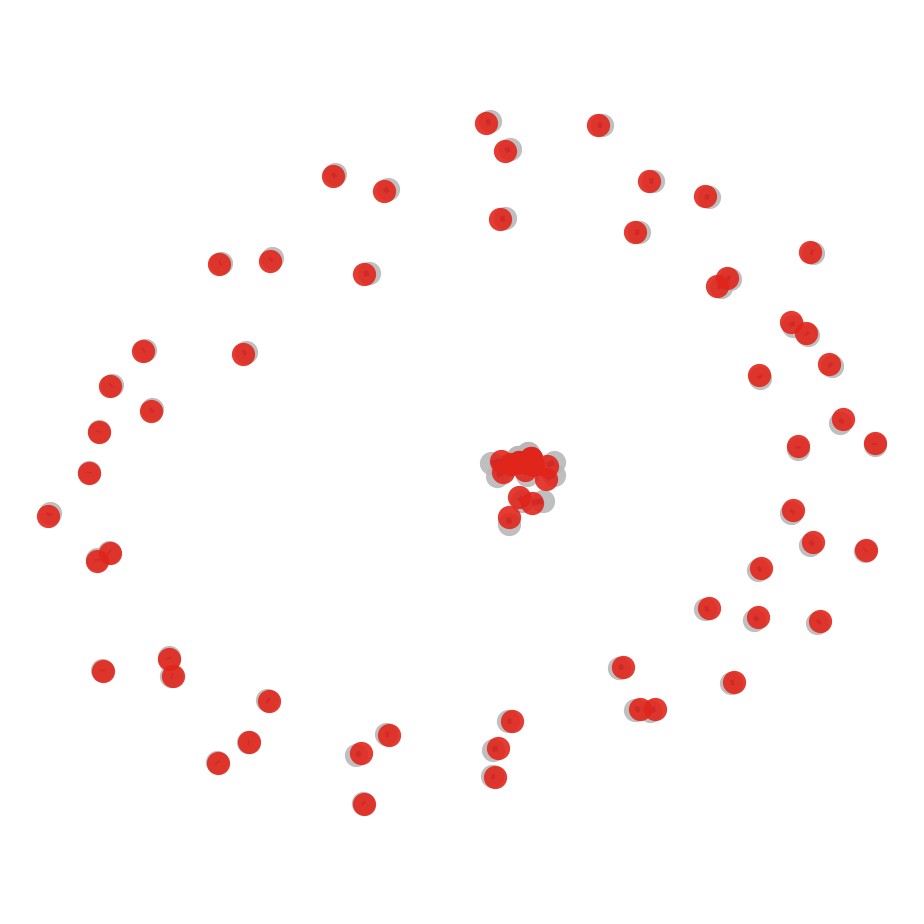}{67.7} & \mapauc{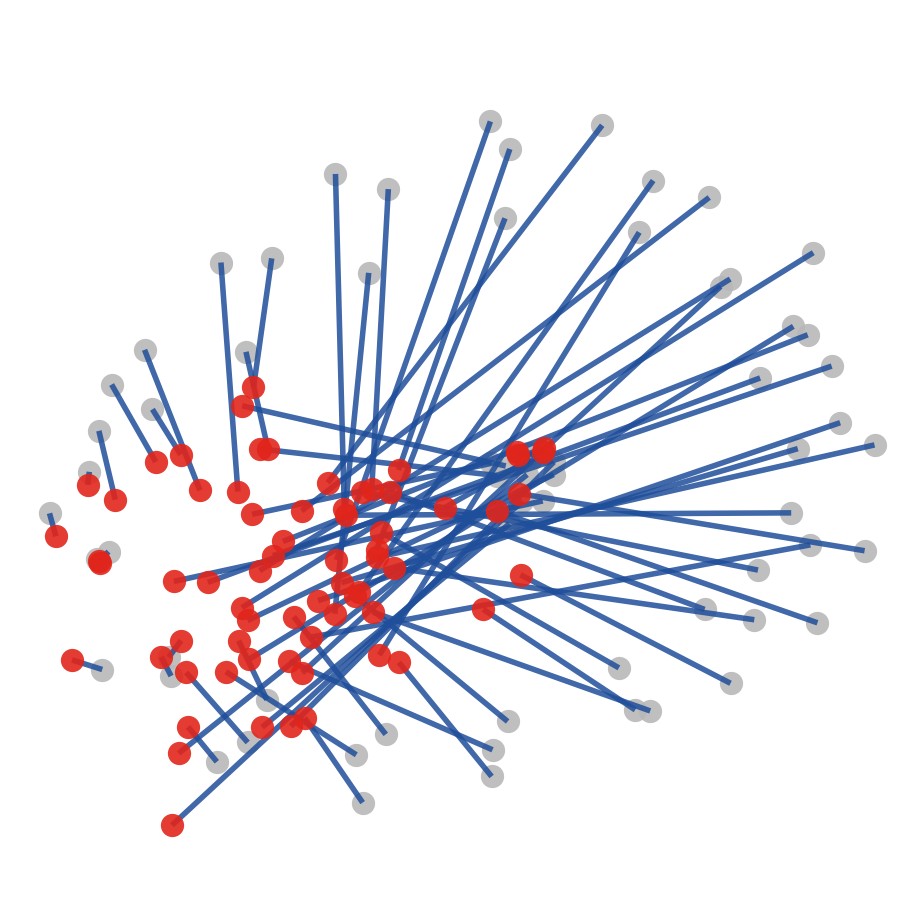}{0.4} & \mapauc{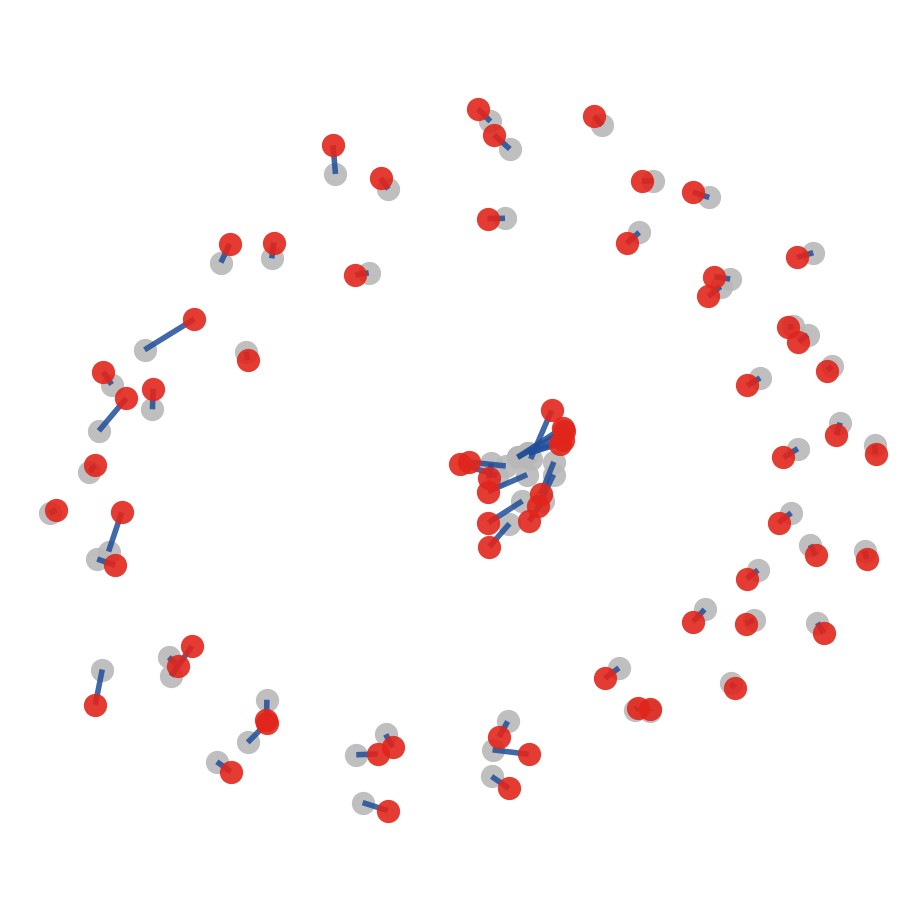}{10.3} \\[0pt]
\maprow{\textbf{\textbf{+ \method}}} & \mapname{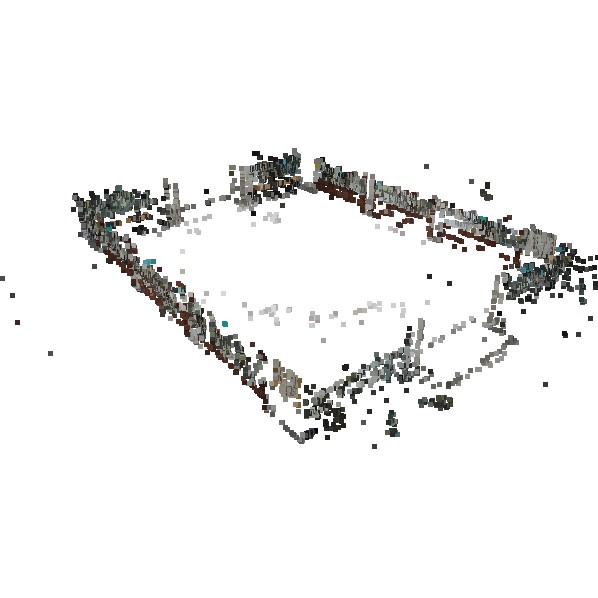}{exhibition hall} & \mapaucbf{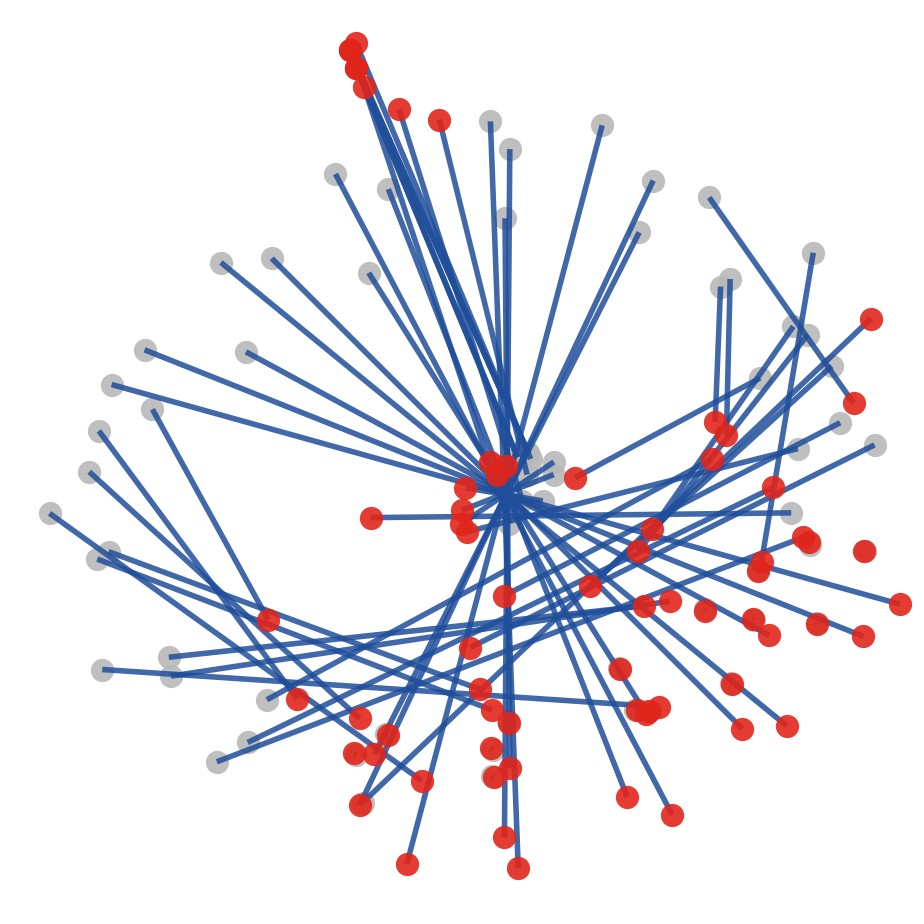}{19.0} & \mapaucbf{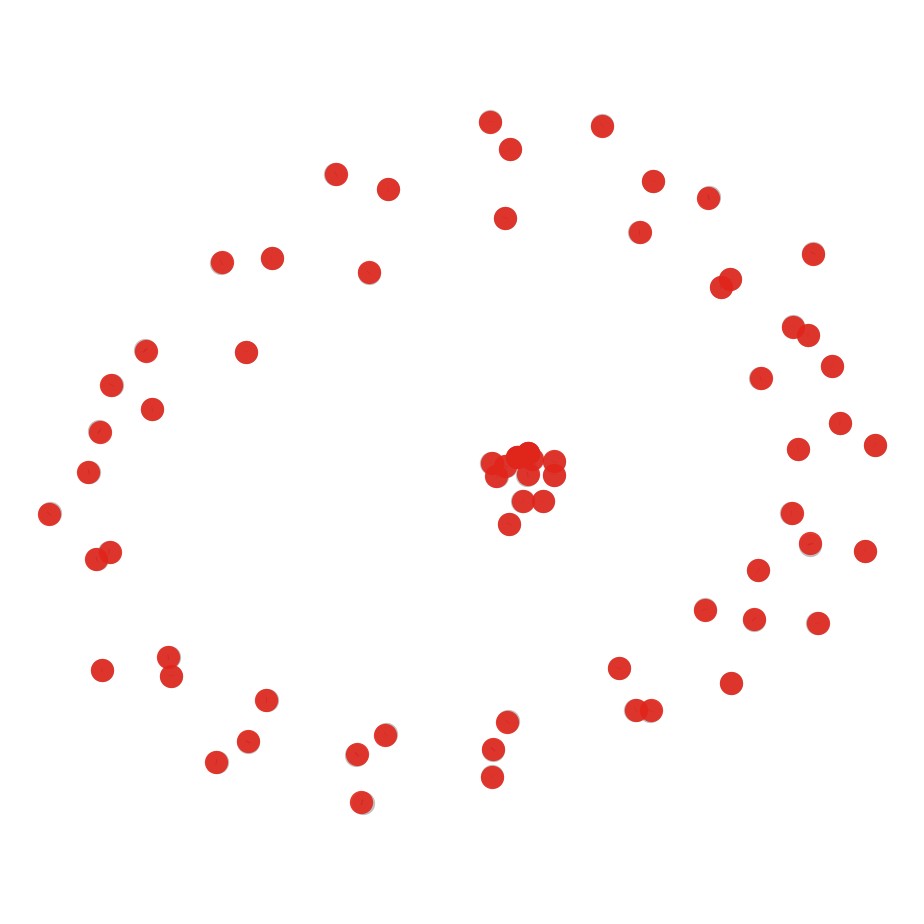}{91.8} & \mapaucbf{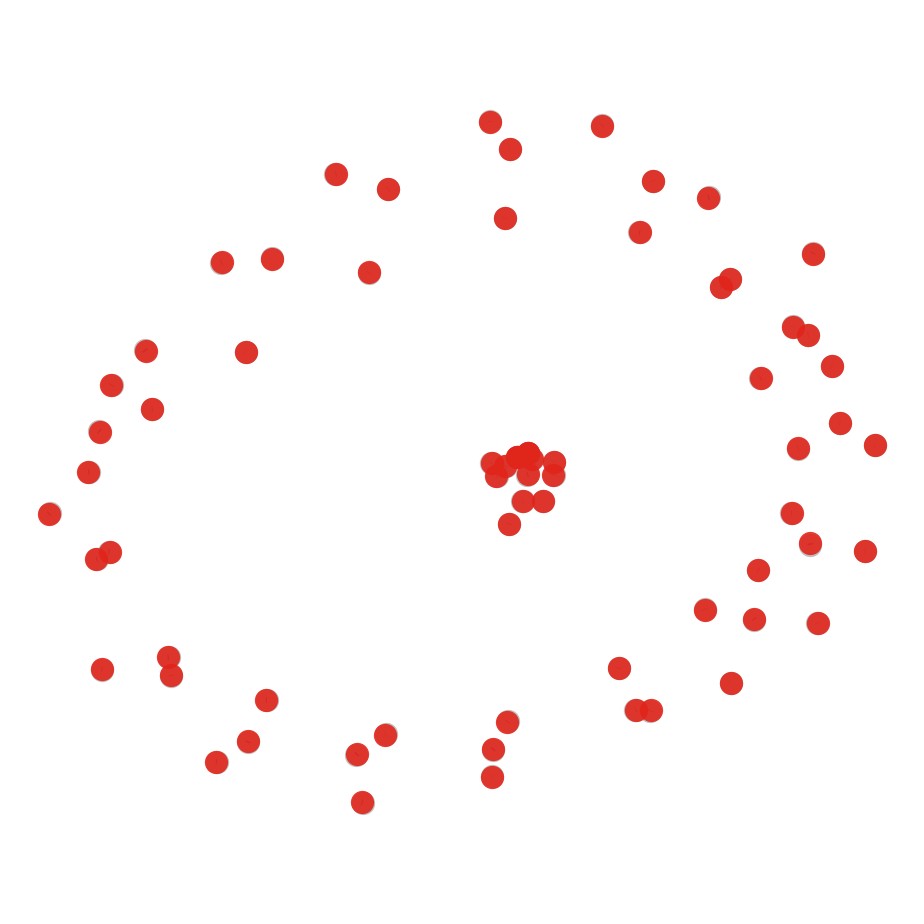}{91.8} & \mapaucbf{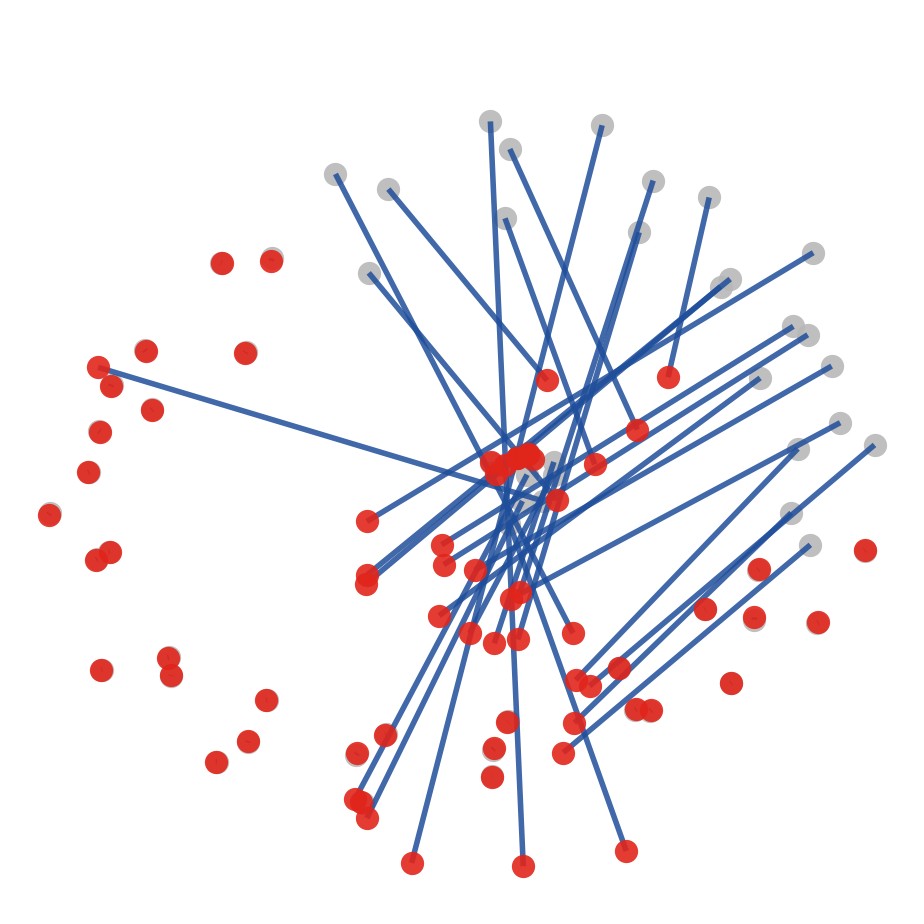}{30.9} & \mapaucbf{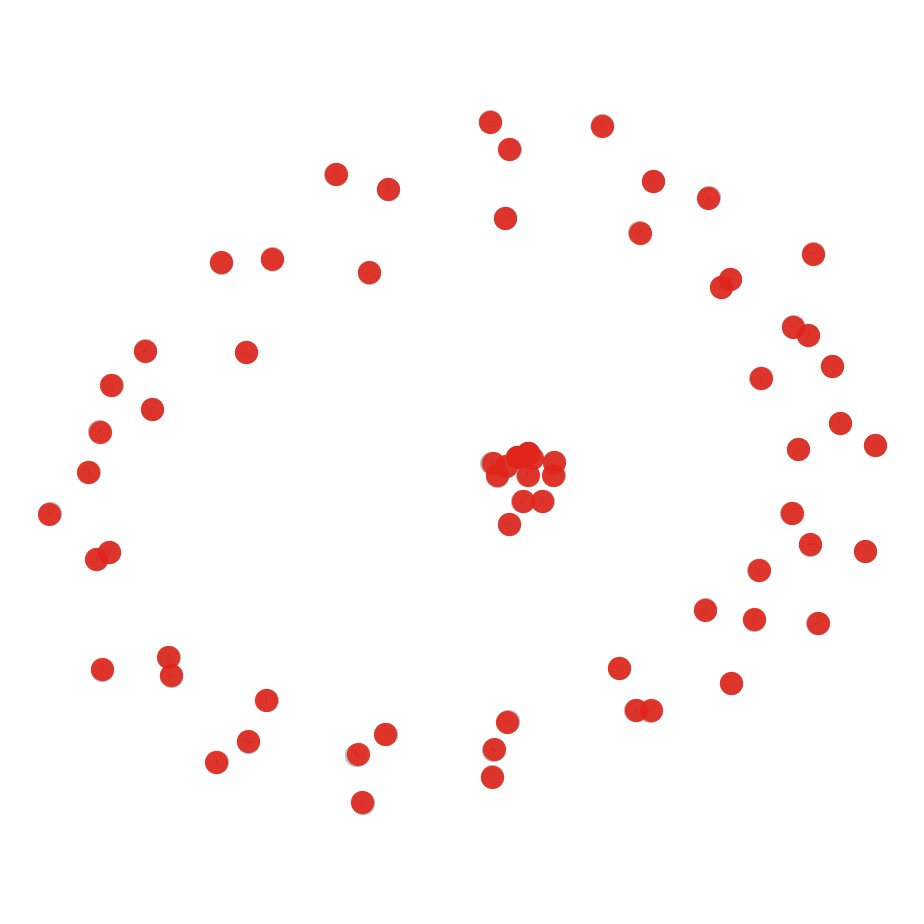}{91.4} \\[4pt]
\maprow{Initialization} & \mappanel{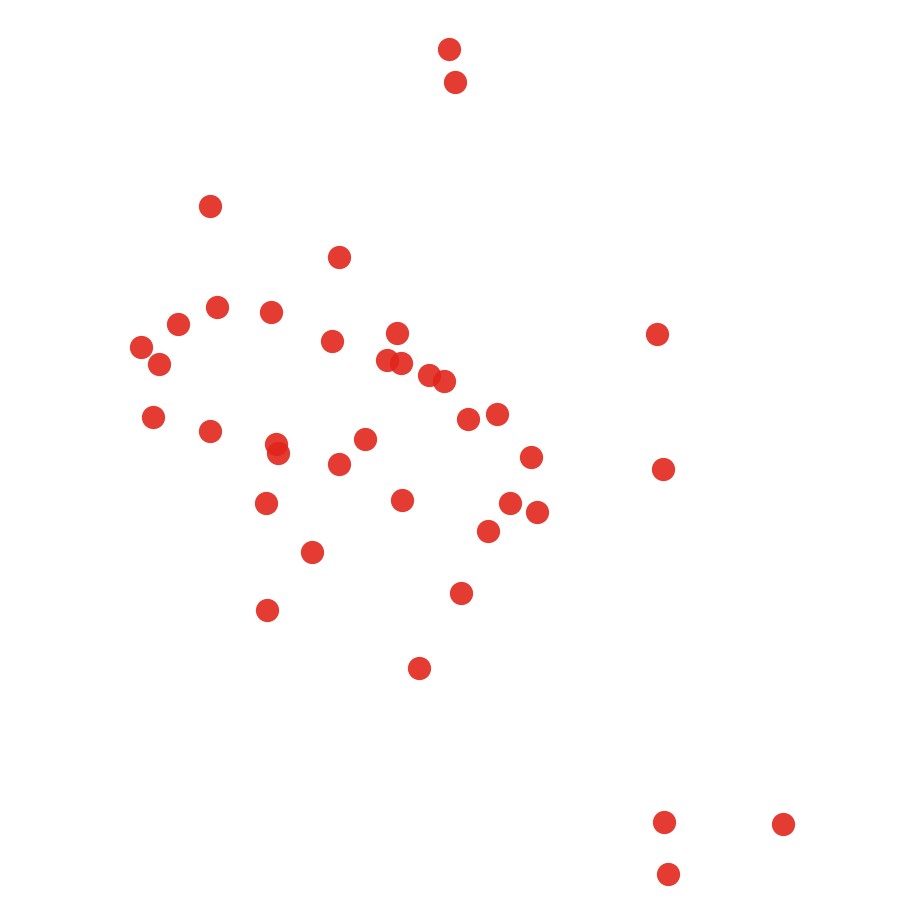} & \mapauc{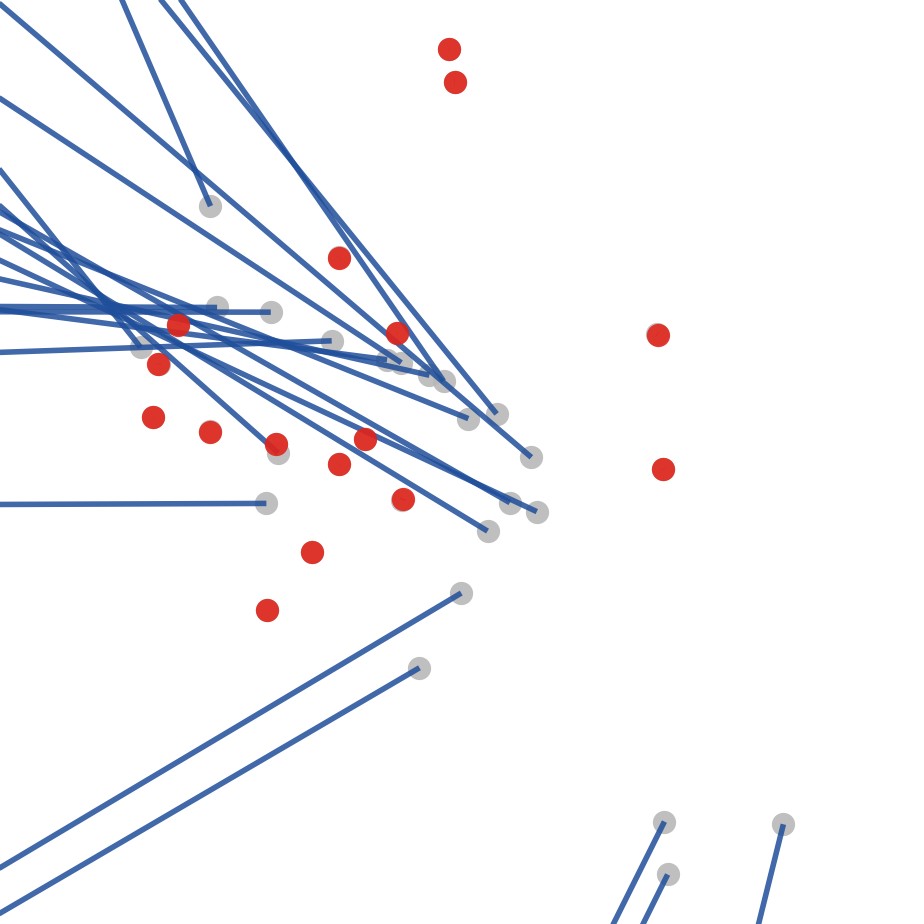}{21.4} & \mapauc{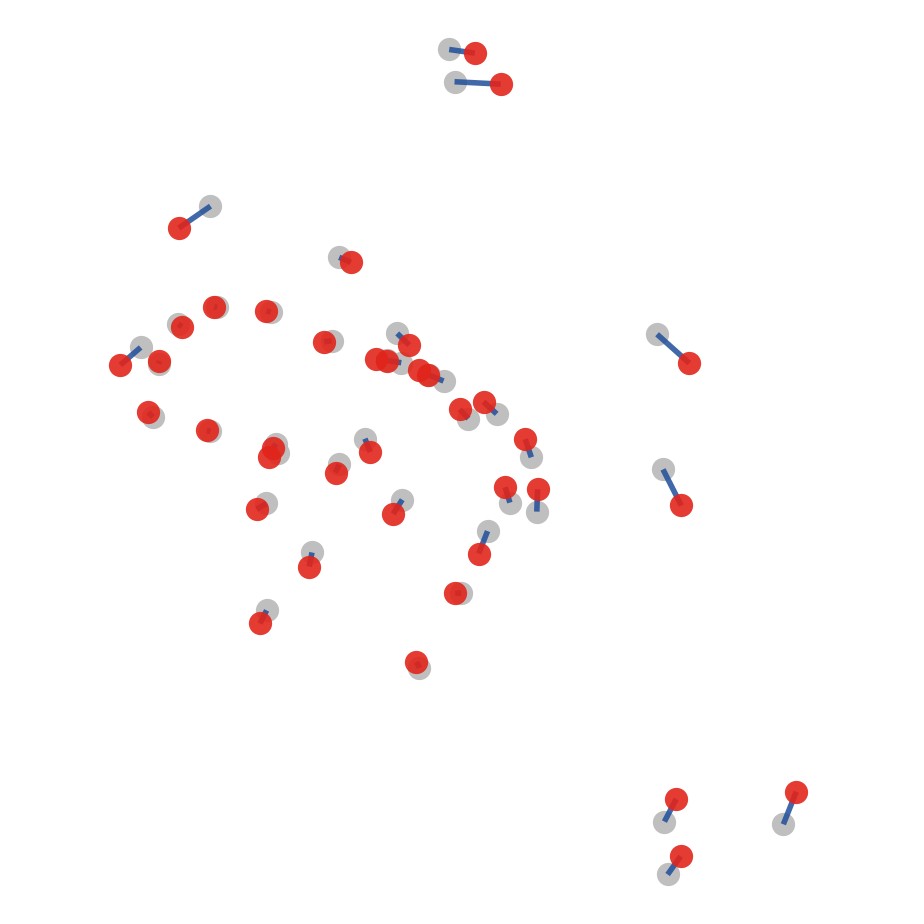}{13.9} & \mapauc{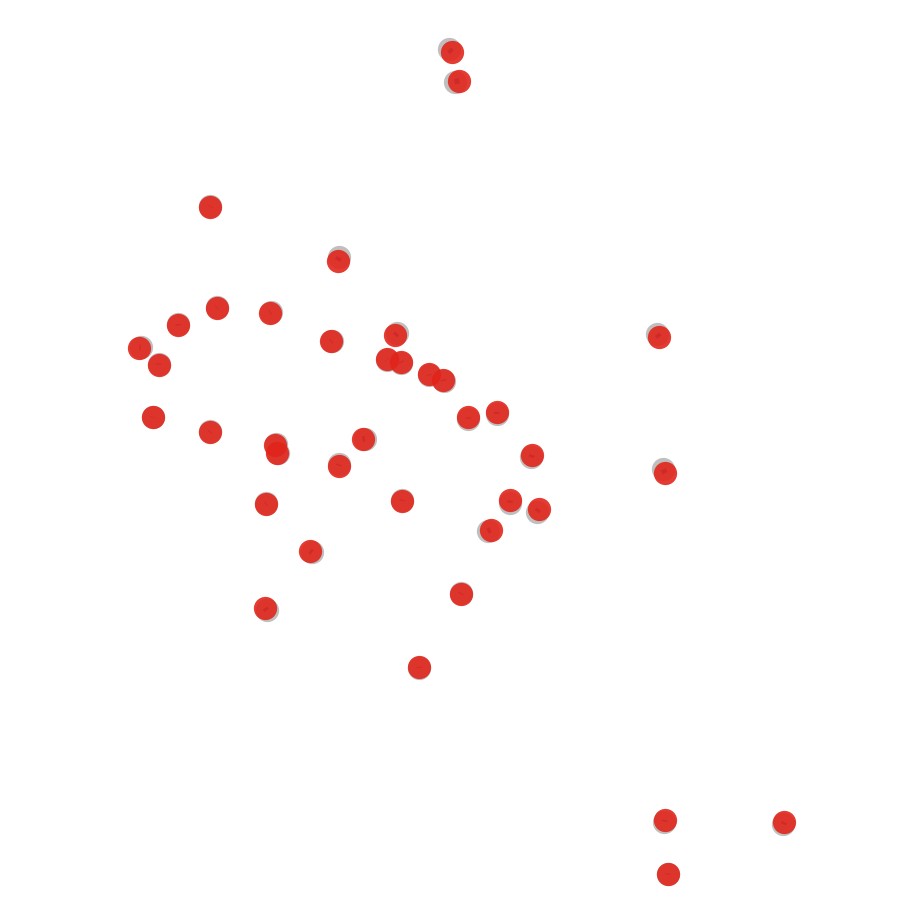}{75.1} & \mapauc{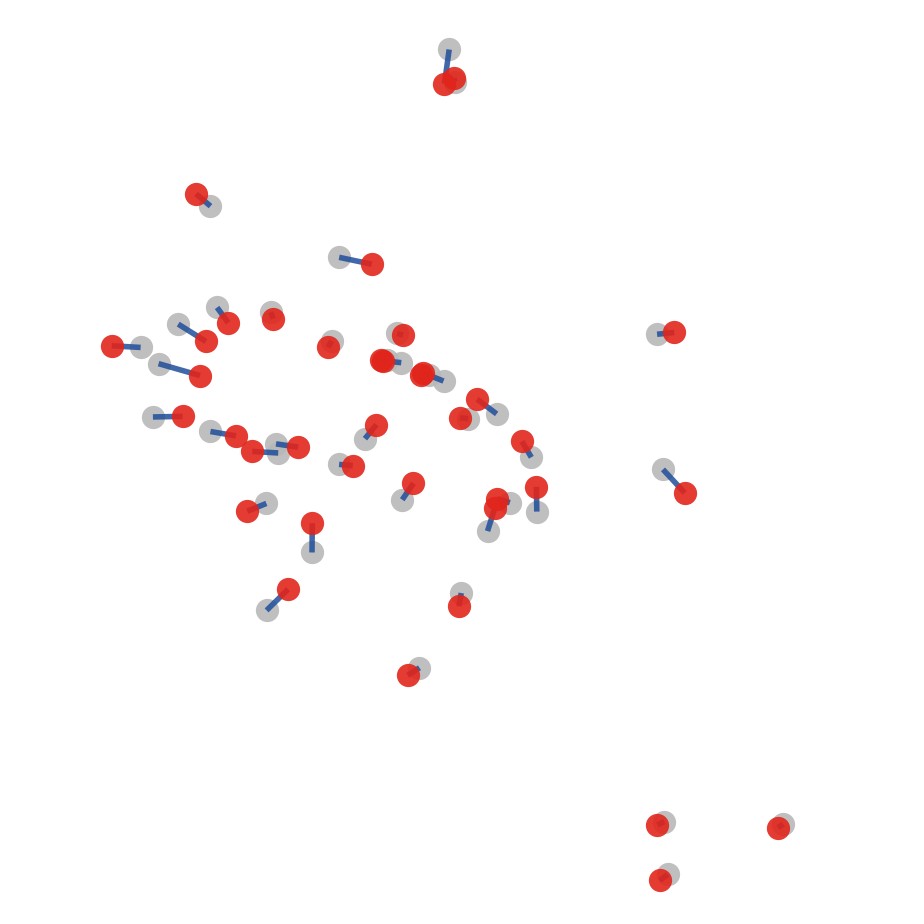}{13.7} & \mapauc{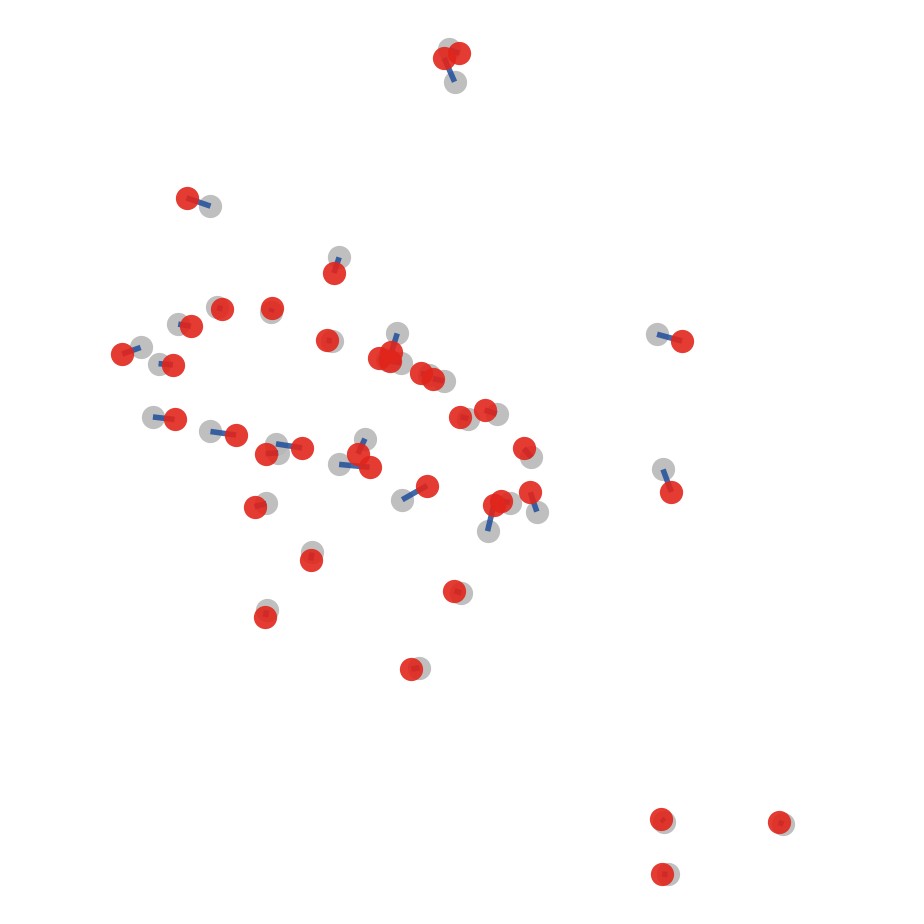}{15.2} \\[0pt]
\maprow{\textbf{\textbf{+ \method}}} & \mapname{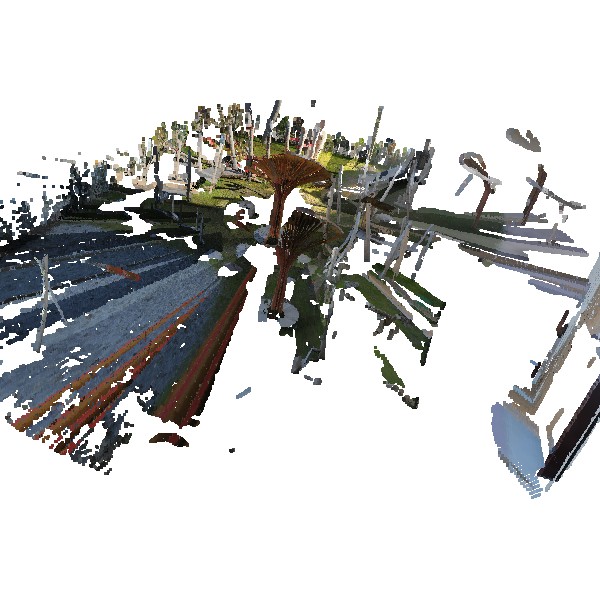}{playground} & \mapaucbf{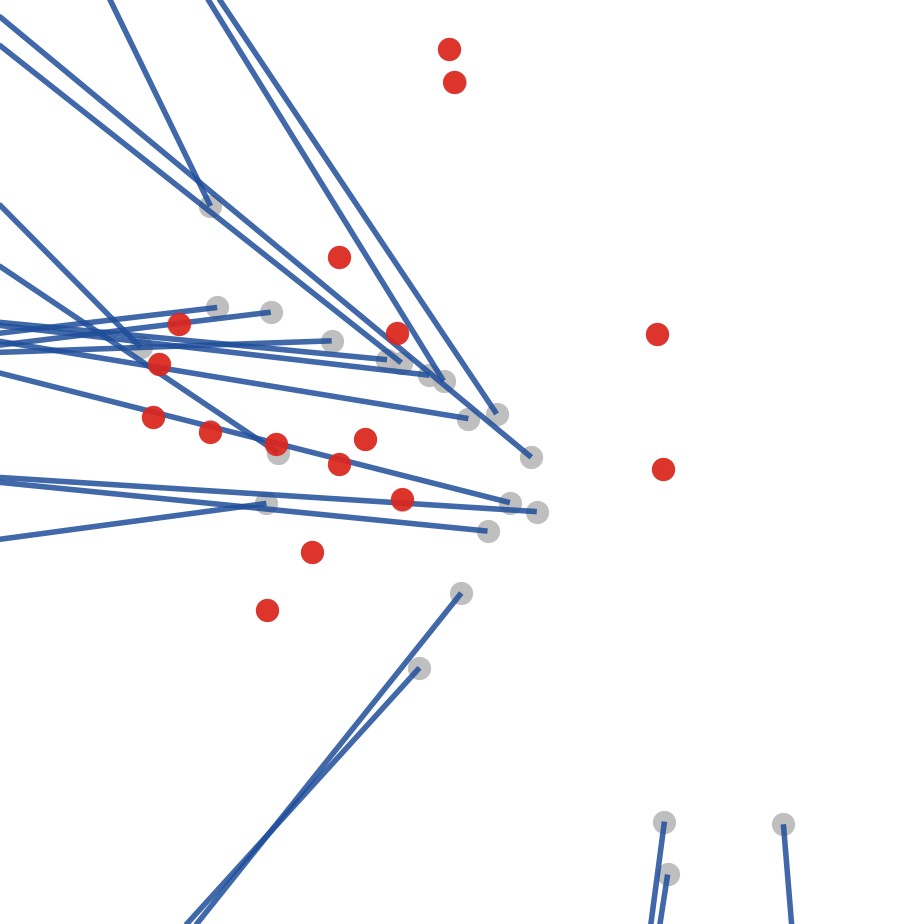}{24.1} & \mapaucbf{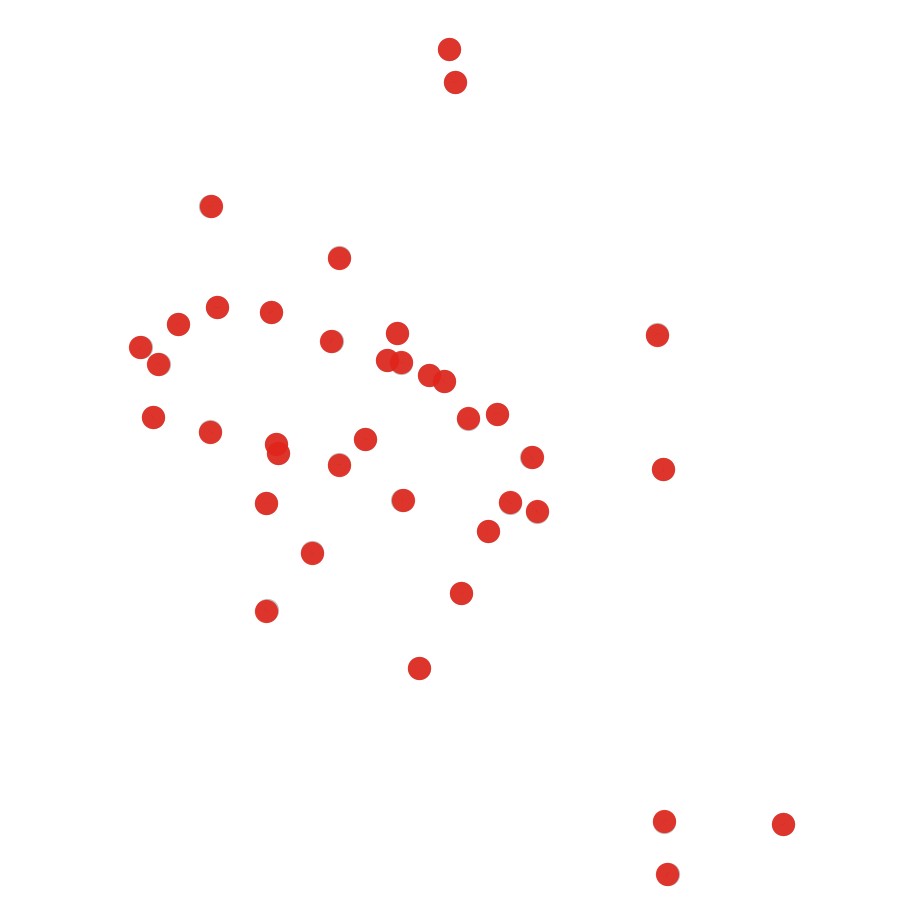}{94.8} & \mapaucbf{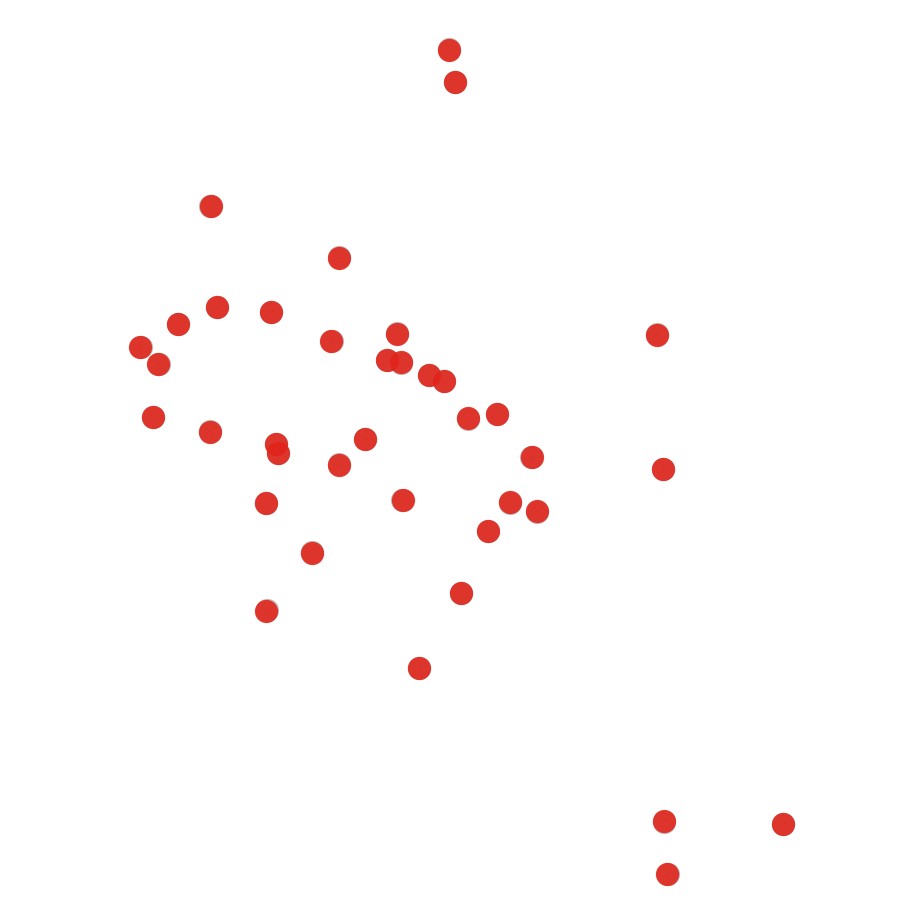}{94.6} & \mapaucbf{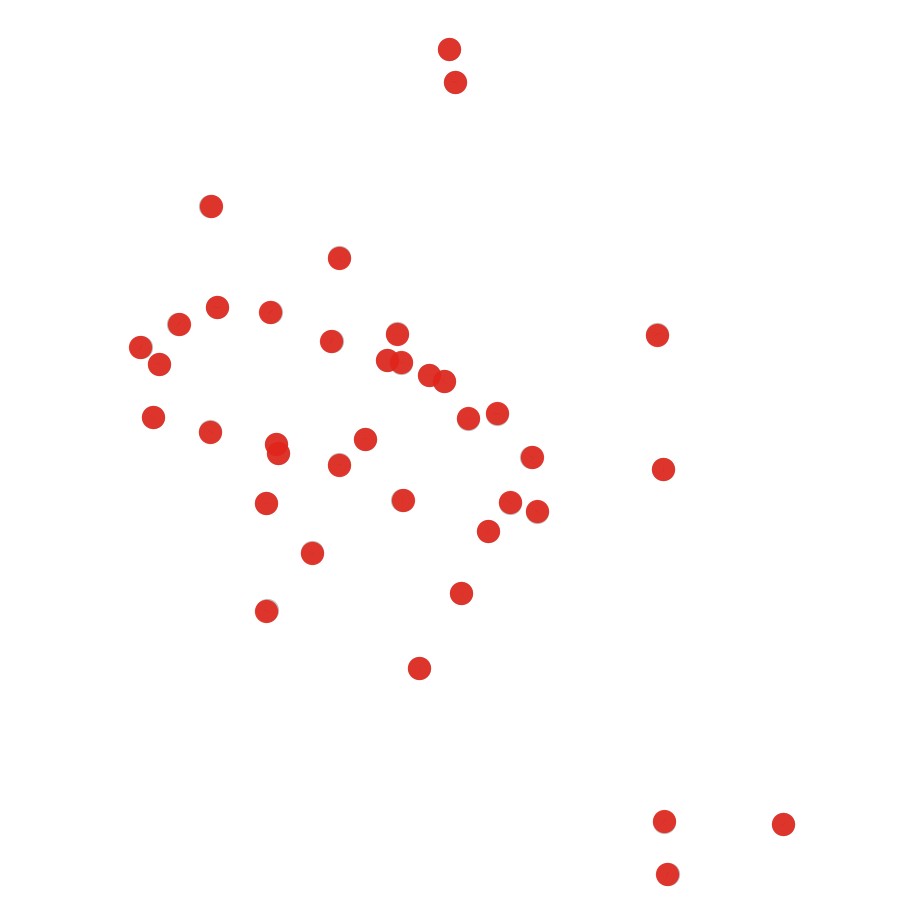}{94.7} & \mapaucbf{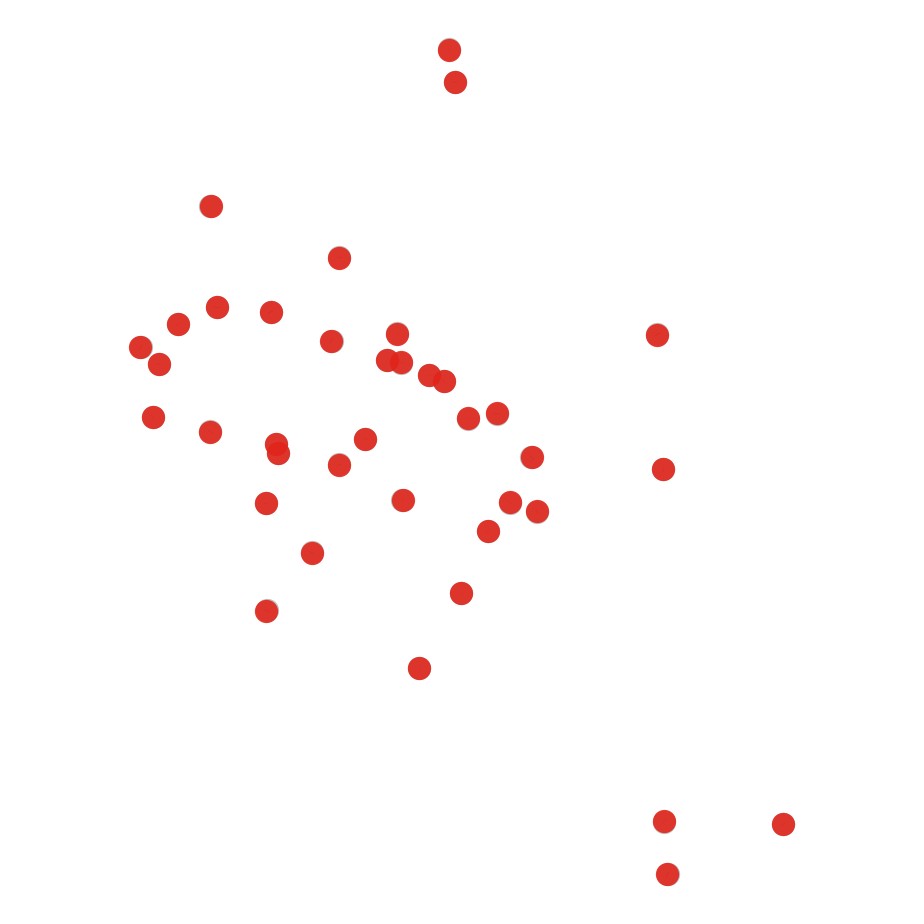}{94.5} \\[4pt]
\maprow{Initialization} & \mappanel{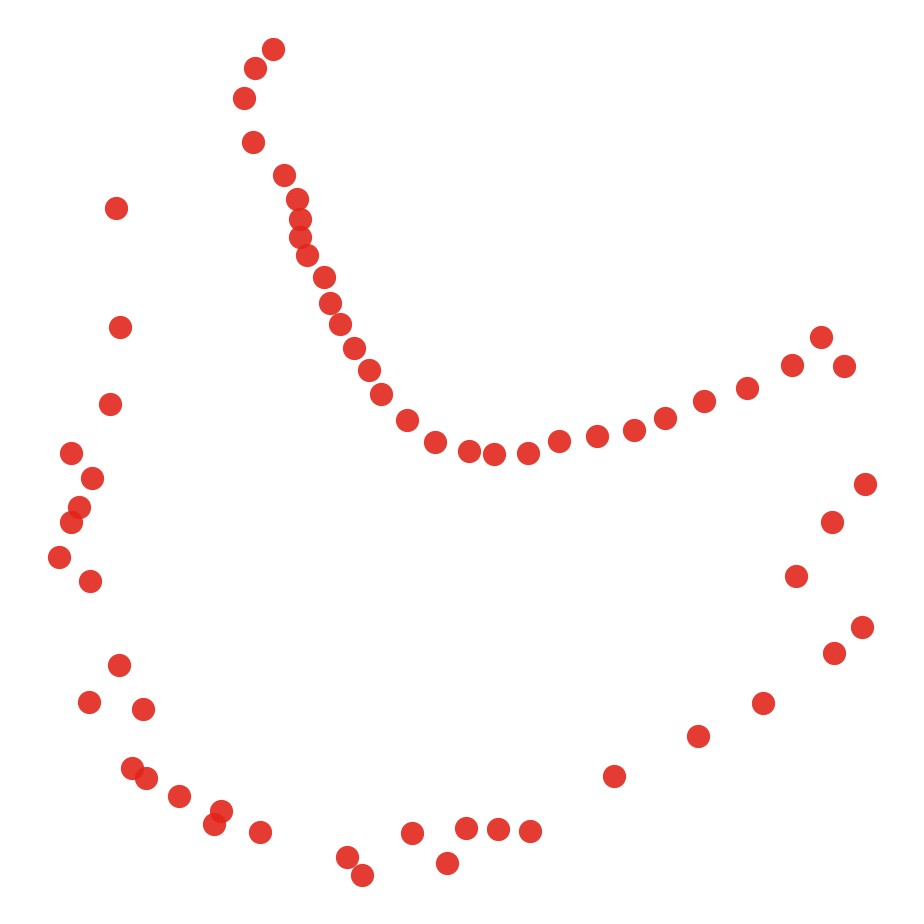} & \mapauc{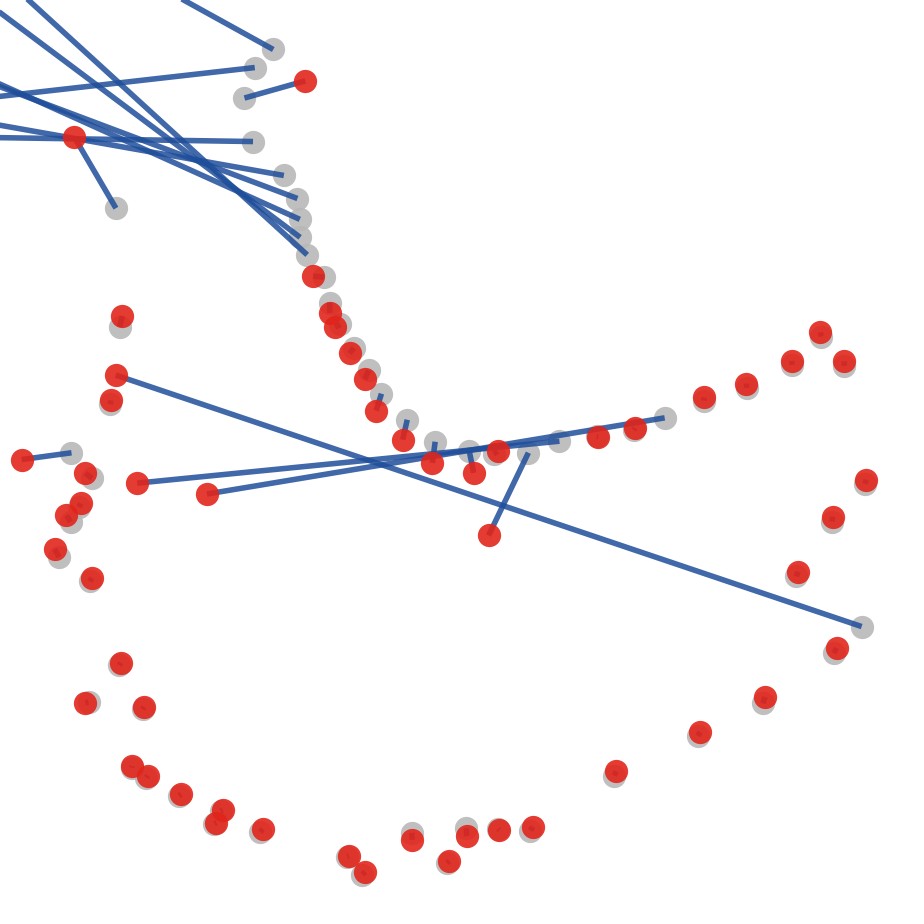}{30.5} & \mapauc{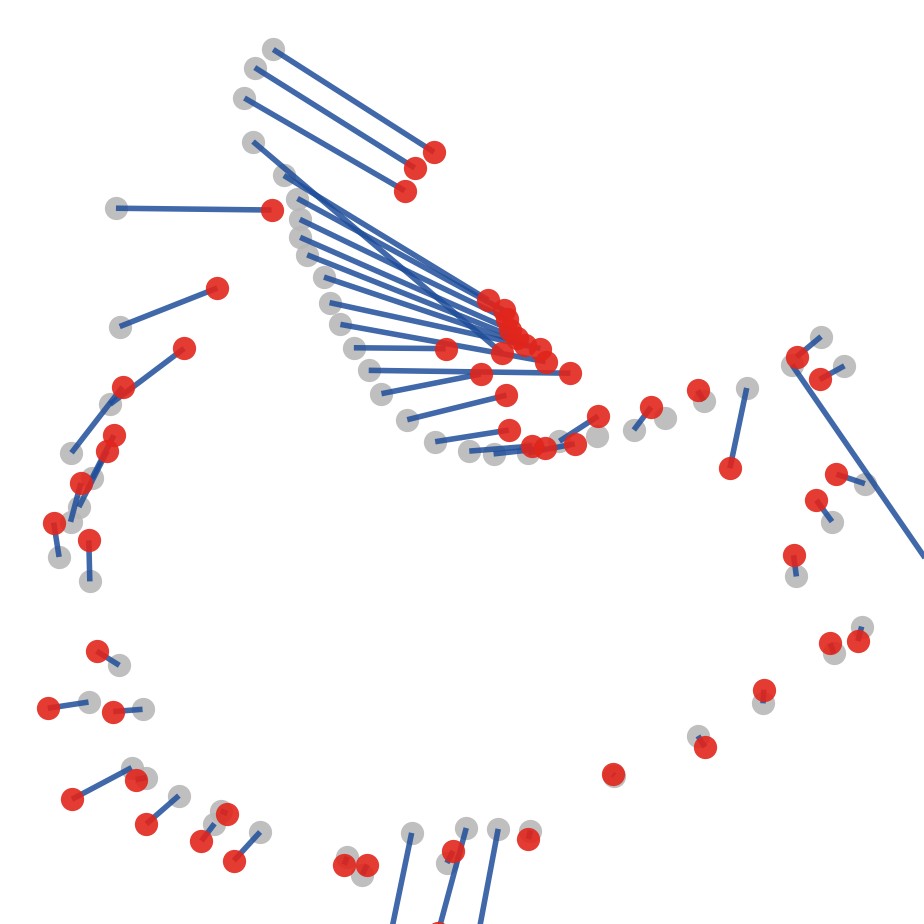}{0.4} & \mapauc{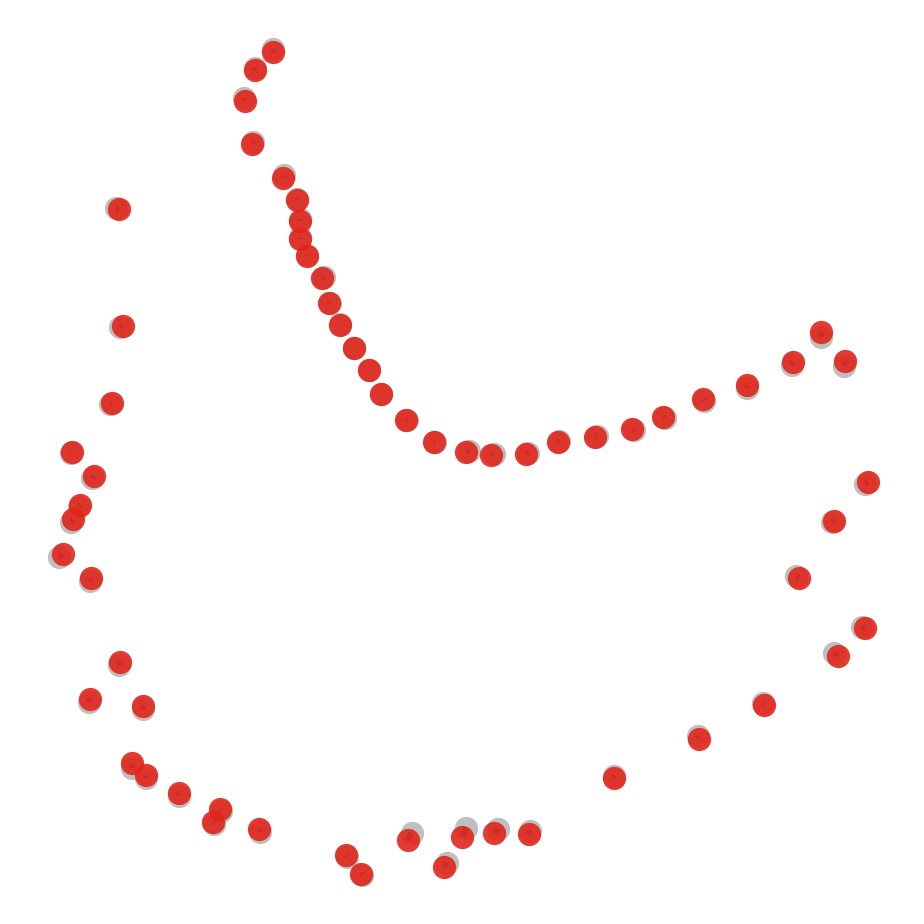}{79.7} & \mapauc{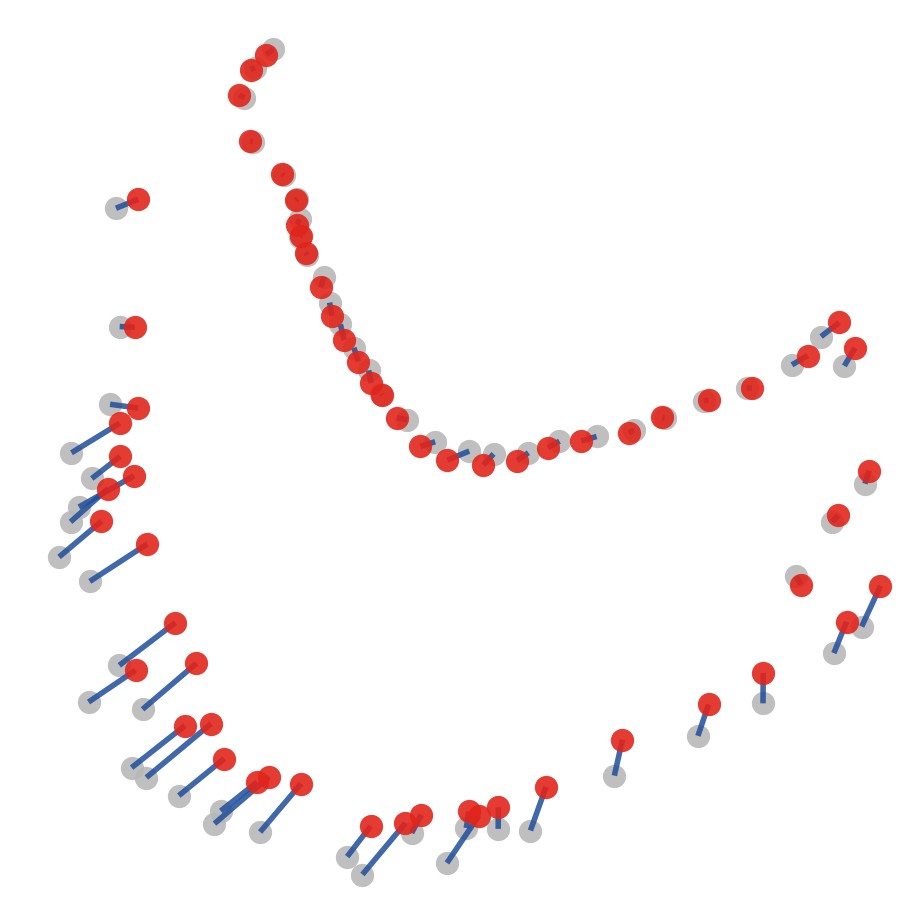}{6.4} & \mapauc{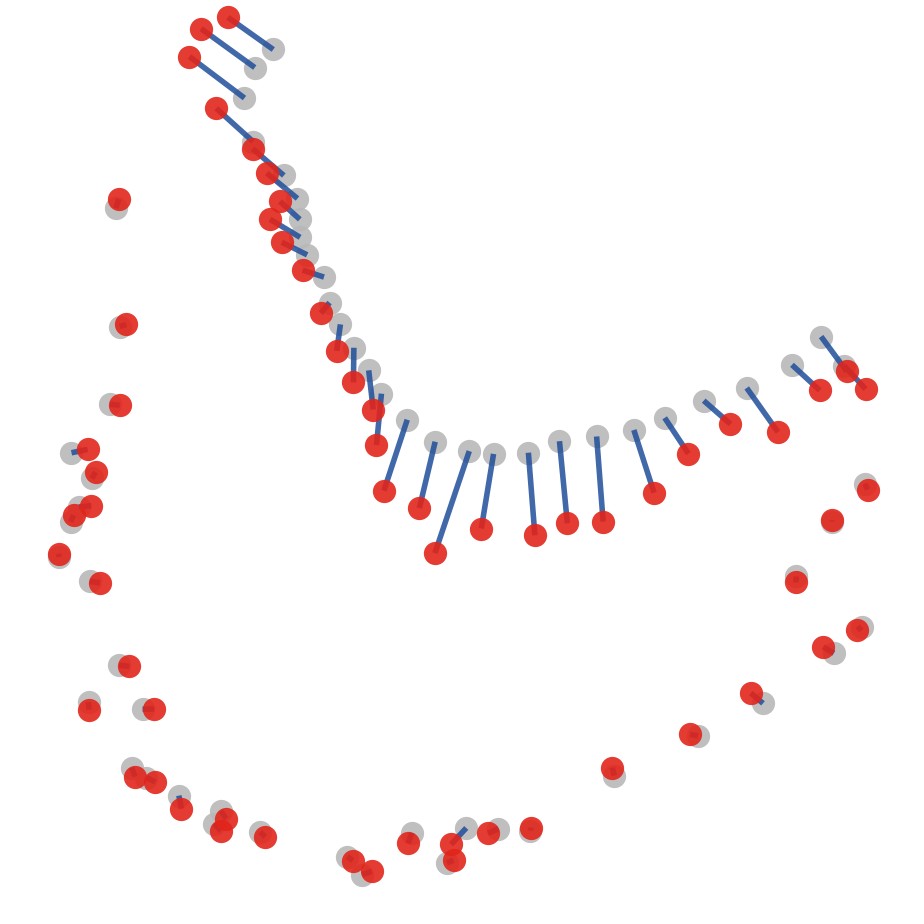}{10.6} \\[0pt]
\maprow{\textbf{\textbf{+ \method}}} & \mapname{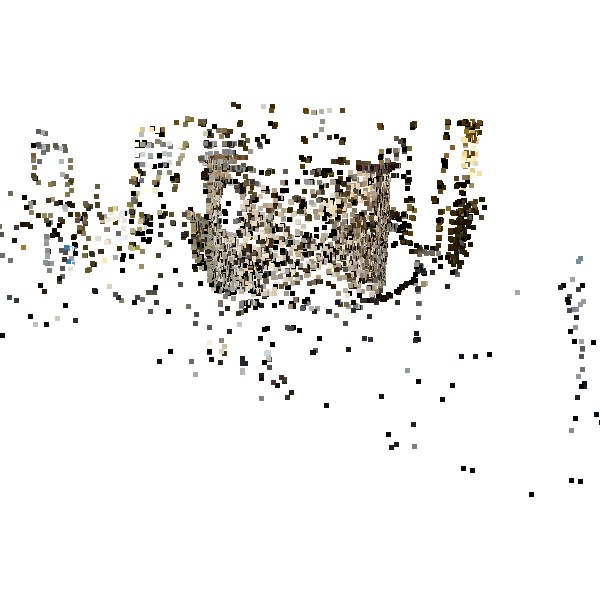}{Rotunda} & \mapaucbf{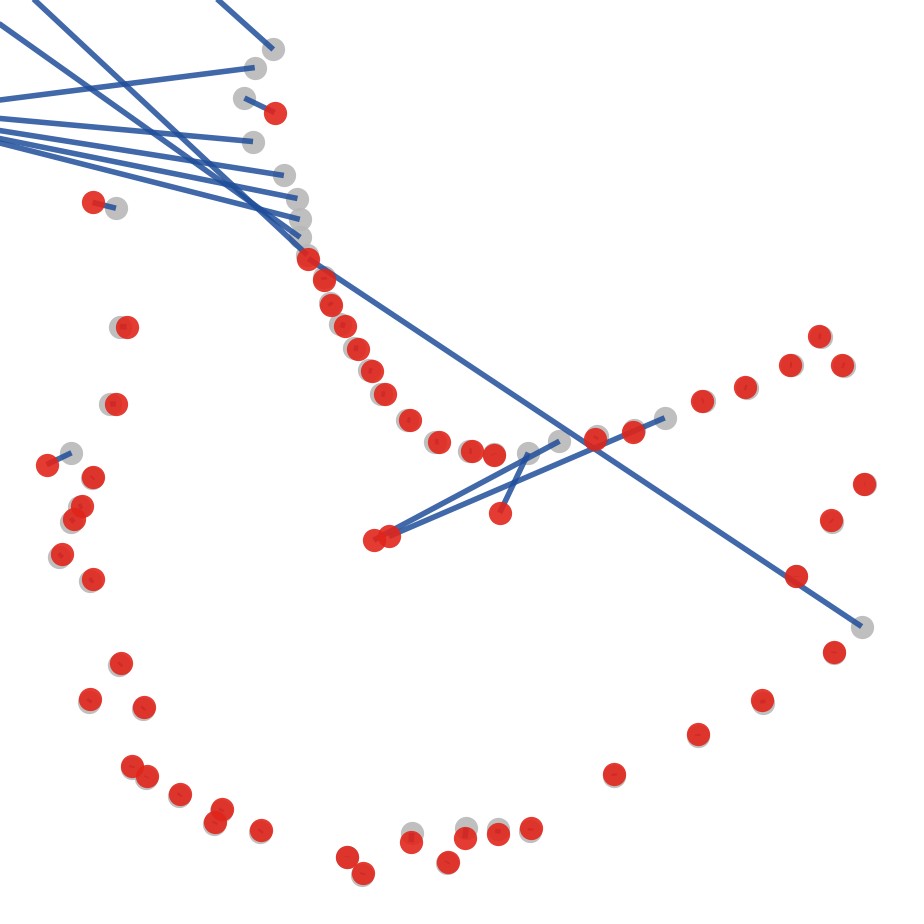}{44.4} & \mapaucbf{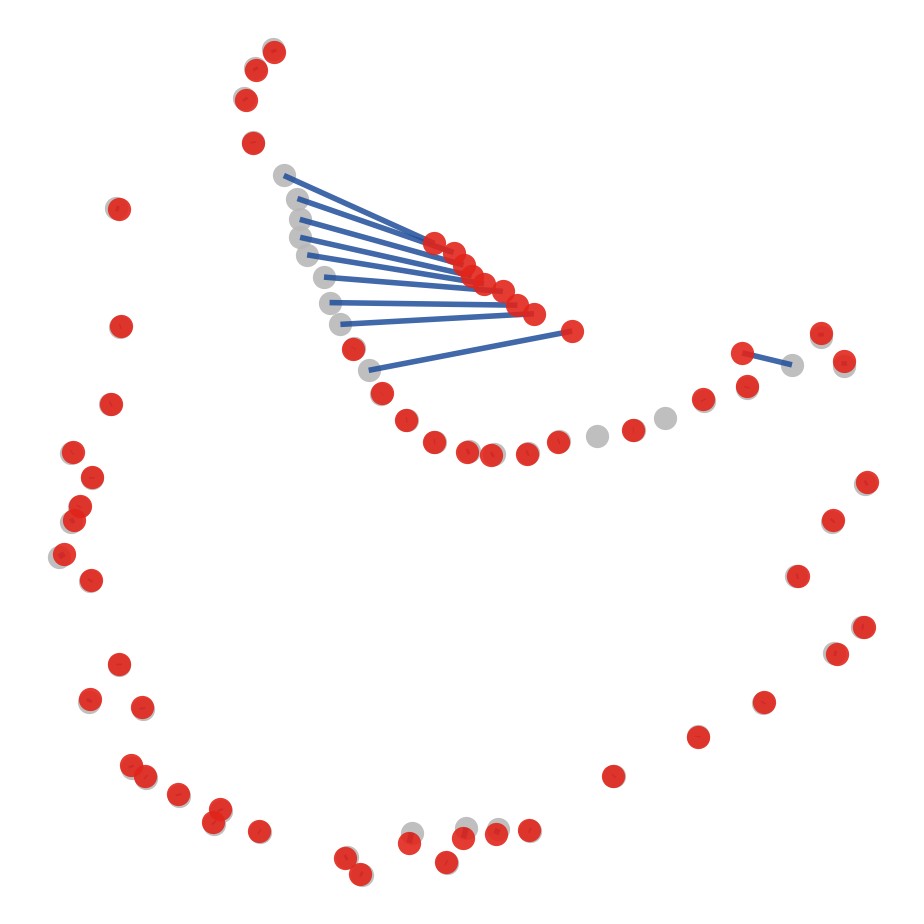}{53.1} & \mapaucbf{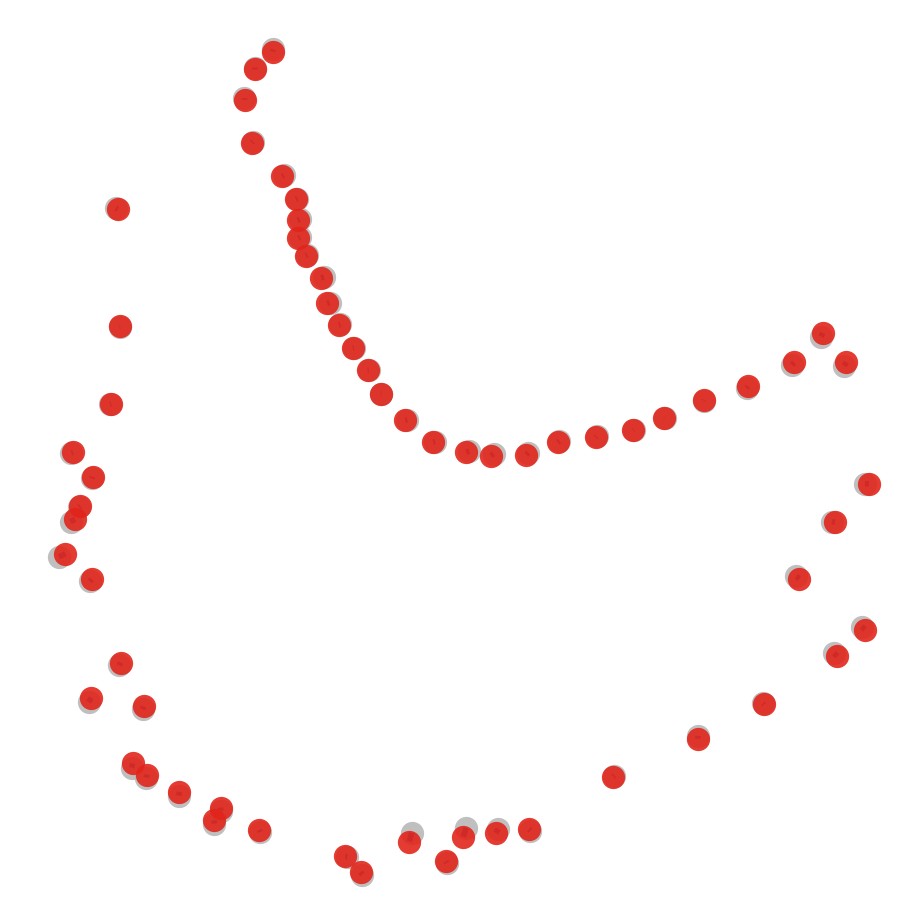}{83.3} & \mapaucbf{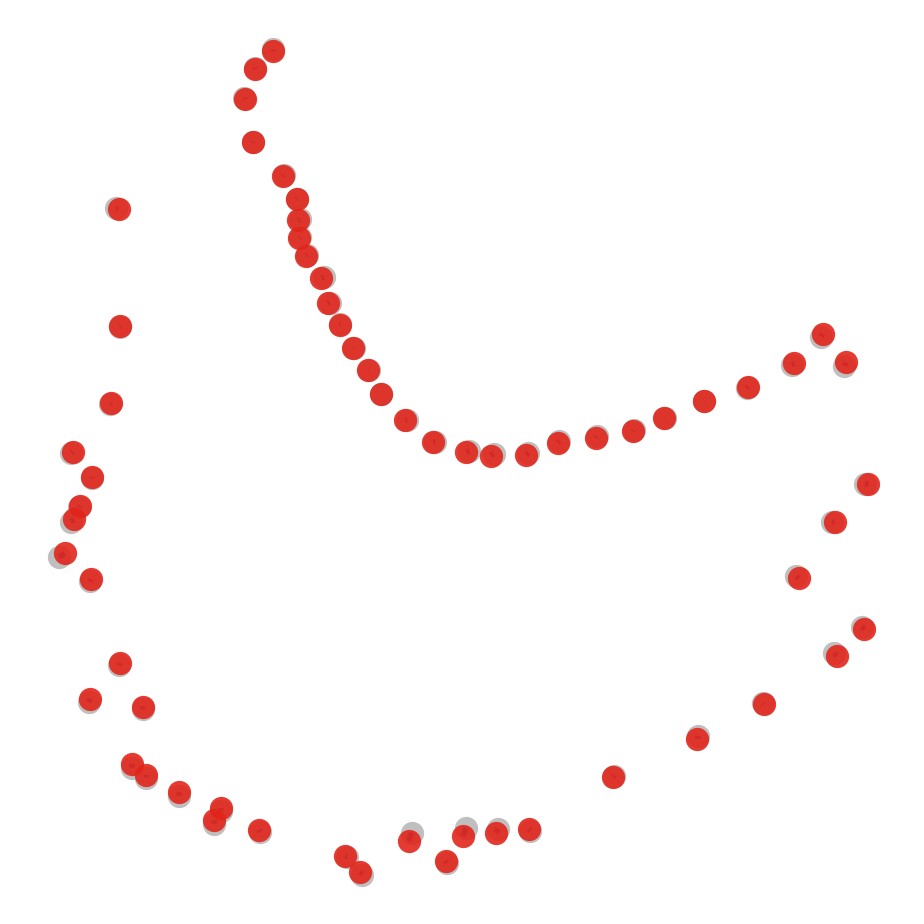}{83.0} & \mapaucbf{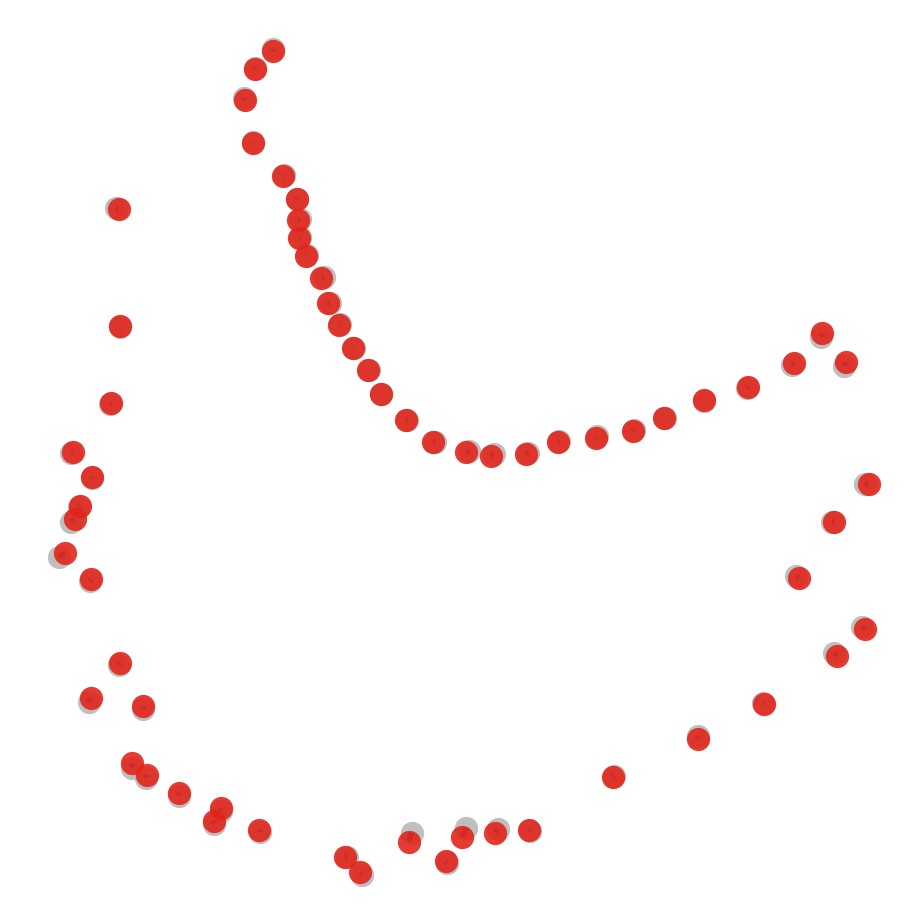}{82.6} \\[4pt]
\end{tabular}
\caption{Camera positions on further ETH3D scenes with lens distortion and on Rotunda scene: Initialization cameras (upper row) and after \method (lower row), each joined to its ground-truth position by a blue line.}
\label{fig:distorted_map_b}
\end{figure*}

\ifdefined\imcW\else\newlength{\imcW}\fi\setlength{\imcW}{0.93in}
\begin{figure*}[p]
\def\mappanel#1{{\setlength{\fboxsep}{0pt}\setlength{\fboxrule}{0.3pt}\fcolorbox{black!20}{white}{\includegraphics[width=\dimexpr\imcW-0.6pt\relax]{fig/imc_map/#1}}}}
\def\mapauc#1#2{\mappanel{#1}\llap{\raisebox{1.2pt}{\setlength{\fboxsep}{0.8pt}\colorbox{white}{\tiny AUC@3$^\circ$ #2}}\hspace{1.2pt}}}
\def\mapaucbf#1#2{\mappanel{#1}\llap{\raisebox{1.2pt}{\setlength{\fboxsep}{0.8pt}\colorbox{white}{\tiny\bfseries AUC@3$^\circ$ #2}}\hspace{1.2pt}}}
\def\maprow#1{\rotatebox{90}{\scriptsize #1}}
\def\mapname#1#2{\raisebox{0pt}[0pt][0pt]{\parbox{\imcW}{\centering\includegraphics[width=0.66\imcW]{fig/imc_map/#1}\\[1pt]\footnotesize\bfseries #2}}}  
\centering
\setlength{\tabcolsep}{0.8pt}
\renewcommand{\arraystretch}{0.7}
\begin{tabular}{@{}>{\centering\arraybackslash}m{0.9em} >{\centering\arraybackslash}m{\imcW}@{\hspace{4pt}} *{6}{>{\centering\arraybackslash}m{\imcW}} @{}}
 & \small\bfseries GT & \small\bfseries COLMAP & \small\bfseries GGPT & \small\bfseries DGSfM & \small\bfseries Solv3R & \small\bfseries DVLT & \small\bfseries Pi3X (calib.) \\[2pt]

\maprow{Initialization} & \mappanel{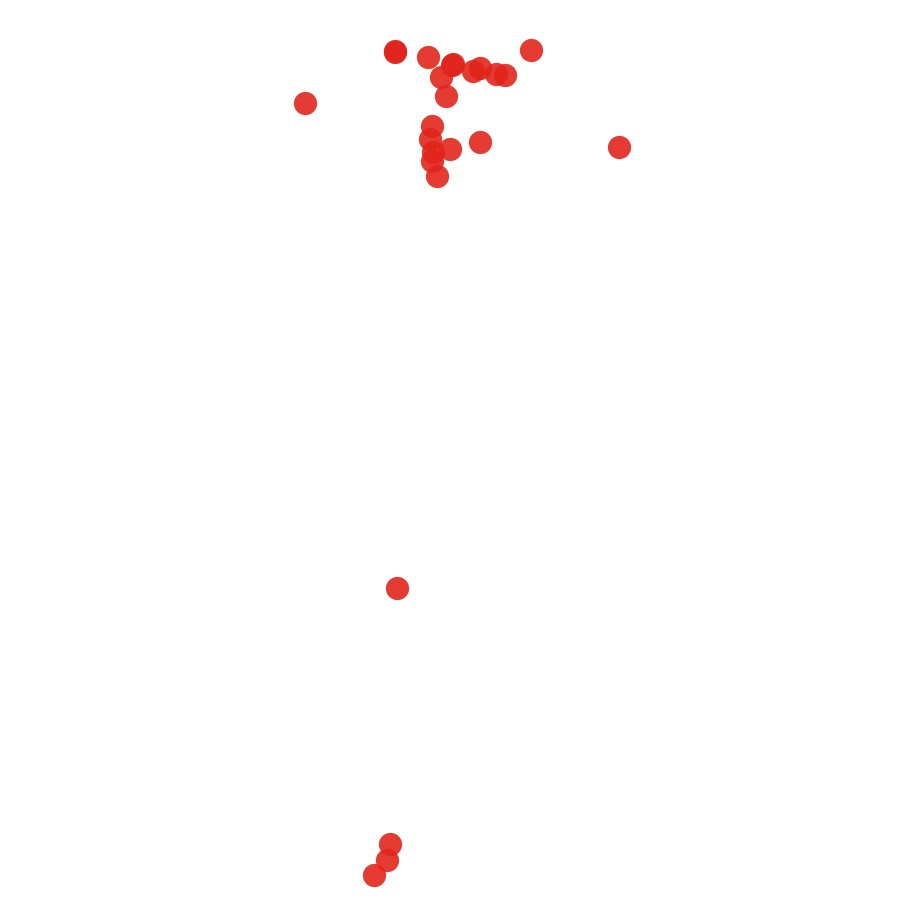} & \mapauc{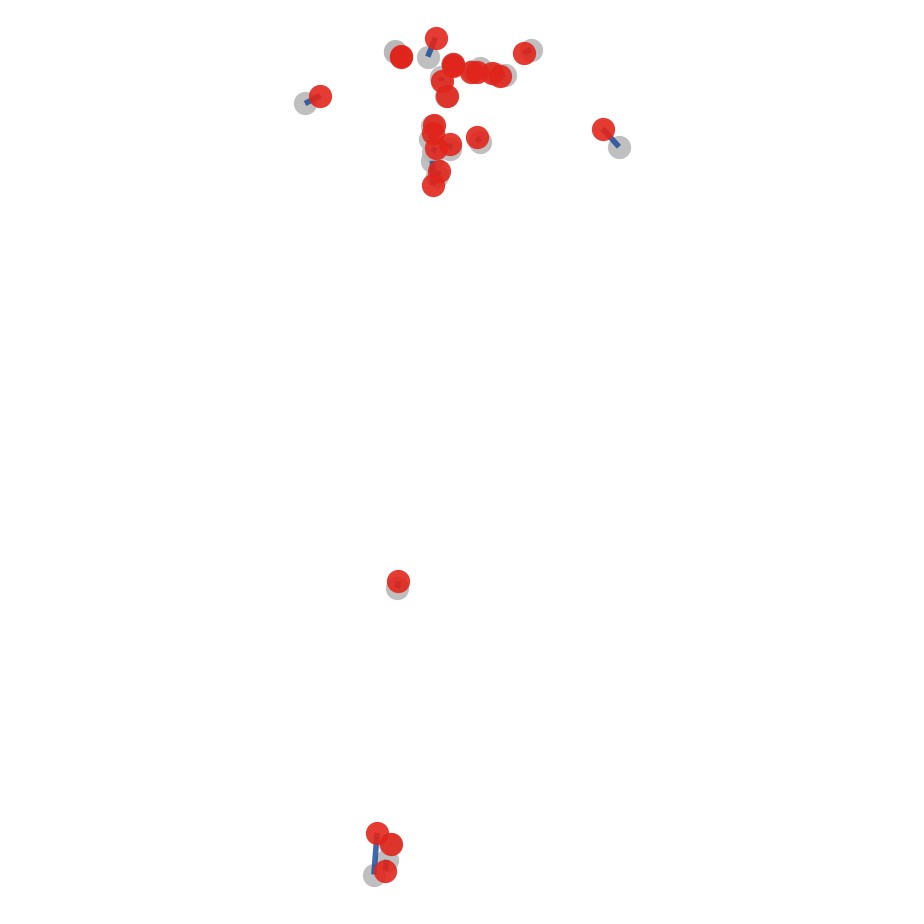}{47.3} & \mapauc{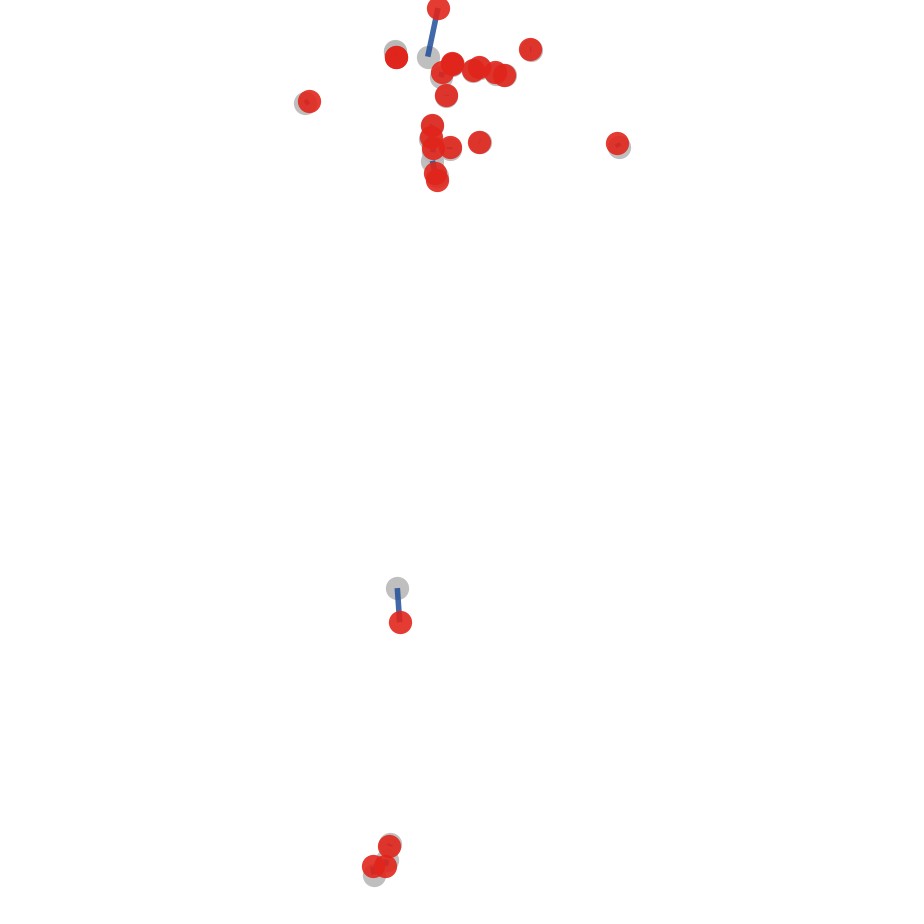}{49.6} & \mapauc{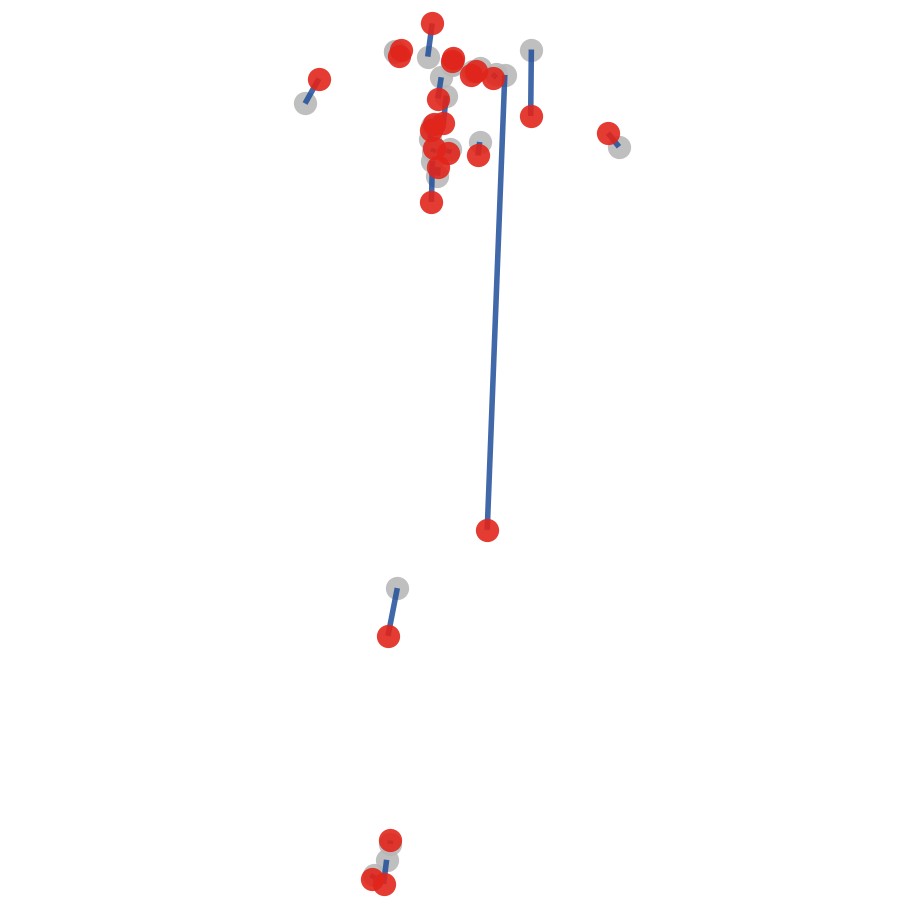}{22.8} & \mapauc{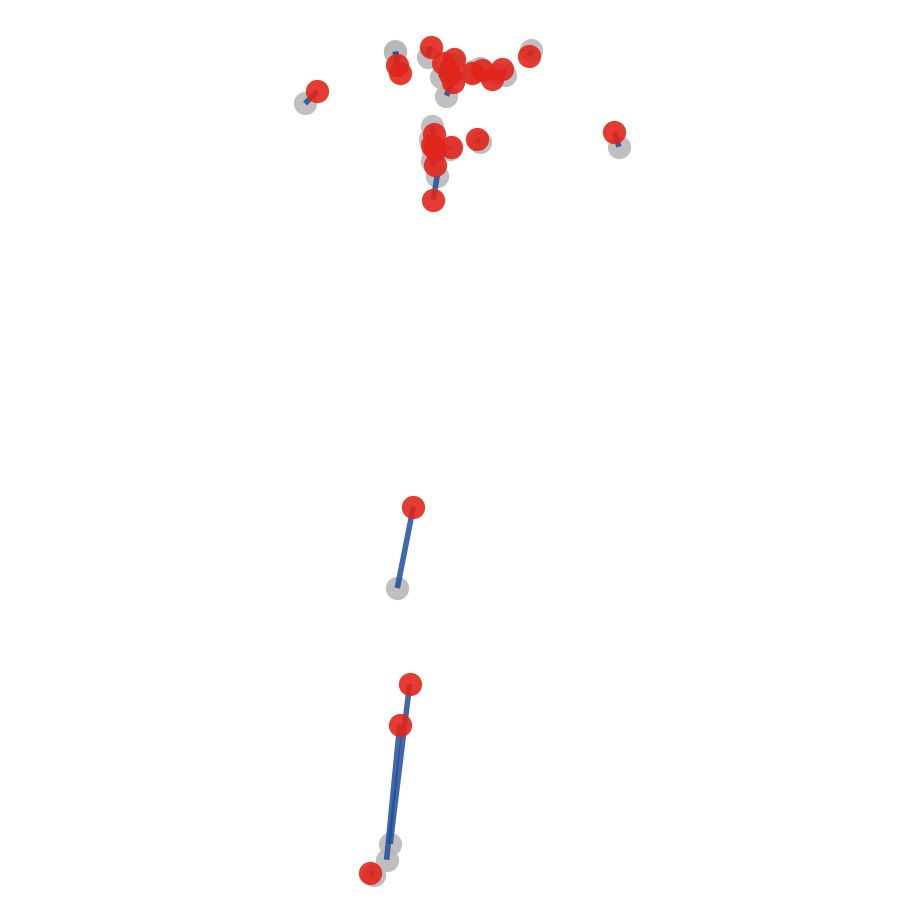}{15.8} & \mapauc{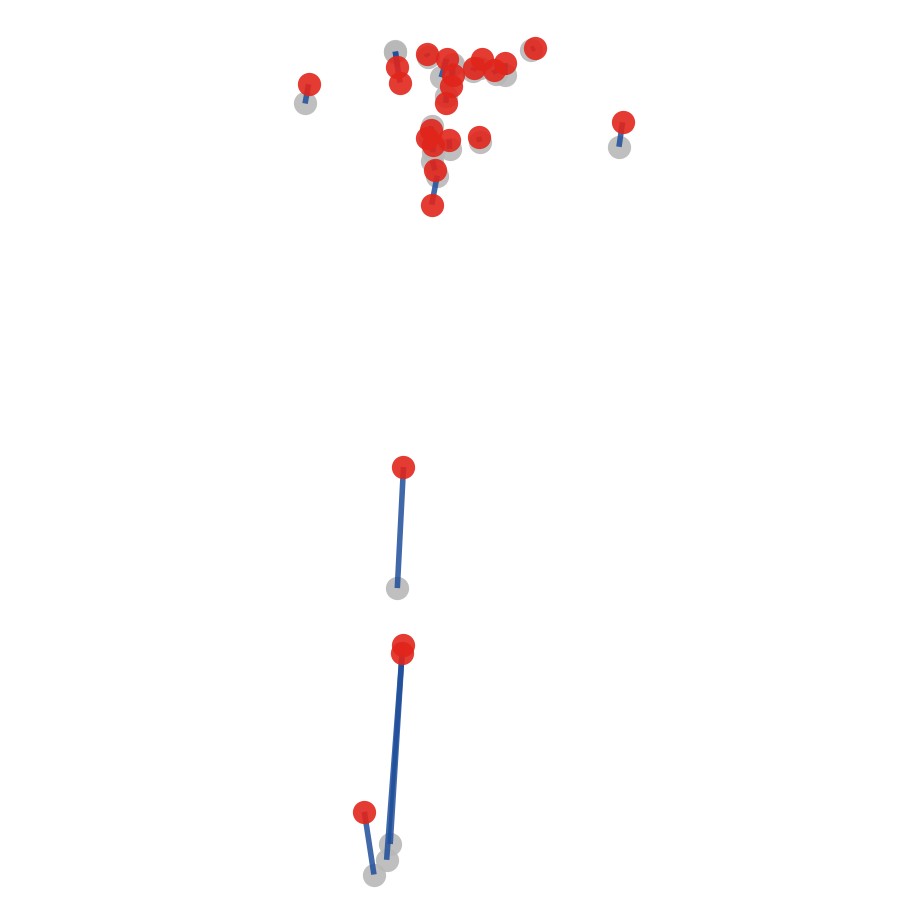}{8.9} & \mapauc{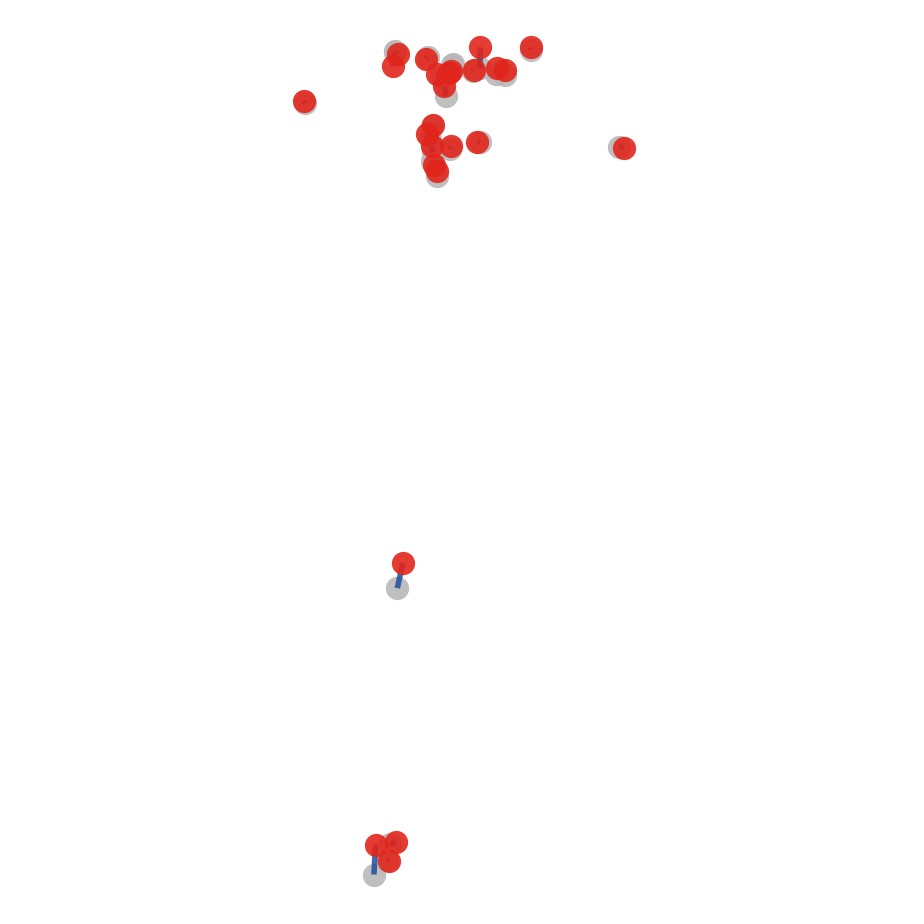}{30.6} \\[0pt]
\maprow{\textbf{+ \method}} & \mapname{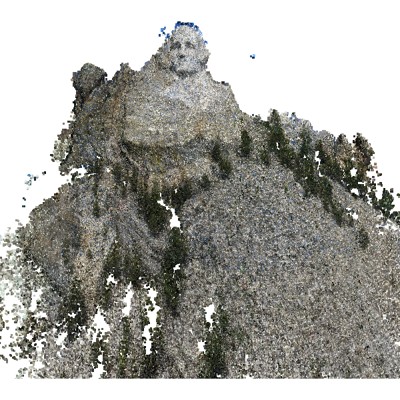}{\shortstack{Mt.\ Rushmore\\bag 10}} & \mapaucbf{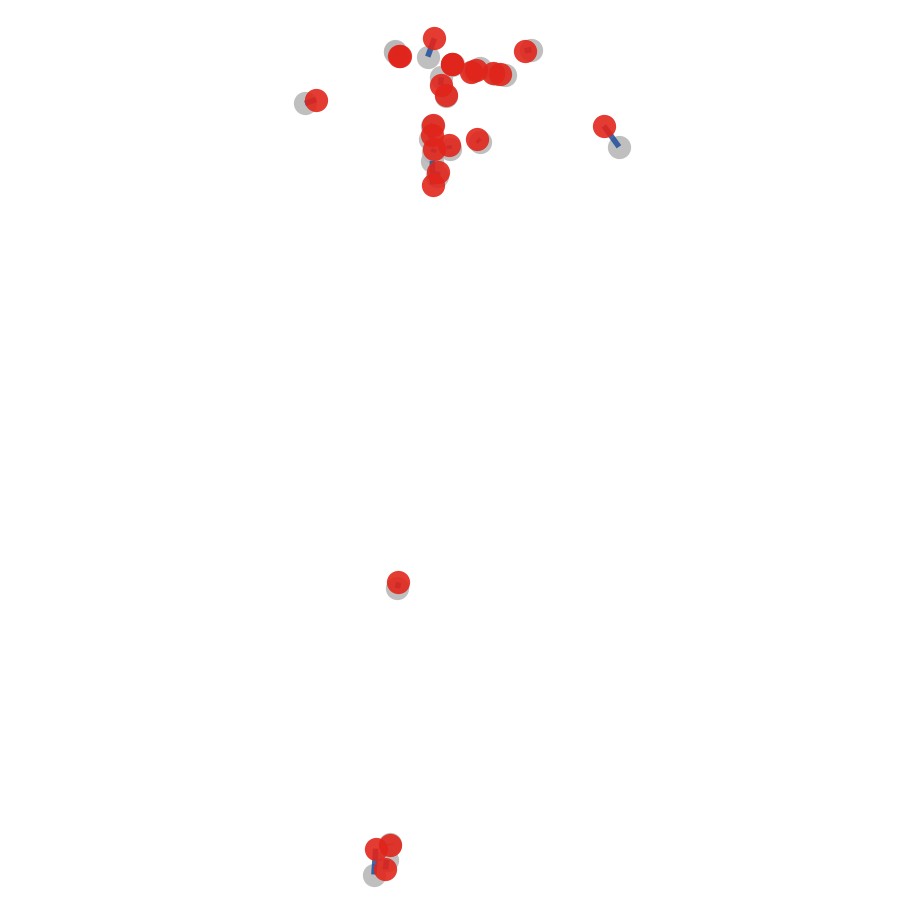}{50.6} & \mapaucbf{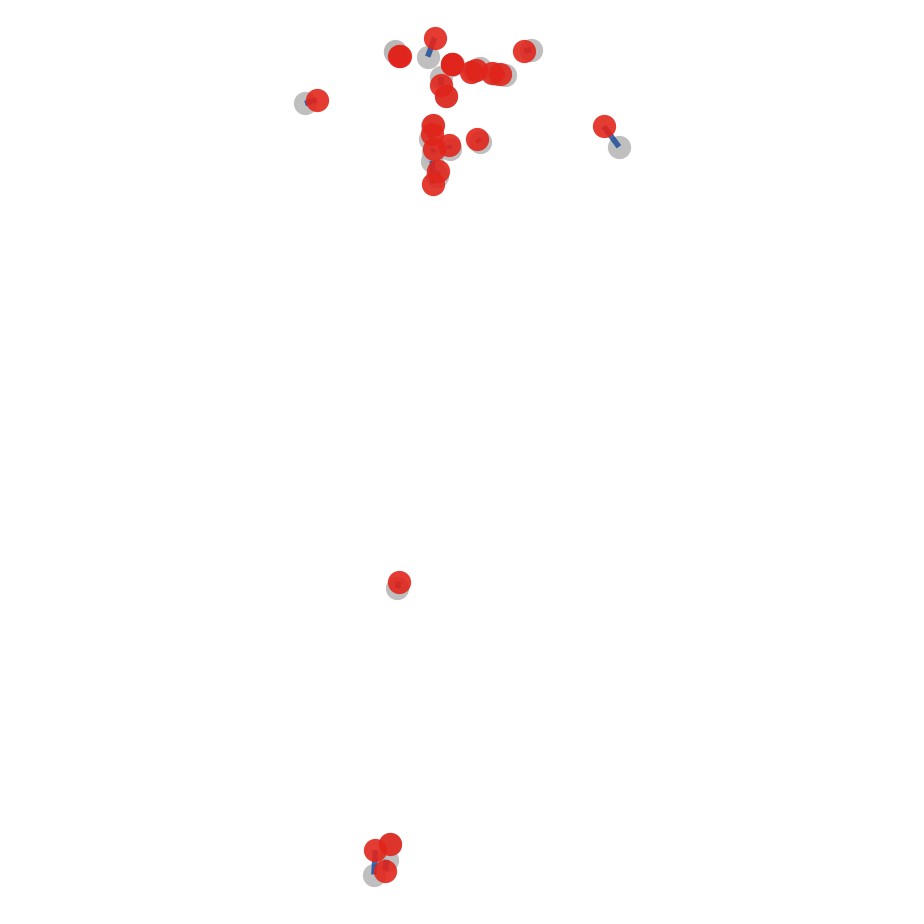}{50.4} & \mapaucbf{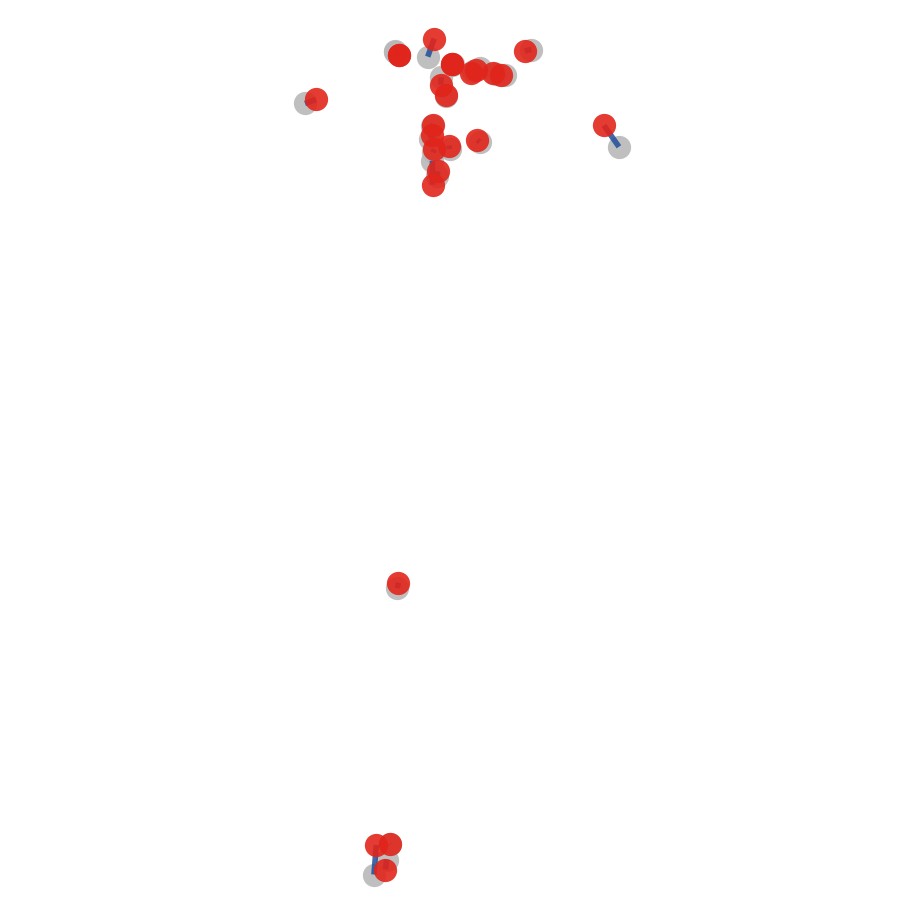}{51.3} & \mapaucbf{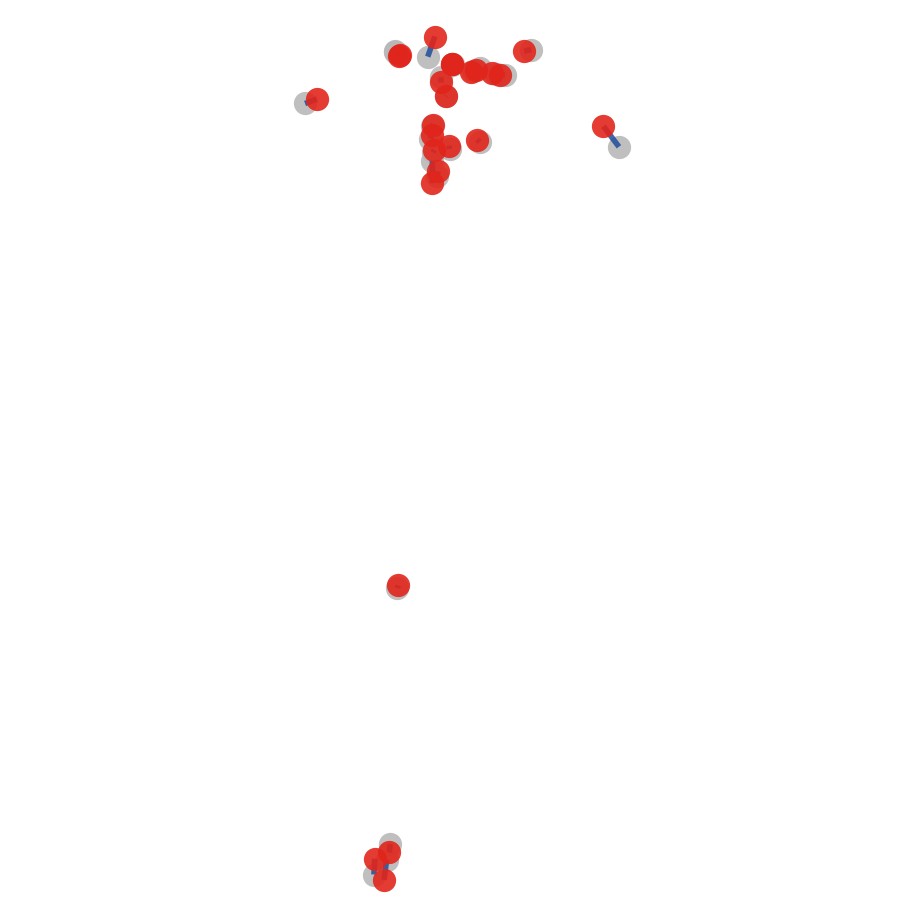}{48.5} & \mapaucbf{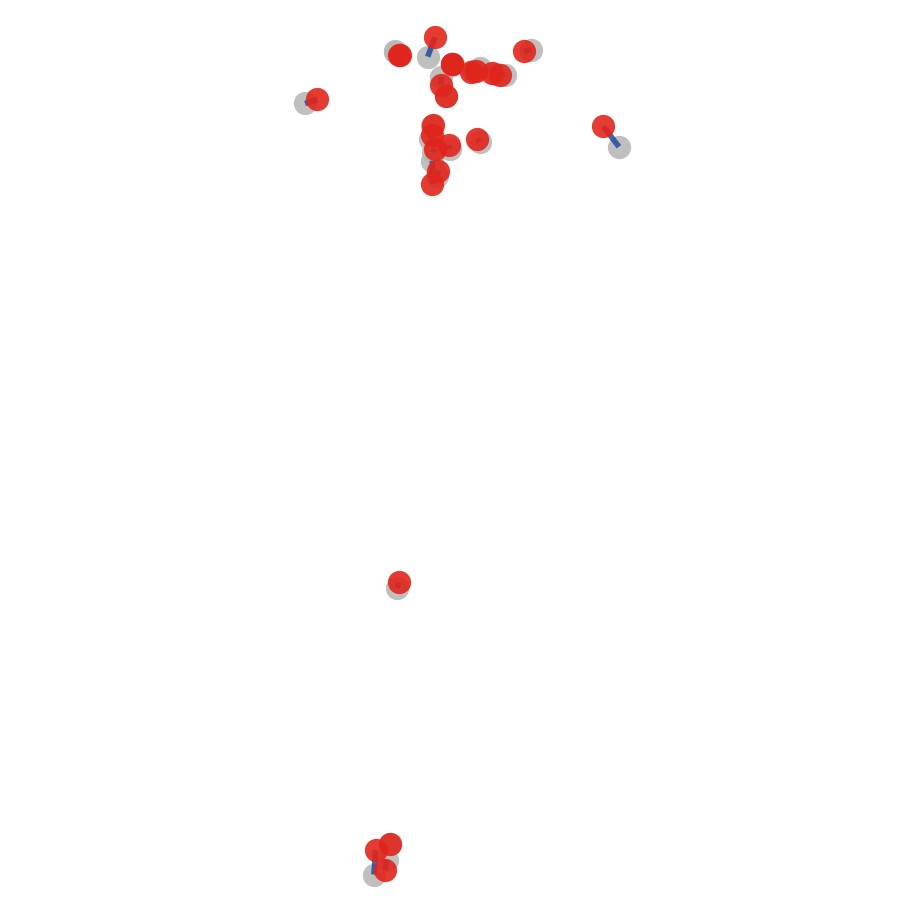}{50.7} & \mapaucbf{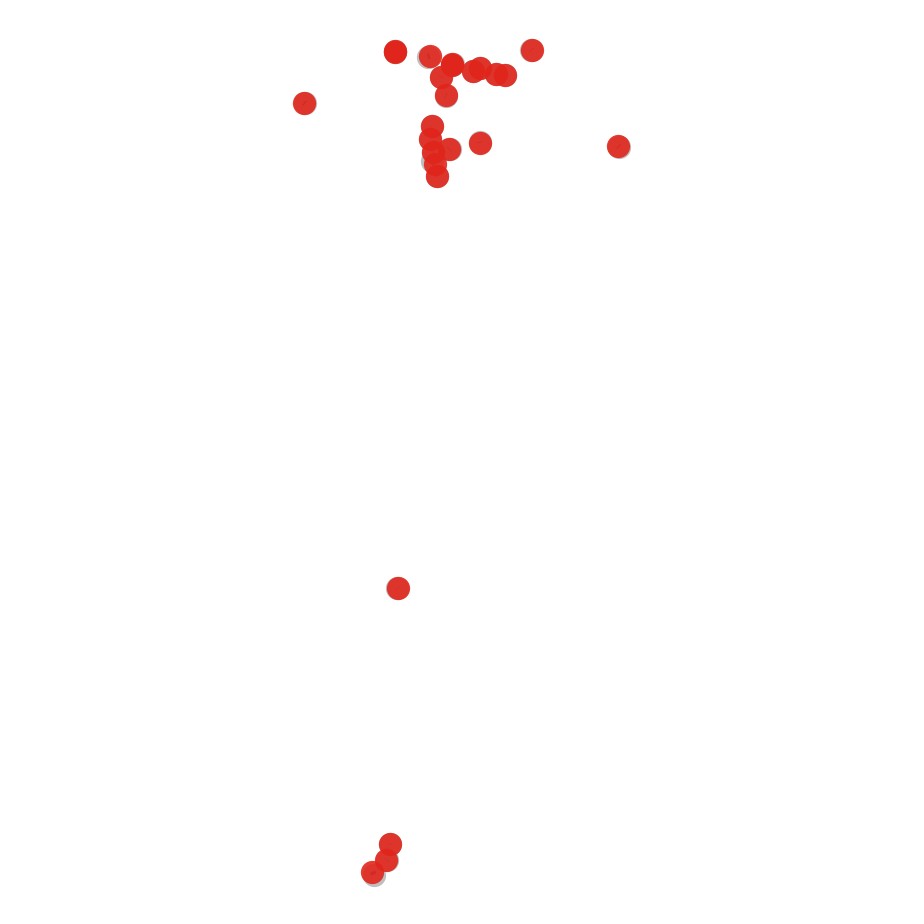}{79.1} \\[4pt]
\maprow{Initialization} & \mappanel{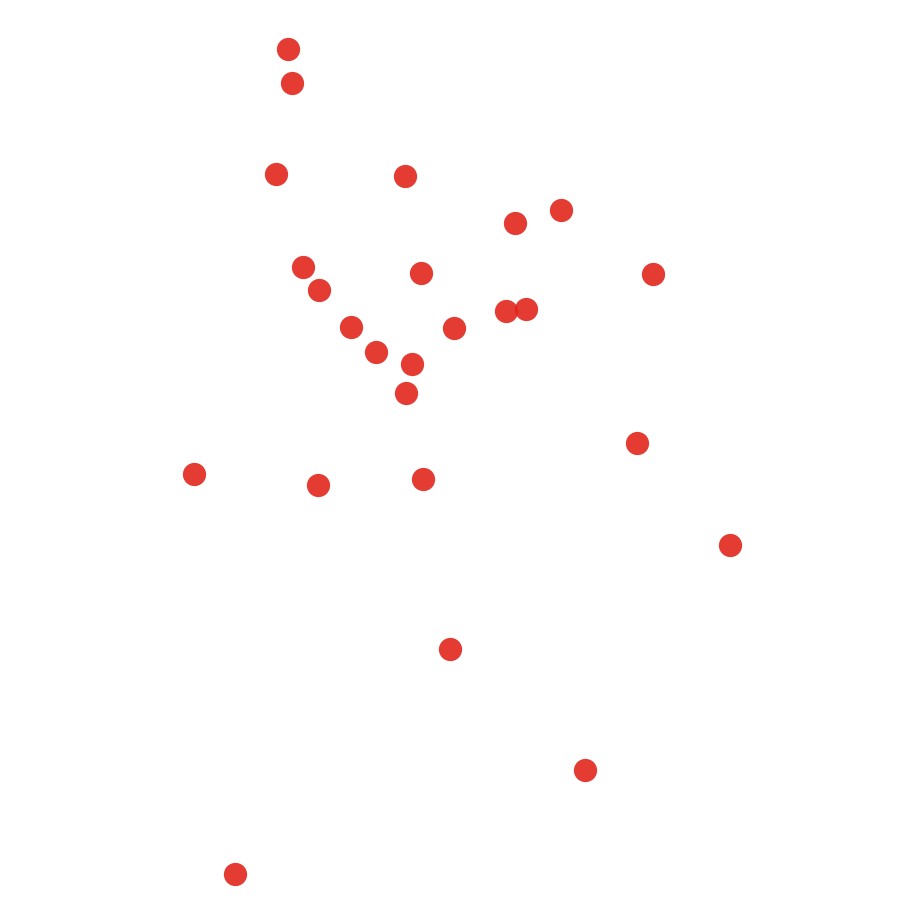} & \mapauc{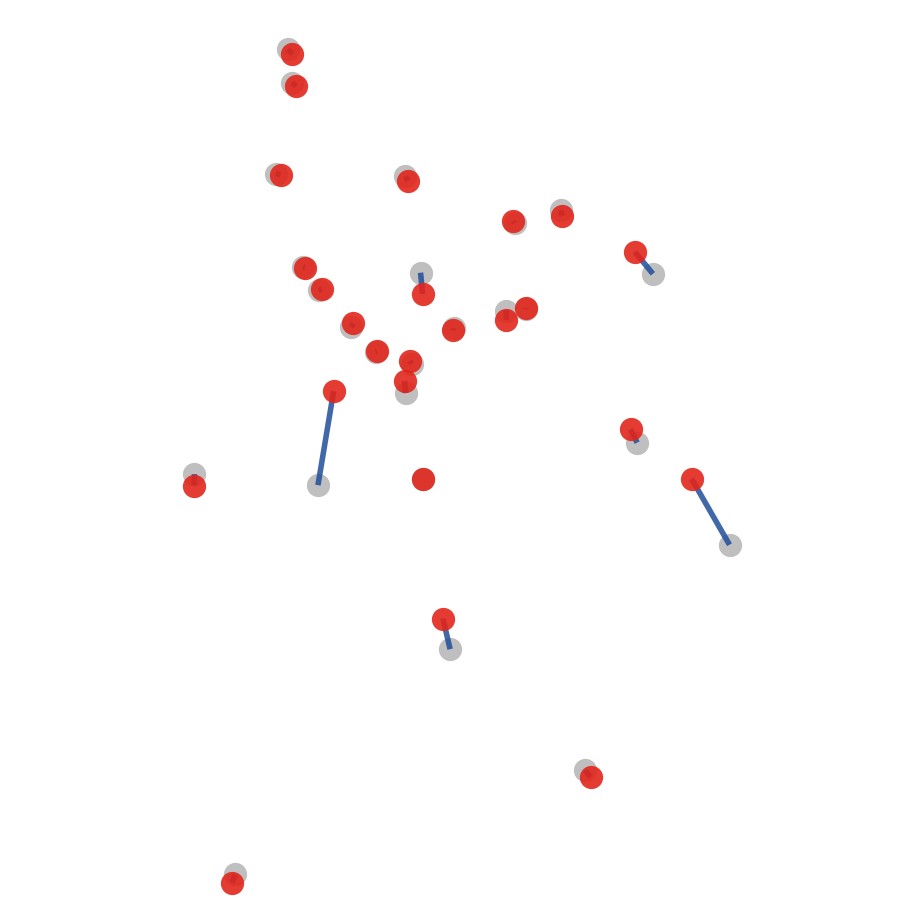}{42.6} & \mapauc{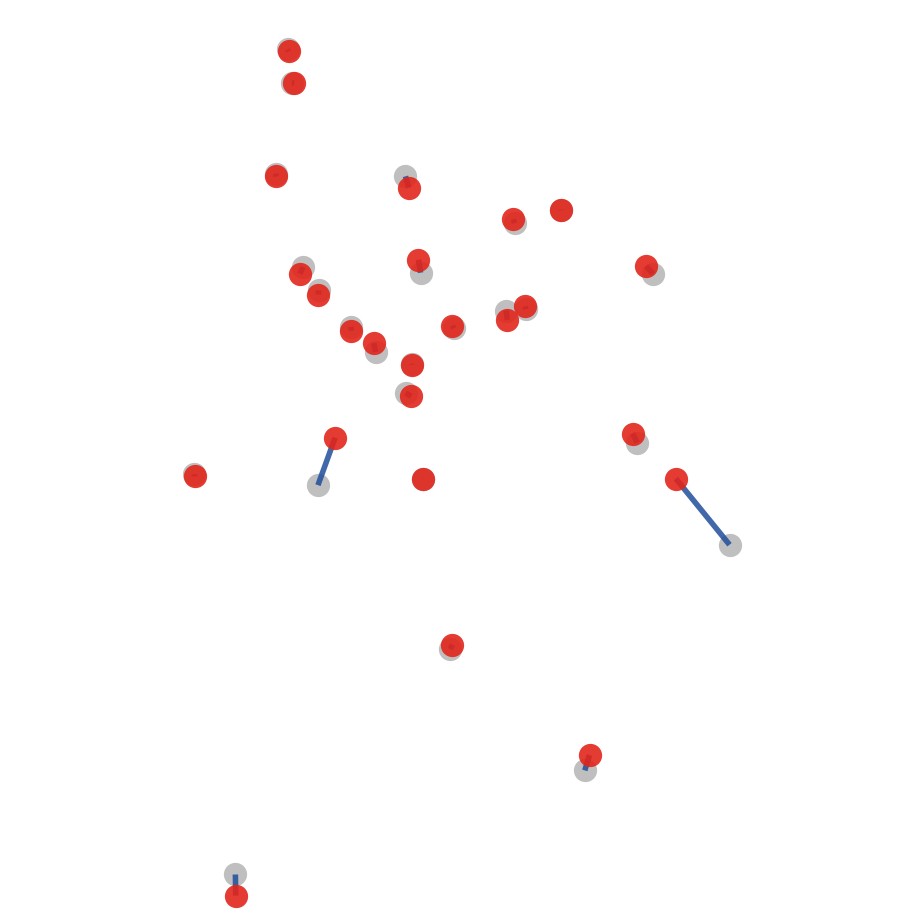}{35.4} & \mapauc{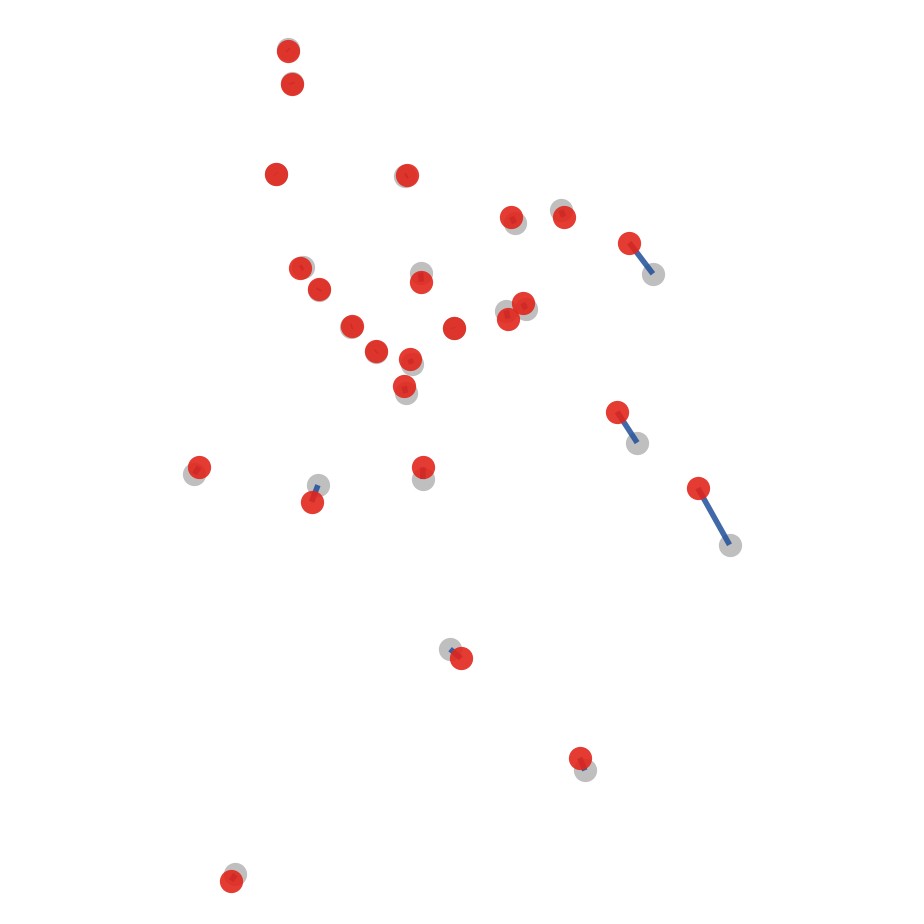}{43.5} & \mapauc{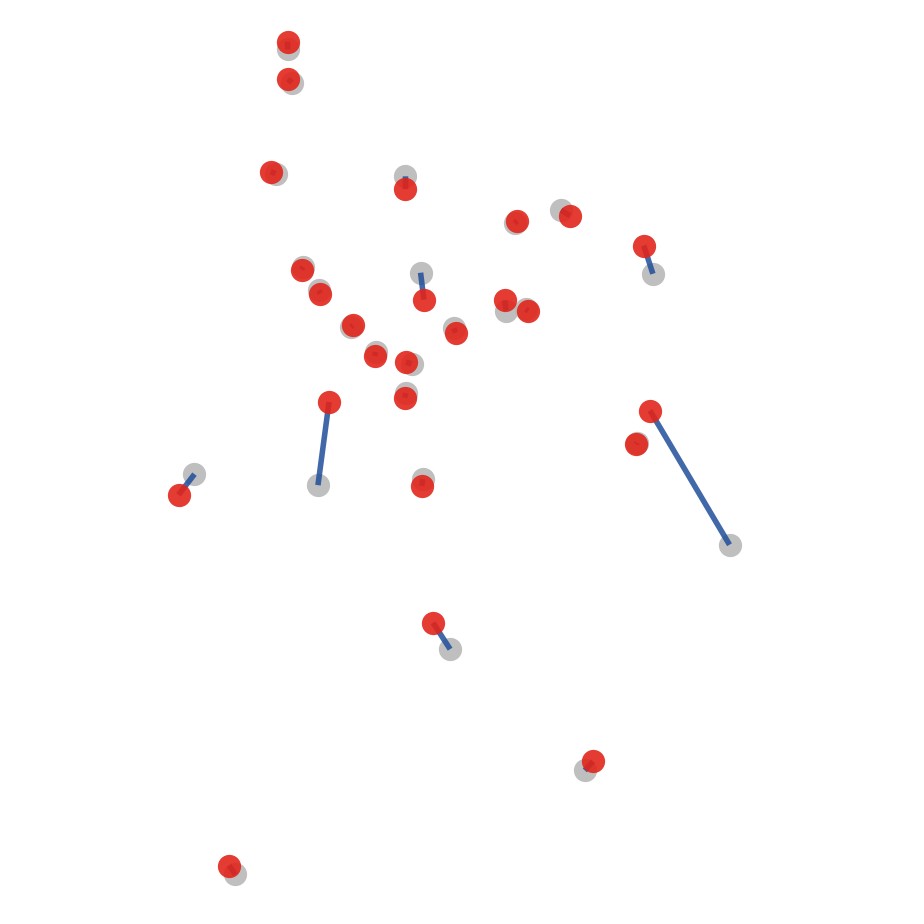}{28.9} & \mapauc{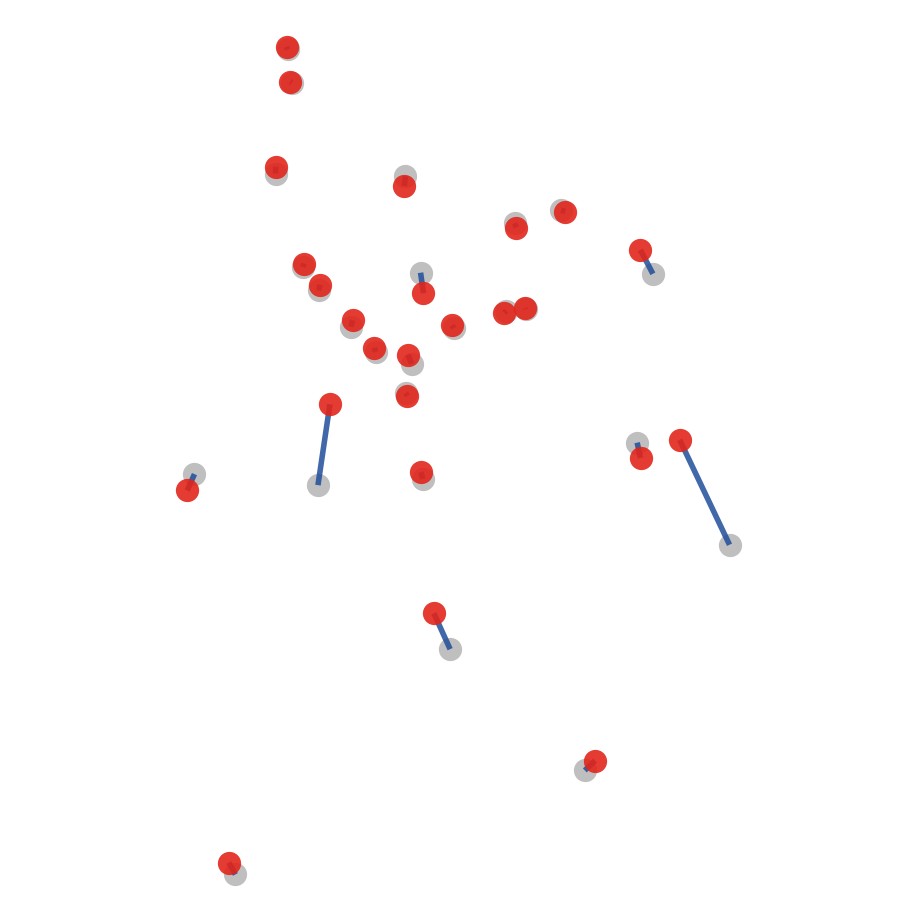}{29.8} & \mapauc{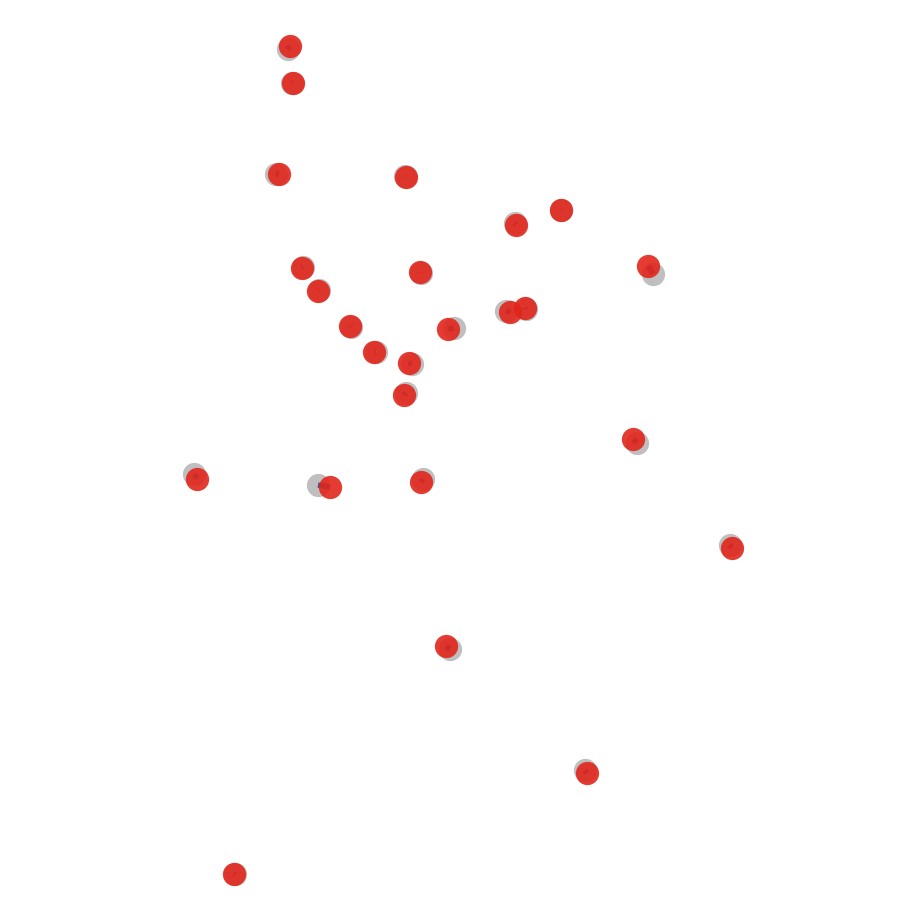}{62.5} \\[0pt]
\maprow{\textbf{+ \method}} & \mapname{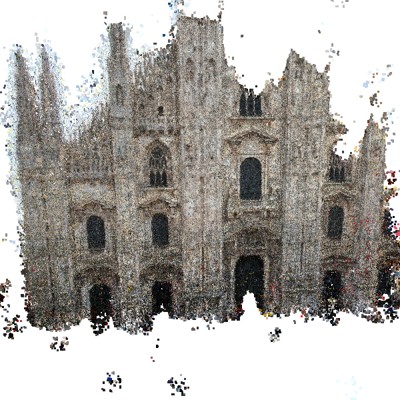}{\shortstack{Milan Cathedral\\bag 22}} & \mapaucbf{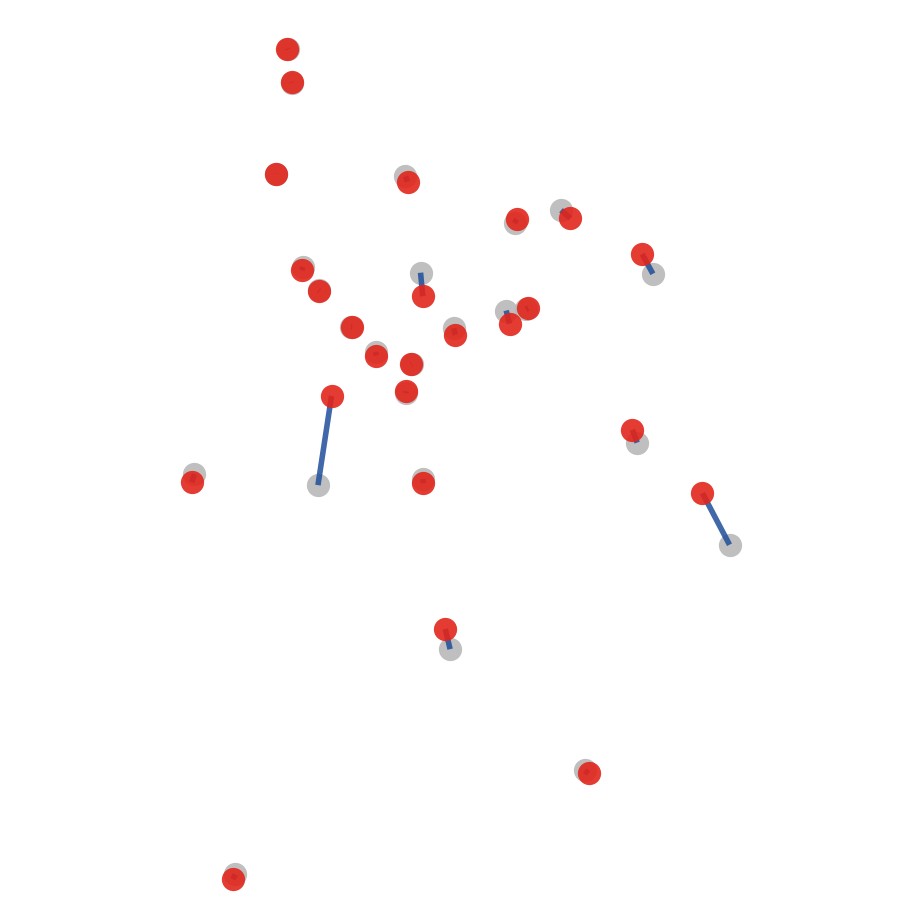}{43.7} & \mapaucbf{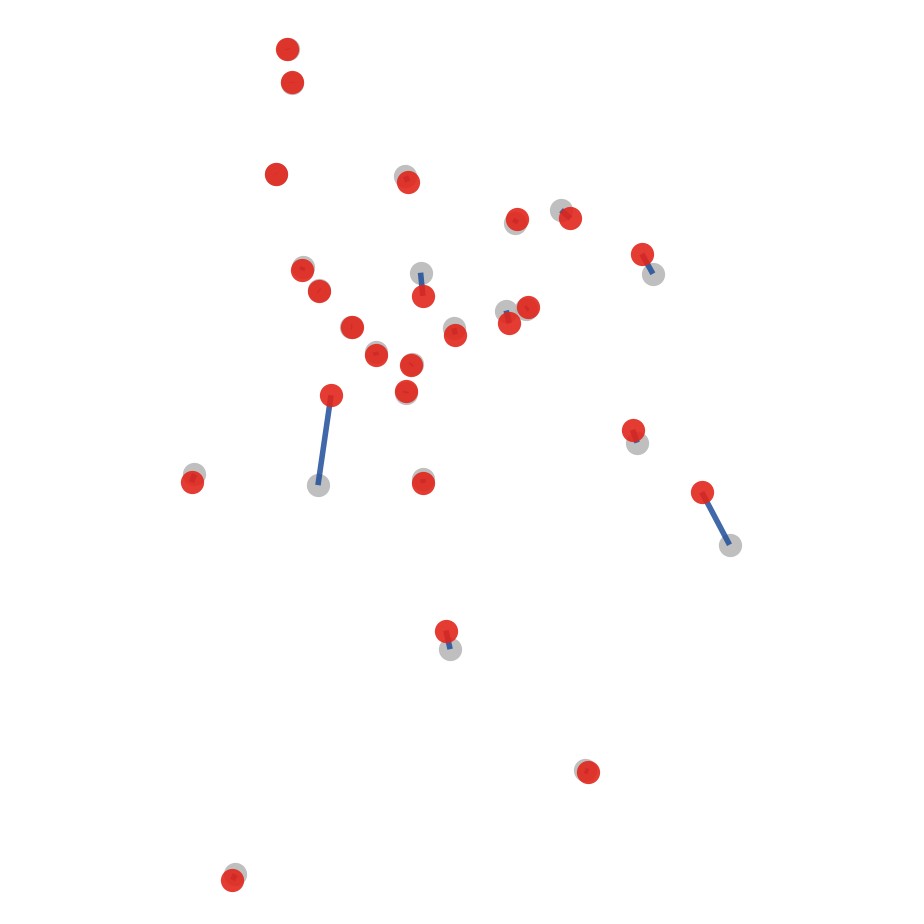}{43.6} & \mapaucbf{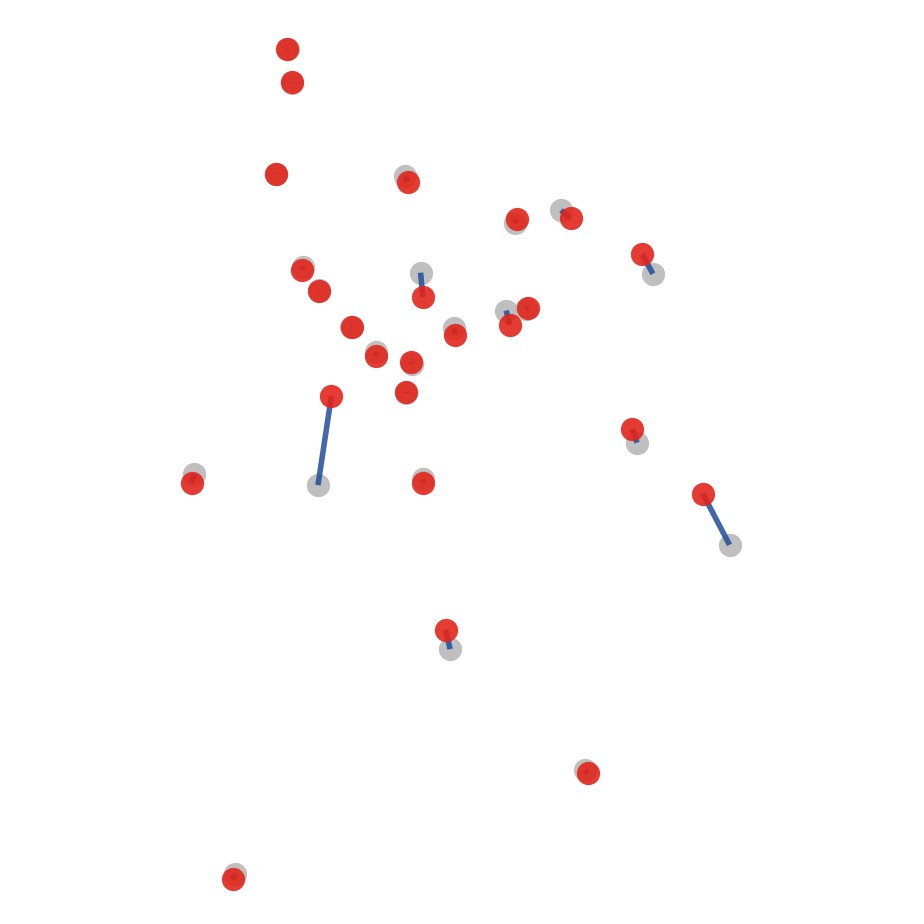}{43.2} & \mapaucbf{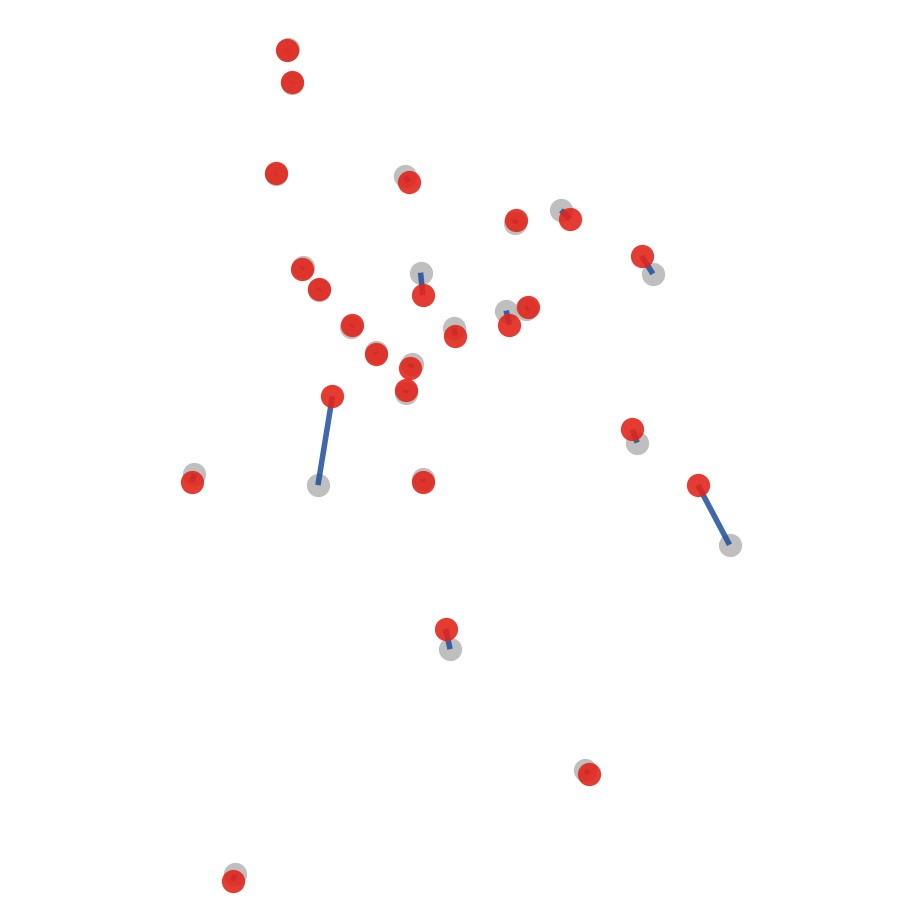}{40.7} & \mapaucbf{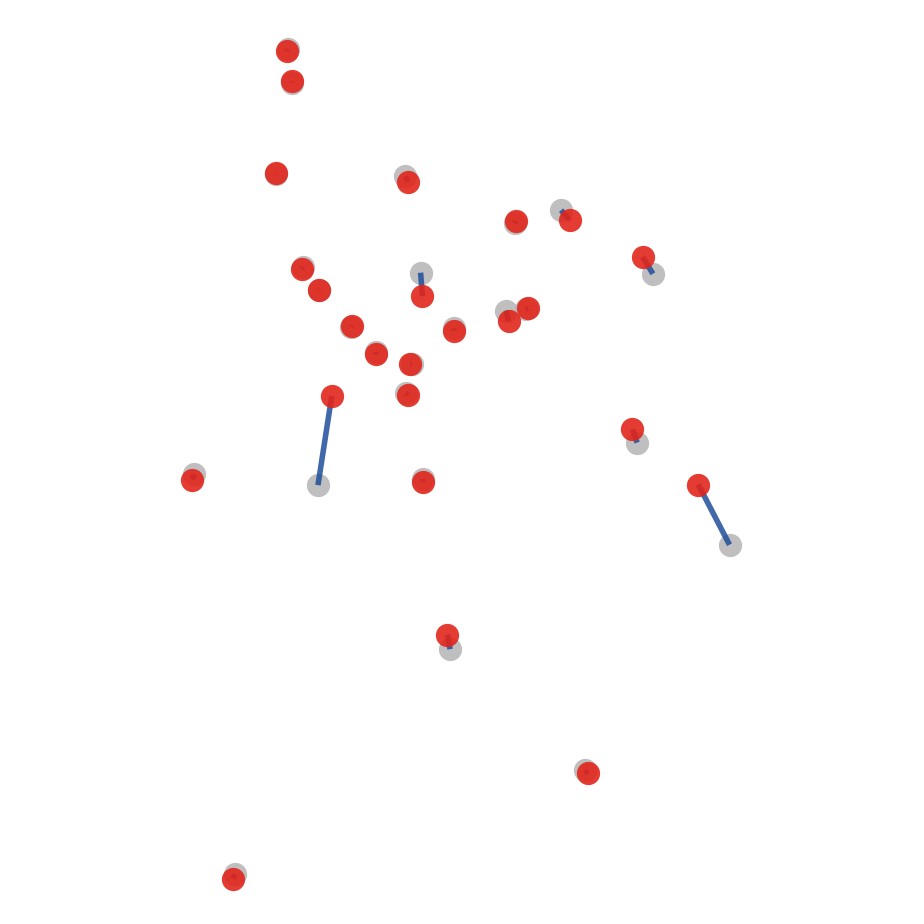}{42.9} & \mapaucbf{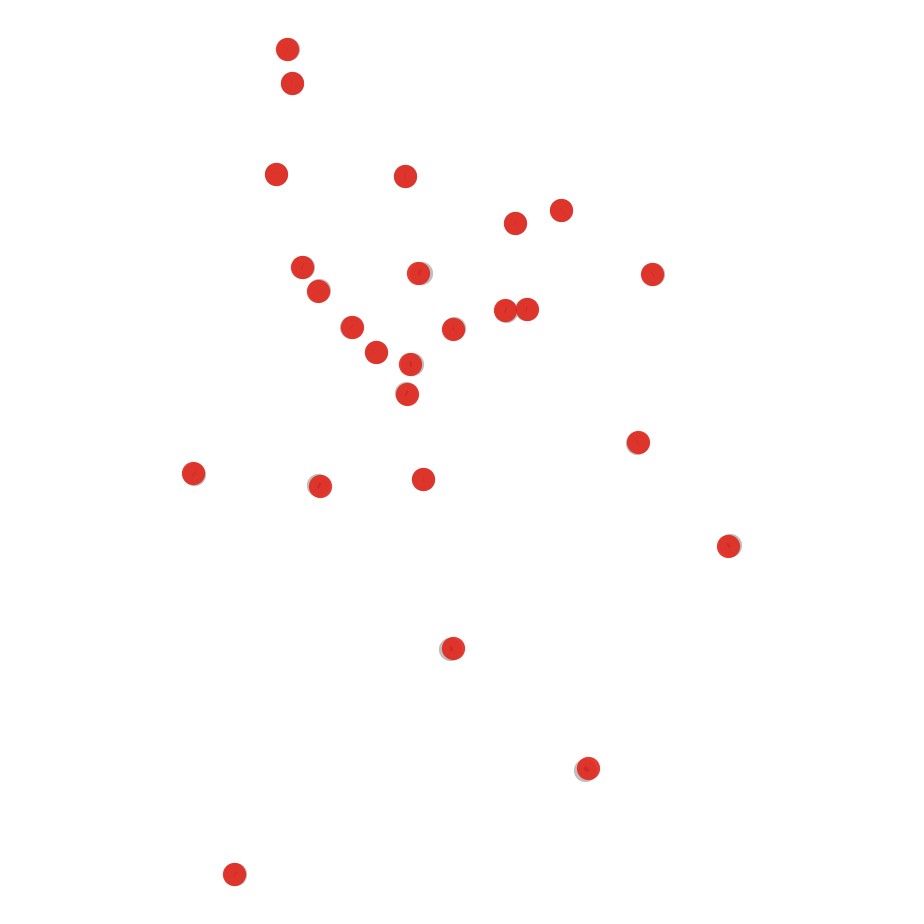}{81.8} \\[4pt]
\maprow{Initialization} & \mappanel{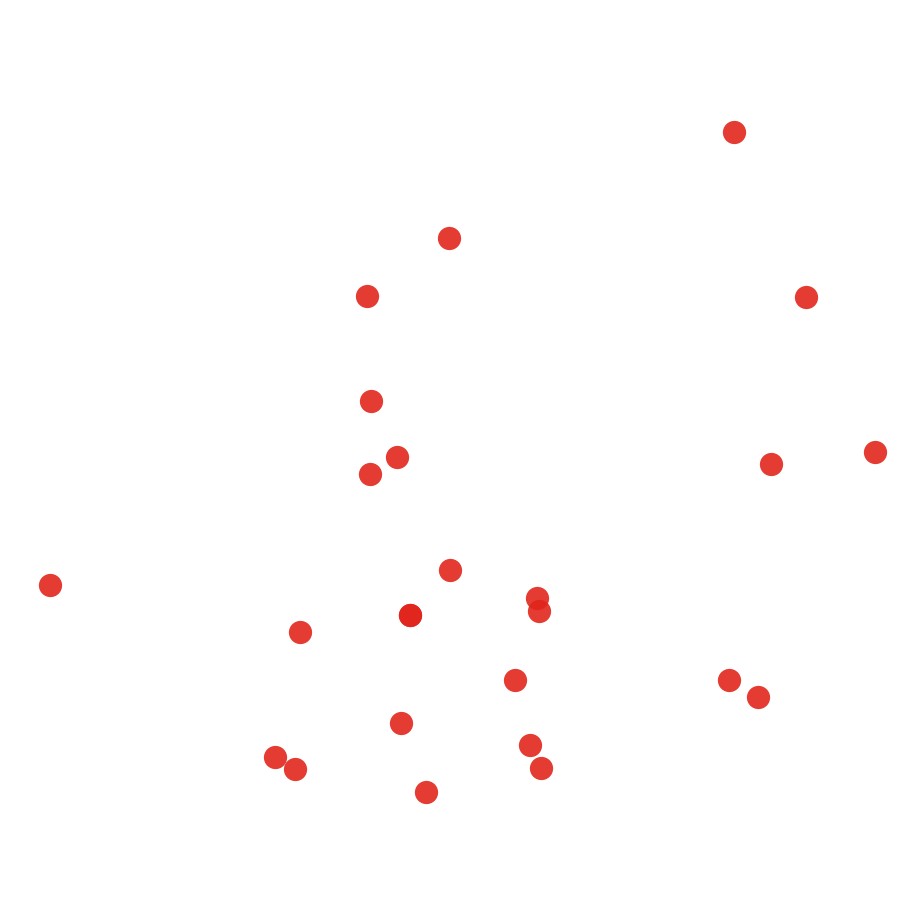} & \mapauc{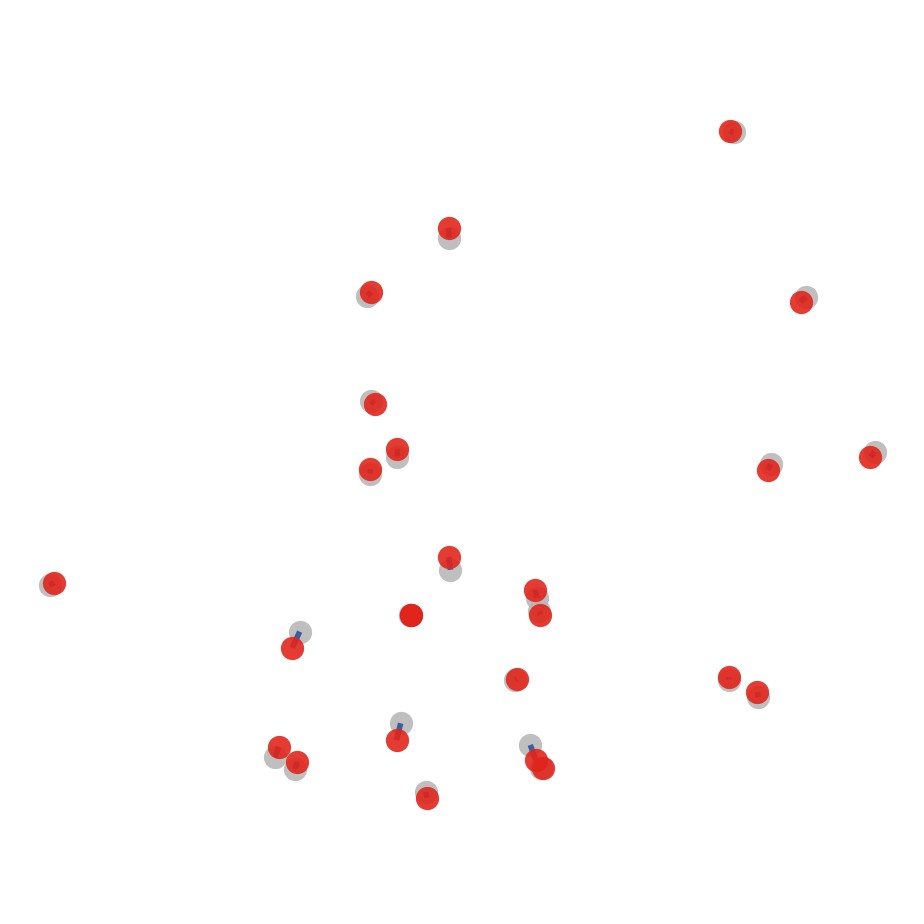}{56.1} & \mapauc{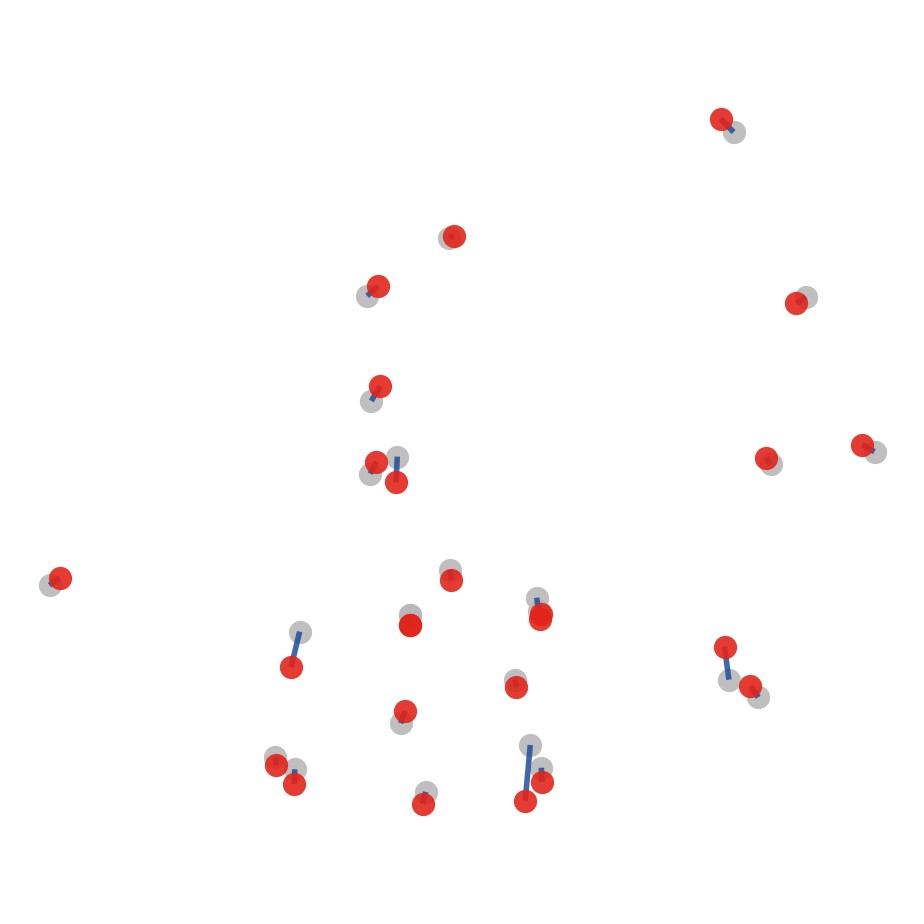}{29.6} & \mapauc{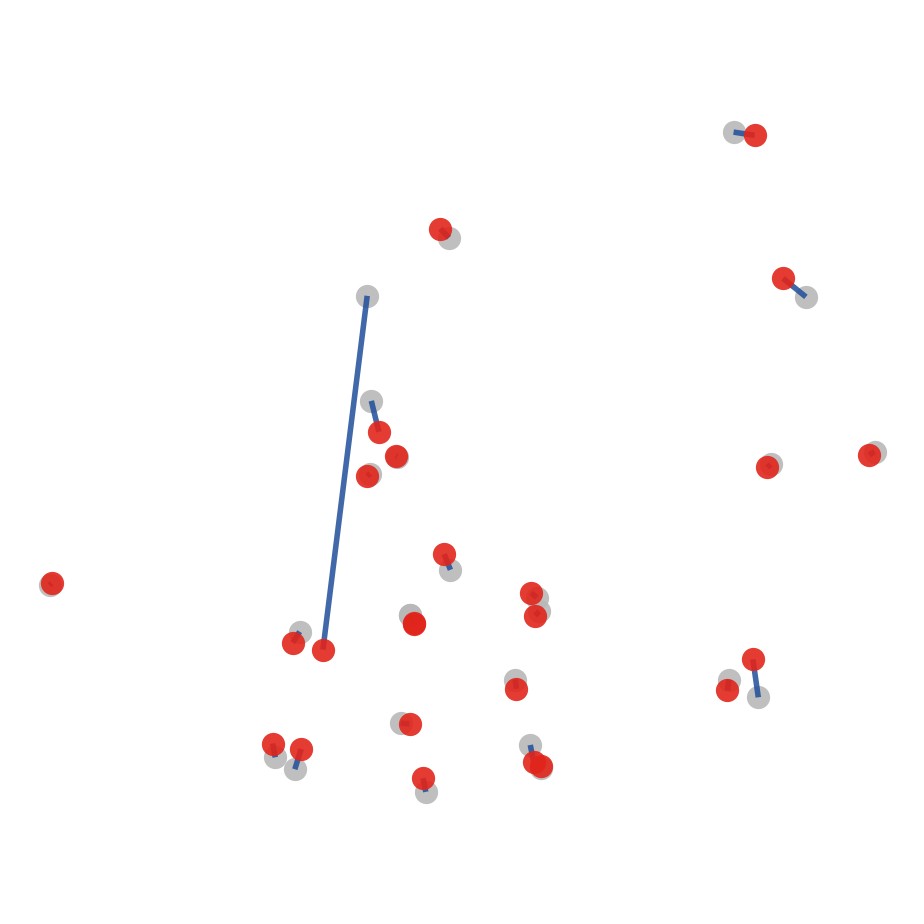}{28.3} & \mapauc{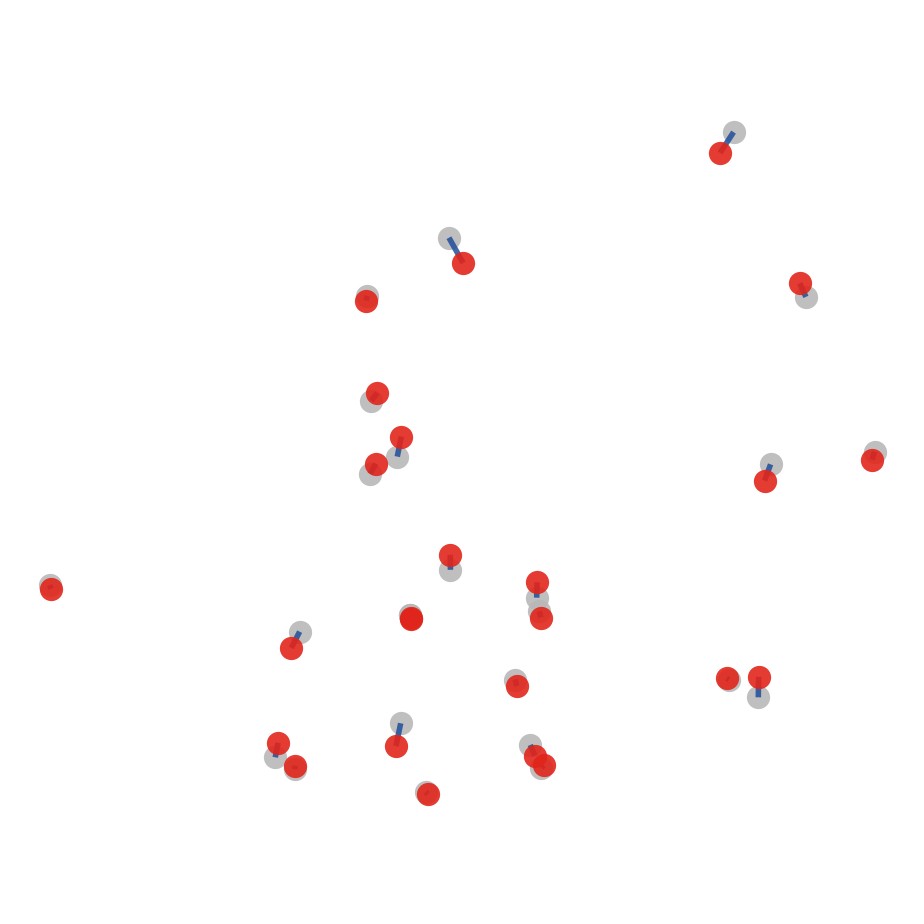}{31.2} & \mapauc{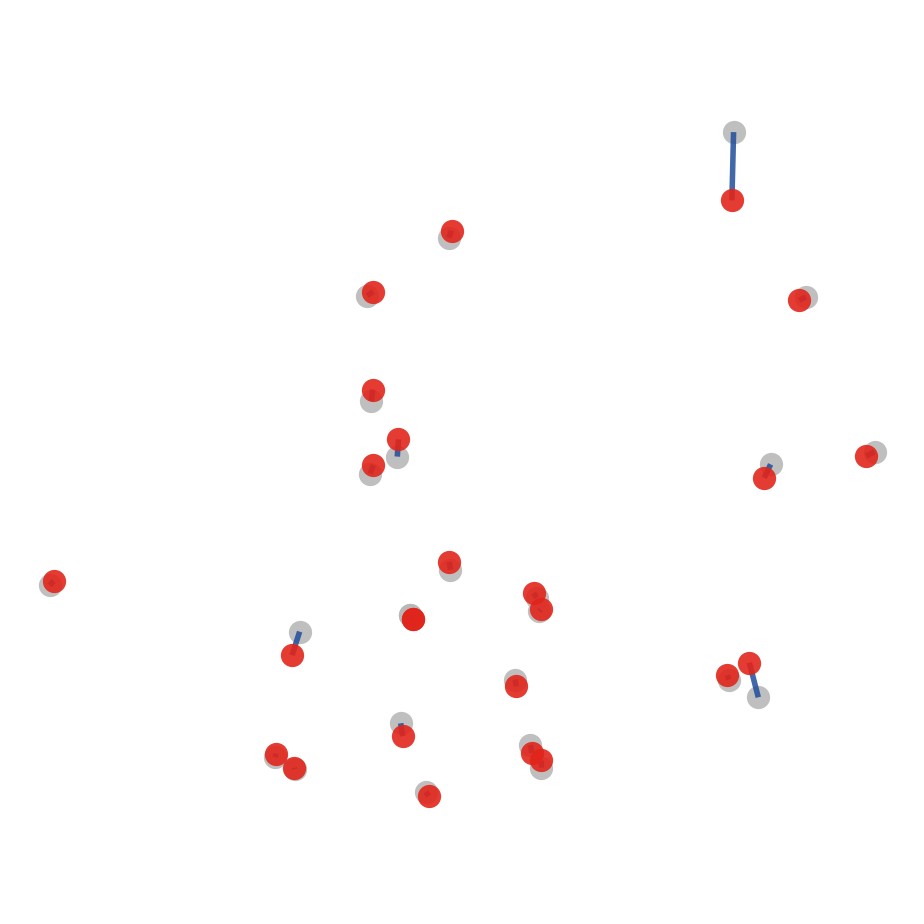}{38.3} & \mapauc{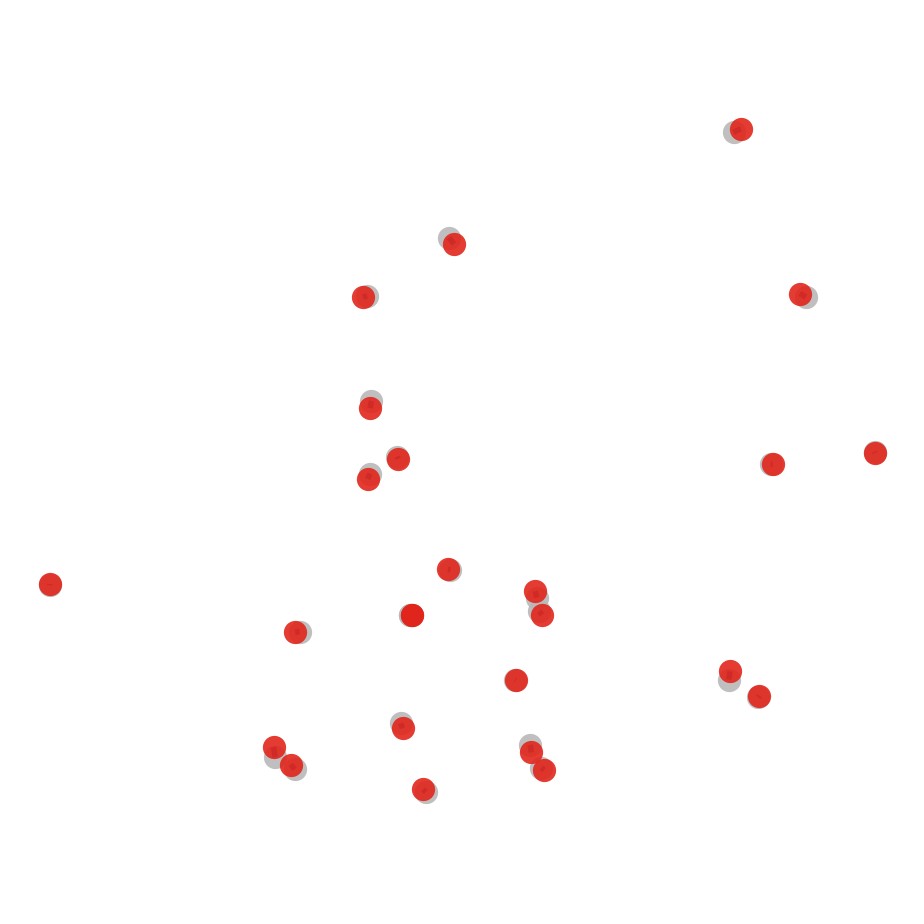}{59.1} \\[0pt]
\maprow{\textbf{+ \method}} & \mapname{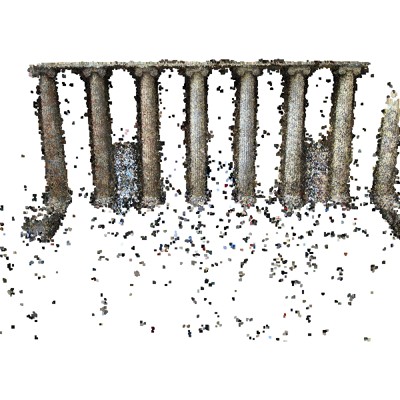}{\shortstack{British Museum\\bag 13}} & \mapaucbf{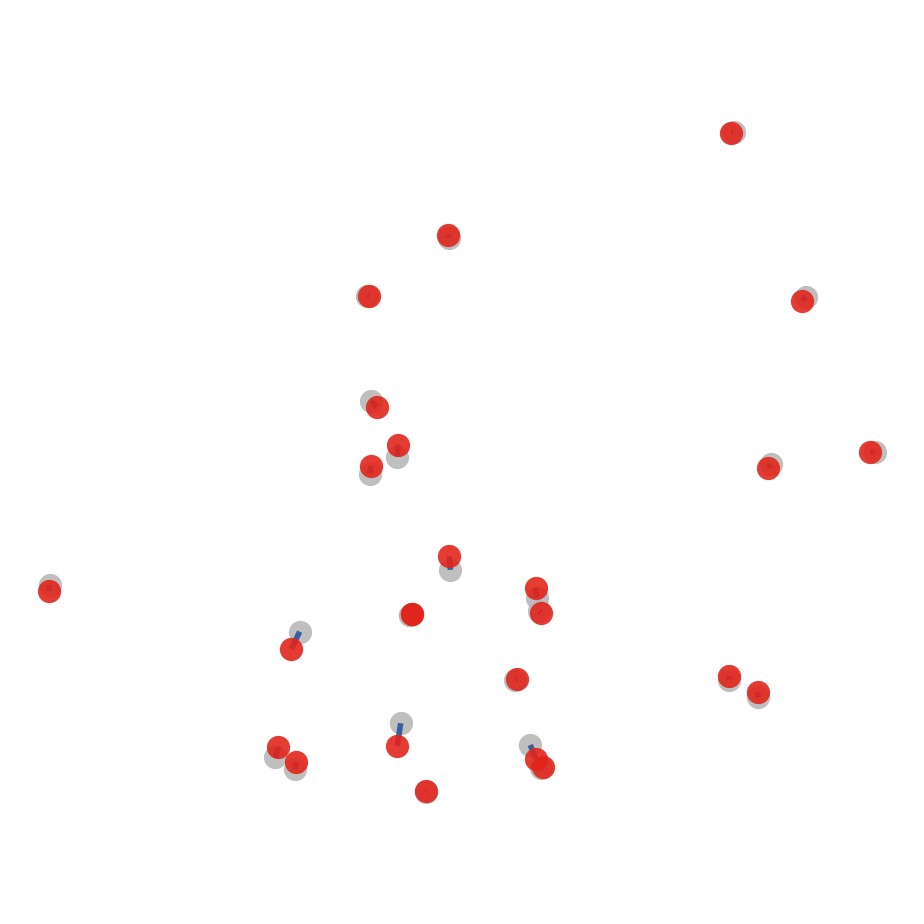}{57.0} & \mapaucbf{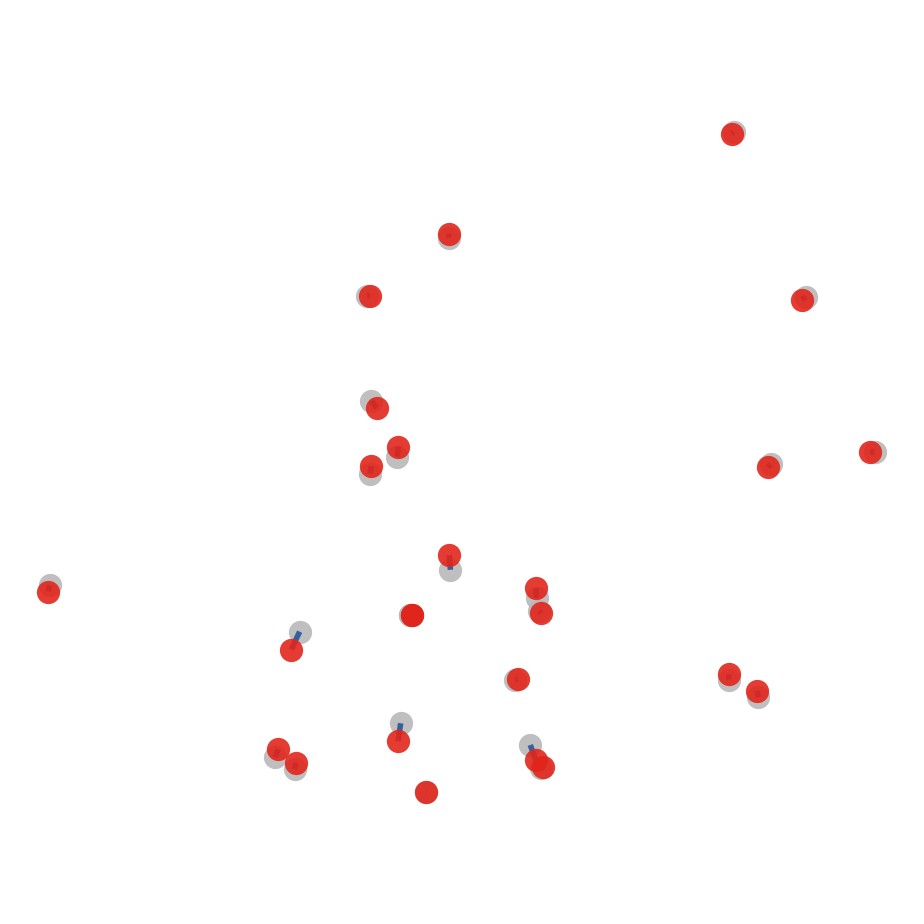}{56.6} & \mapaucbf{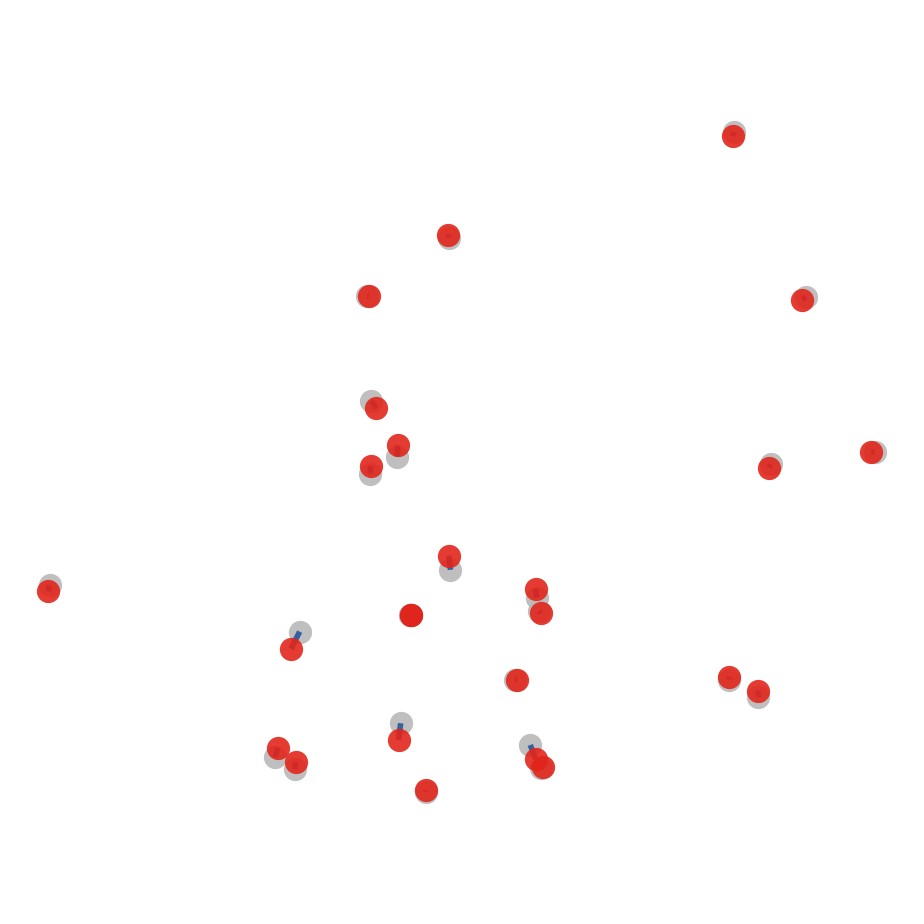}{56.8} & \mapaucbf{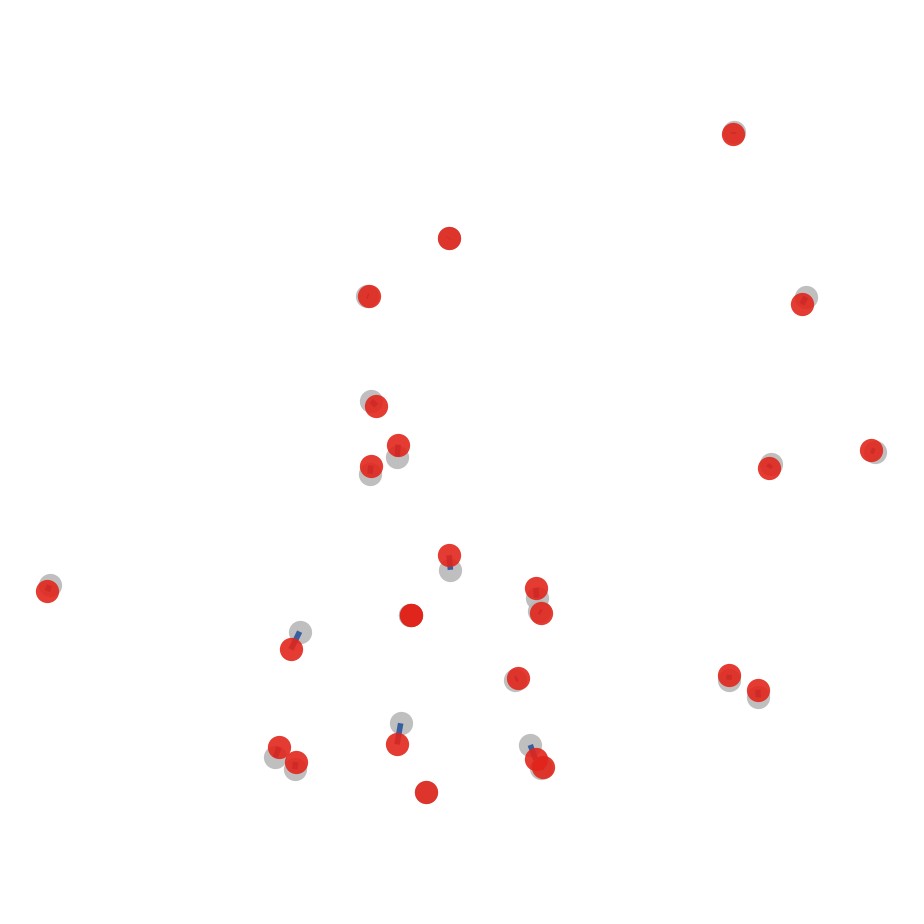}{55.5} & \mapaucbf{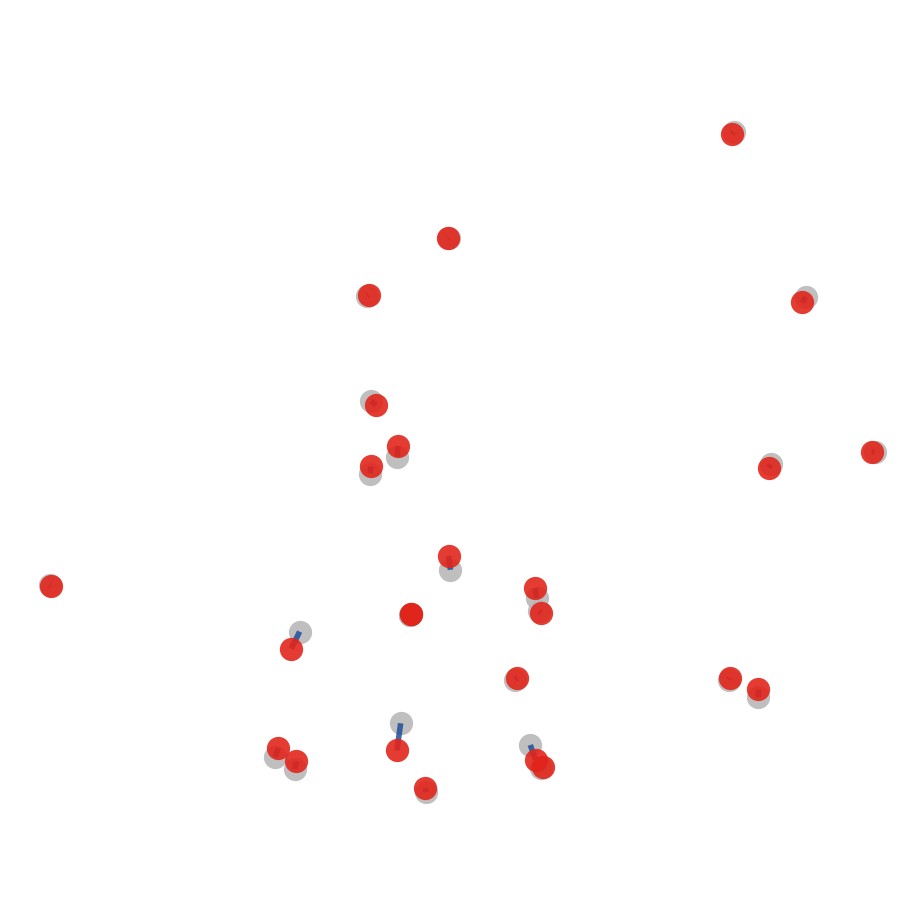}{57.5} & \mapaucbf{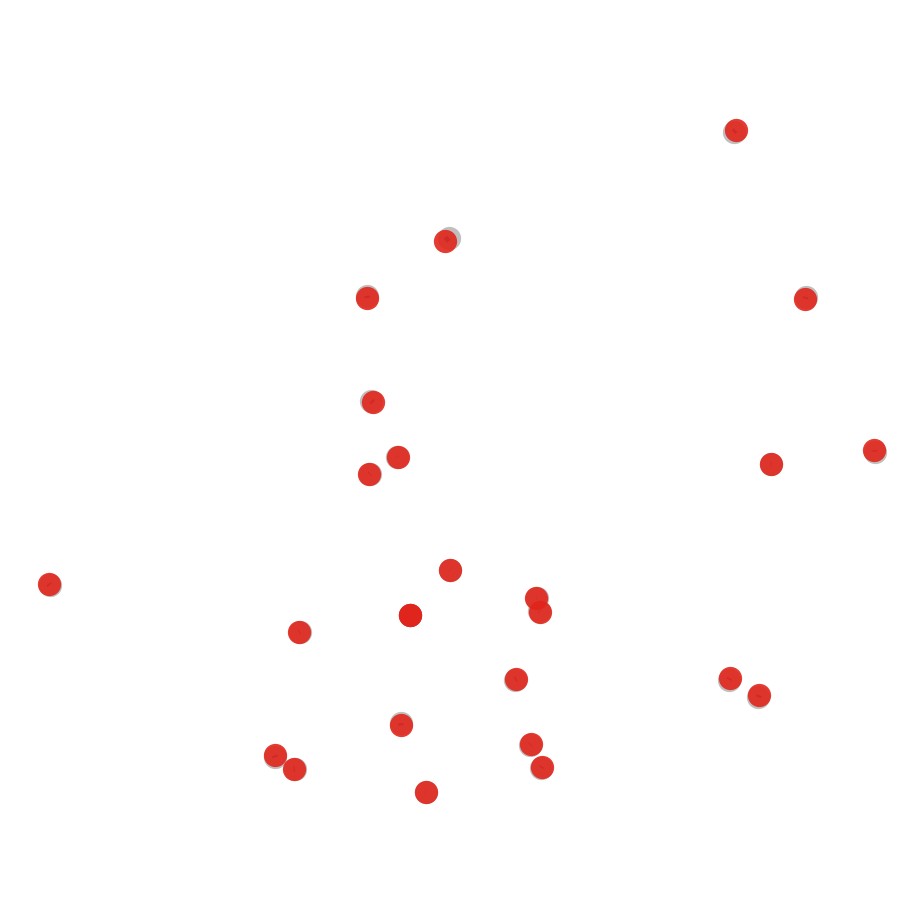}{82.1} \\[4pt]
\maprow{Initialization} & \mappanel{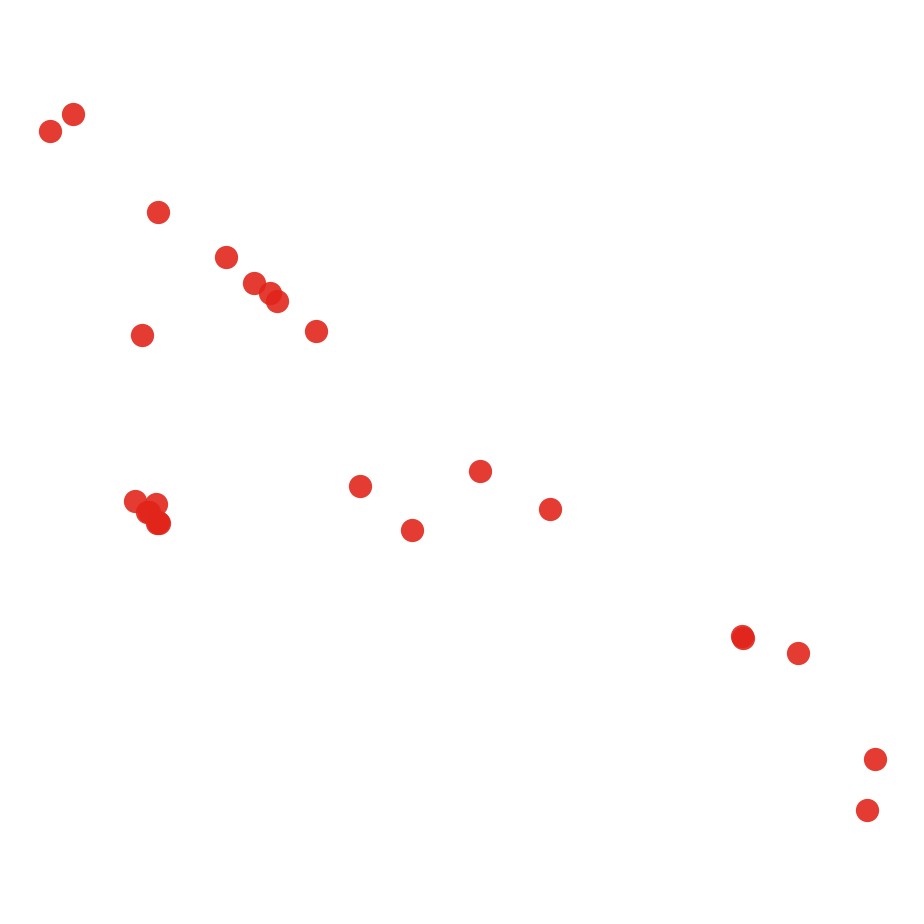} & \mapauc{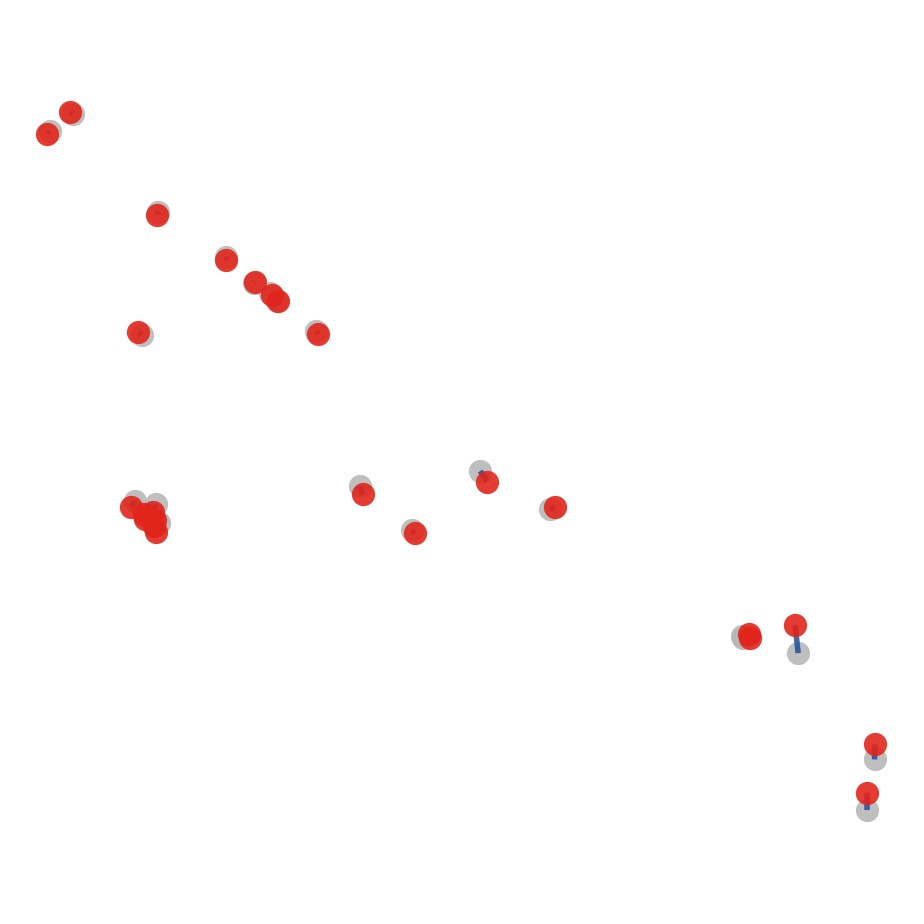}{57.3} & \mapauc{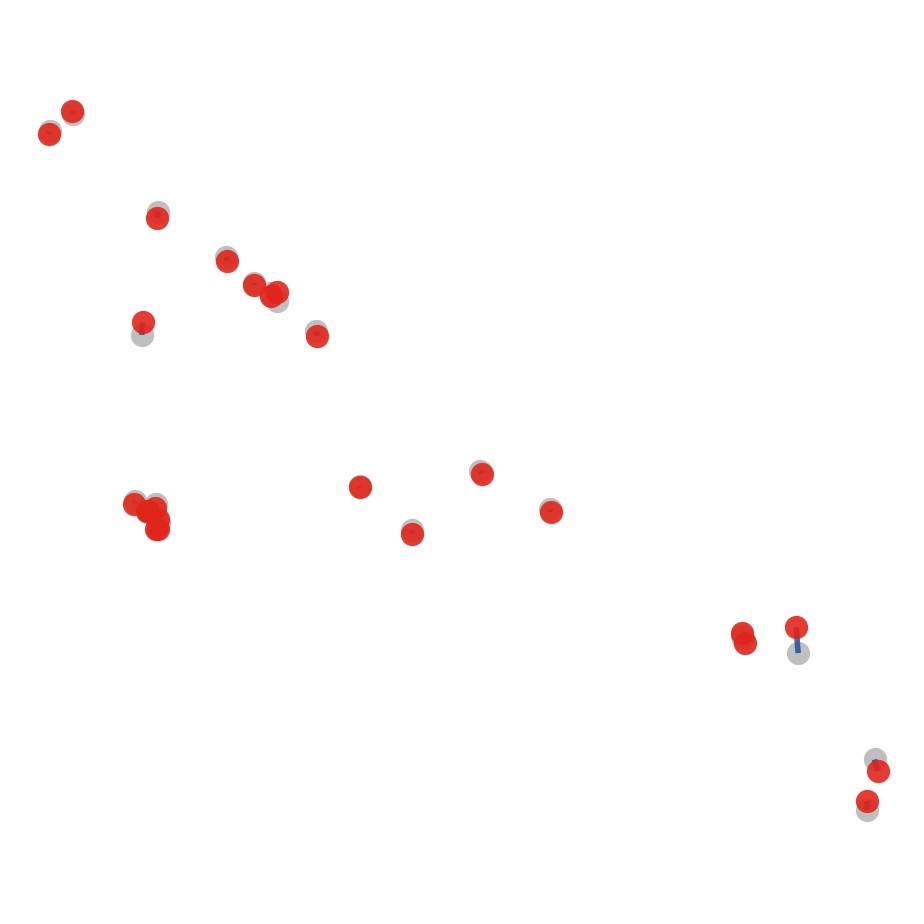}{64.2} & \mapauc{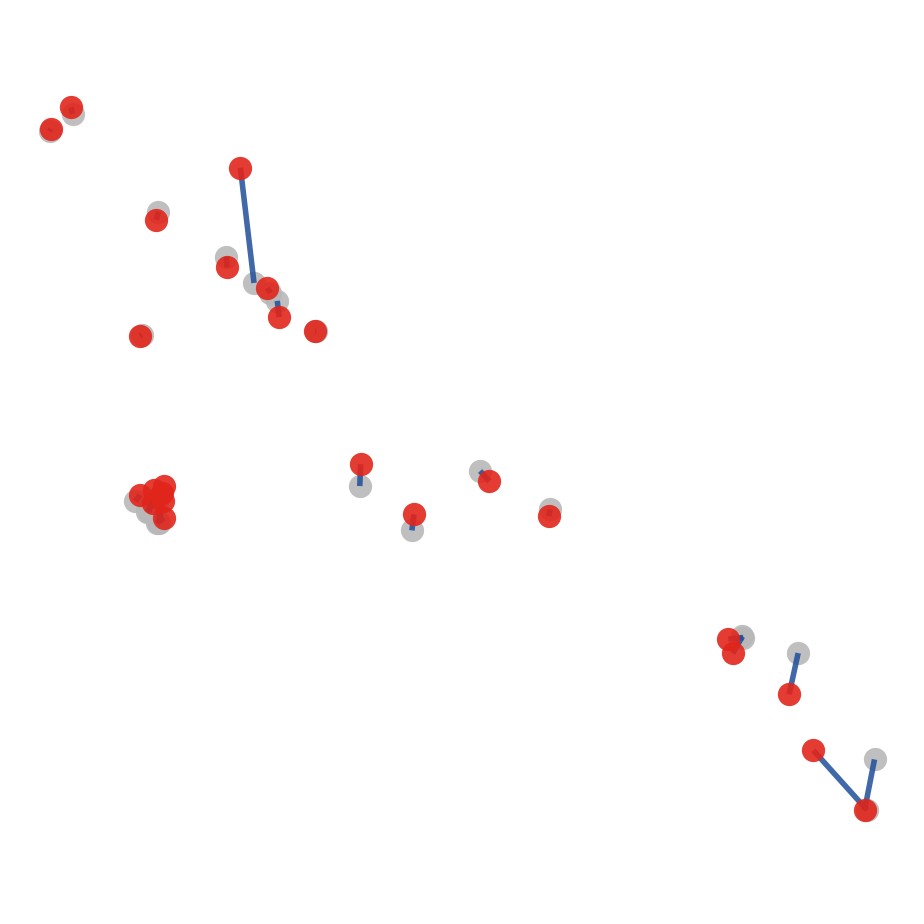}{23.8} & \mapauc{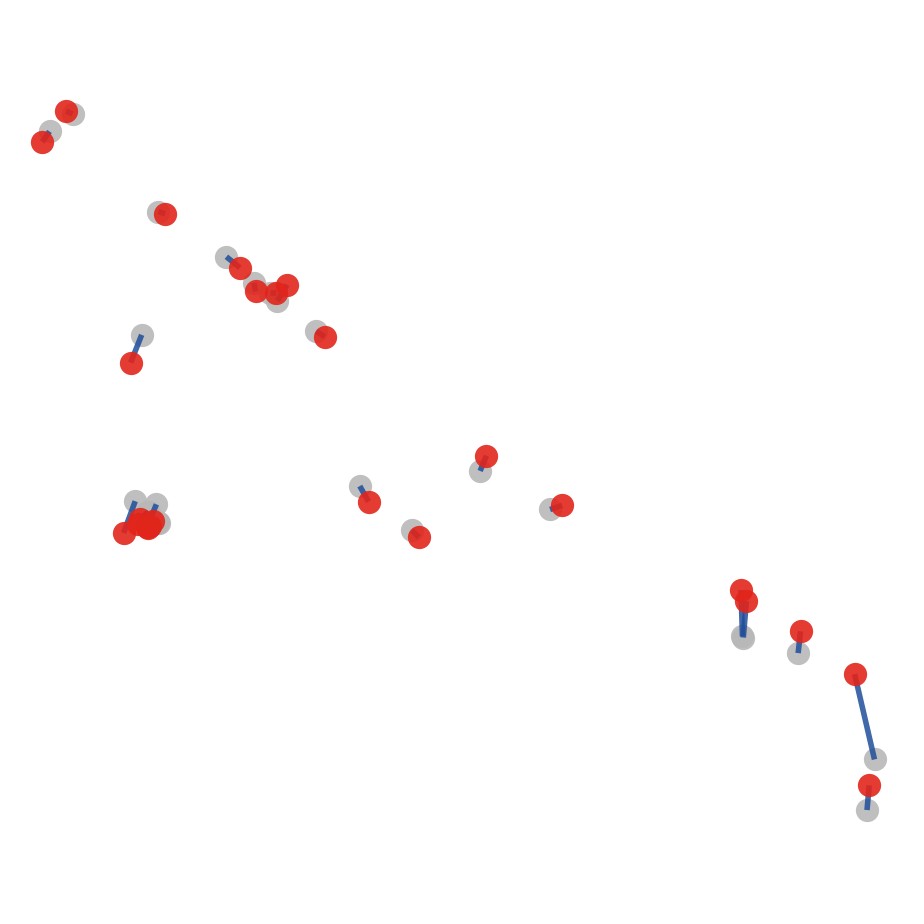}{19.6} & \mapauc{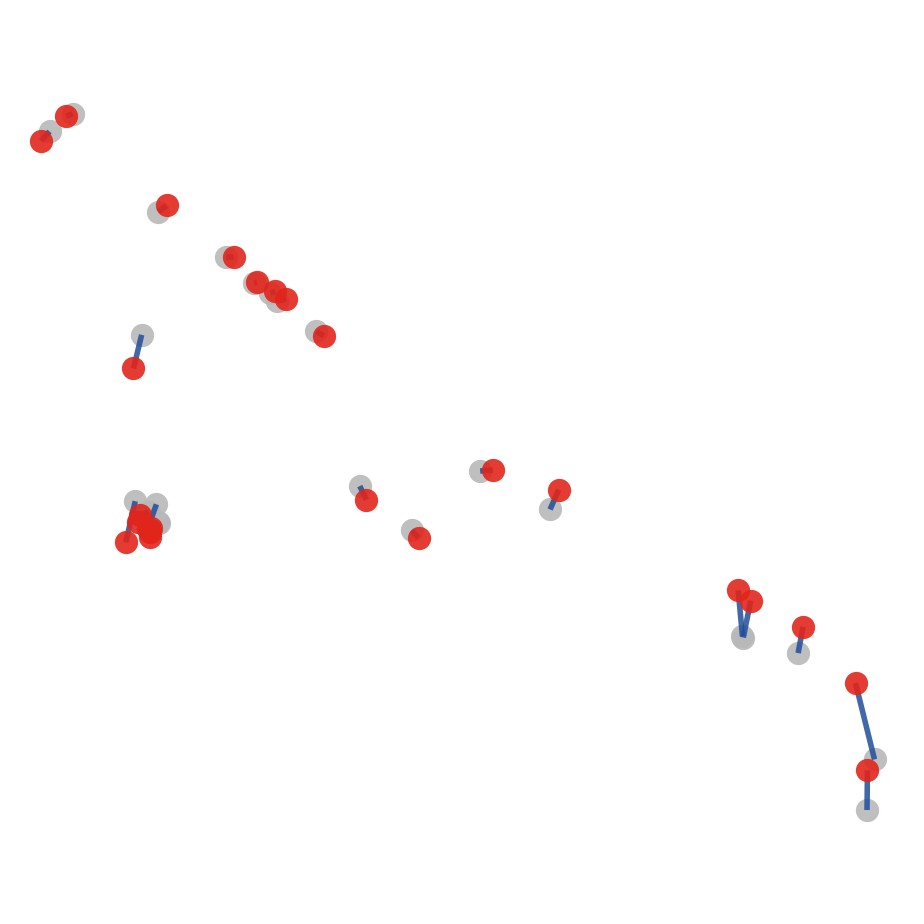}{19.5} & \mapauc{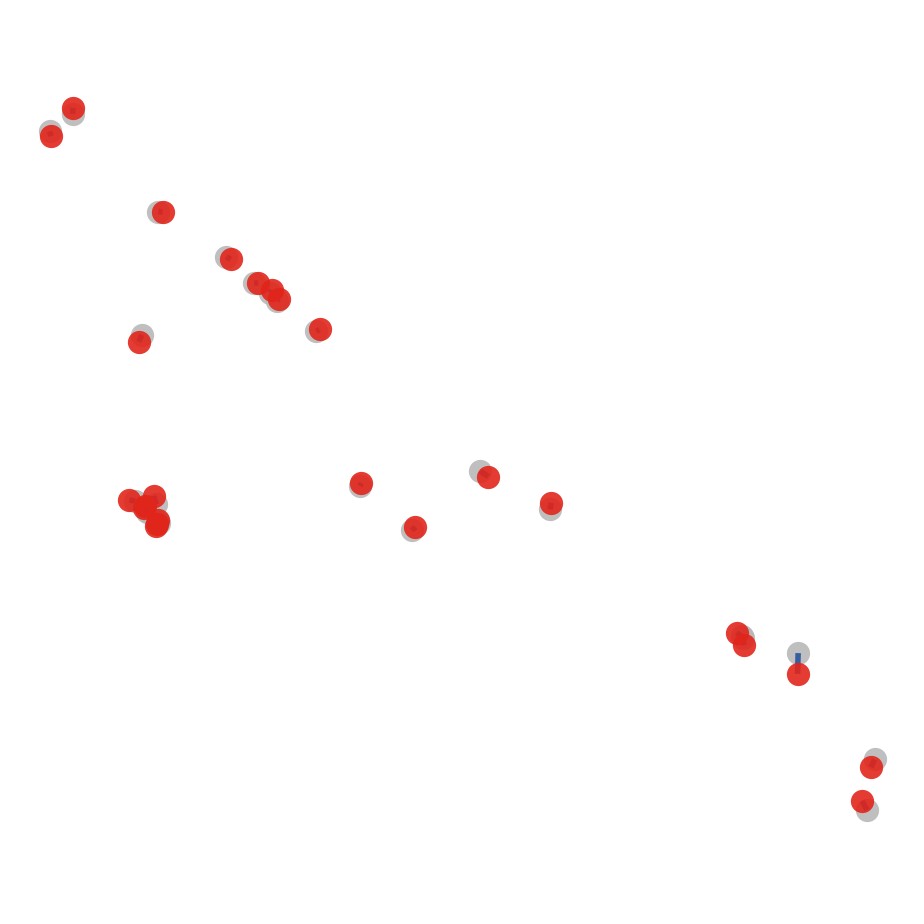}{54.0} \\[0pt]
\maprow{\textbf{+ \method}} & \mapname{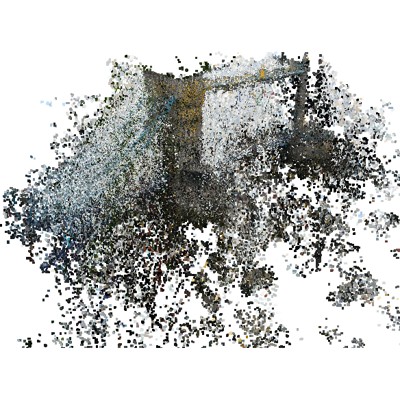}{\shortstack{London Bridge\\bag 18}} & \mapaucbf{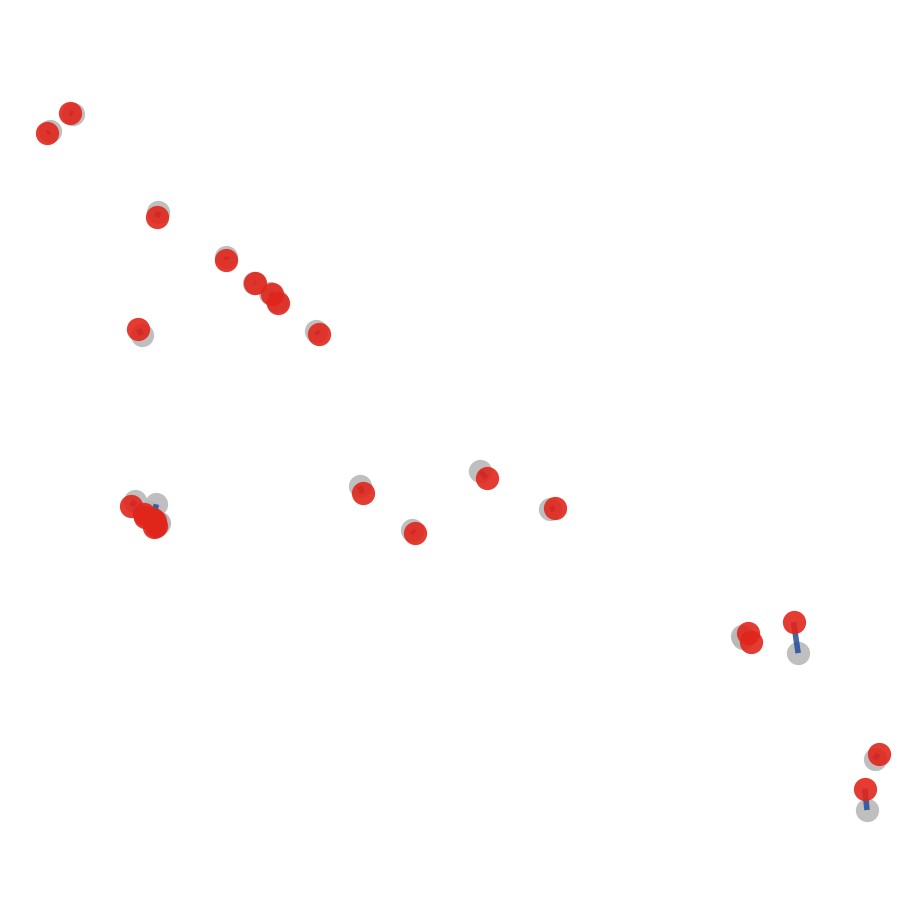}{57.1} & \mapaucbf{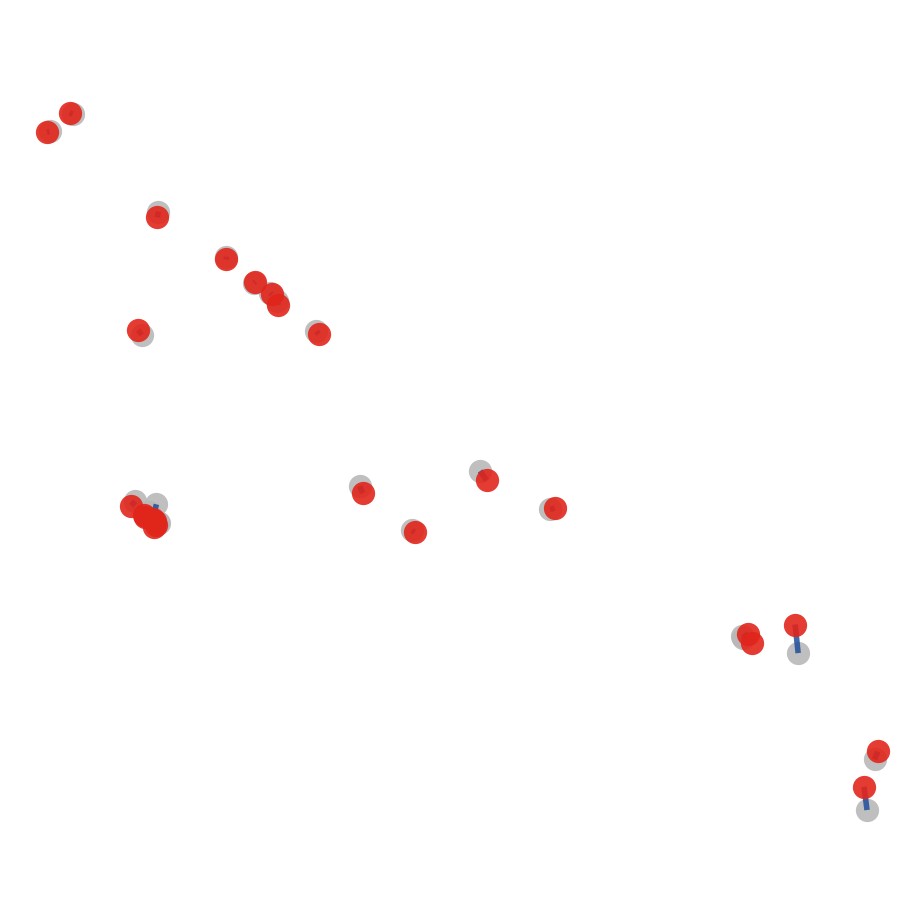}{56.5} & \mapaucbf{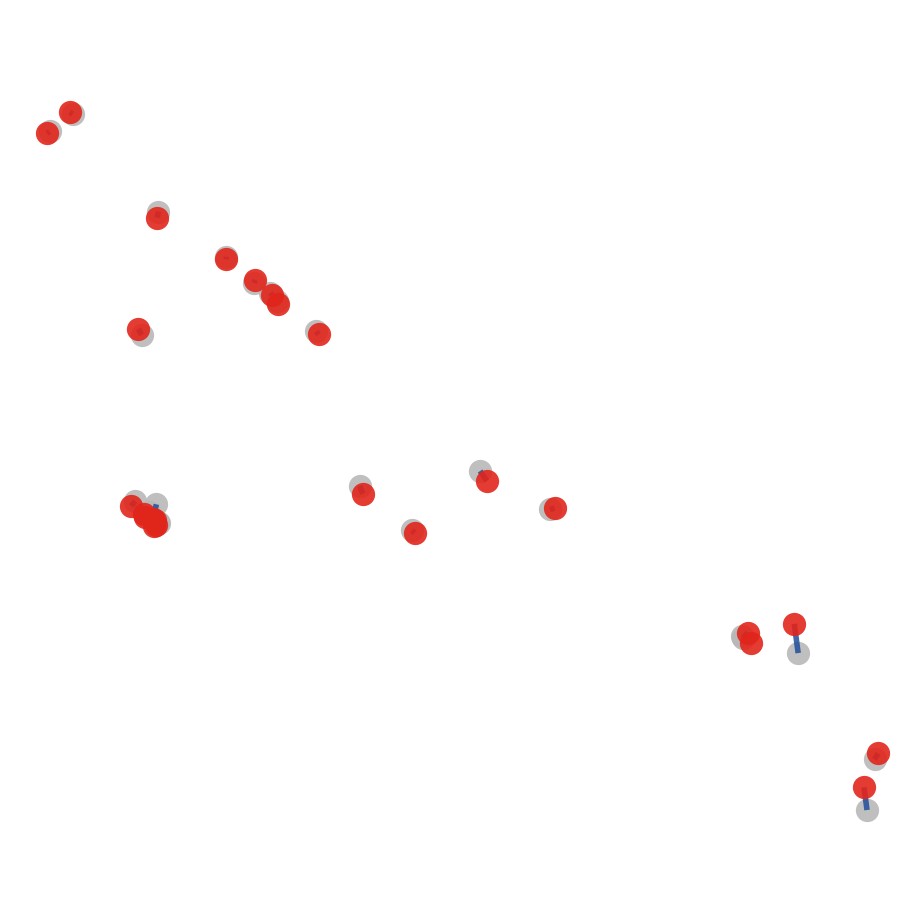}{56.2} & \mapaucbf{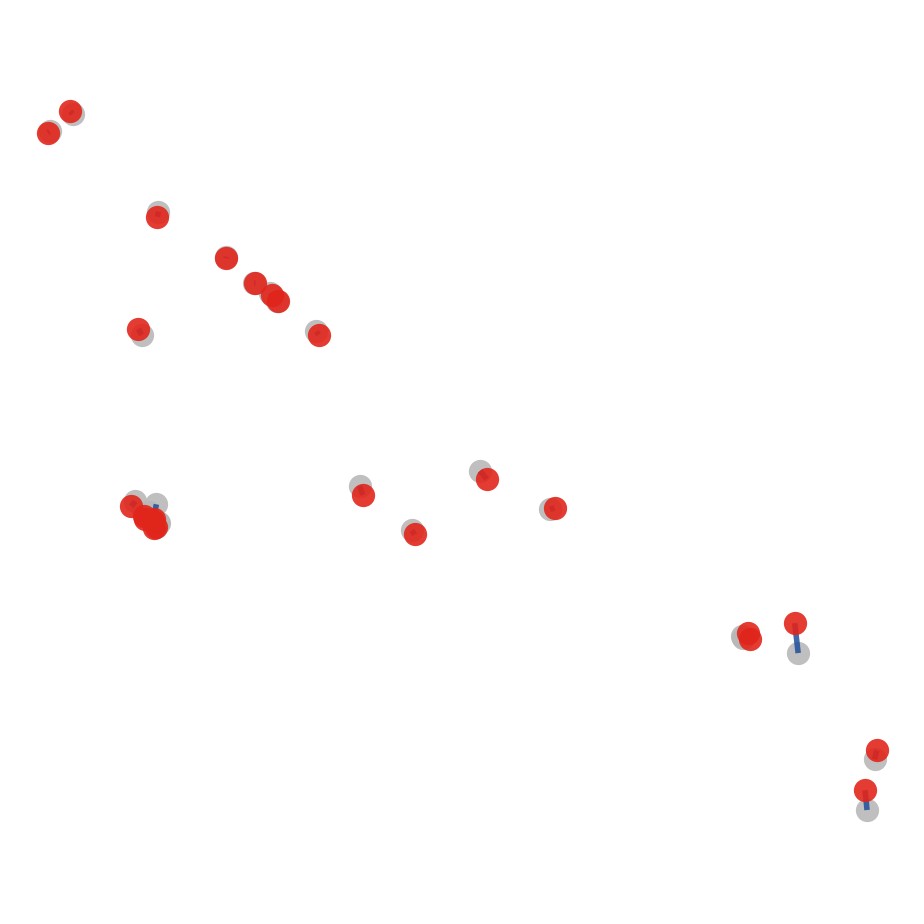}{56.6} & \mapaucbf{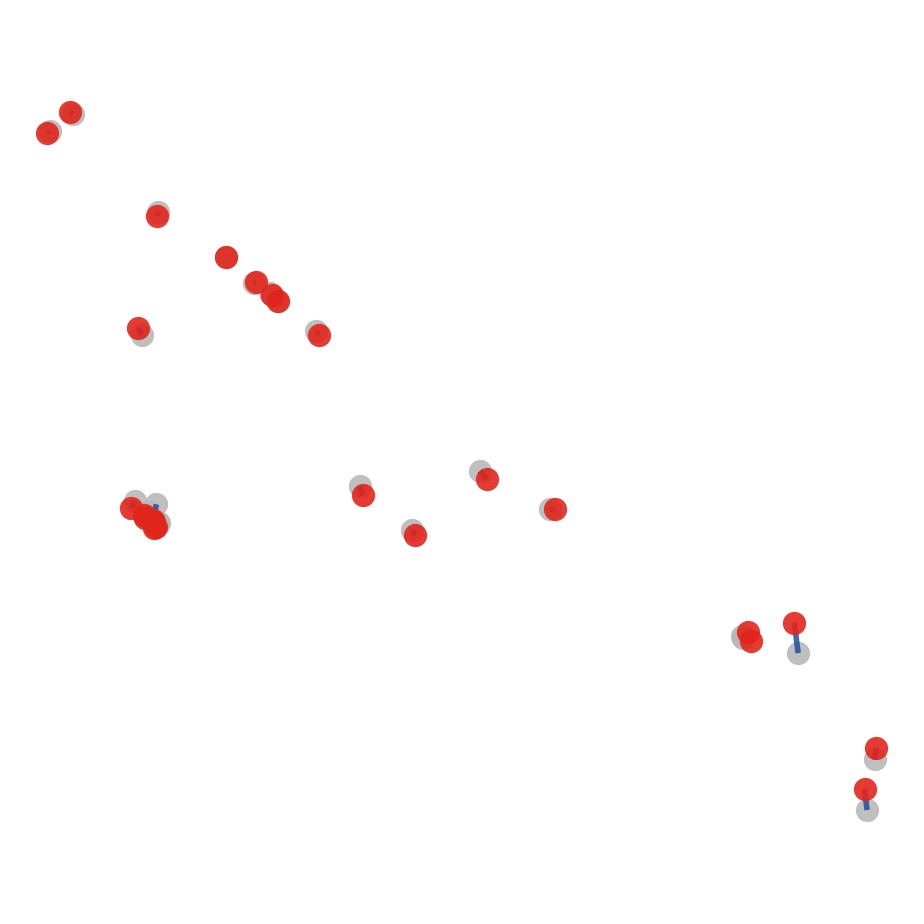}{55.6} & \mapaucbf{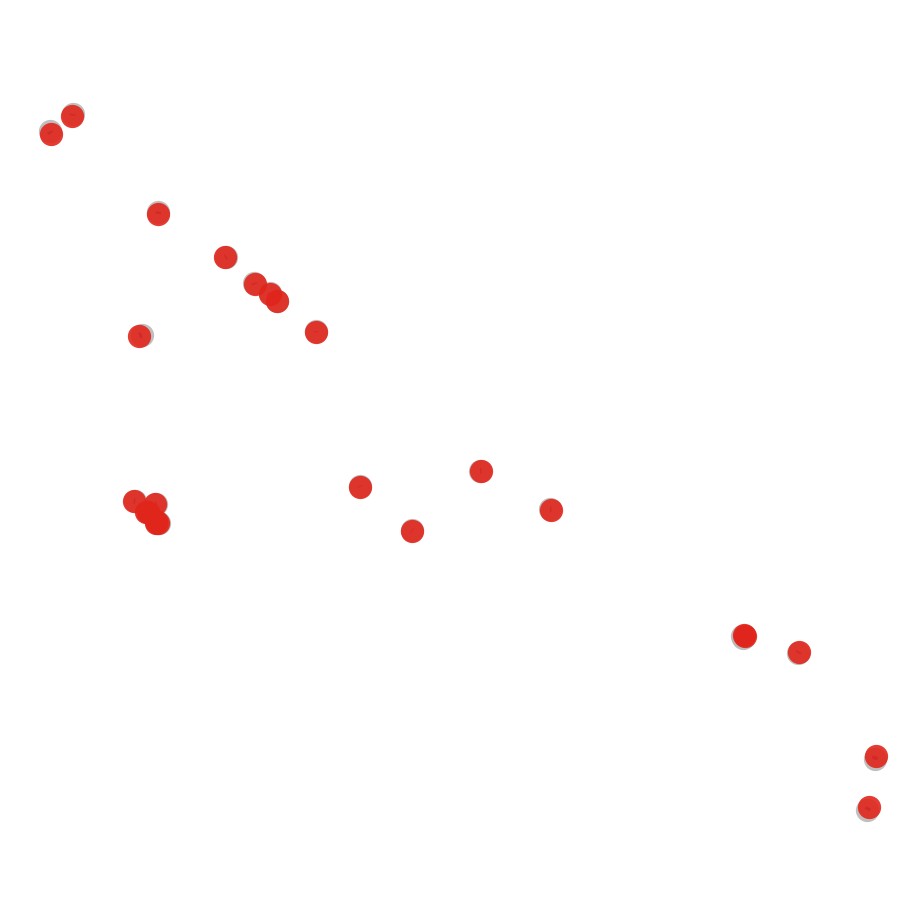}{85.1} \\[4pt]
\end{tabular}
\caption{Camera positions on IMC2021 bags of 25 images: the Initialization cameras of each method (upper row of each bag) and the same cameras after \method (lower row), aligned to the ground truth by Sim(3); each estimated camera (red) is joined to its ground-truth position (grey) by a blue line. Pi3X (calib.): with the ground-truth intrinsics.}
\label{fig:imc_map}
\end{figure*}

\ifdefined\smerfW\else\newlength{\smerfW}\fi\setlength{\smerfW}{1.08in}
\begin{figure*}[p]
\def\mappanel#1{{\setlength{\fboxsep}{0pt}\setlength{\fboxrule}{0.3pt}\fcolorbox{black!20}{white}{\includegraphics[width=\dimexpr\smerfW-0.6pt\relax]{fig/smerf_map/#1}}}}
\def\mapauc#1#2{\mappanel{#1}\llap{\raisebox{1.2pt}{\setlength{\fboxsep}{0.8pt}\colorbox{white}{\tiny AUC@5$^\circ$ #2}}\hspace{1.2pt}}}
\def\mapaucbf#1#2{\mappanel{#1}\llap{\raisebox{1.2pt}{\setlength{\fboxsep}{0.8pt}\colorbox{white}{\tiny\bfseries AUC@5$^\circ$ #2}}\hspace{1.2pt}}}
\def\maprow#1{\rotatebox{90}{\scriptsize #1}}
\def\mapname#1#2{\raisebox{0pt}[0pt][0pt]{\parbox{\smerfW}{\centering\includegraphics[width=0.66\smerfW]{fig/smerf_map/#1}\\[1pt]\footnotesize\bfseries #2}}}  
\centering
\setlength{\tabcolsep}{0.8pt}
\renewcommand{\arraystretch}{0.7}
\begin{tabular}{@{}>{\centering\arraybackslash}m{0.9em} >{\centering\arraybackslash}m{\smerfW}@{\hspace{4pt}} *{5}{>{\centering\arraybackslash}m{\smerfW}} @{}}
 & \small\bfseries GT & \small\bfseries GLUEMAP & \small\bfseries DGSfM & \small\bfseries Solv3R & \small\bfseries DVLT & \small\bfseries VGGT-$\Omega$ \\[2pt]

\maprow{Initialization} & \mappanel{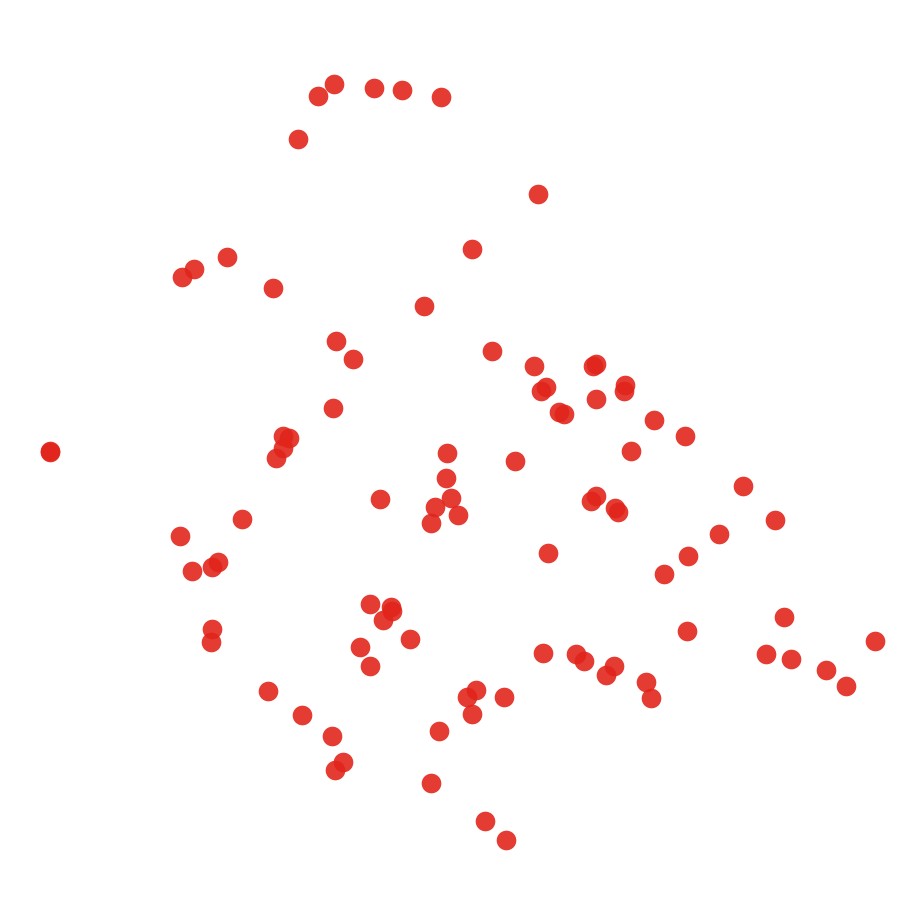} & \mapauc{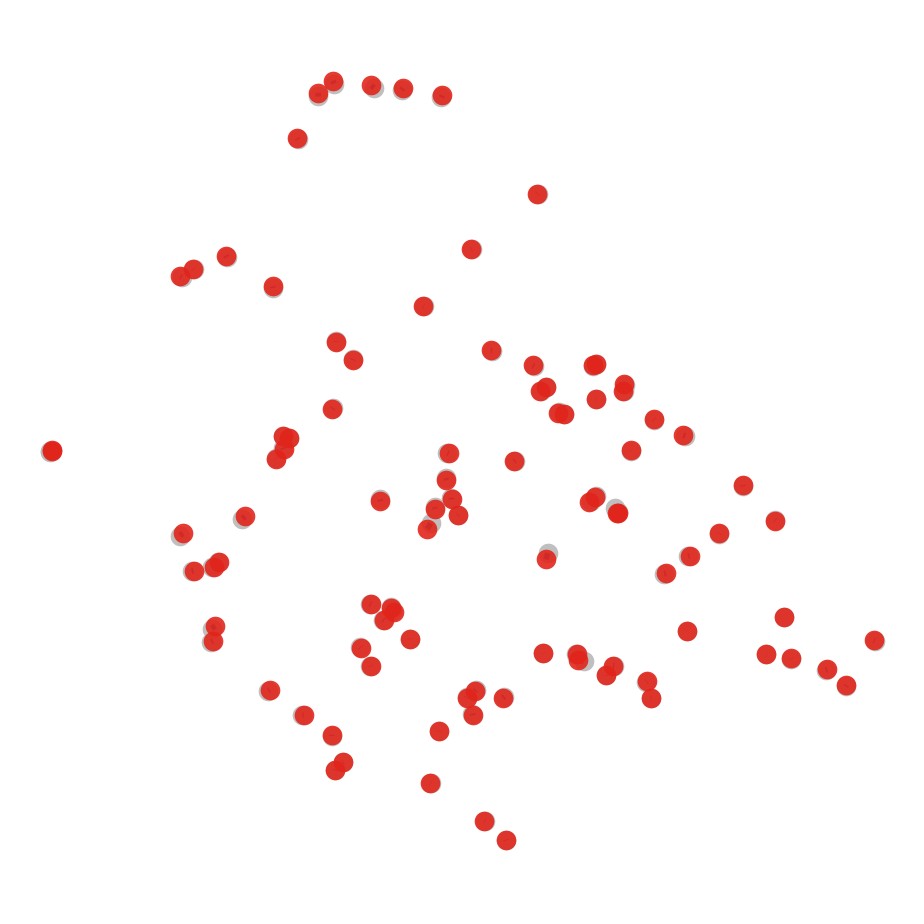}{82.0} & \mapauc{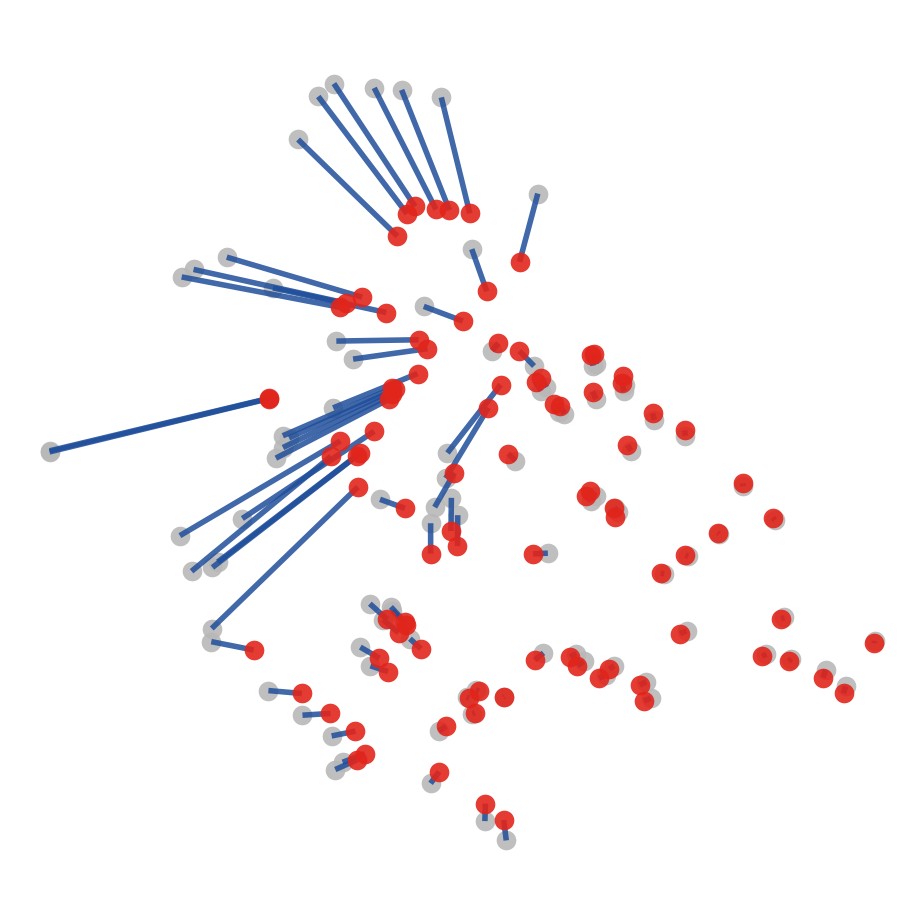}{37.1} & \mapauc{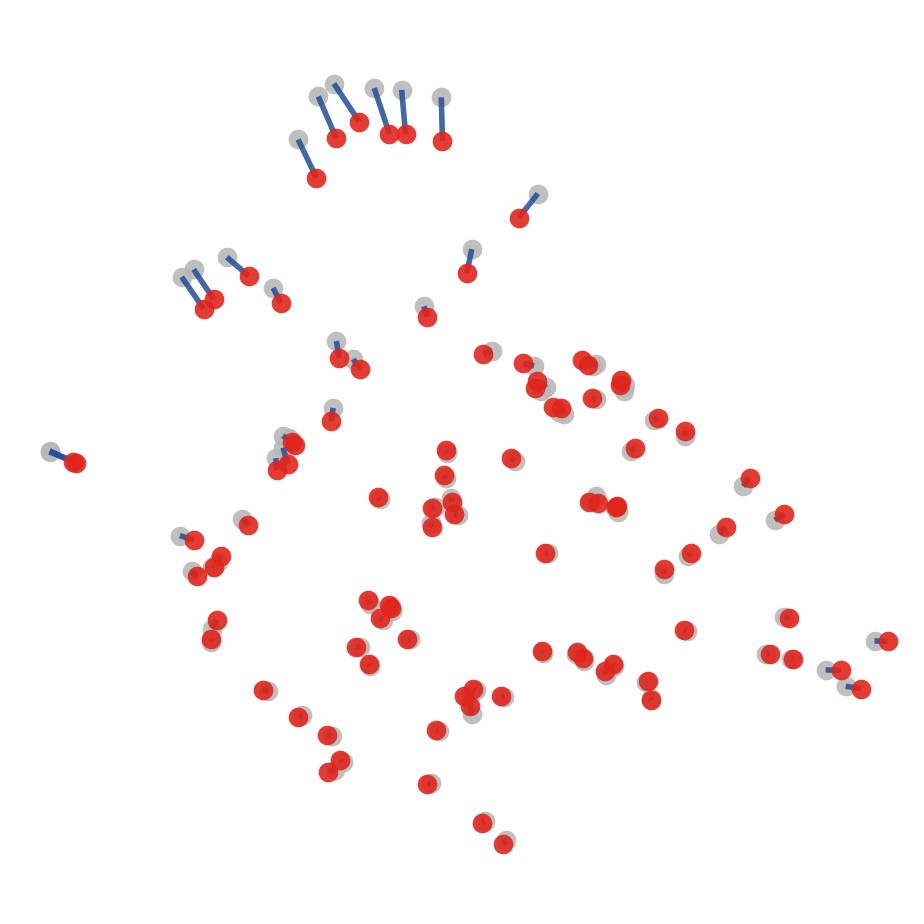}{45.7} & \mapauc{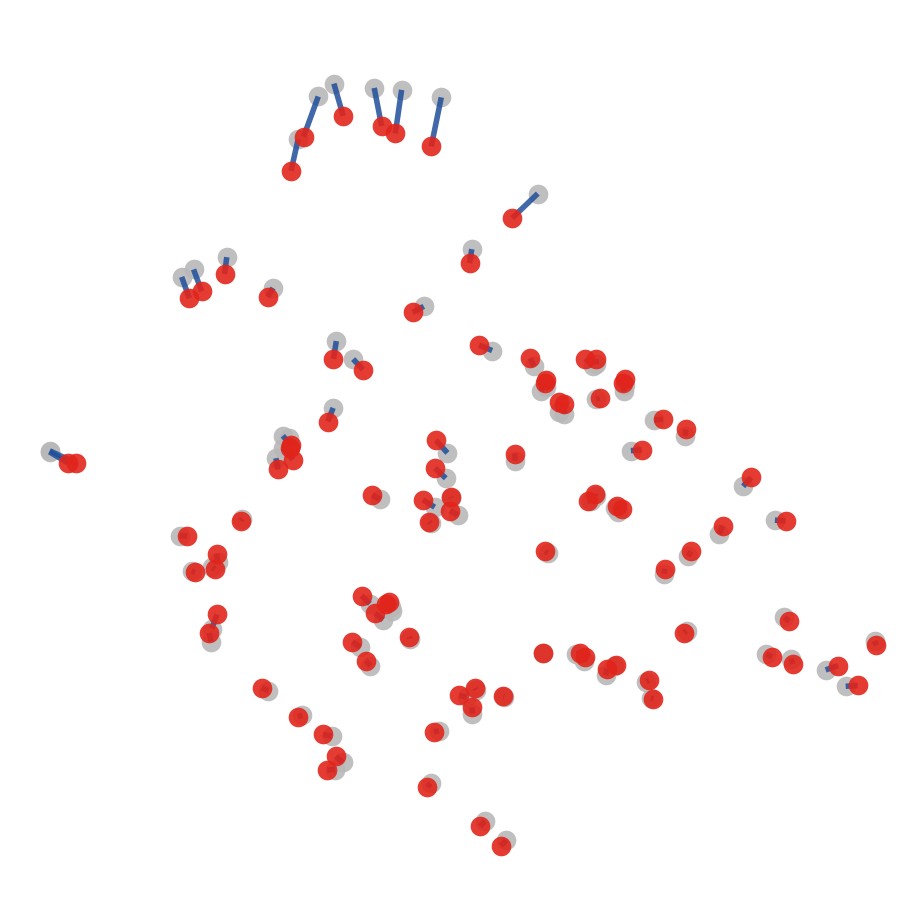}{39.3} & \mapauc{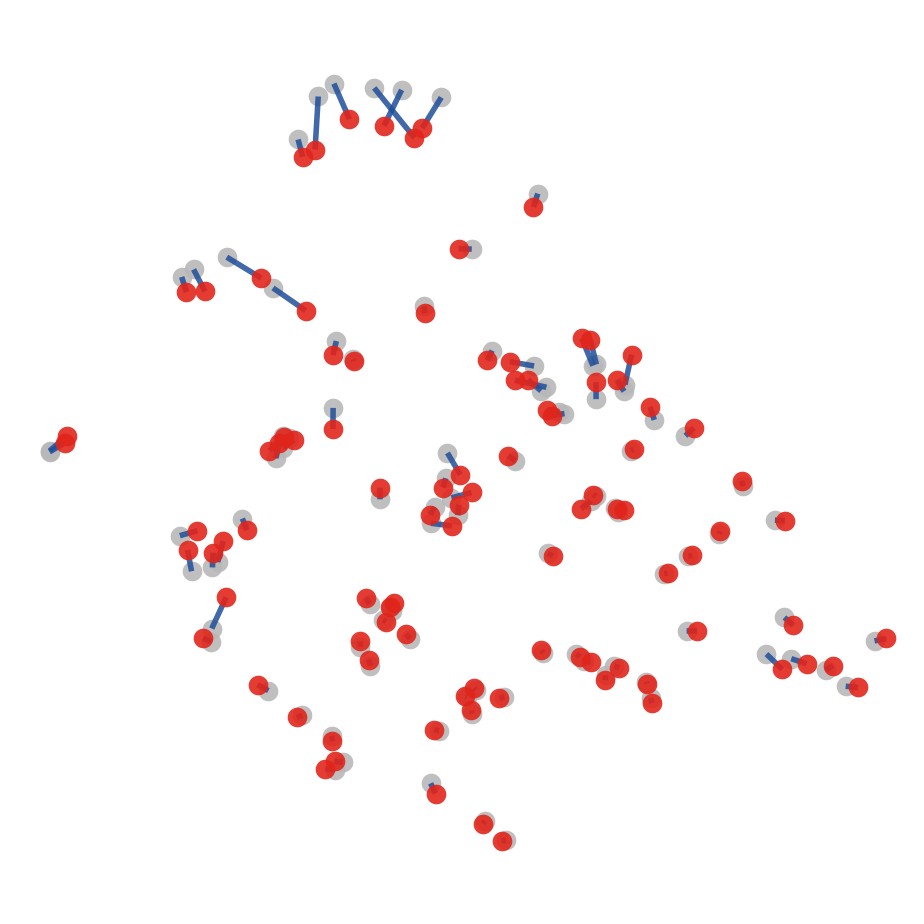}{31.9} \\[0pt]
\maprow{\textbf{+ \method}} & \mapname{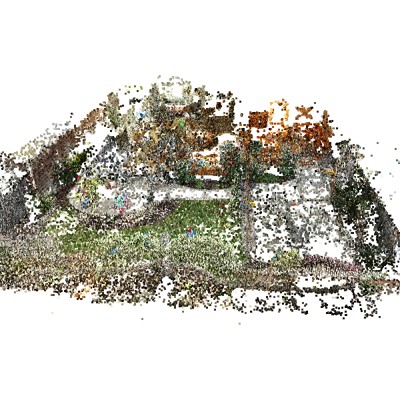}{\shortstack{Alameda 2\\low overlap}} & \mapaucbf{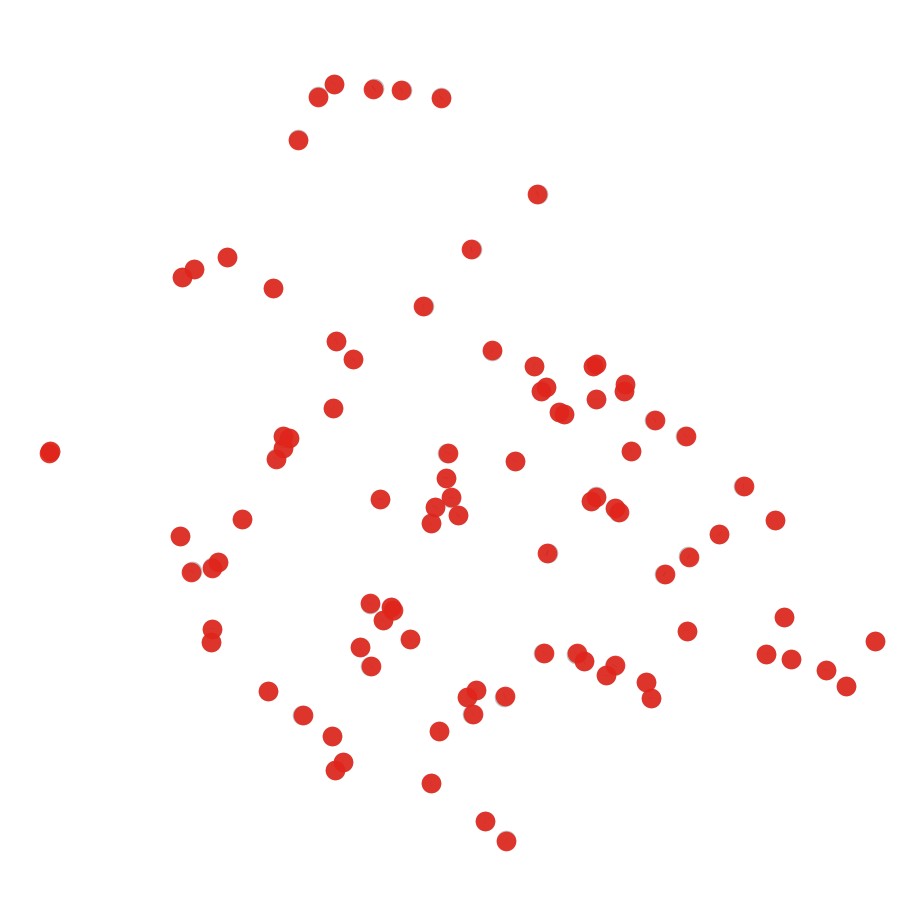}{95.6} & \mapaucbf{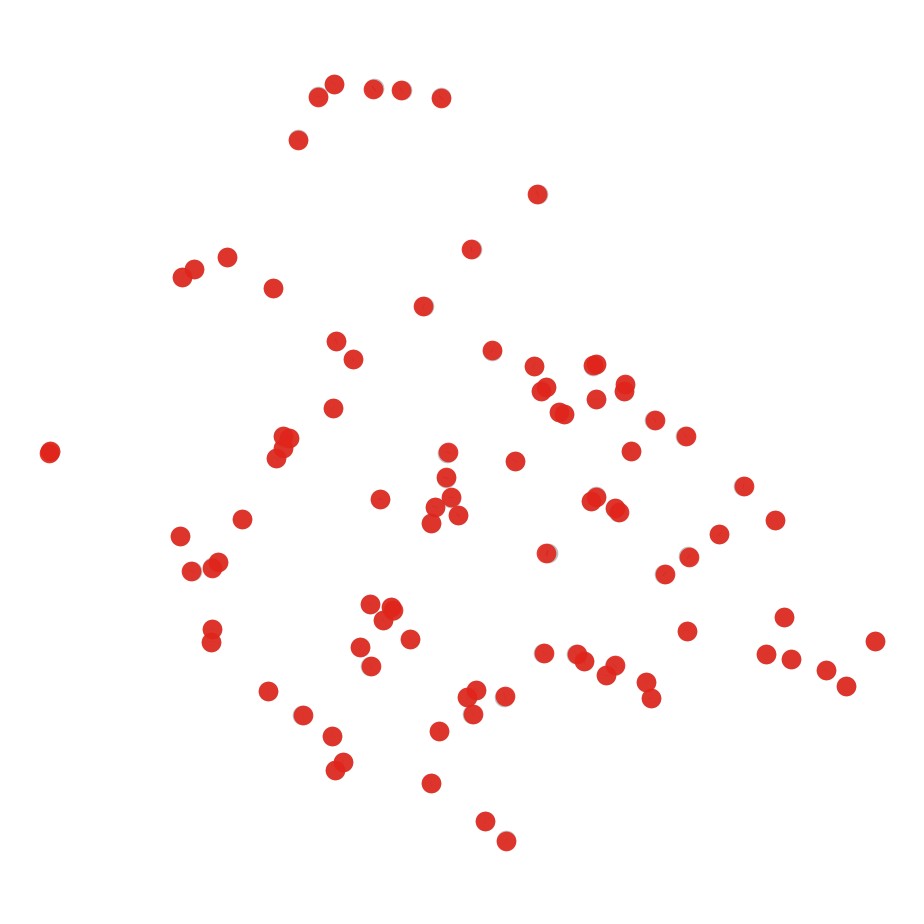}{95.5} & \mapaucbf{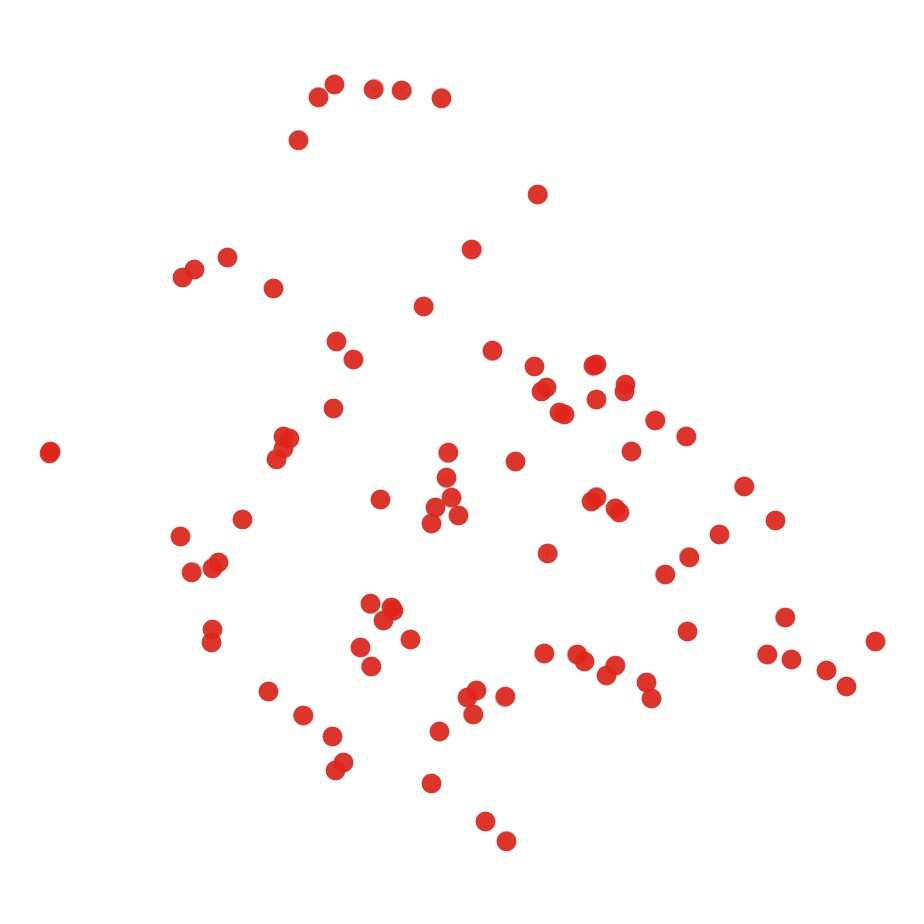}{95.5} & \mapaucbf{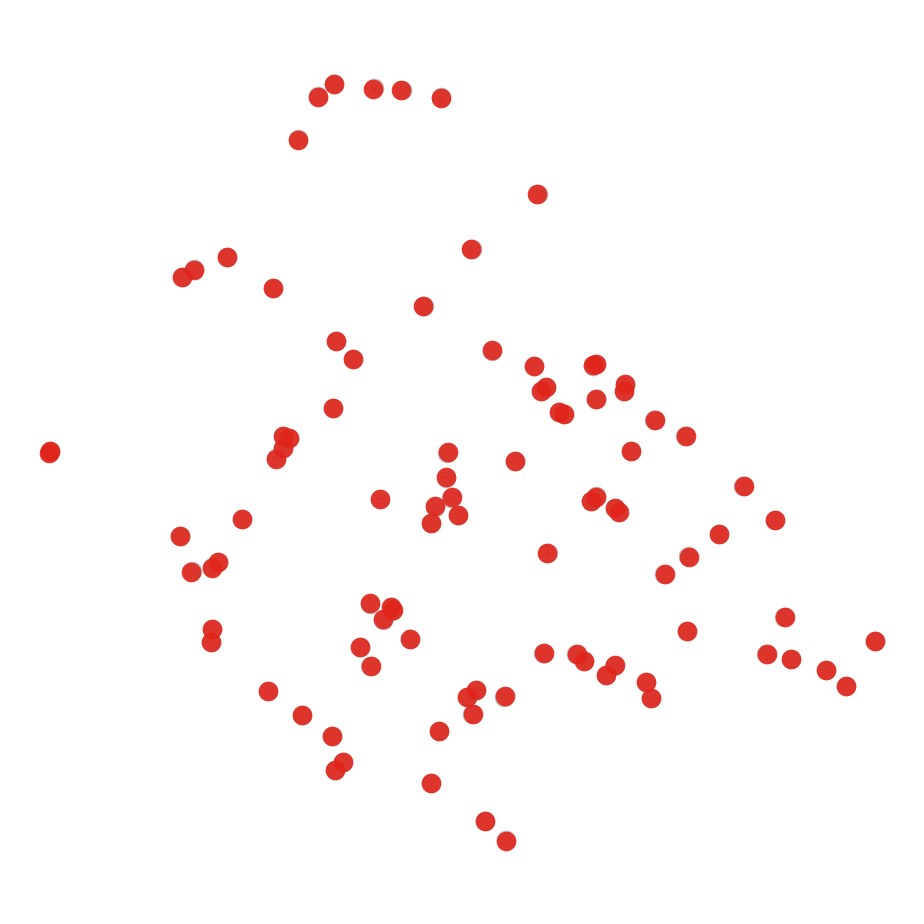}{95.4} & \mapaucbf{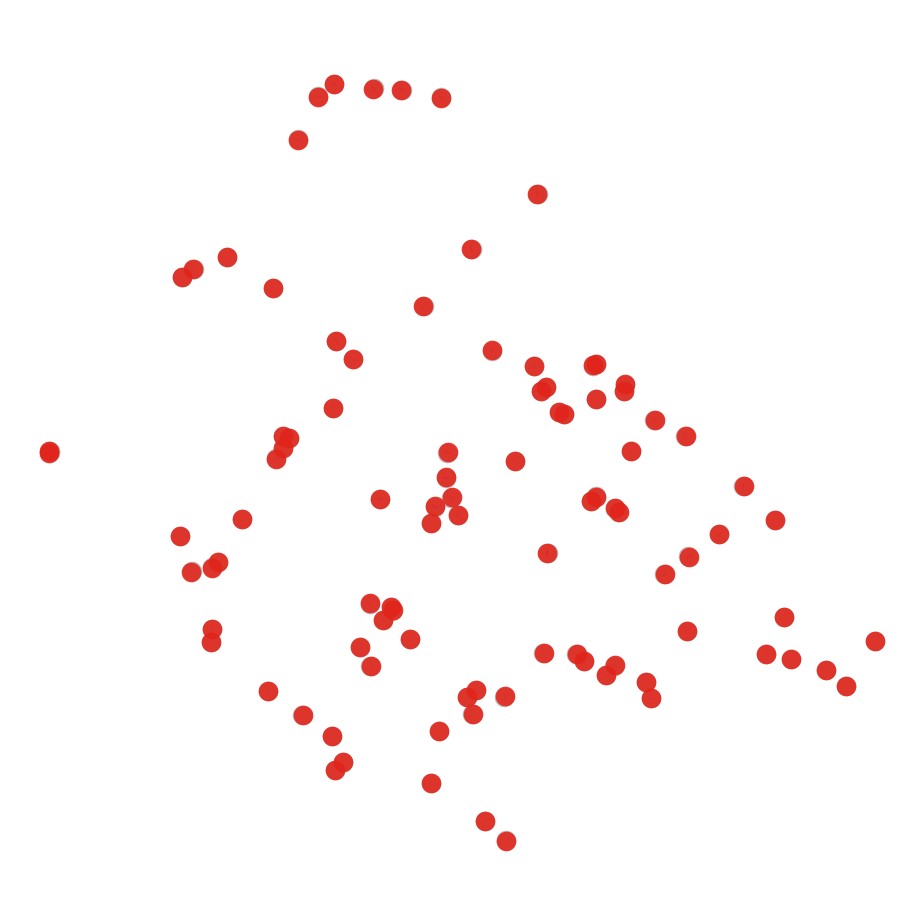}{95.3} \\[4pt]
\maprow{Initialization} & \mappanel{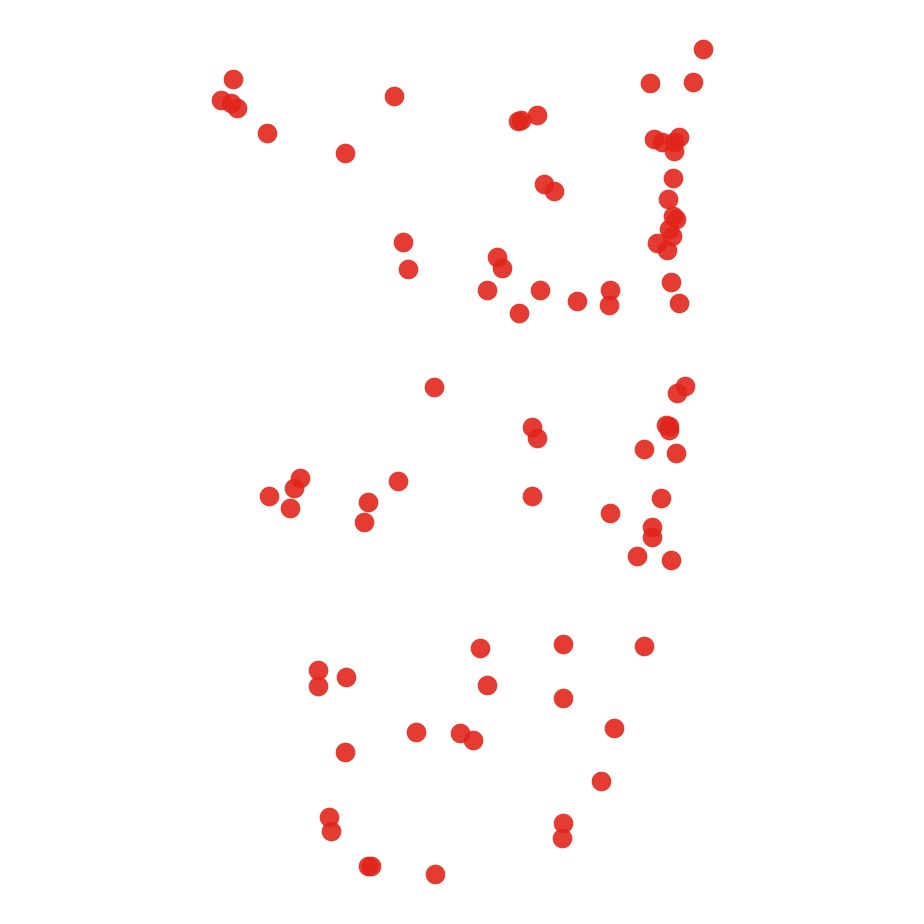} & \mapauc{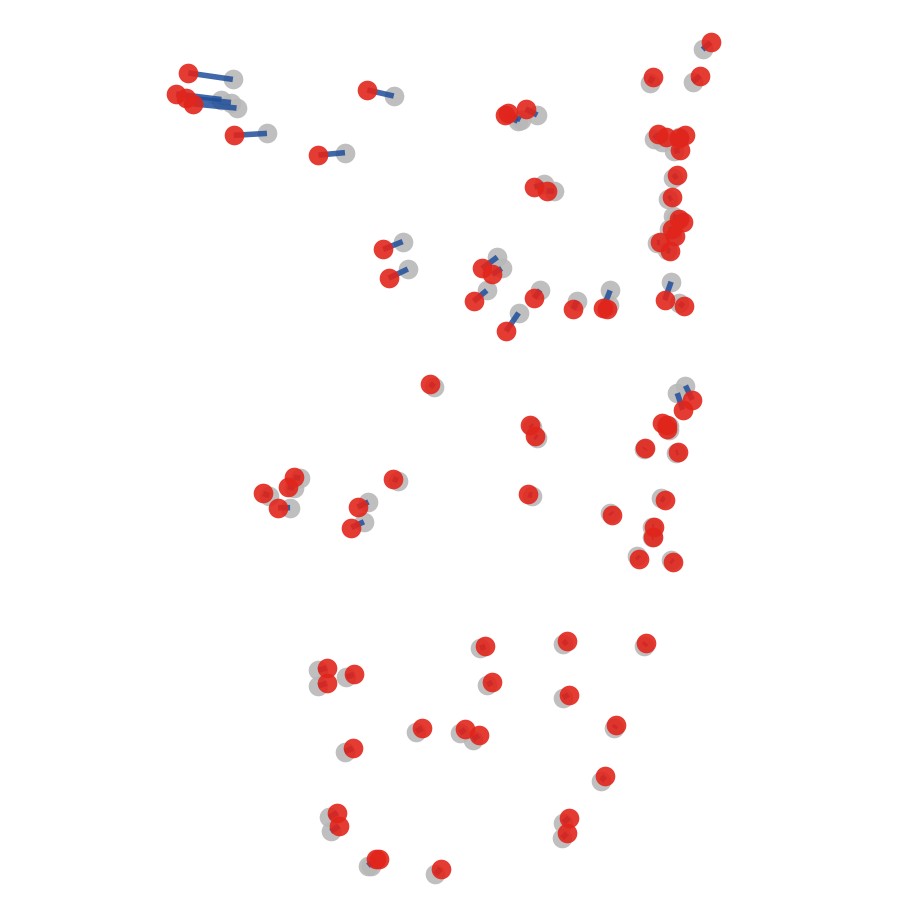}{57.0} & \mapauc{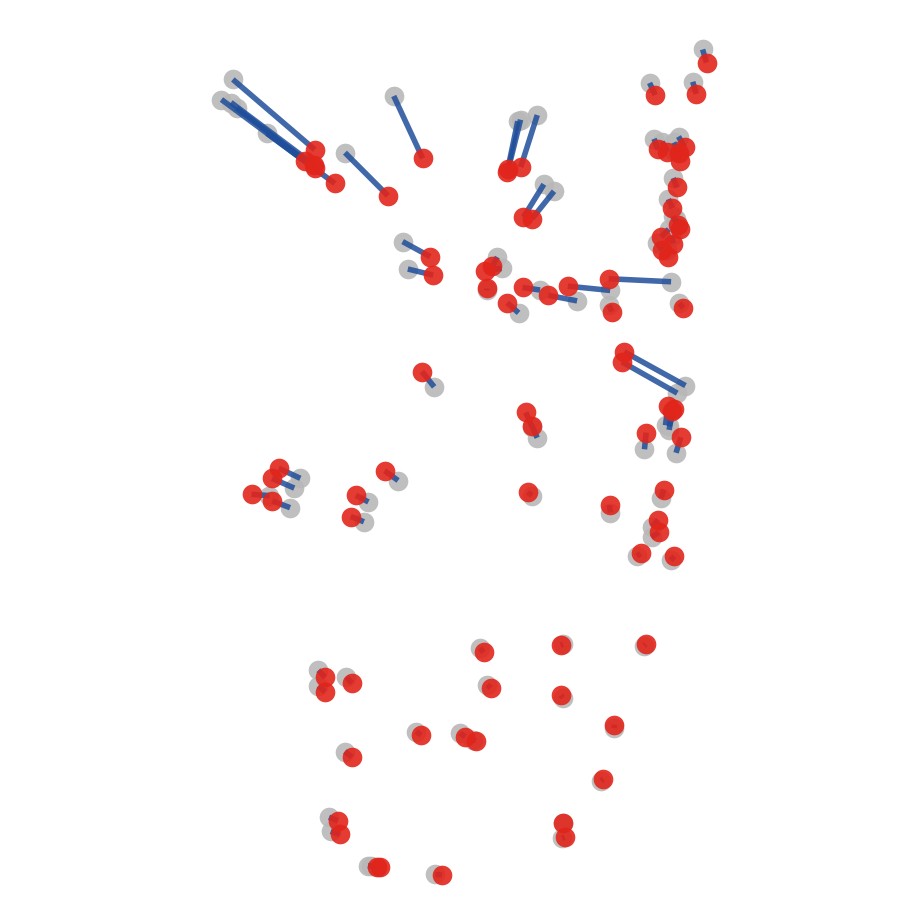}{43.2} & \mapauc{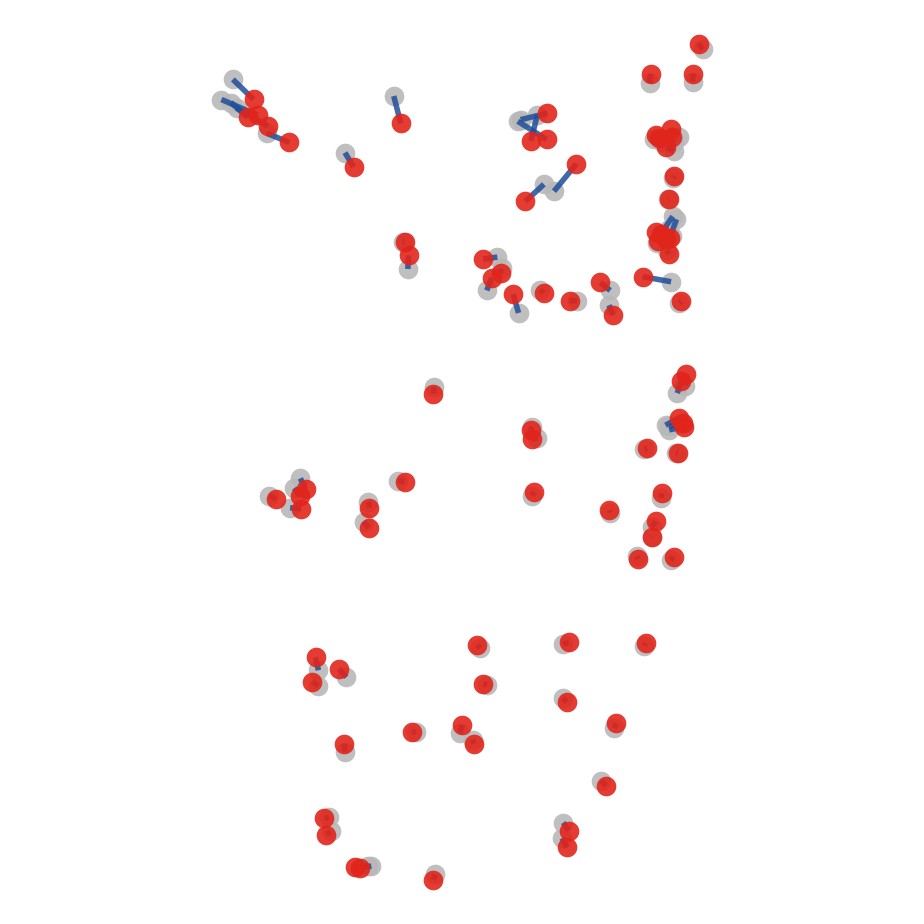}{25.2} & \mapauc{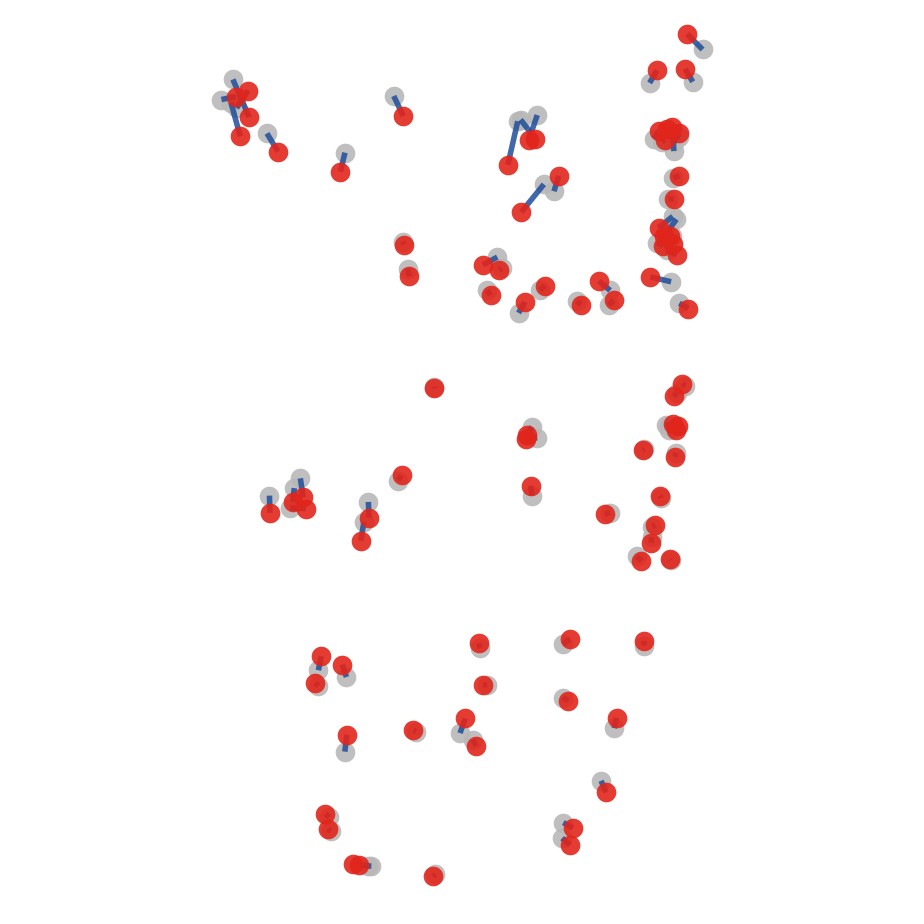}{20.8} & \mapauc{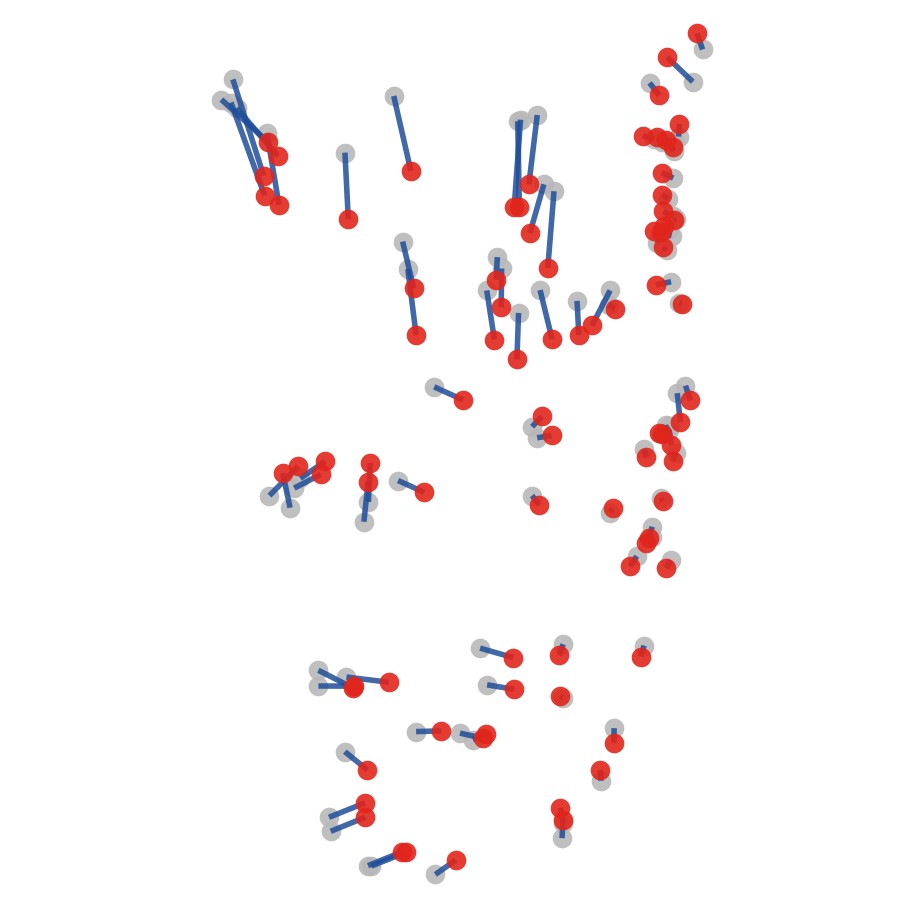}{17.5} \\[0pt]
\maprow{\textbf{+ \method}} & \mapname{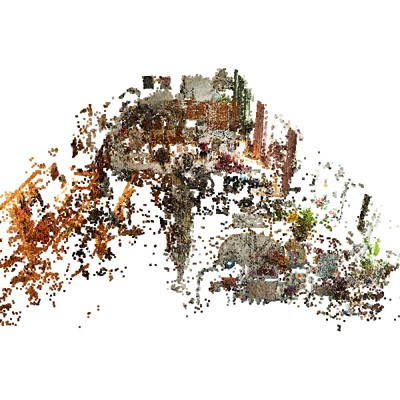}{\shortstack{Berlin 3\\medium overlap}} & \mapaucbf{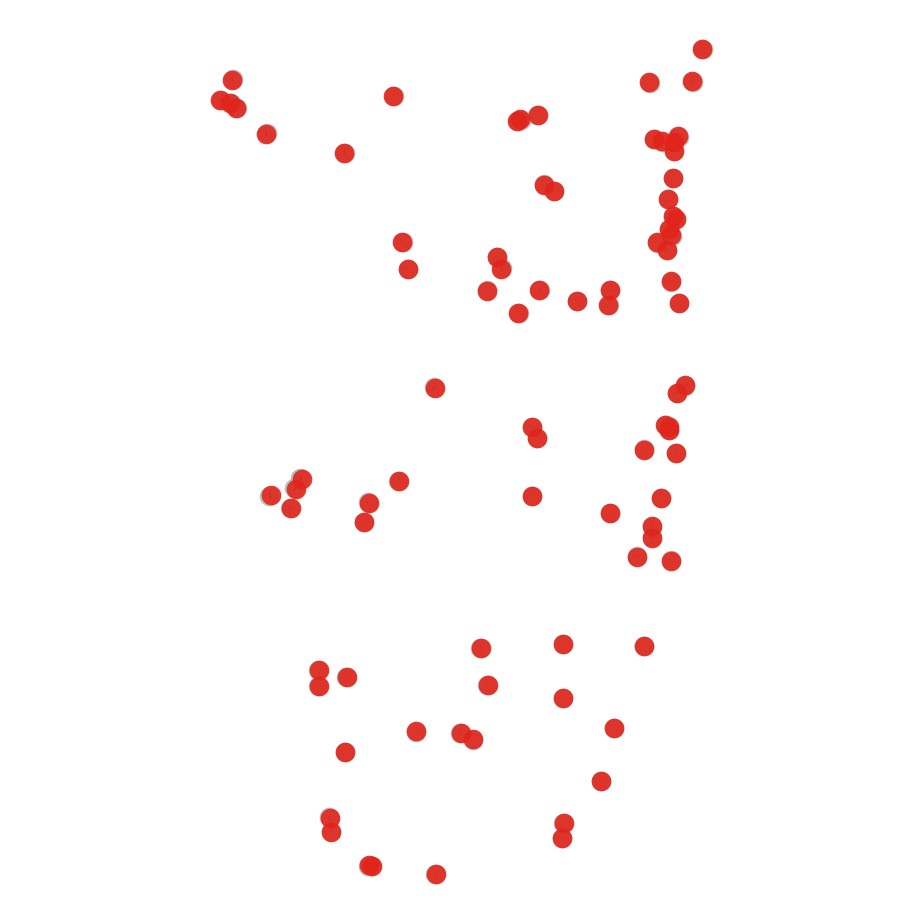}{93.0} & \mapaucbf{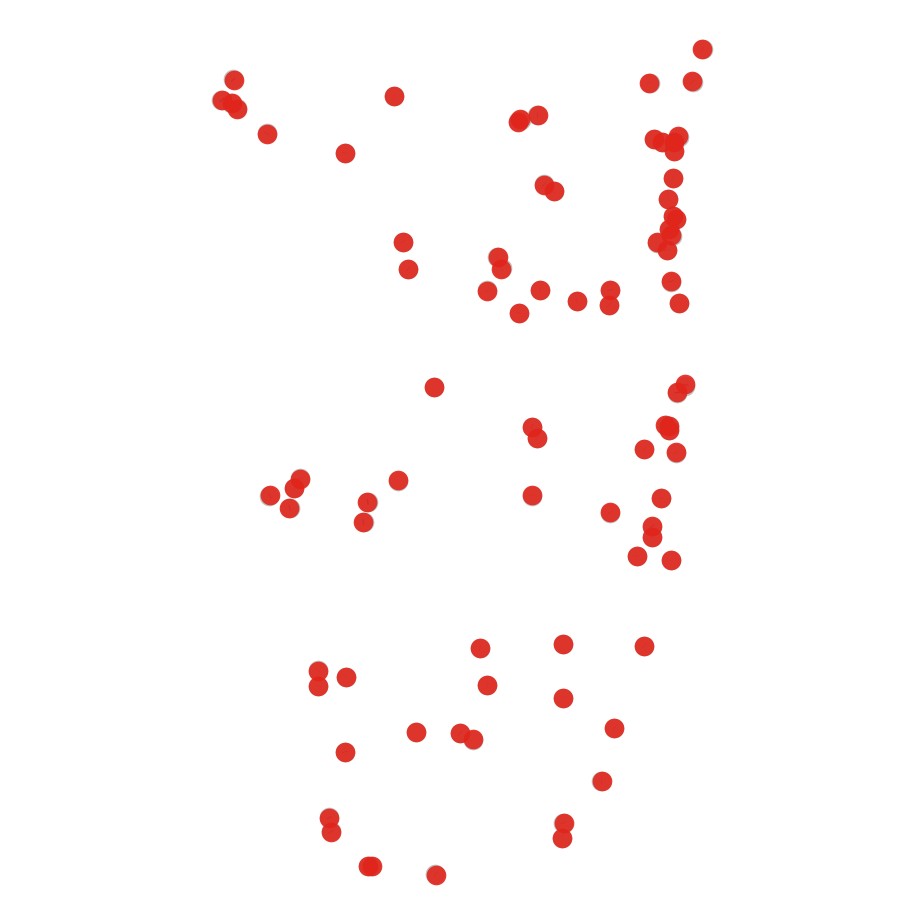}{92.9} & \mapaucbf{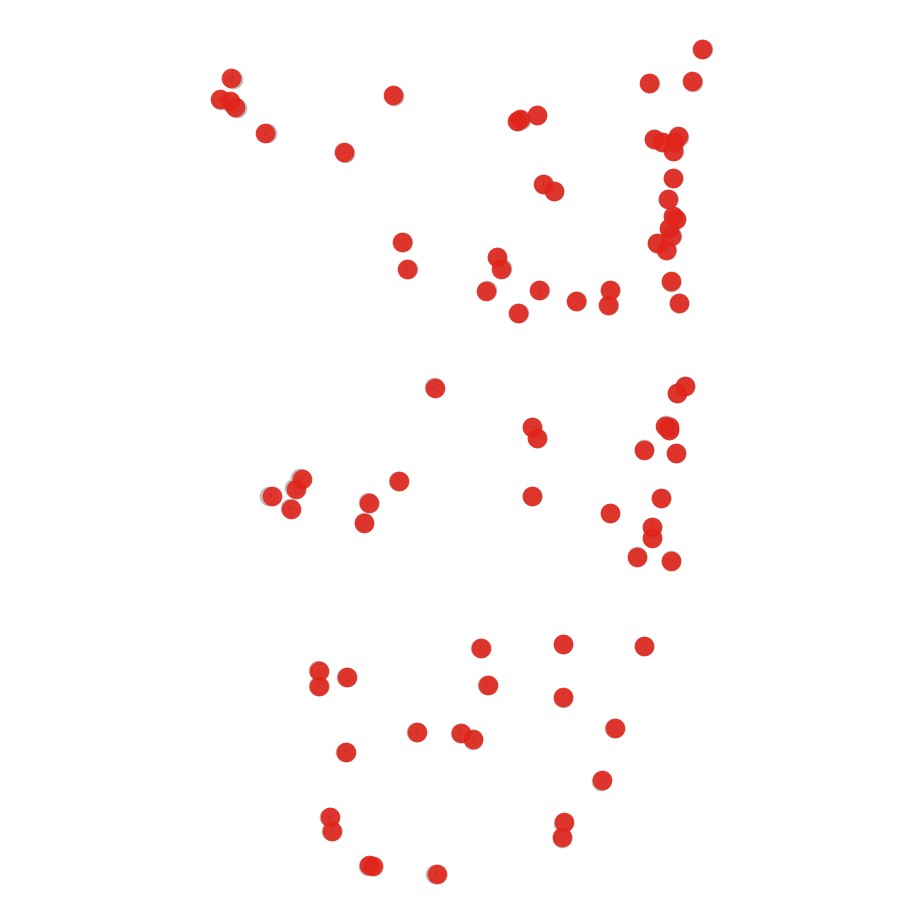}{91.9} & \mapaucbf{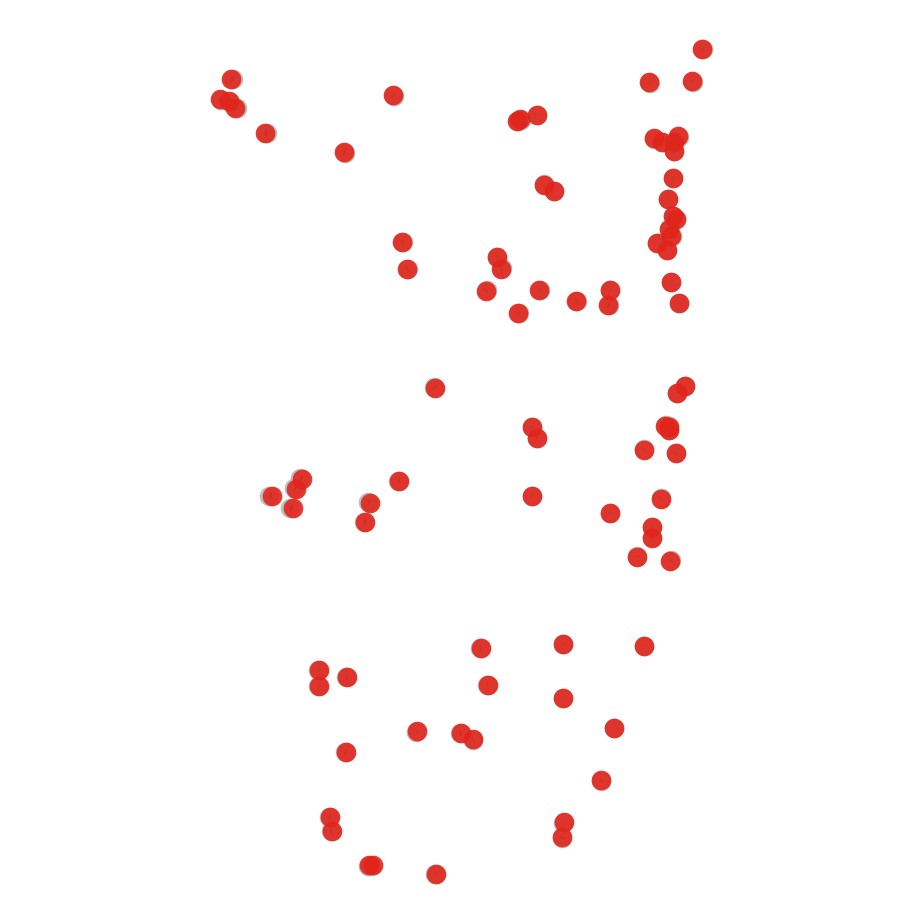}{91.8} & \mapaucbf{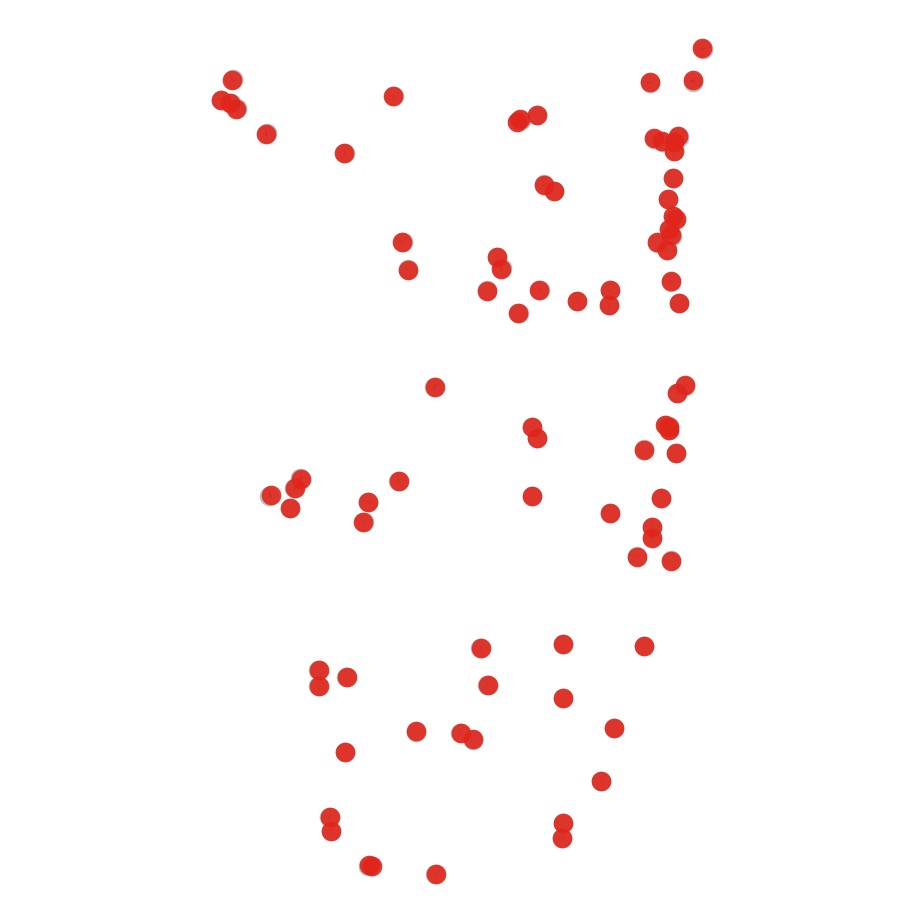}{90.8} \\[4pt]
\maprow{Initialization} & \mappanel{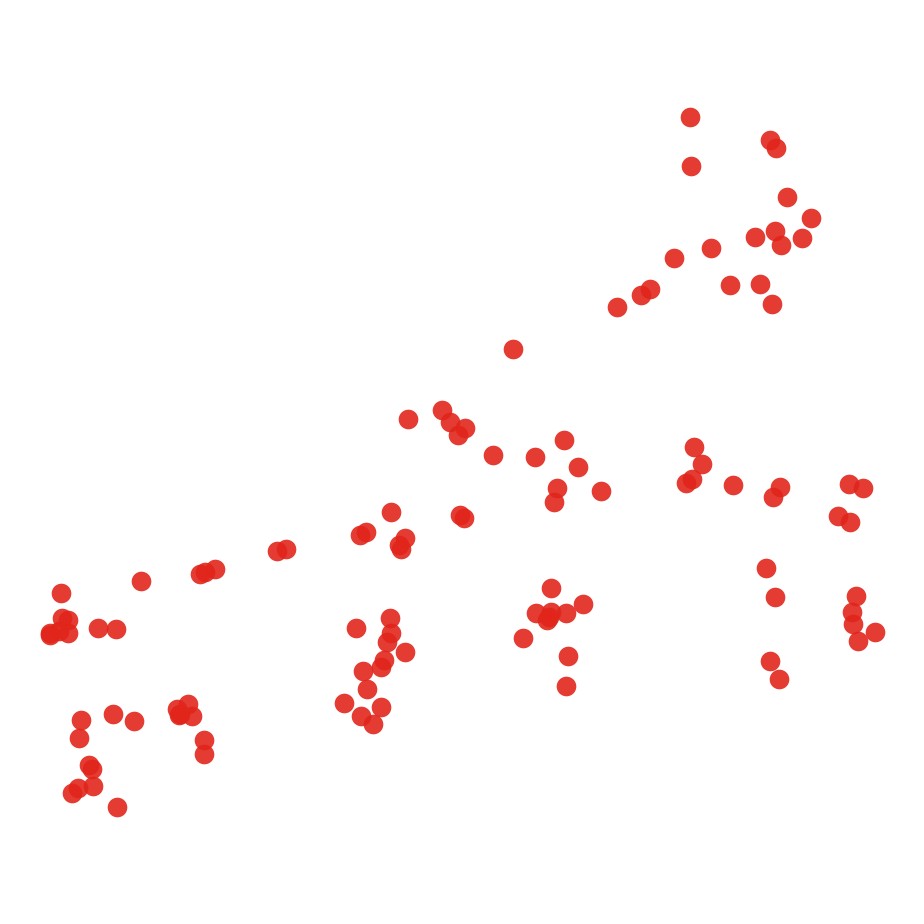} & \mapauc{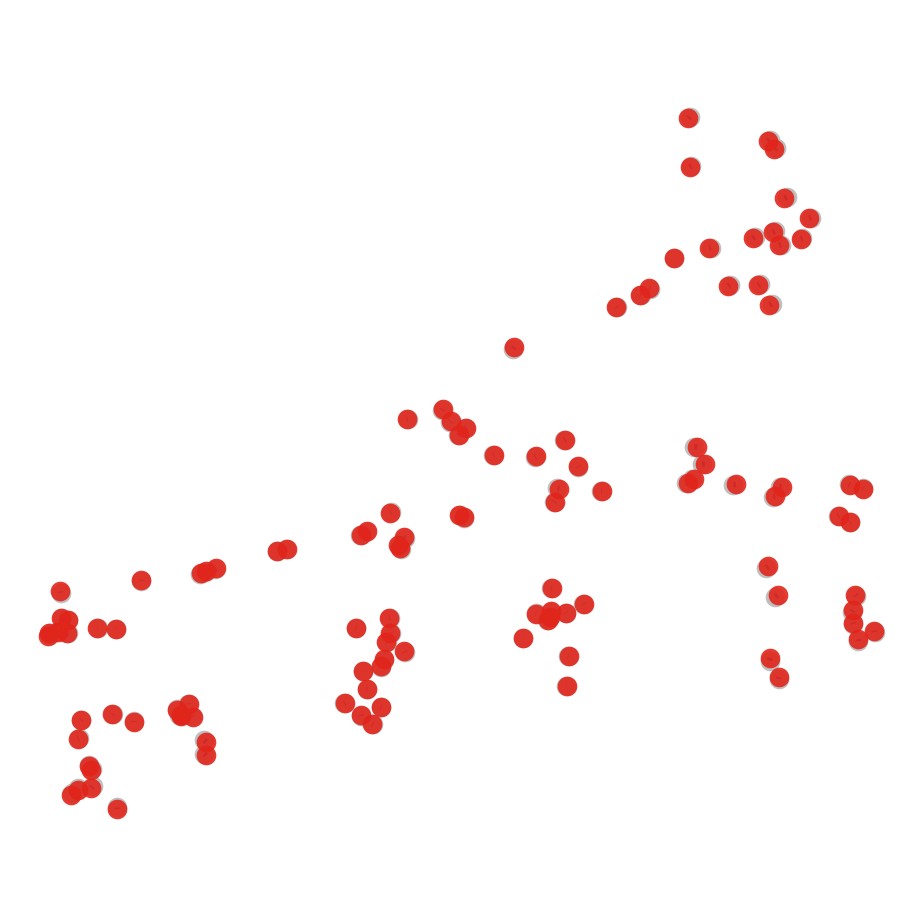}{89.8} & \mapauc{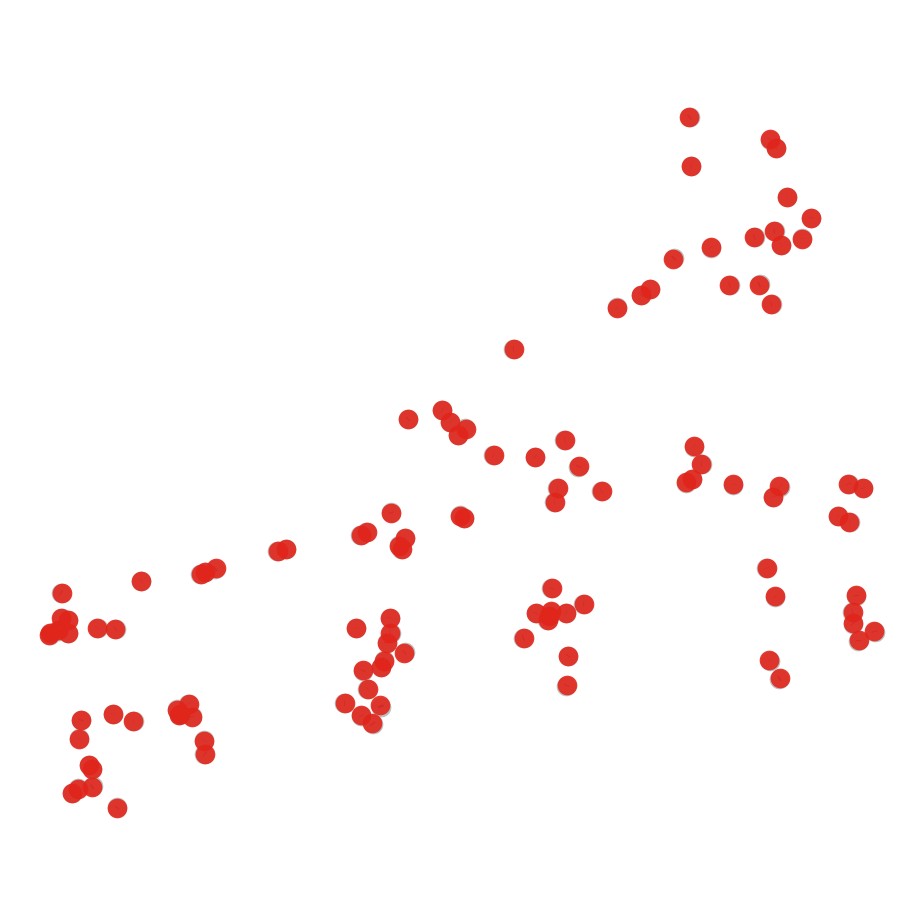}{92.2} & \mapauc{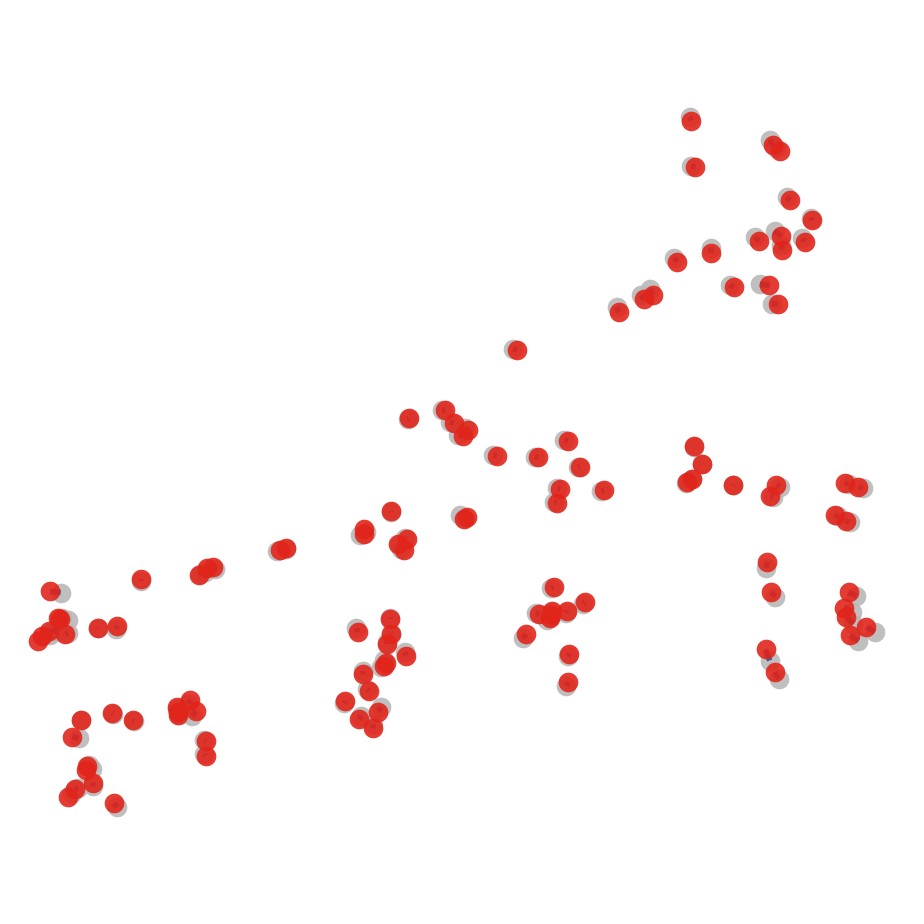}{63.3} & \mapauc{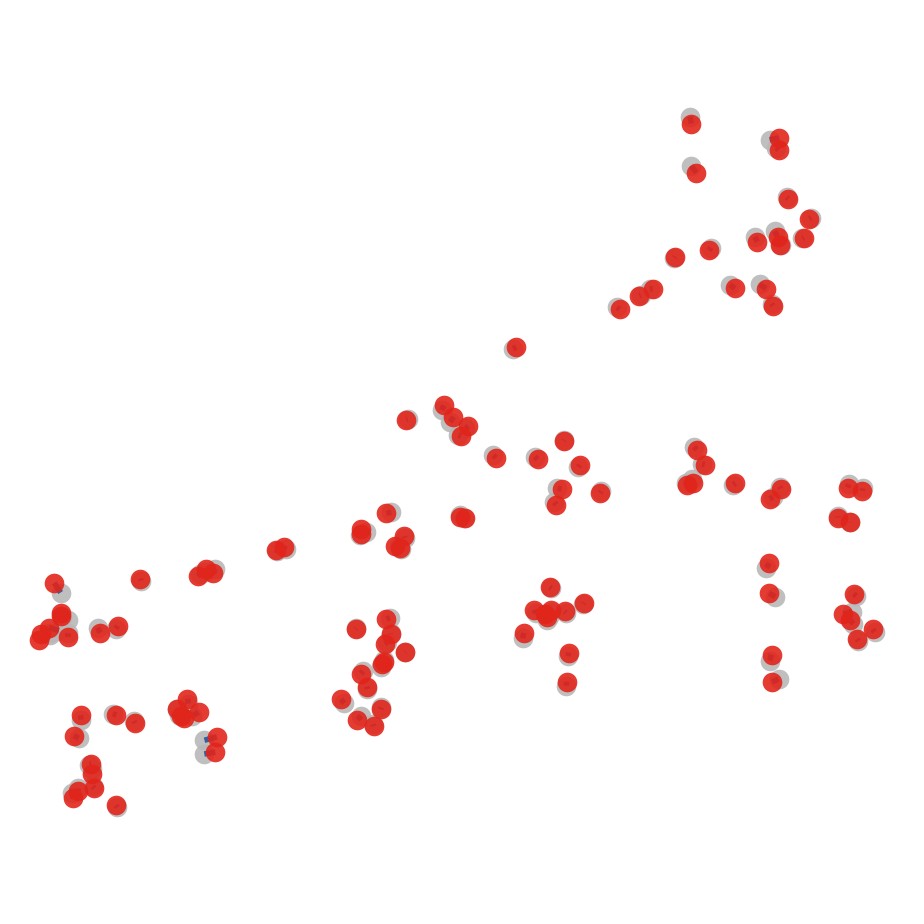}{53.1} & \mapauc{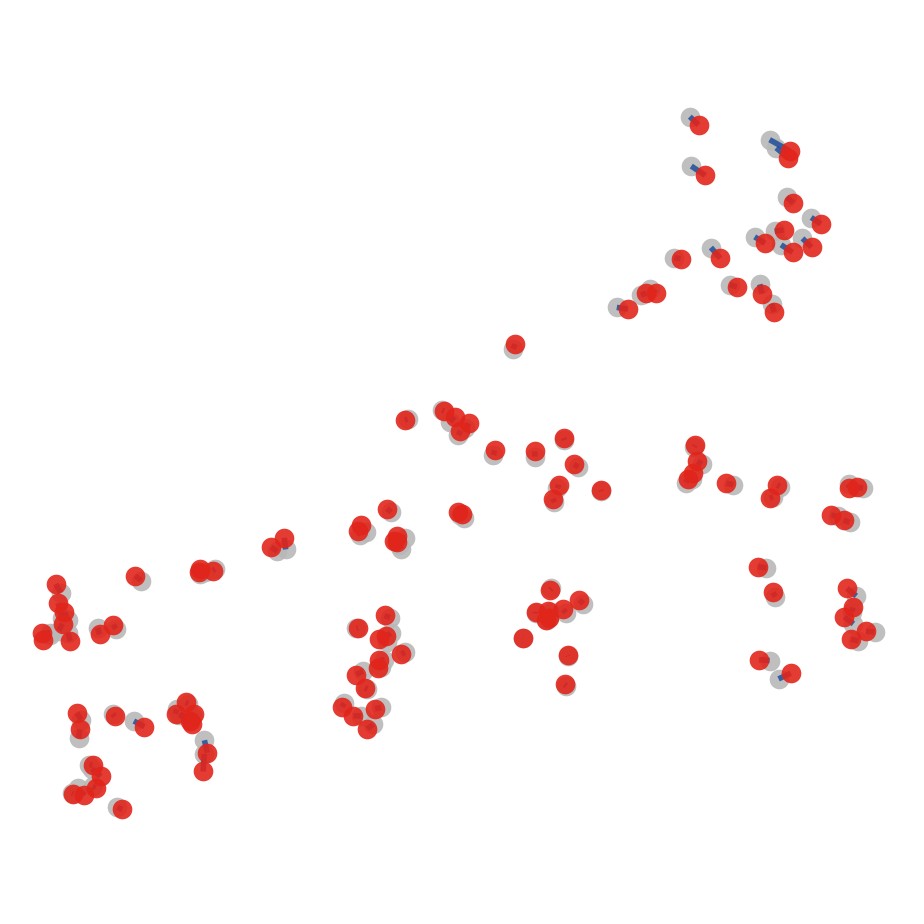}{46.4} \\[0pt]
\maprow{\textbf{+ \method}} & \mapname{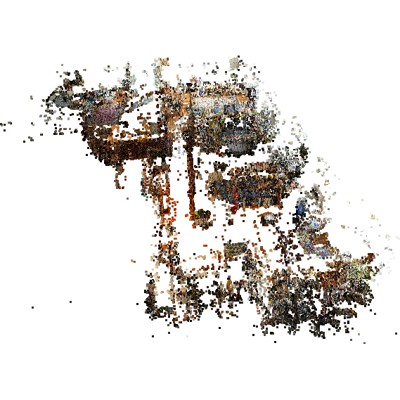}{\shortstack{New York 0\\high overlap}} & \mapaucbf{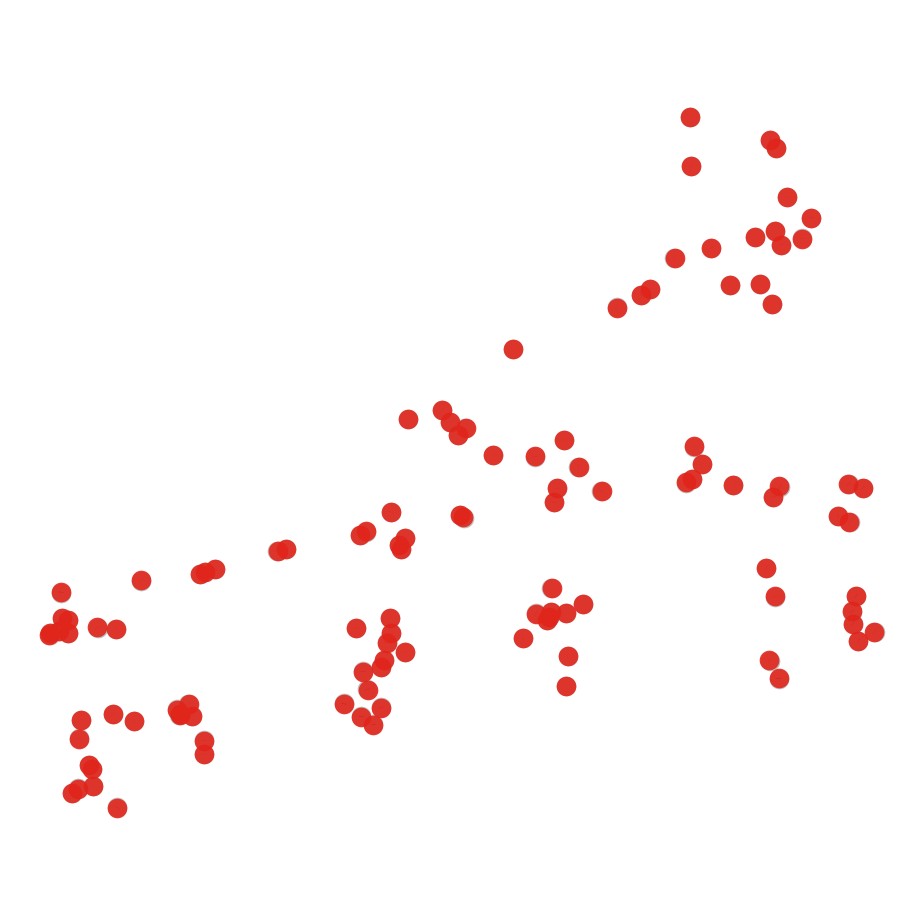}{96.4} & \mapaucbf{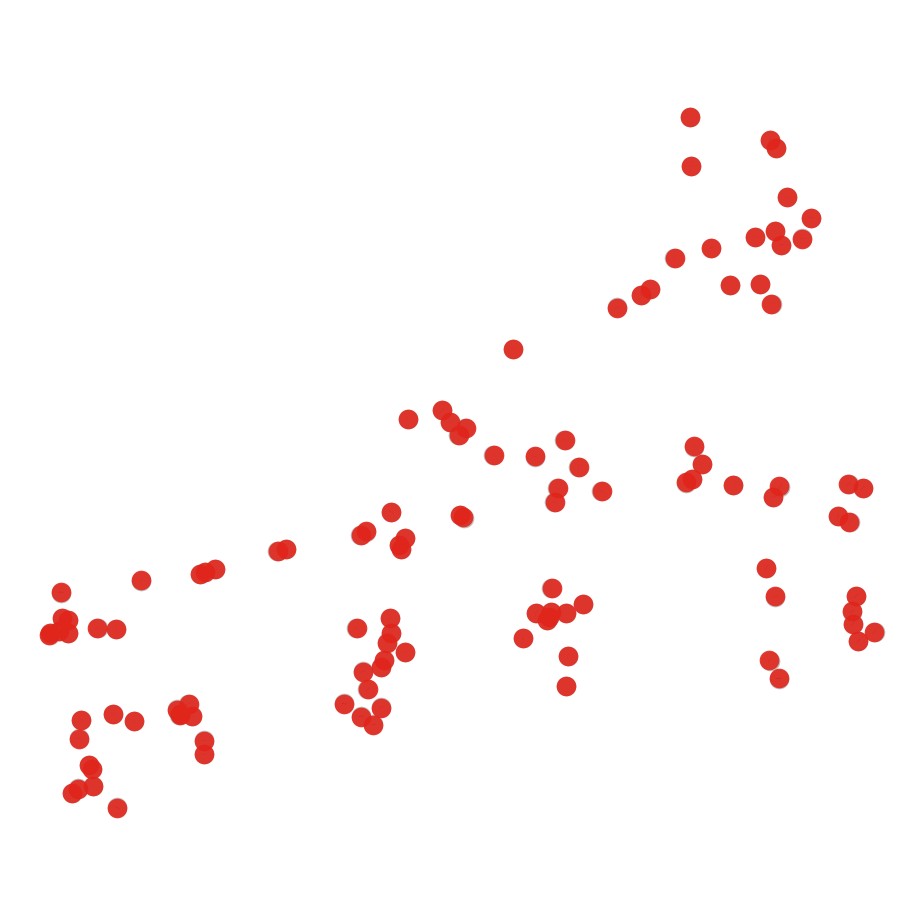}{96.3} & \mapaucbf{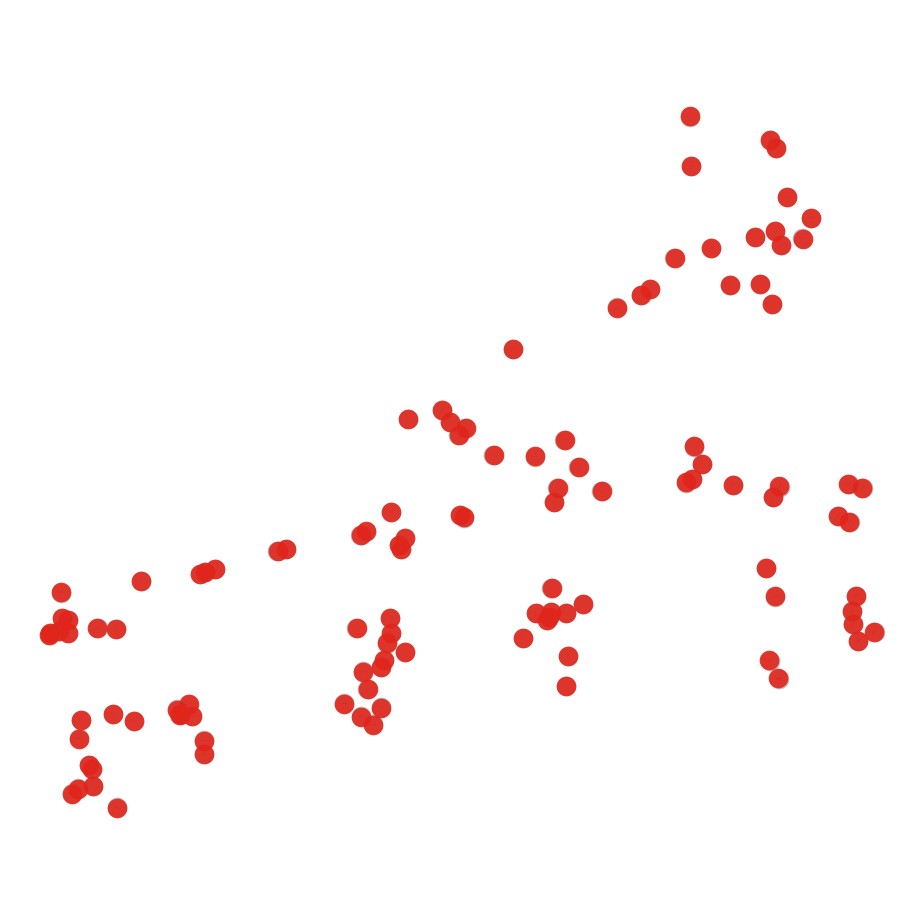}{96.3} & \mapaucbf{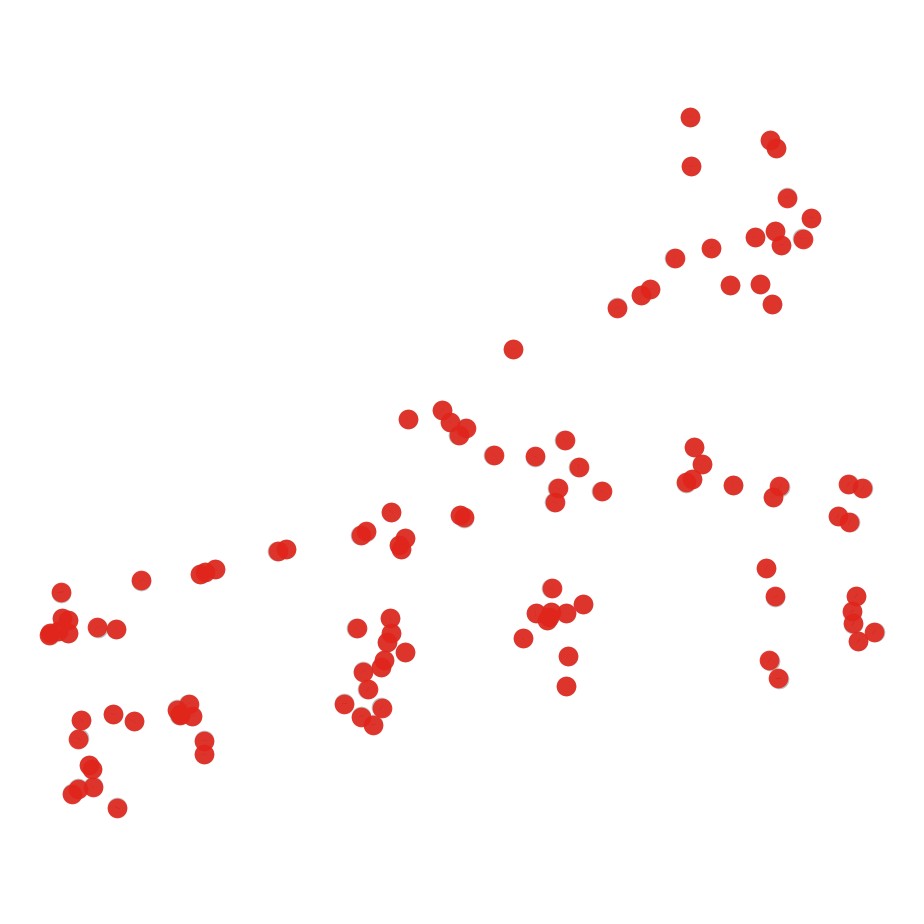}{96.2} & \mapaucbf{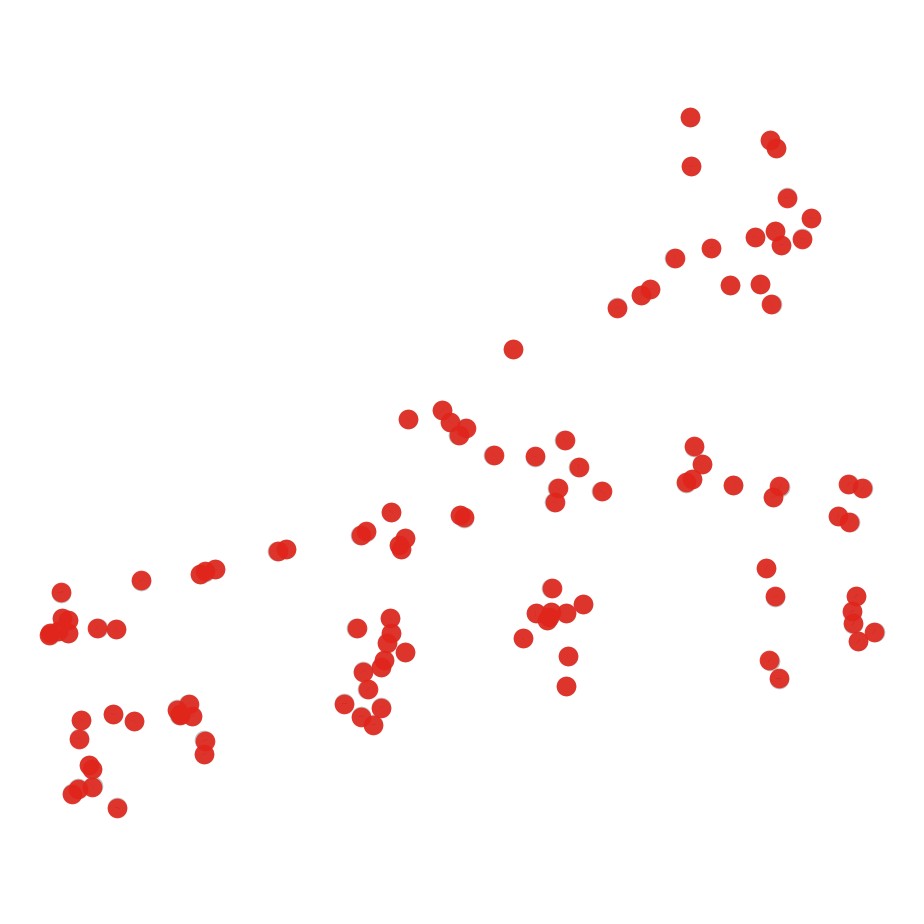}{96.5} \\[4pt]
\end{tabular}
\caption{Camera positions on SMERF collections with low, medium and high overlap: the Initialization cameras of each method (upper row of each collection) and the same cameras after \method (lower row), aligned to the ground truth; each estimated camera (red) is joined to its ground-truth position (grey) by a blue line.}
\label{fig:smerf_map}
\end{figure*}

\ifdefined\slamW\else\newlength{\slamW}\fi\setlength{\slamW}{1.00in}
\begin{figure*}[p]
\def\mappanel#1{{\setlength{\fboxsep}{0pt}\setlength{\fboxrule}{0.3pt}\fcolorbox{black!20}{white}{\includegraphics[width=\dimexpr\slamW-0.6pt\relax]{fig/eth3d_slam2_map/#1}}}}
\def\mapauc#1#2{\mappanel{#1}\llap{\raisebox{1.2pt}{\setlength{\fboxsep}{0.8pt}\colorbox{white}{\tiny AUC@3$^\circ$ #2}}\hspace{1.2pt}}}  
\def\mapaucbf#1#2{\mappanel{#1}\llap{\raisebox{1.2pt}{\setlength{\fboxsep}{0.8pt}\colorbox{white}{\tiny\bfseries AUC@3$^\circ$ #2}}\hspace{1.2pt}}}
\def\maprow#1{\rotatebox{90}{\scriptsize #1}}
\def\mapname#1#2{\raisebox{0pt}[0pt][0pt]{\parbox{\slamW}{\centering\includegraphics[width=0.8\slamW]{fig/eth3d_slam2_map/#1}\\[1pt]\footnotesize\bfseries #2}}}  
\centering
\setlength{\tabcolsep}{0.8pt}
\renewcommand{\arraystretch}{0.7}
\begin{tabular}{@{}>{\centering\arraybackslash}m{0.9em} >{\centering\arraybackslash}m{\slamW}@{\hspace{4pt}} *{5}{>{\centering\arraybackslash}m{\slamW}} @{}}
 & \small\bfseries GT & \small\bfseries GLOMAP & \small\bfseries VidMap & \small\bfseries MERG3R$^\S$ & \small\bfseries DVLT$^\ddagger$ & \small\bfseries VGGT-$\Omega^\ddagger$ \\[2pt]

\maprow{Initialization} & \mappanel{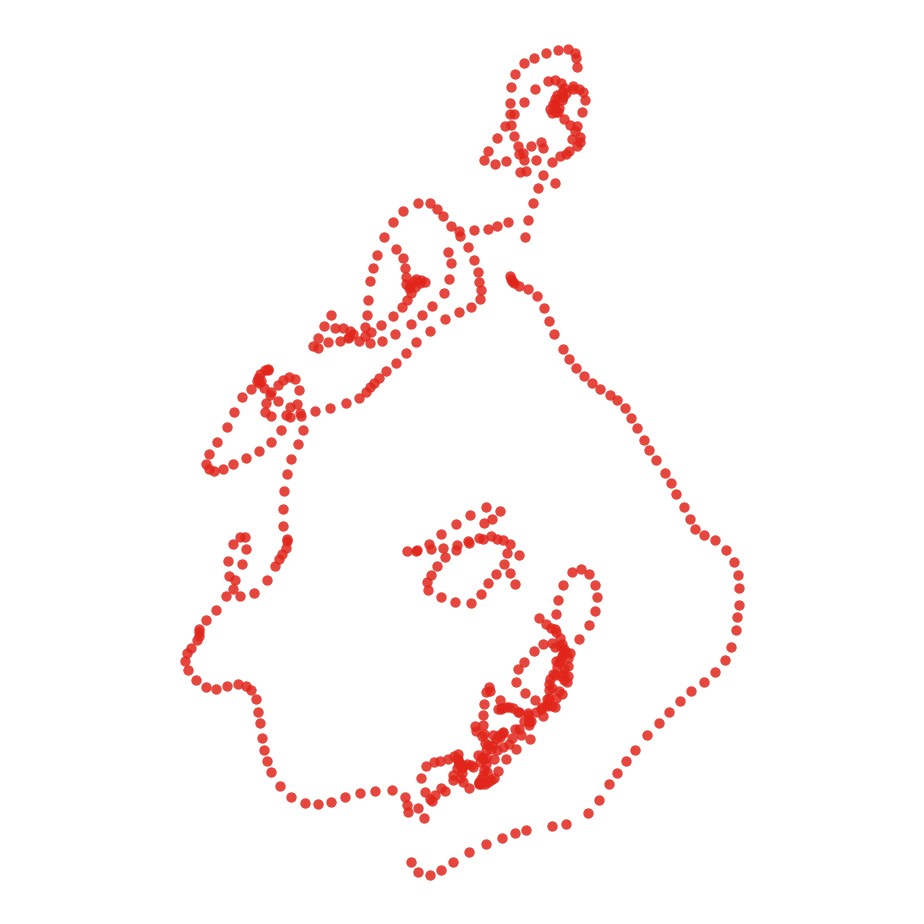} & \mapauc{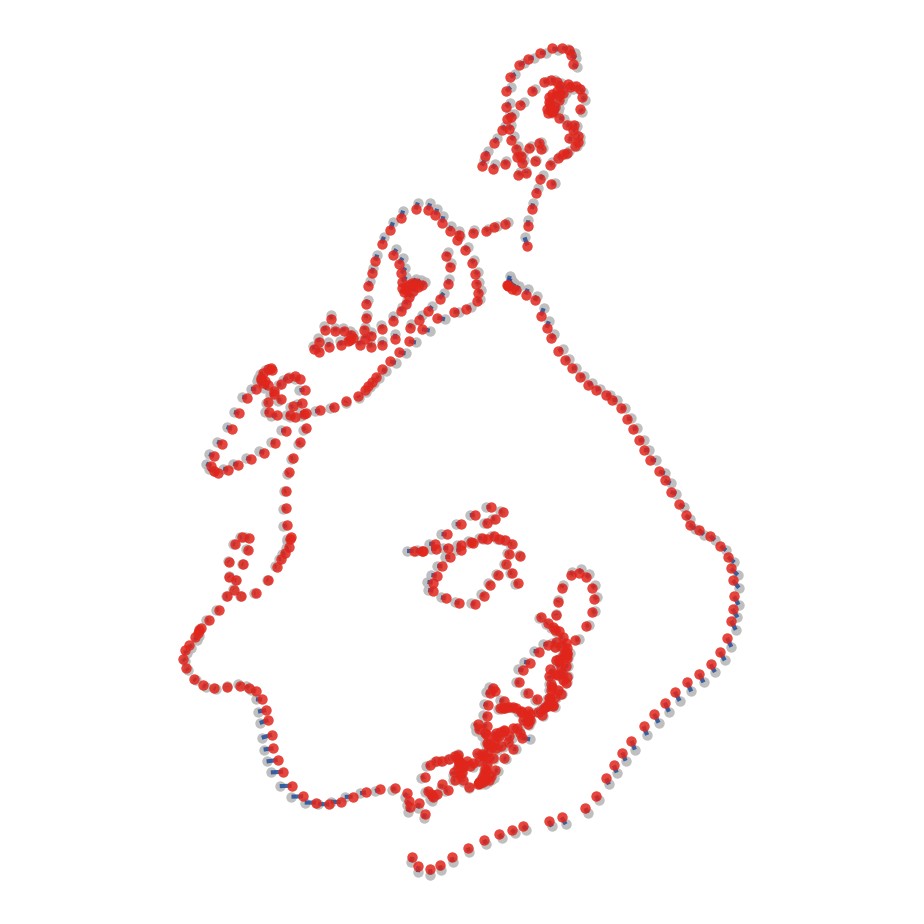}{11.2} & \mapauc{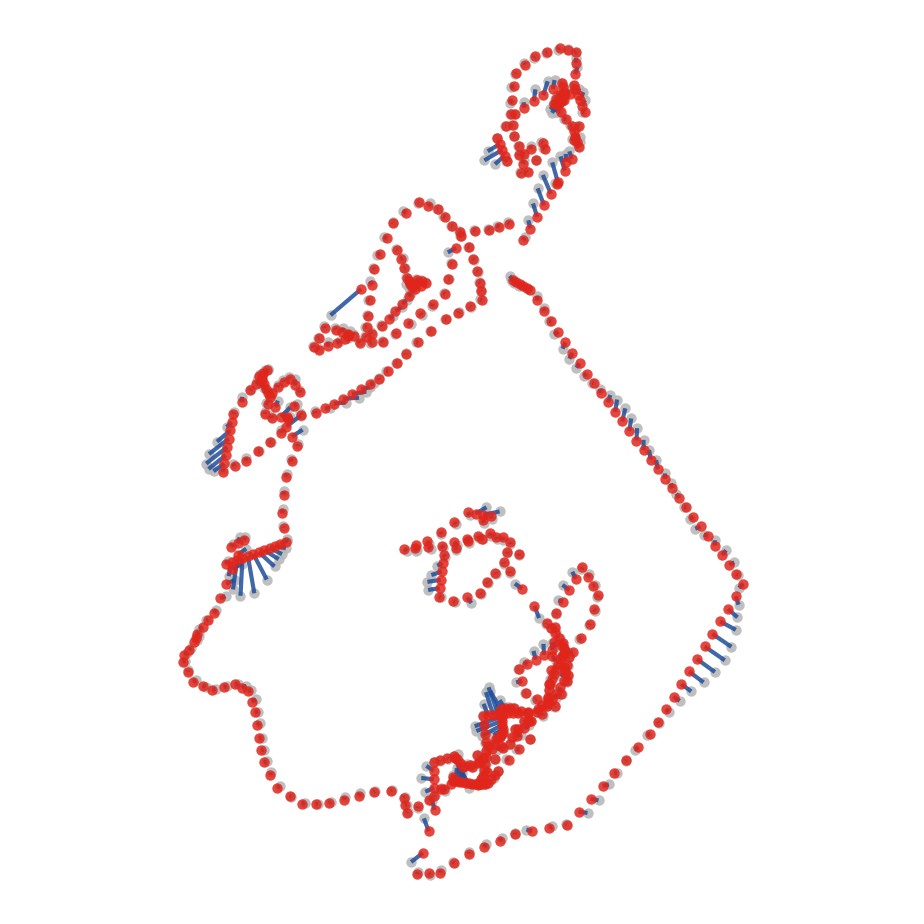}{26.9} & \mapauc{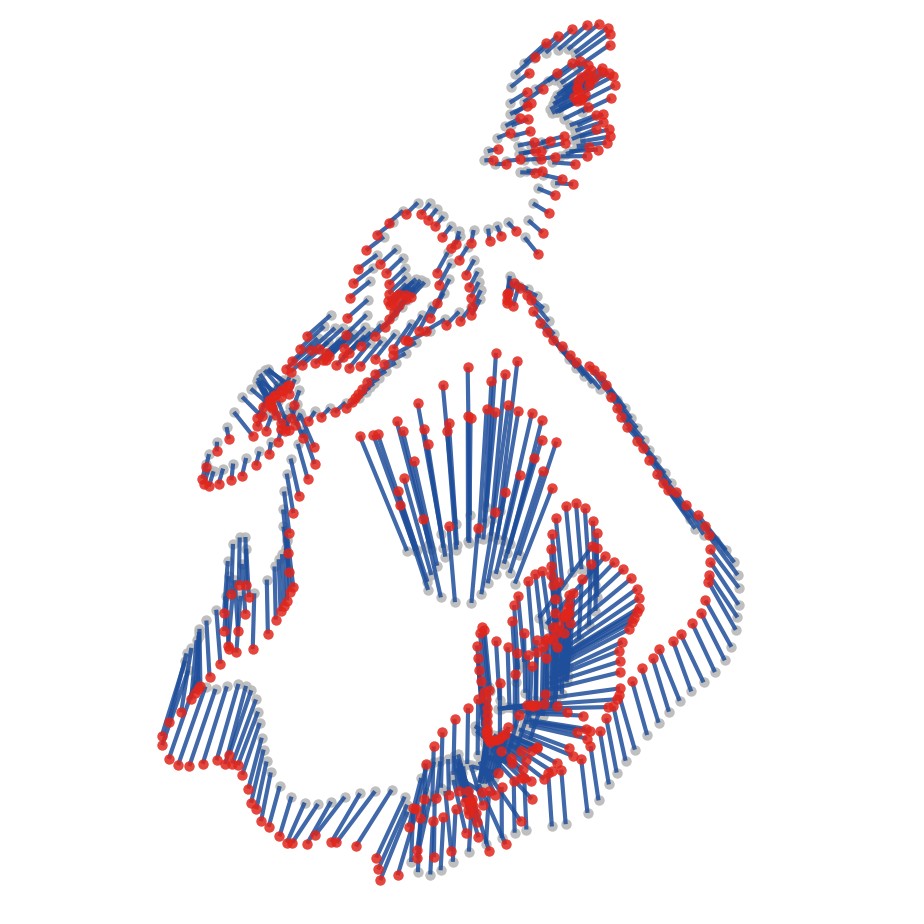}{2.7} & \mapauc{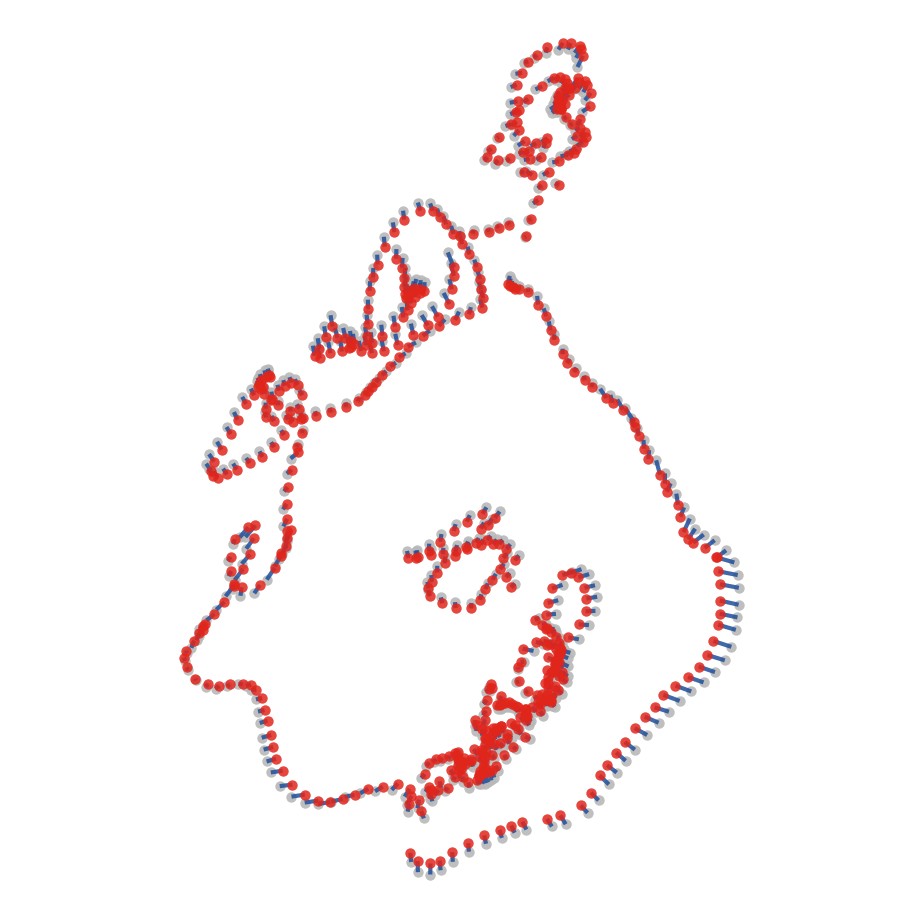}{6.6} & \mapauc{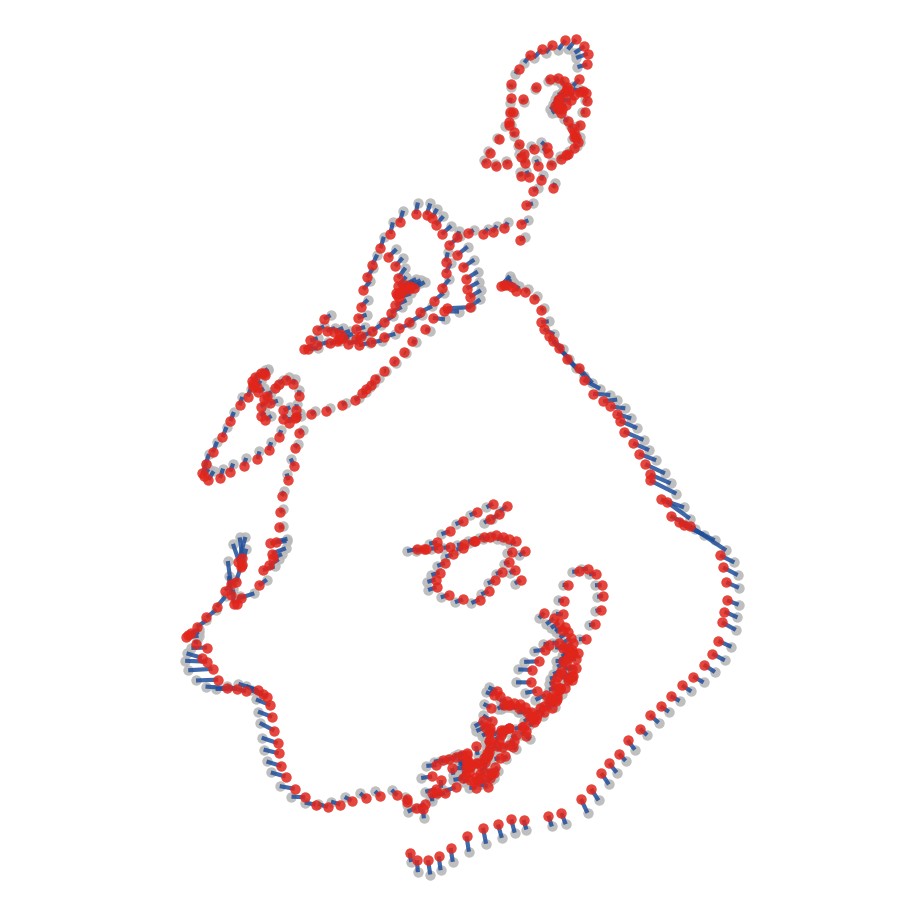}{4.9} \\[0pt]
\maprow{\textbf{+ \method}} & \mapname{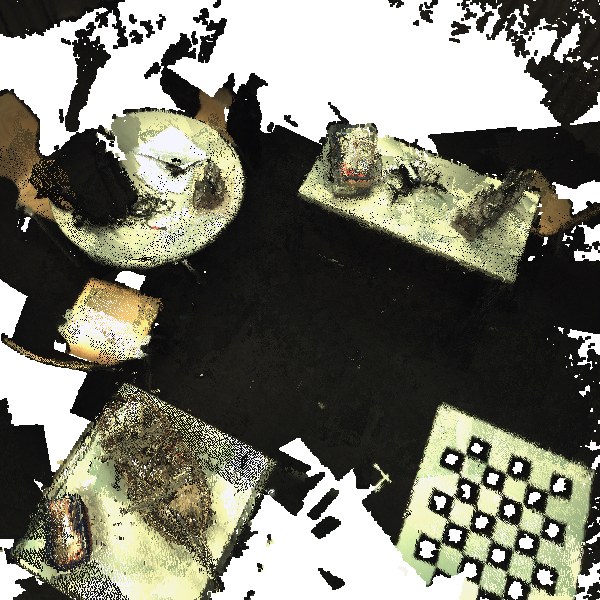}{desk 3} & \mapaucbf{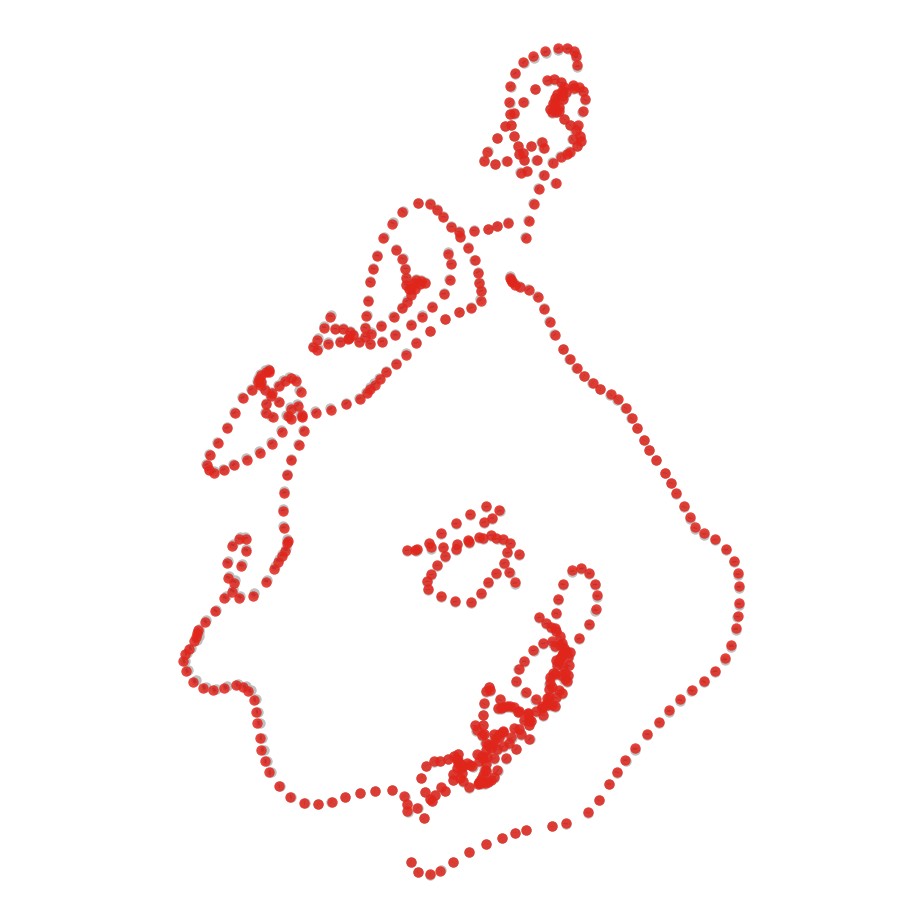}{81.6} & \mapaucbf{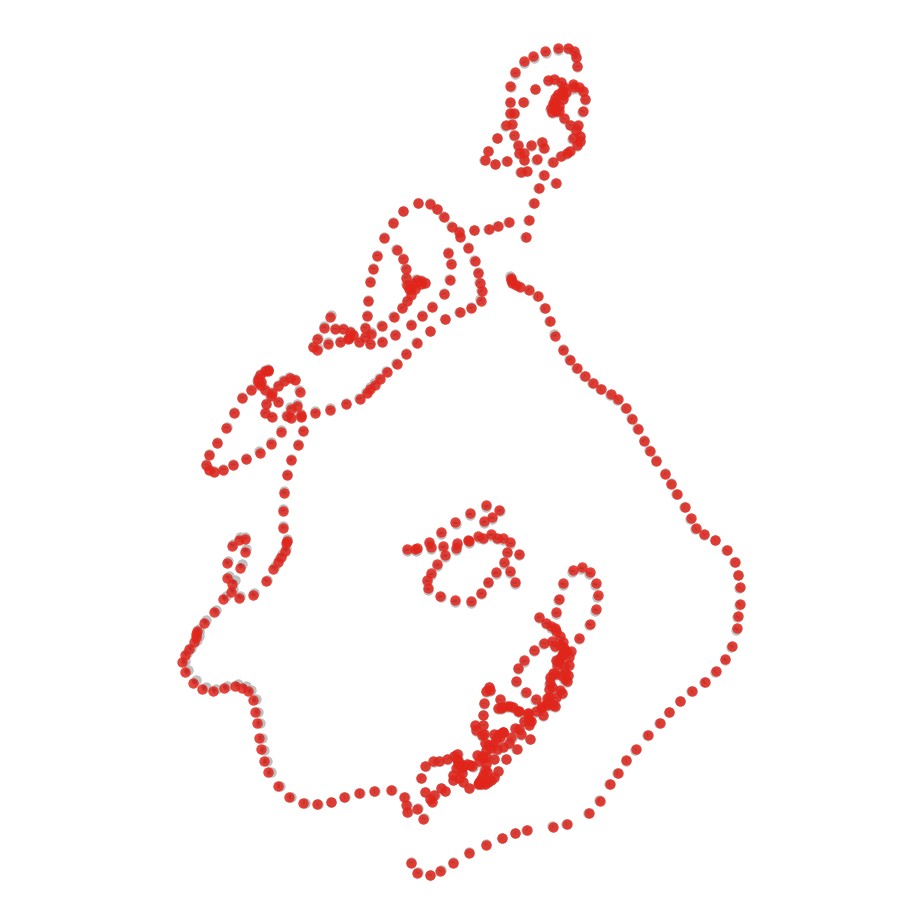}{83.5} & \mapaucbf{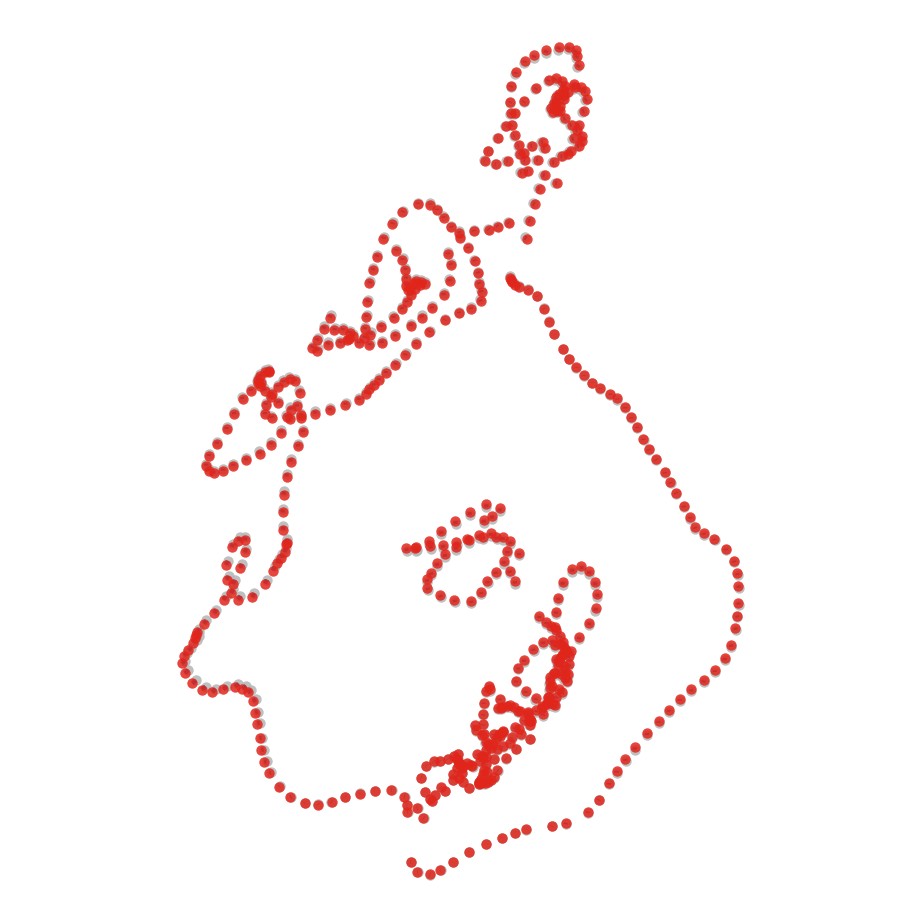}{82.0} & \mapaucbf{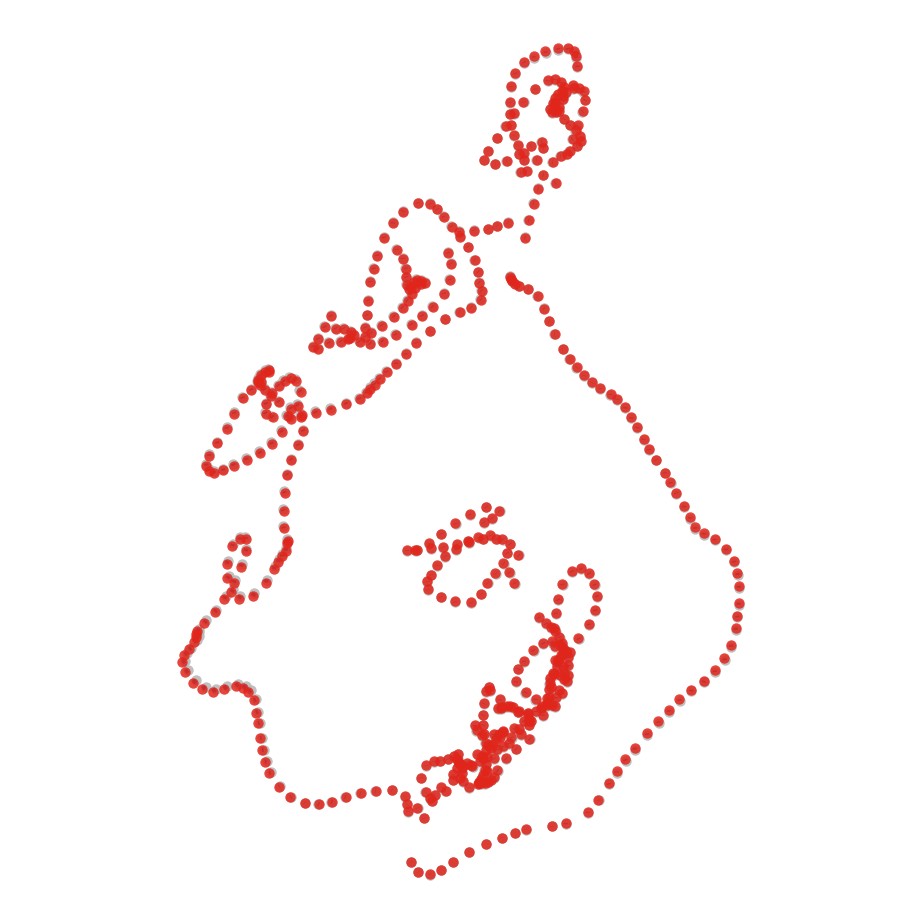}{84.1} & \mapaucbf{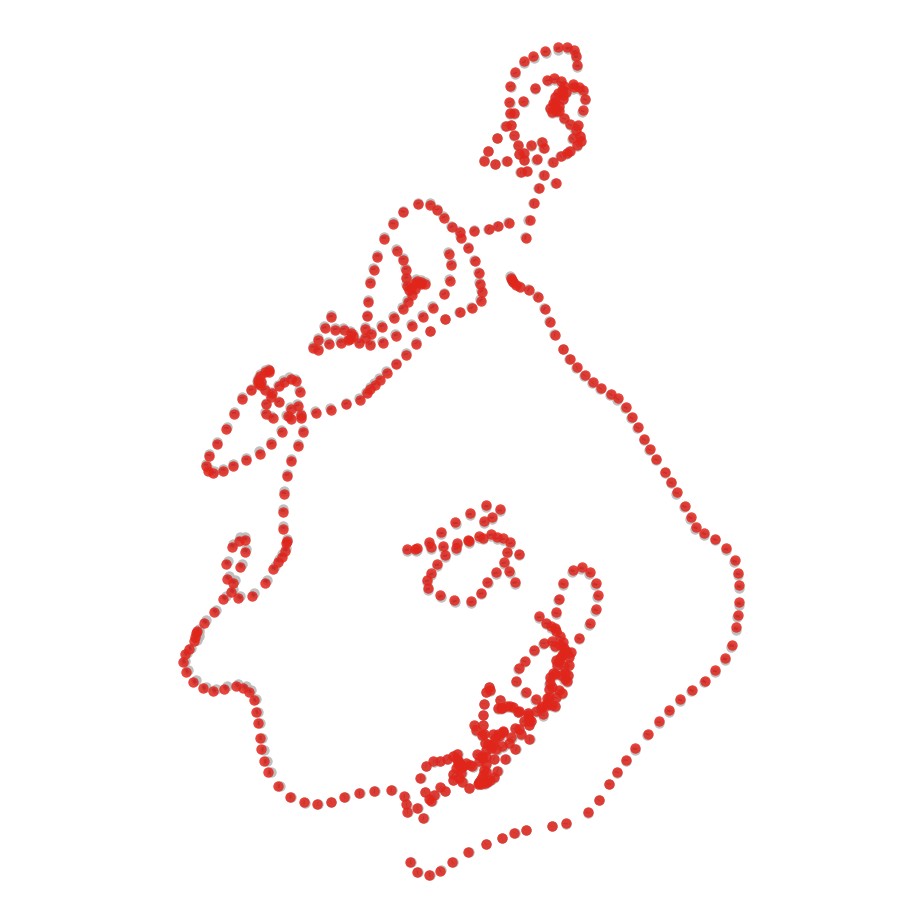}{79.5} \\[4pt]
\maprow{Initialization} & \mappanel{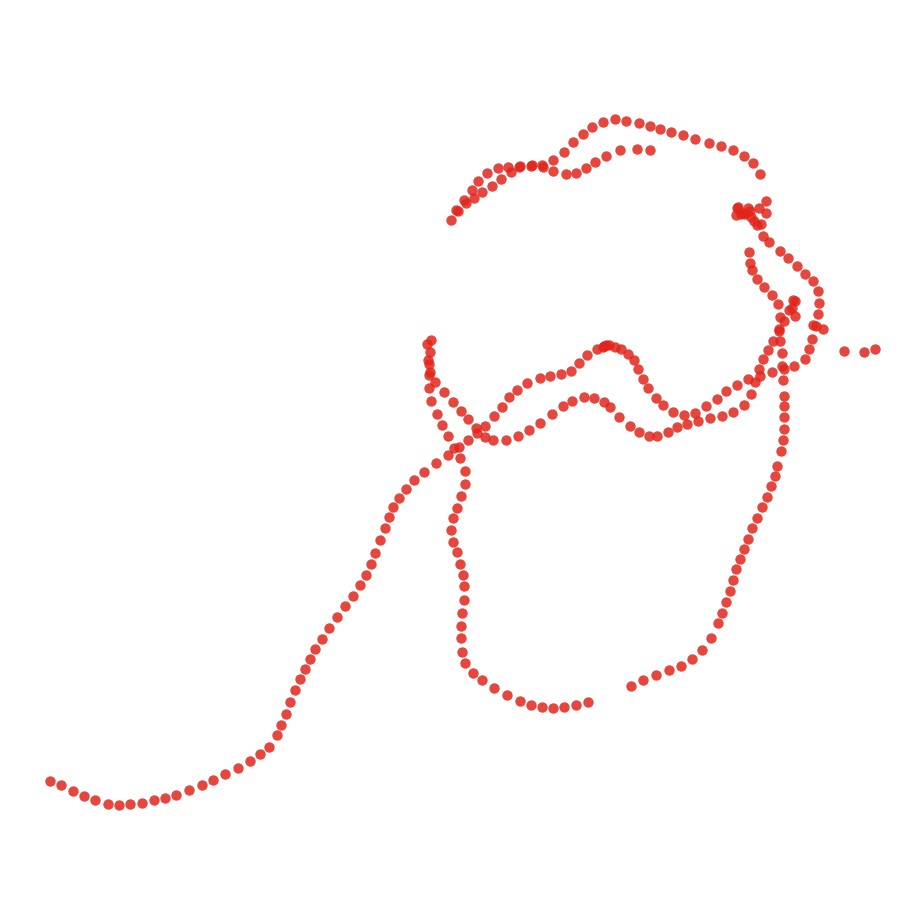} & \mapauc{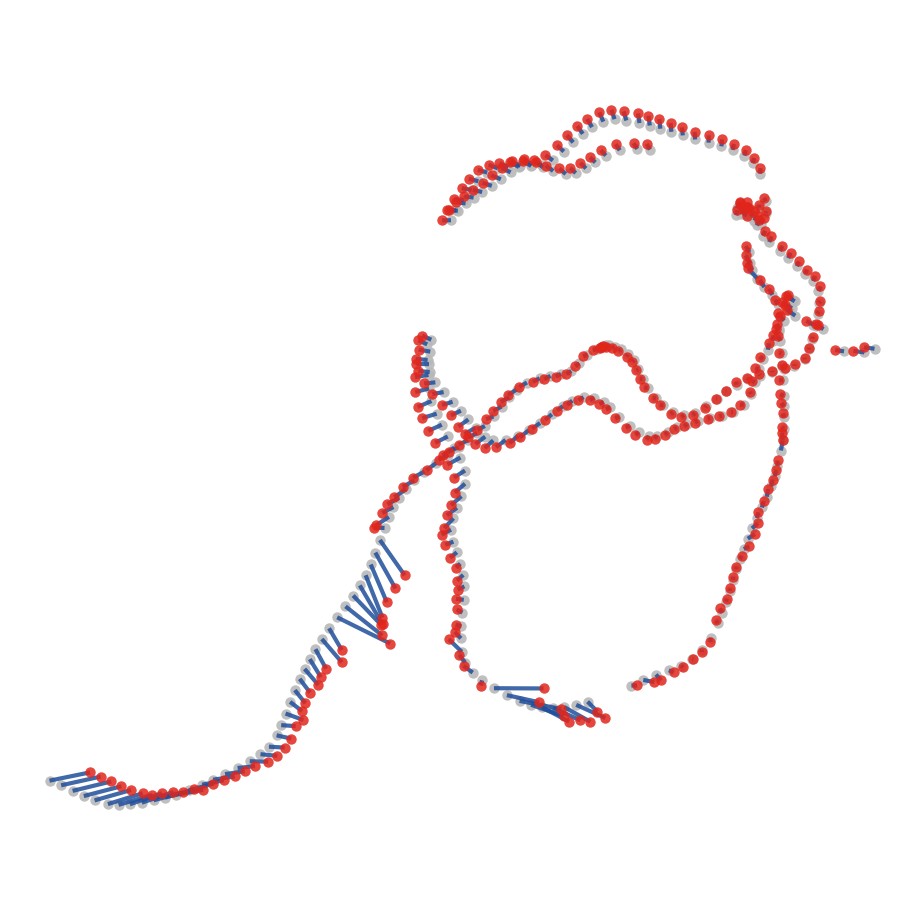}{5.2} & \mapauc{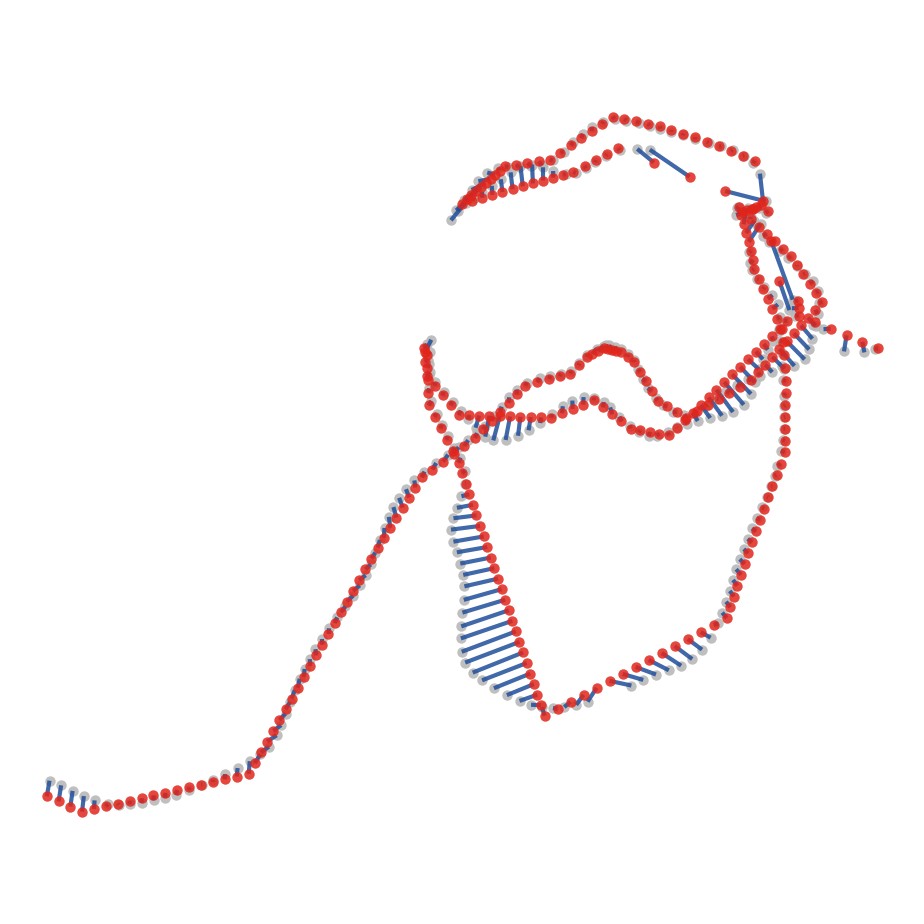}{14.3} & \mapauc{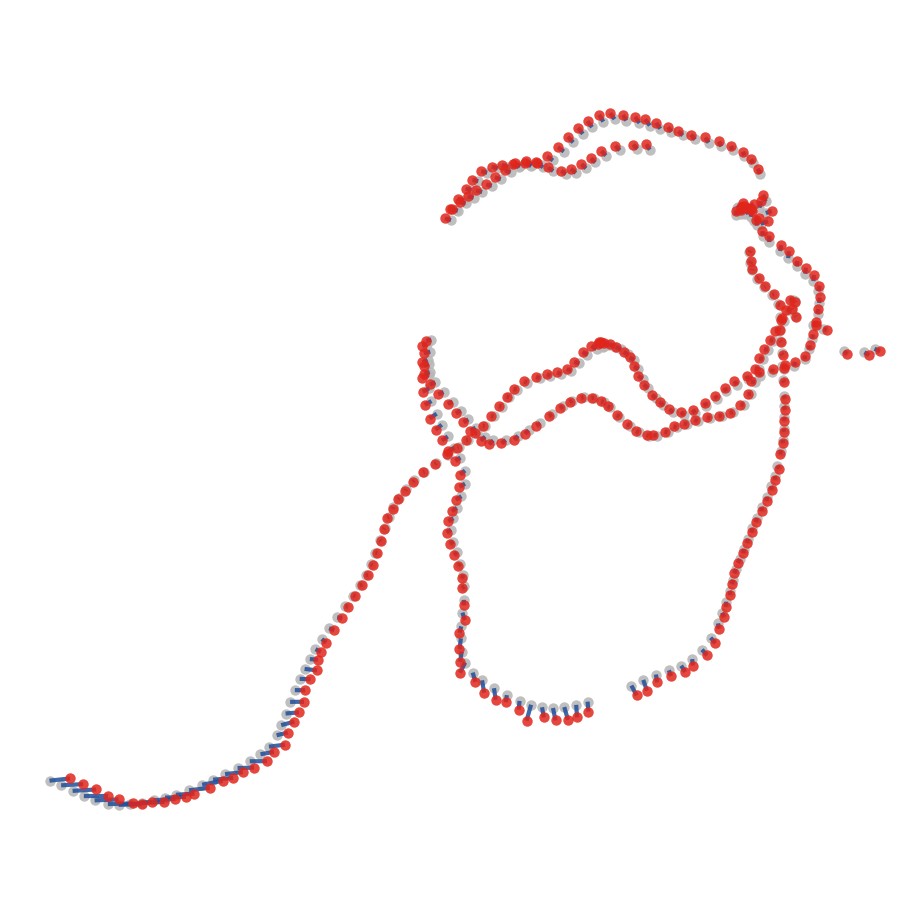}{4.5} & \mapauc{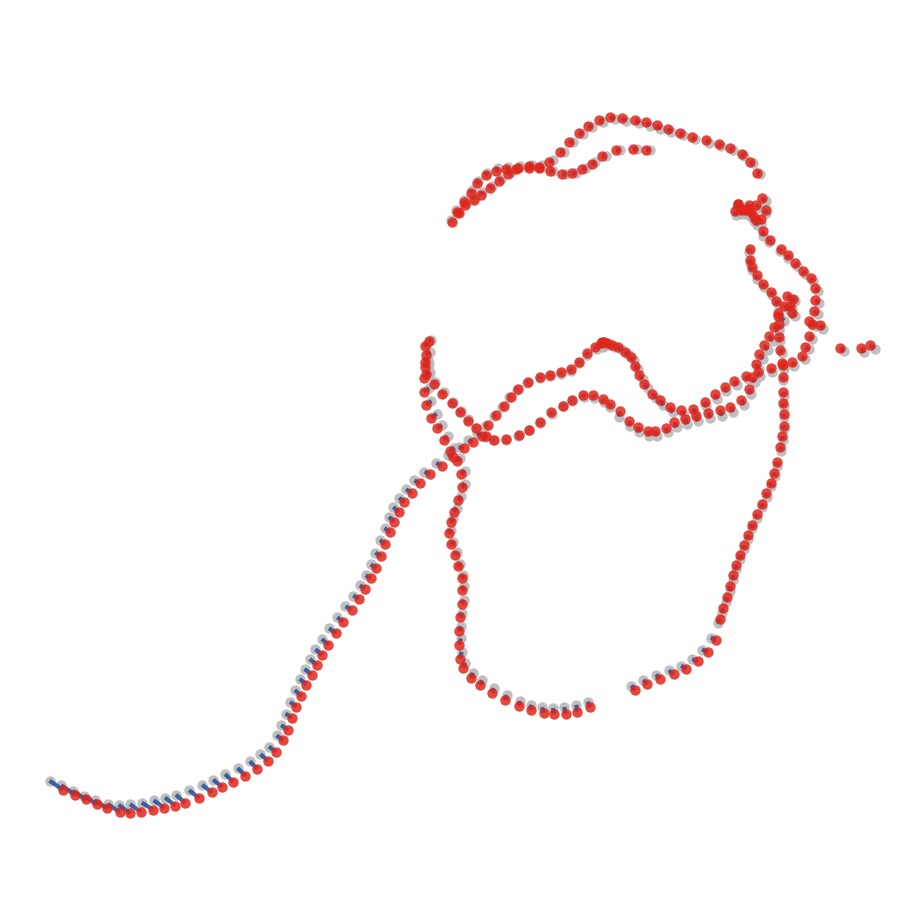}{6.3} & \mapauc{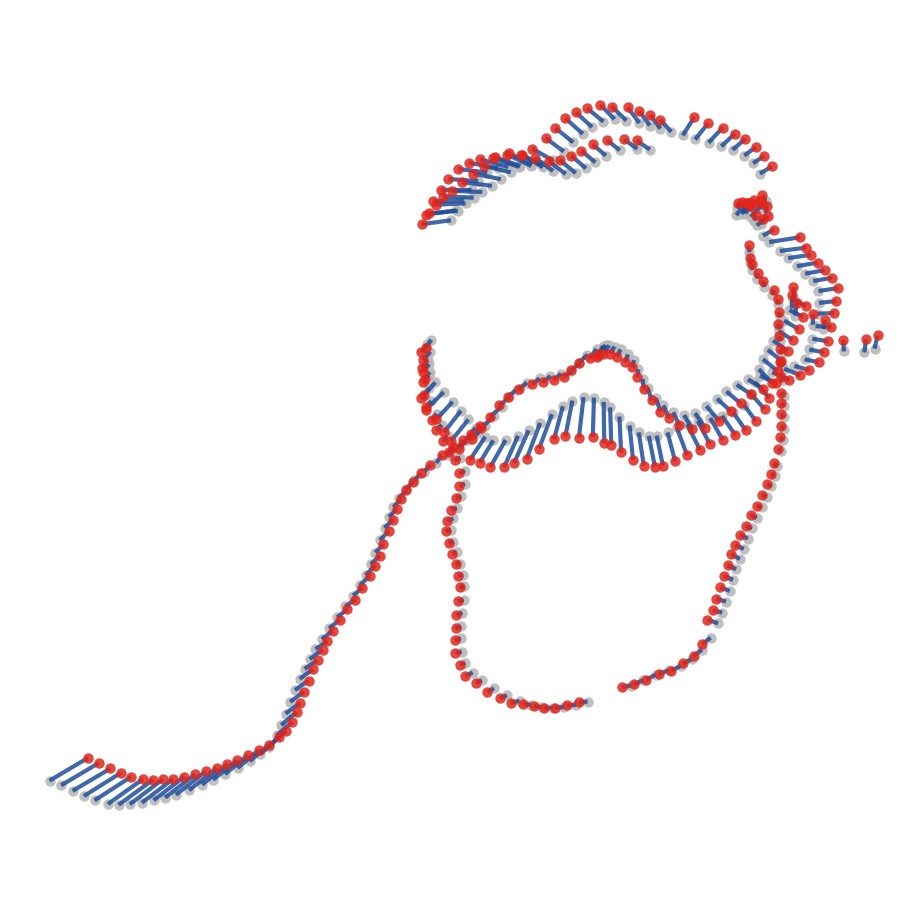}{2.9} \\[0pt]
\maprow{\textbf{+ \method}} & \mapname{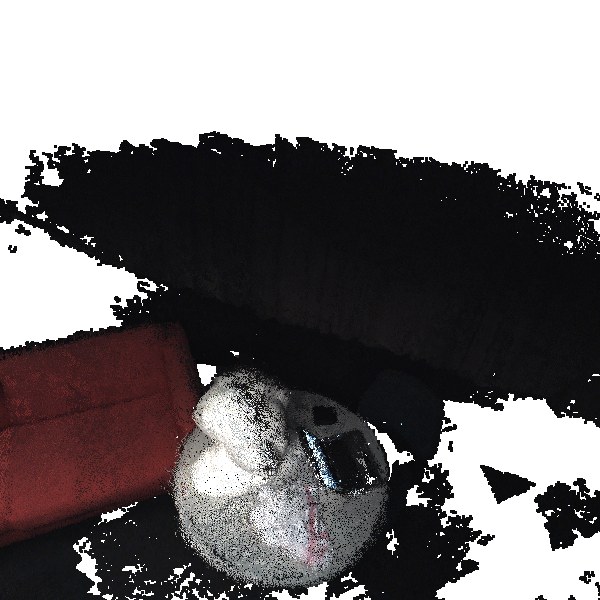}{einstein 2} & \mapaucbf{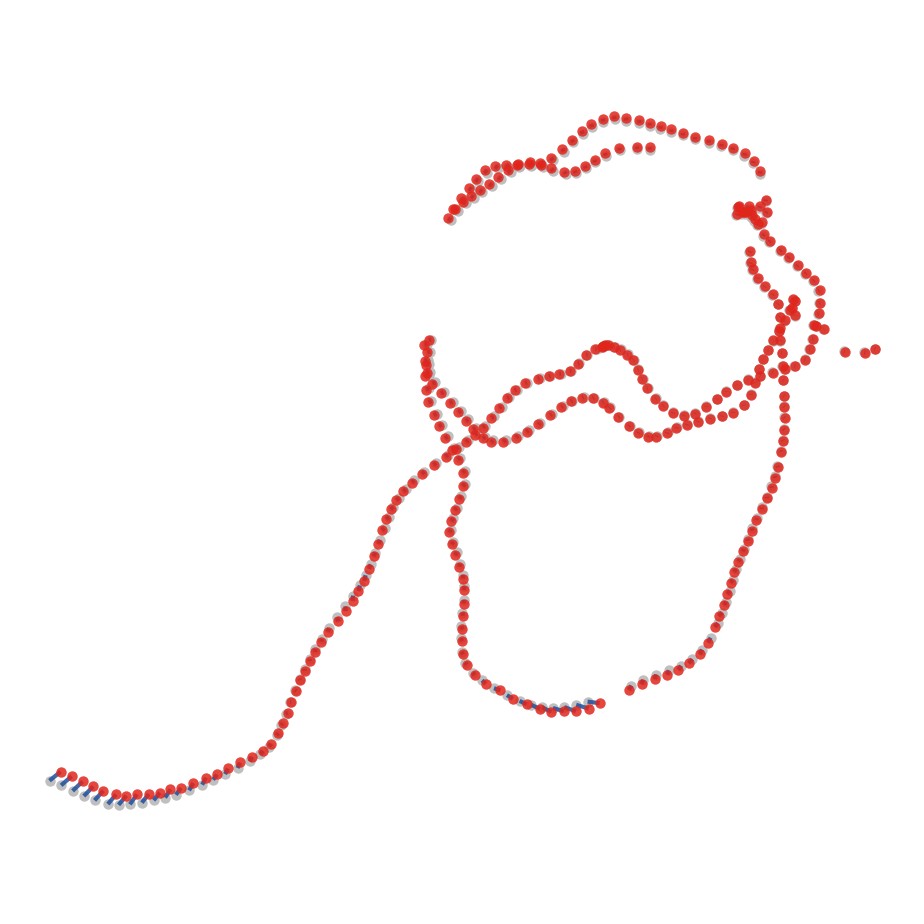}{76.6} & \mapaucbf{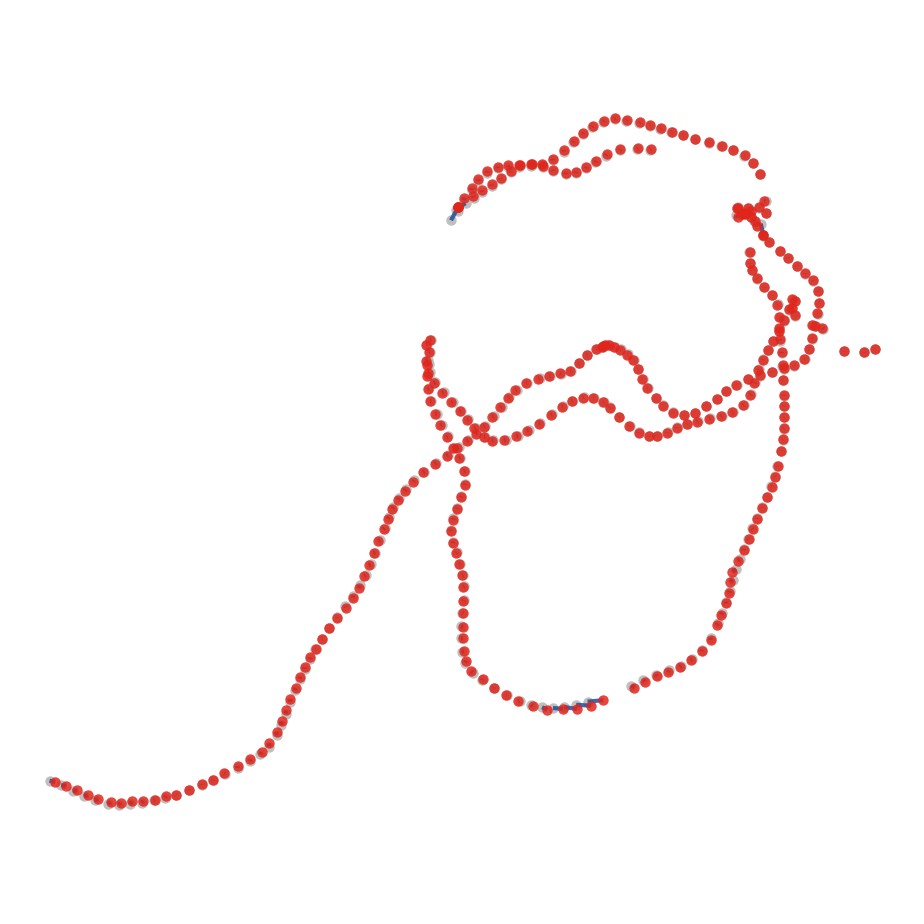}{78.3} & \mapaucbf{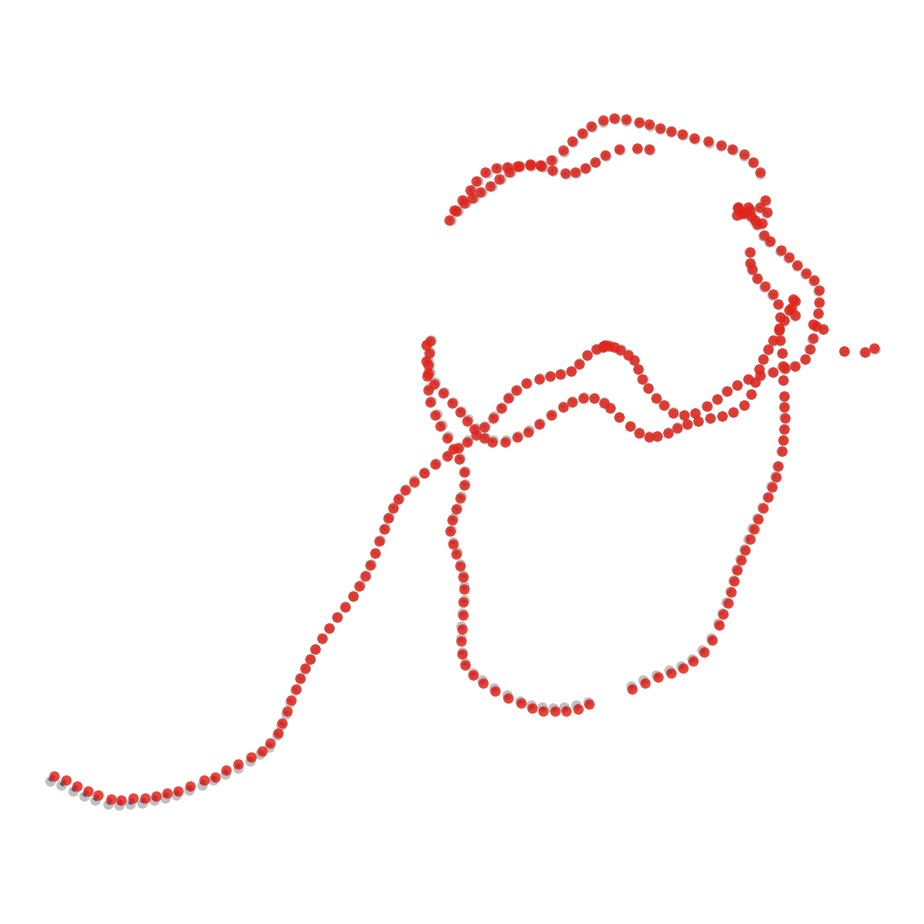}{76.3} & \mapaucbf{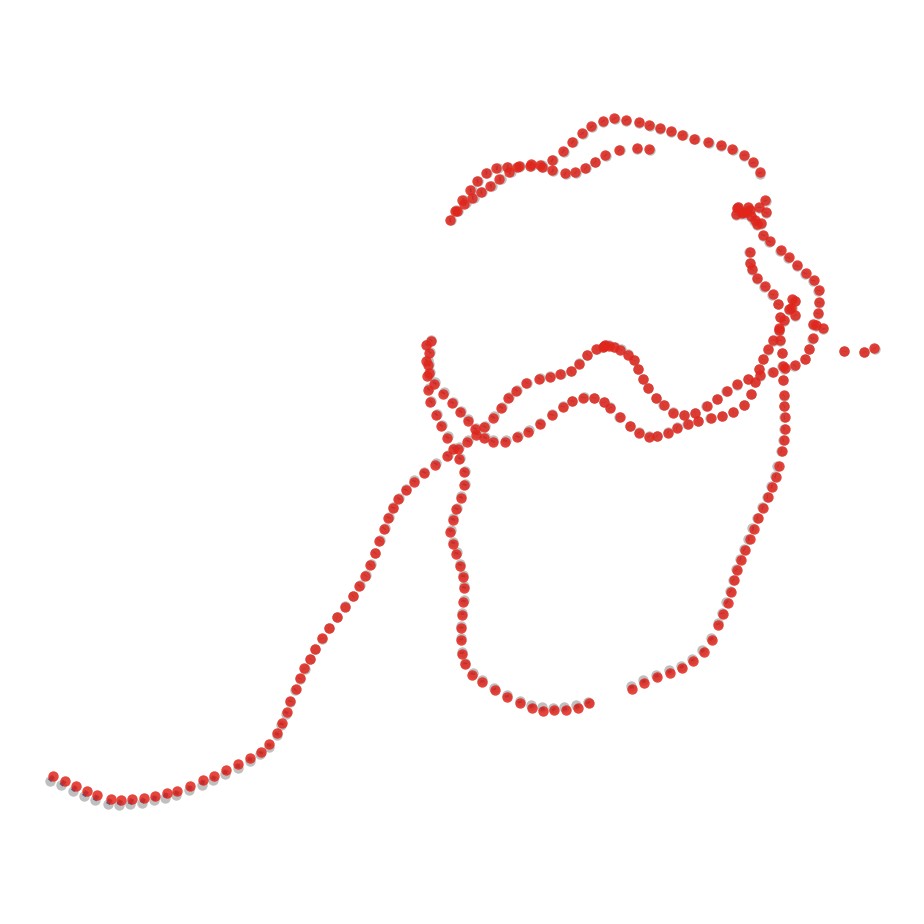}{78.4} & \mapaucbf{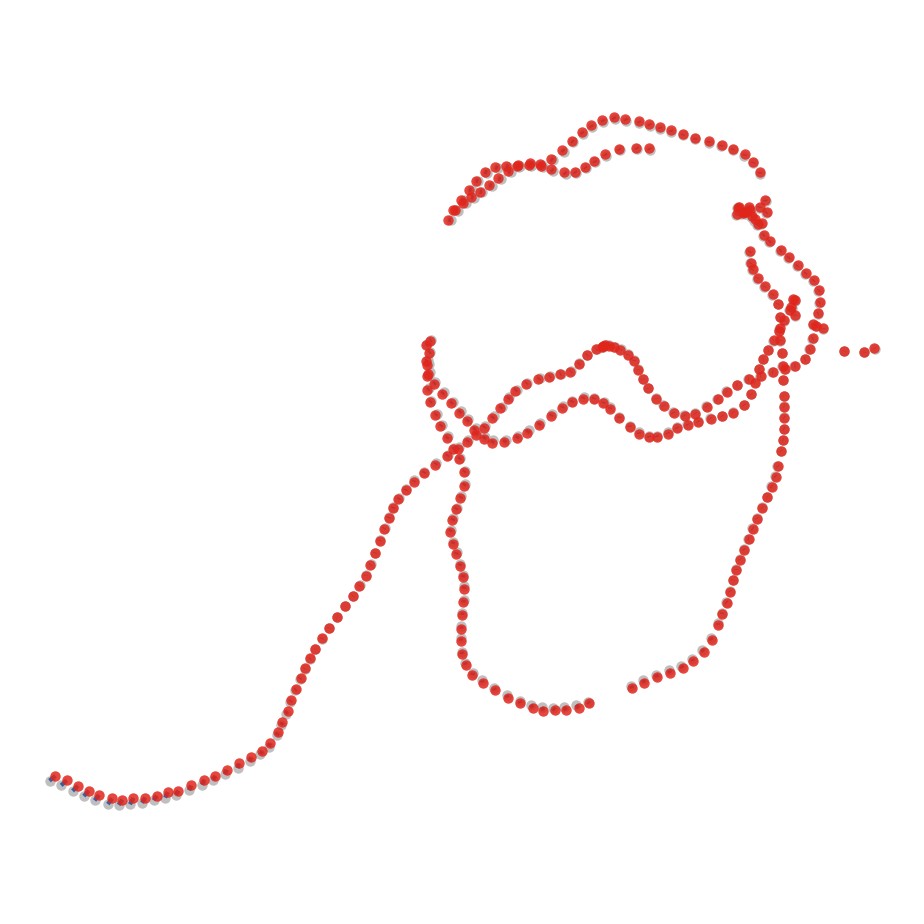}{75.6} \\[4pt]
\maprow{Initialization} & \mappanel{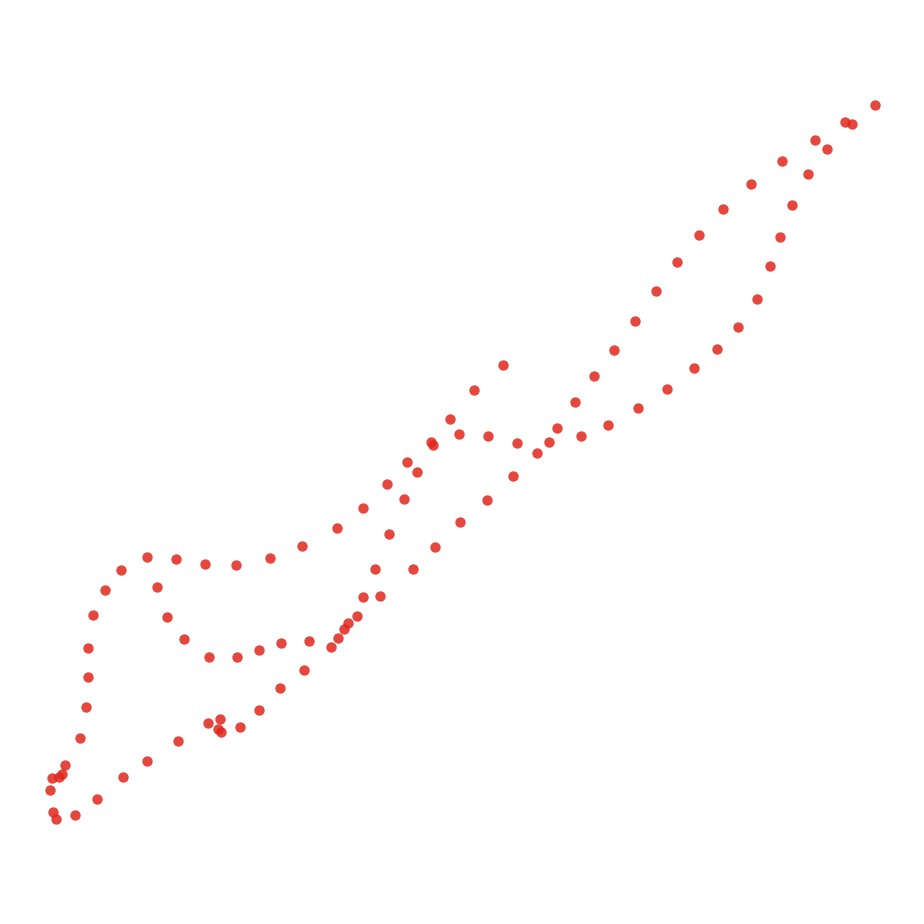} & \mapauc{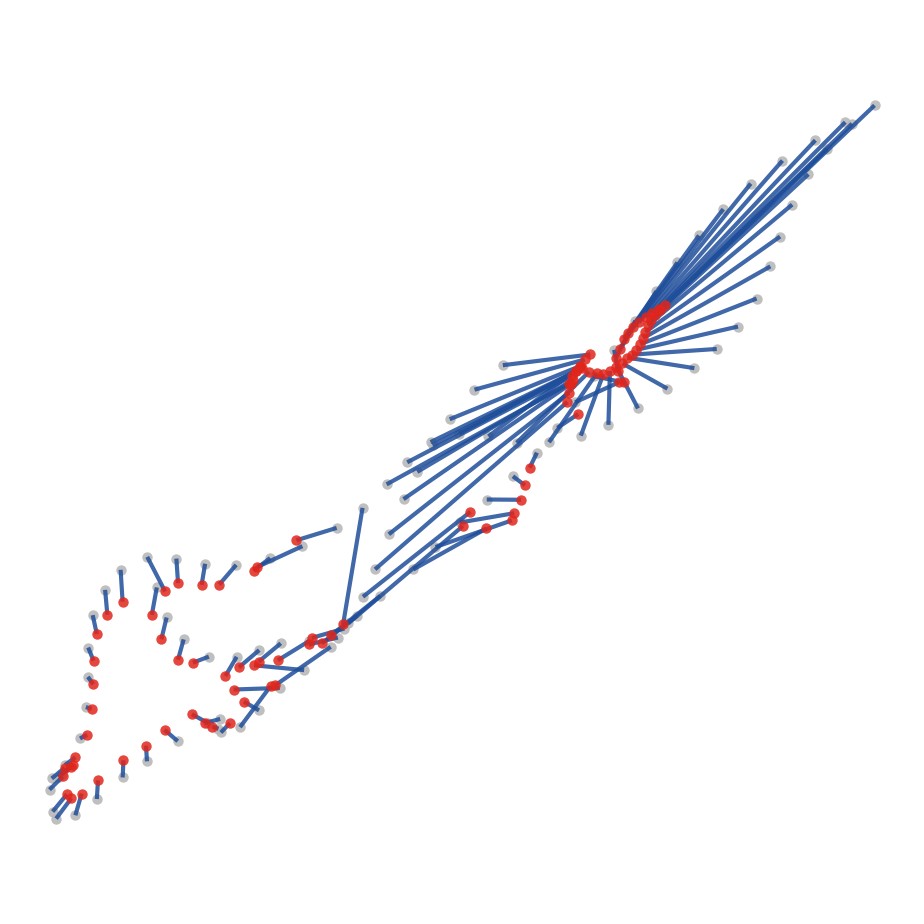}{4.0} & \mapauc{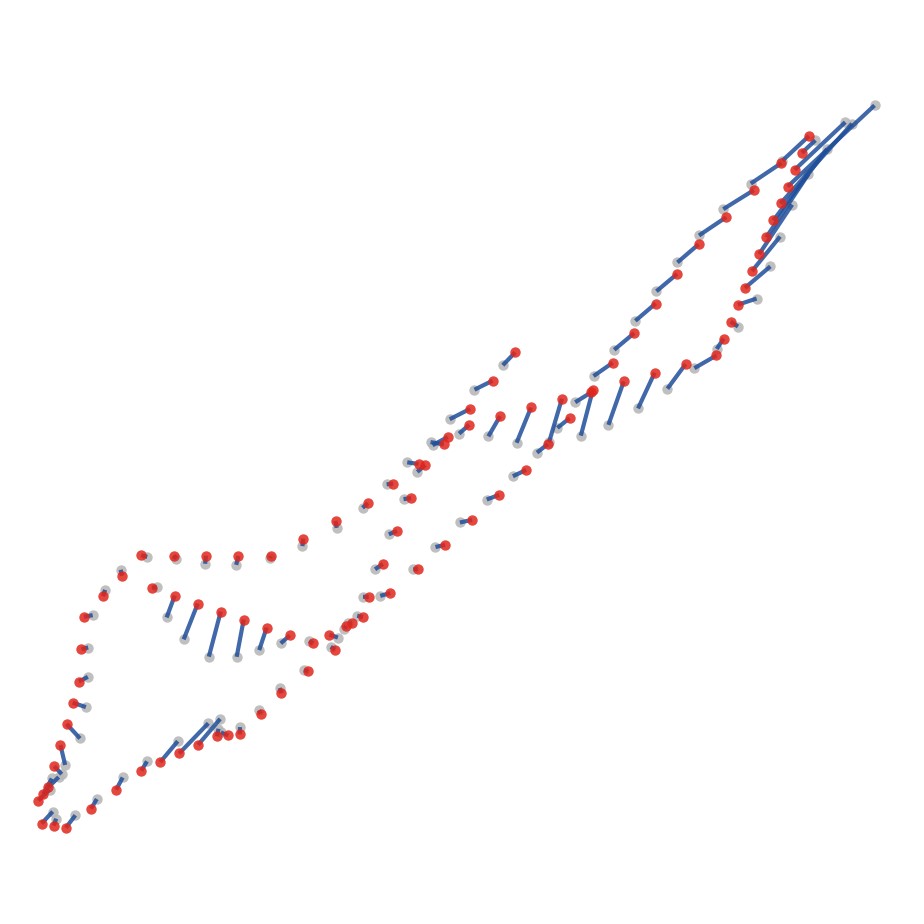}{22.7} & \mapauc{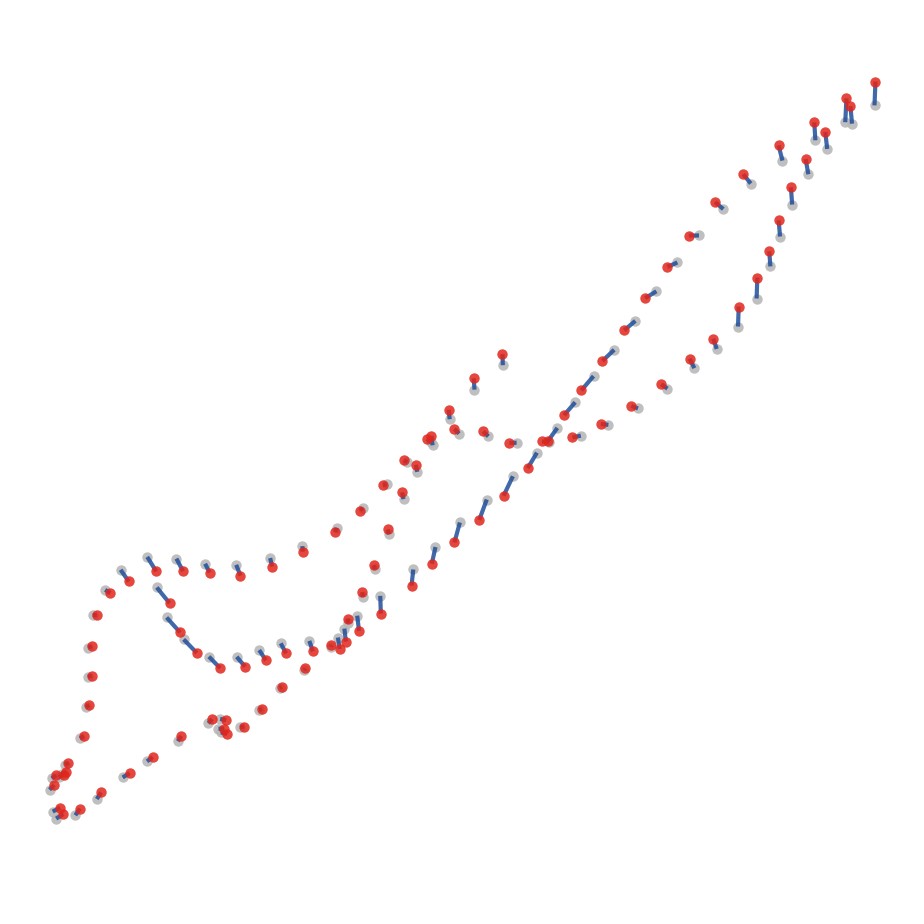}{10.3} & \mapauc{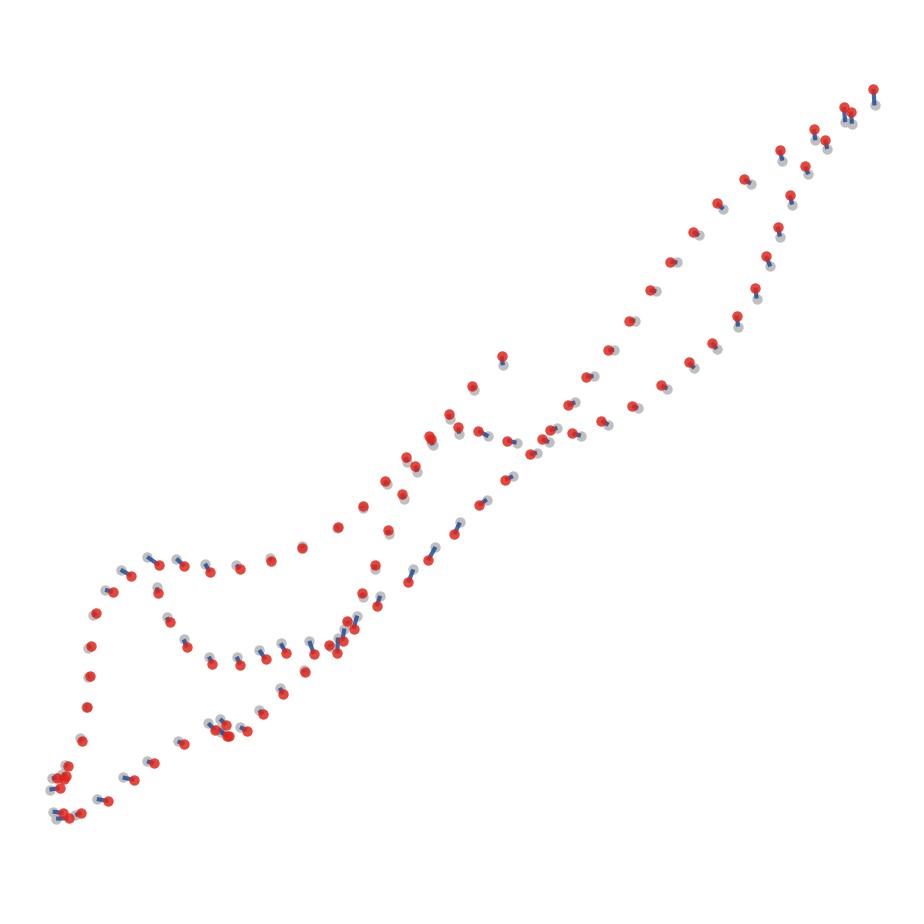}{14.1} & \mapauc{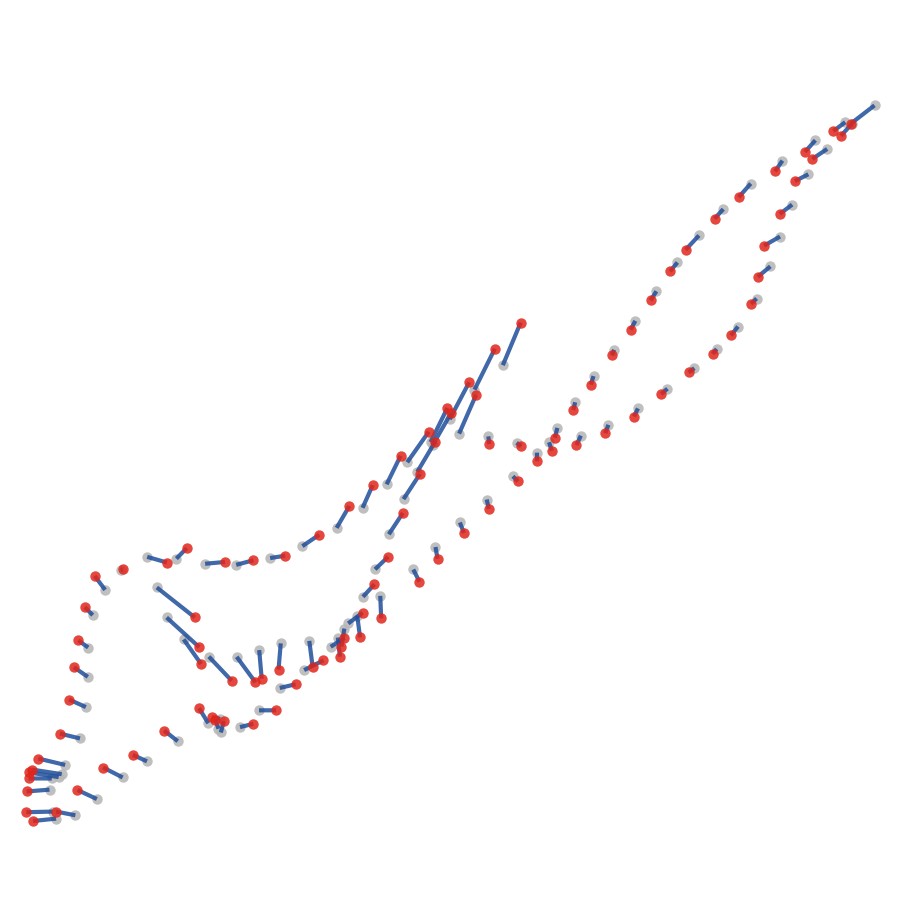}{6.1} \\[0pt]
\maprow{\textbf{+ \method}} & \mapname{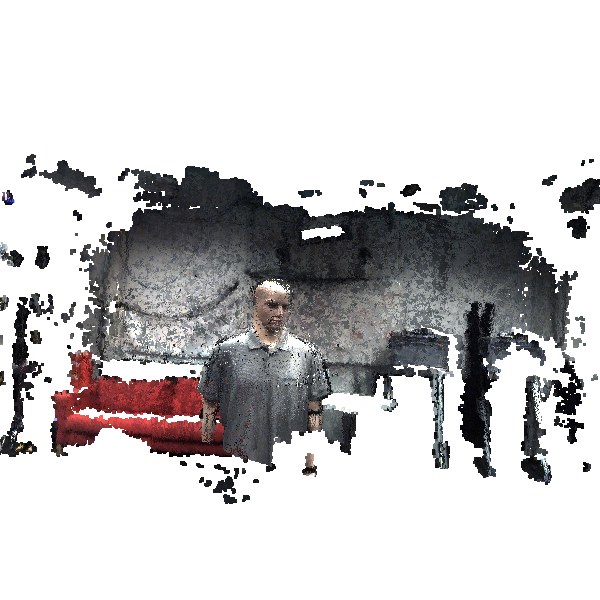}{mannequin face 1} & \mapaucbf{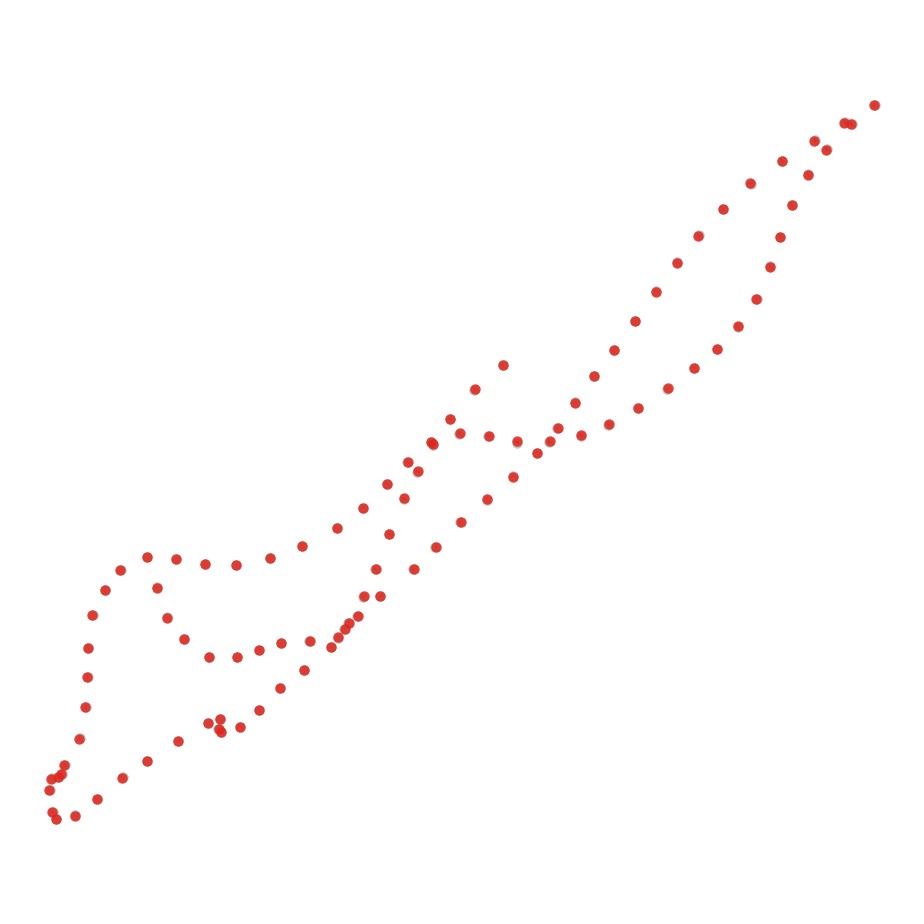}{94.3} & \mapaucbf{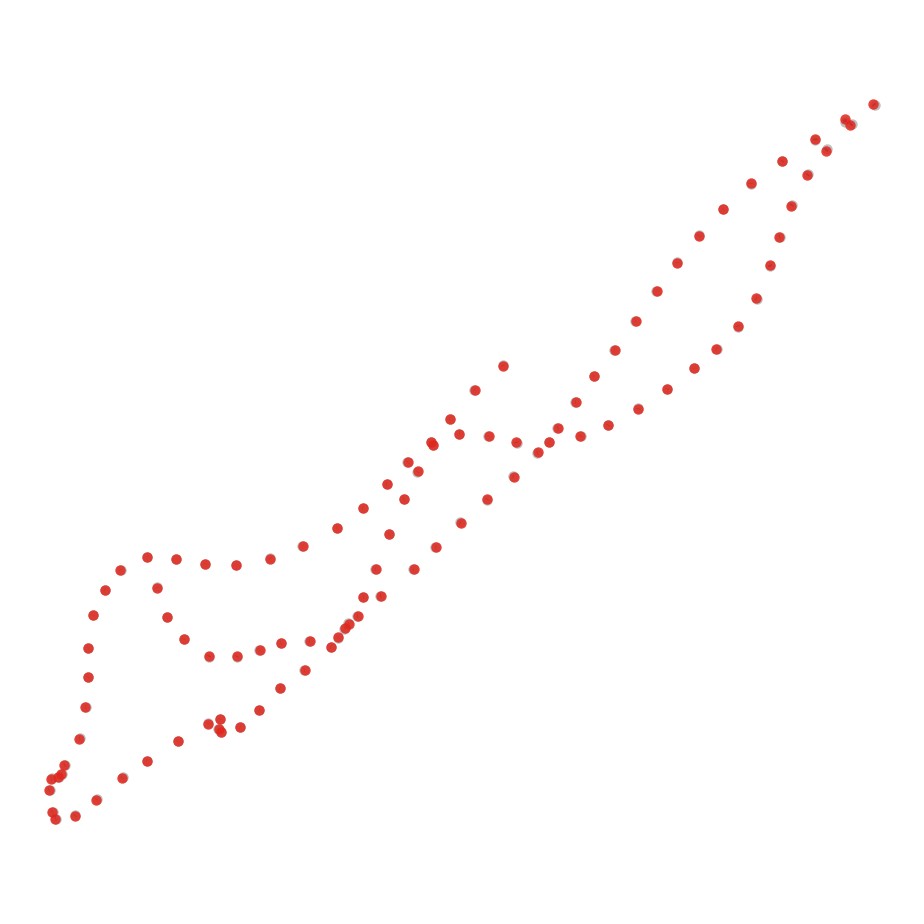}{94.5} & \mapaucbf{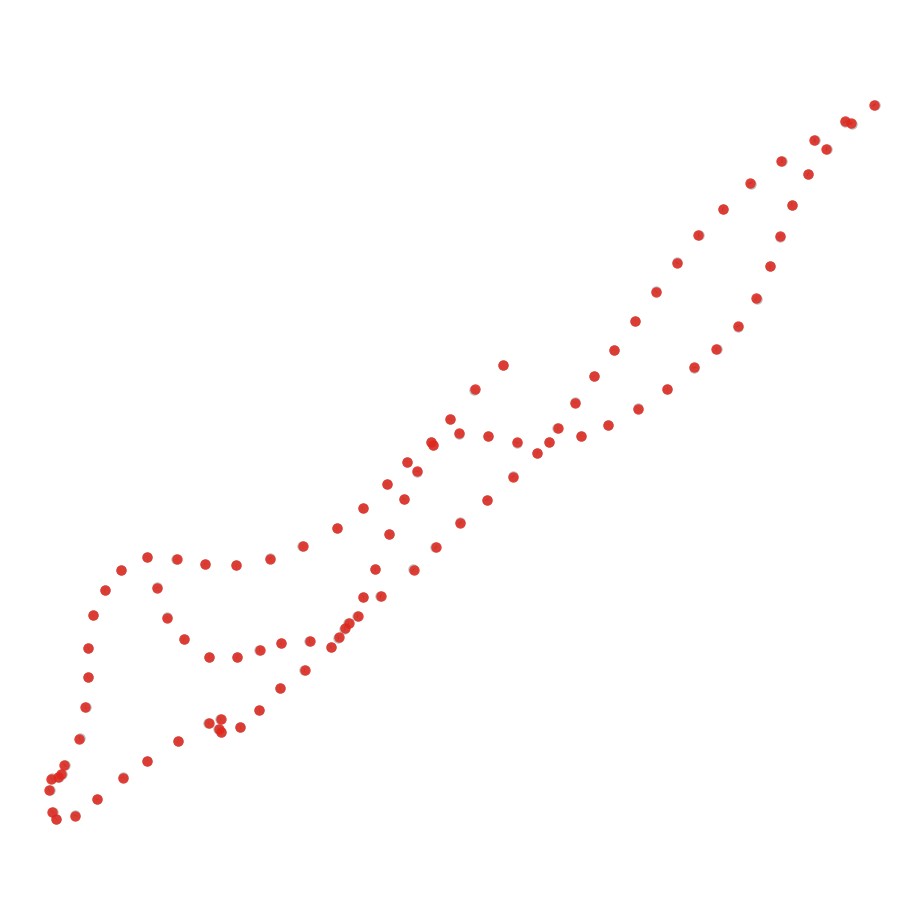}{93.2} & \mapaucbf{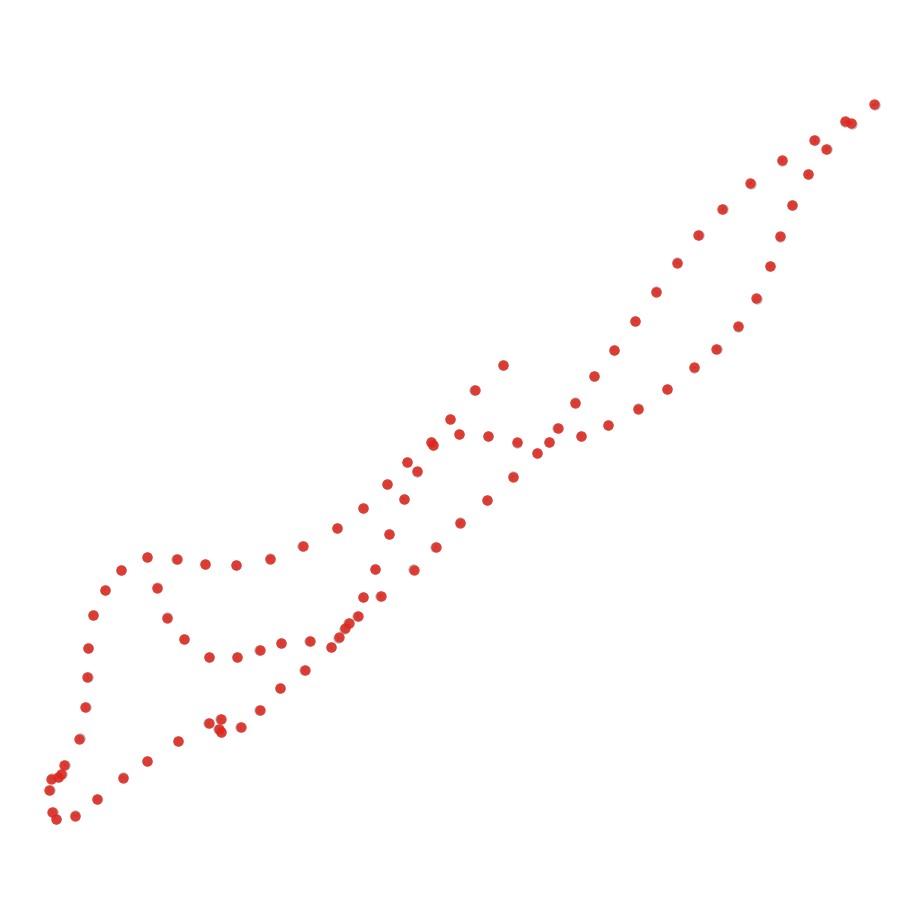}{93.0} & \mapaucbf{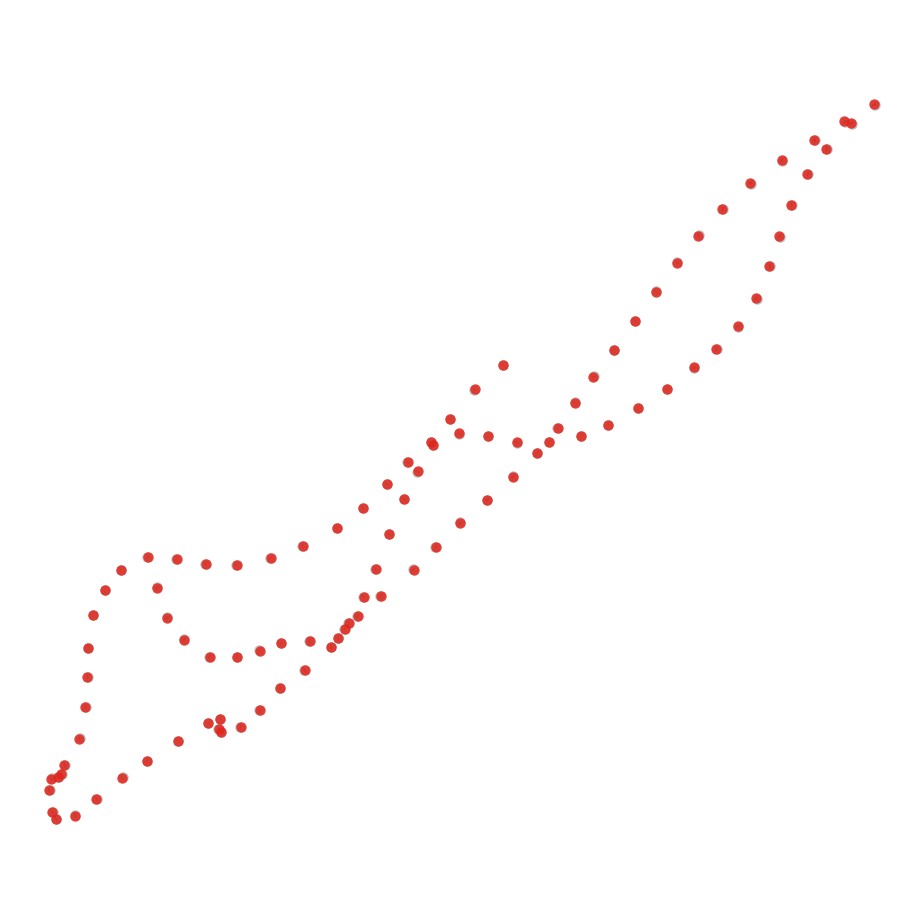}{93.0} \\[4pt]
\maprow{Initialization} & \mappanel{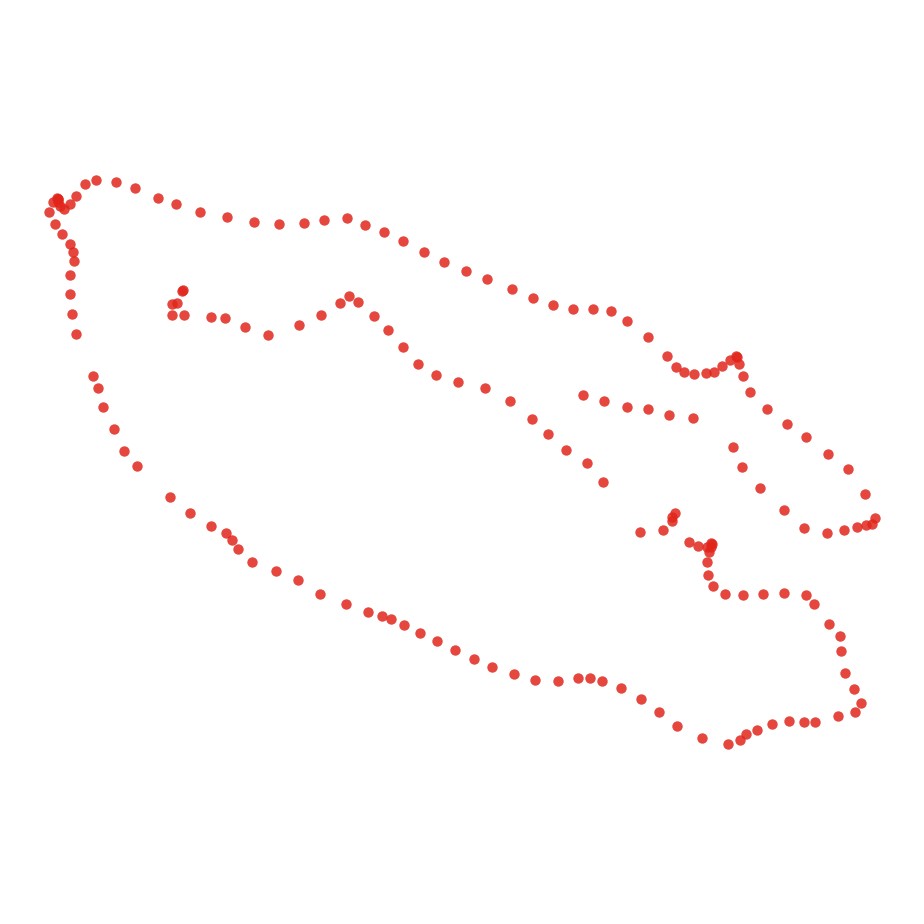} & \mapauc{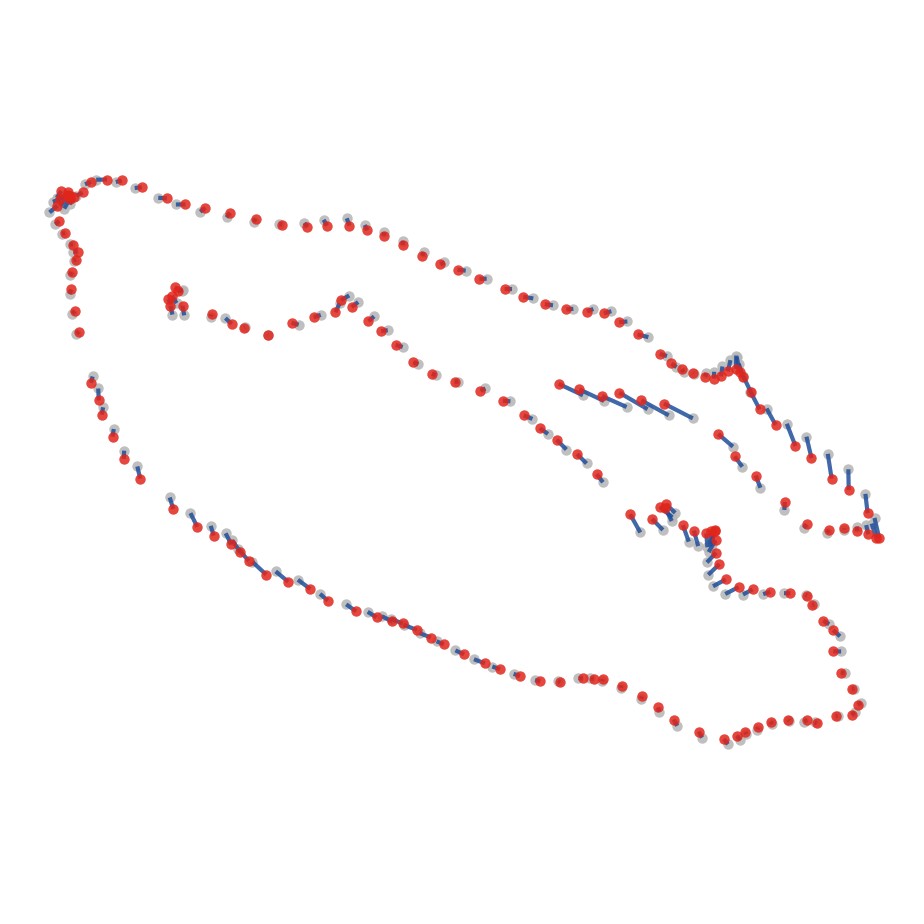}{13.2} & \mapauc{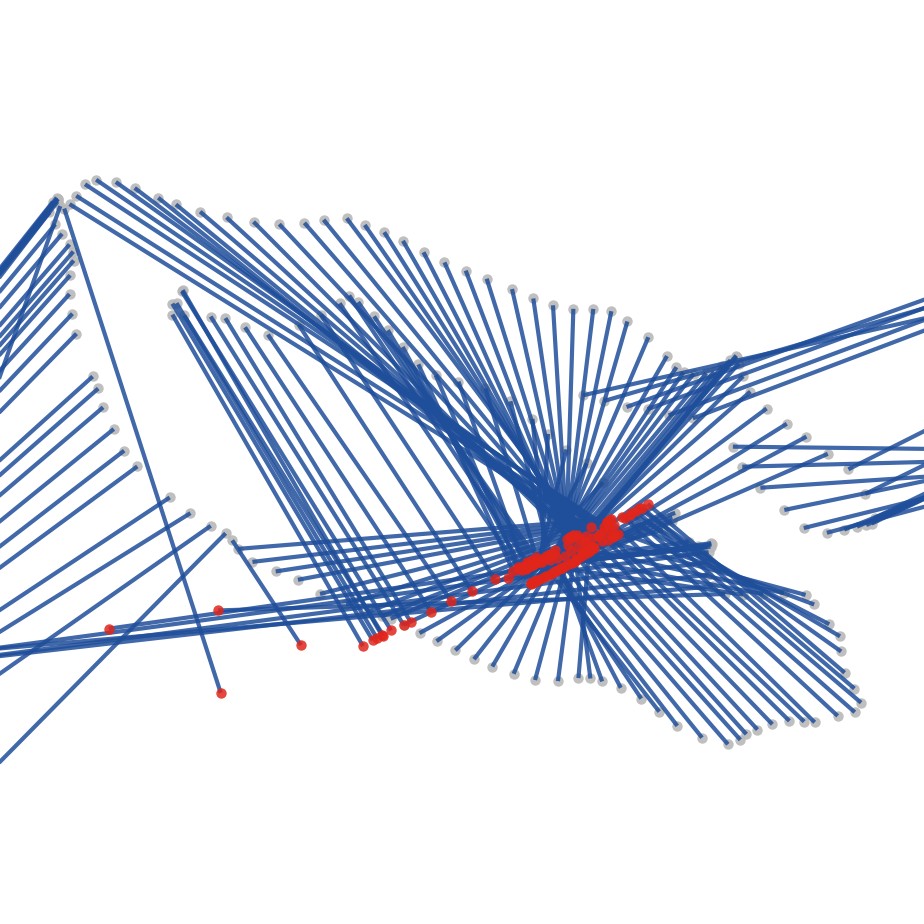}{0.0} & \mapauc{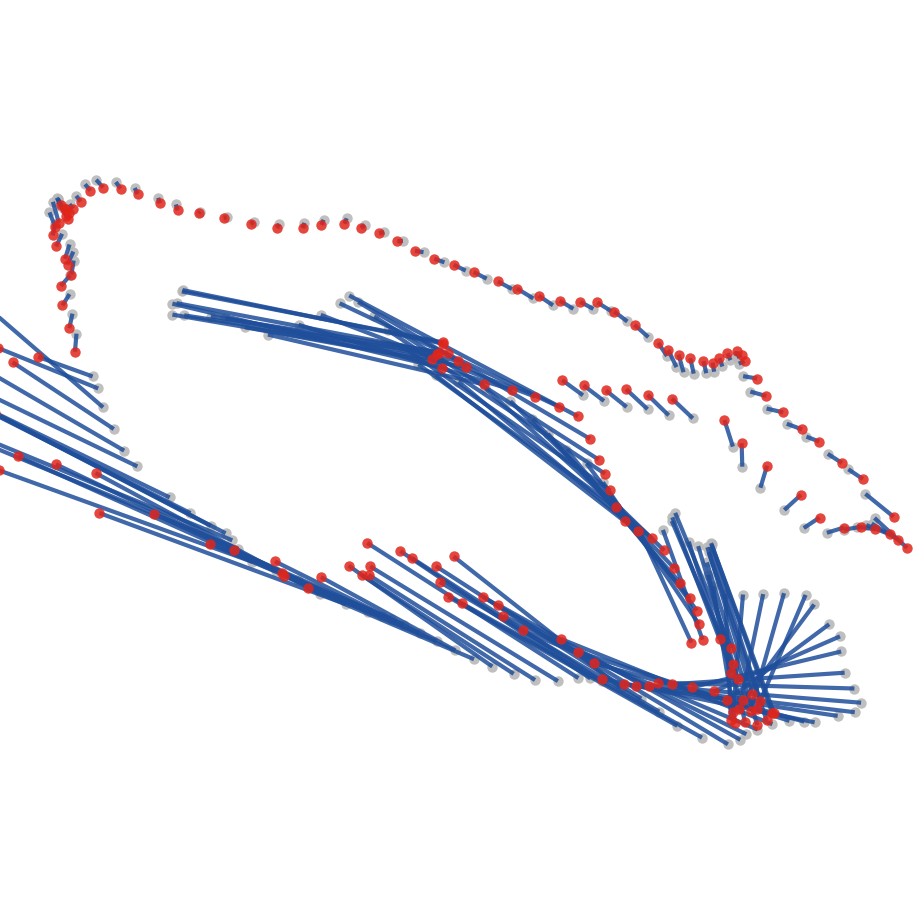}{3.3} & \mapauc{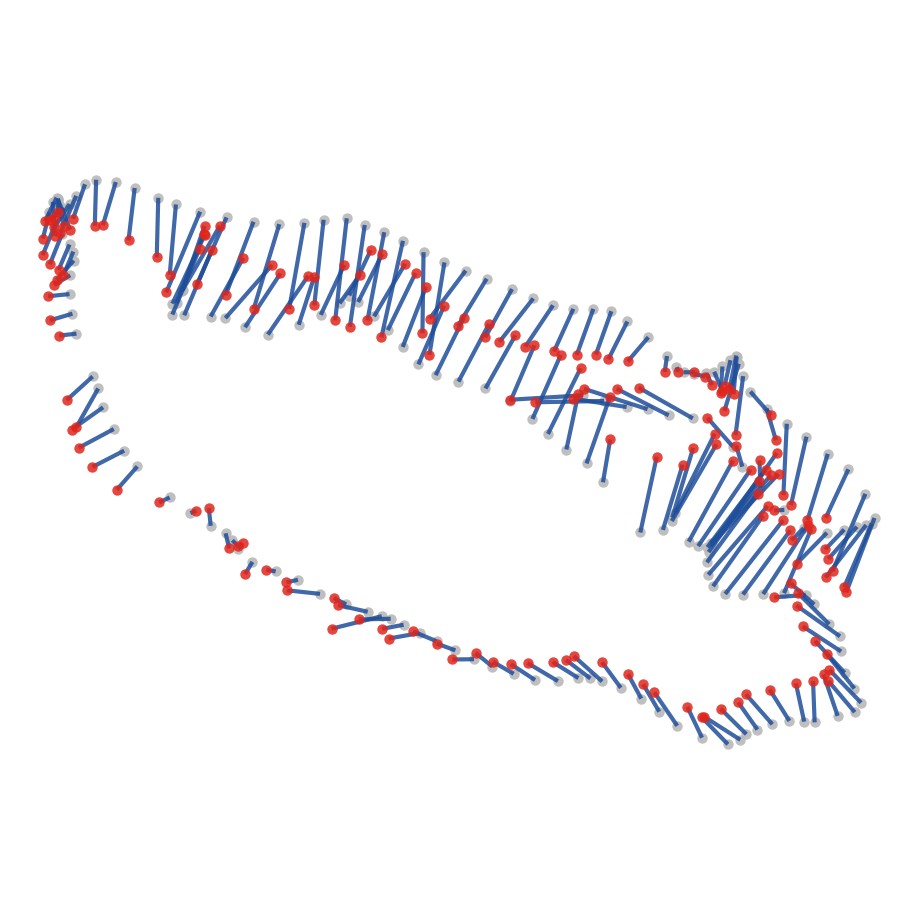}{0.5} & \mapauc{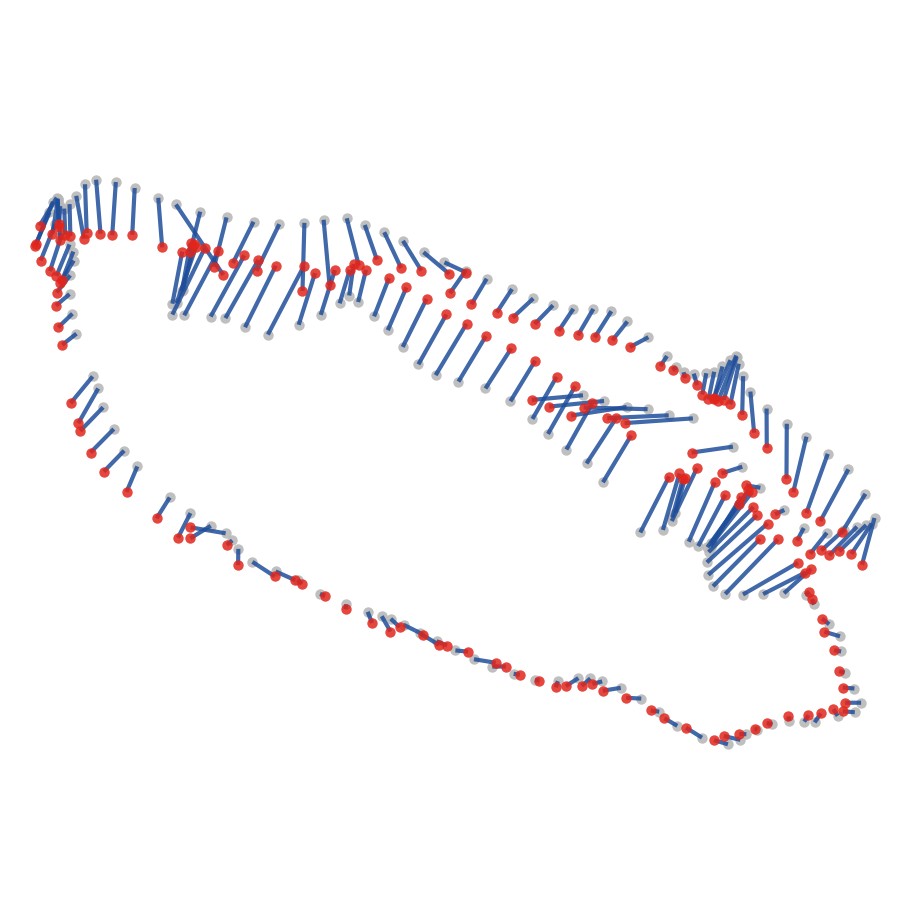}{1.7} \\[0pt]
\maprow{\textbf{+ \method}} & \mapname{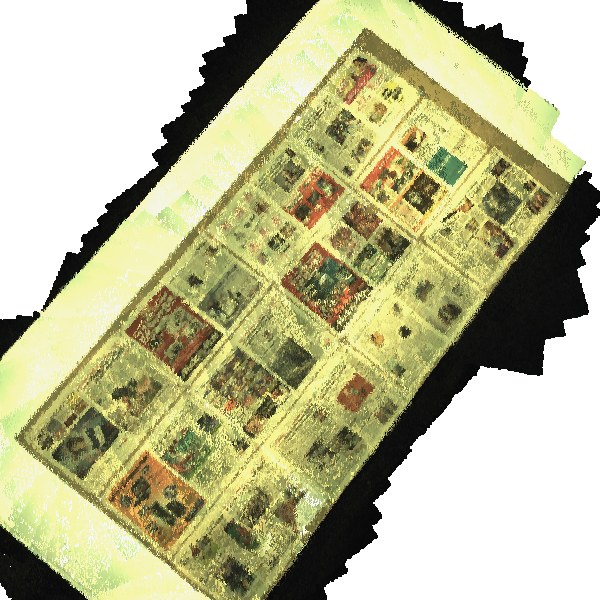}{planar 3} & \mapaucbf{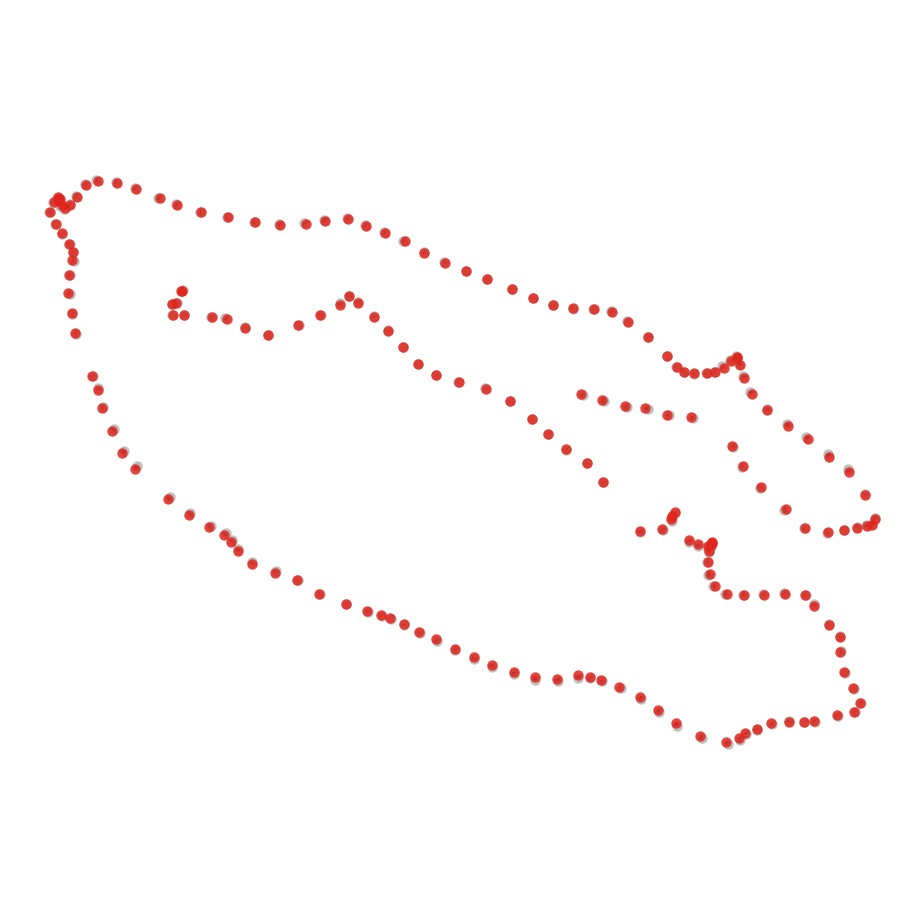}{84.7} & \mapaucbf{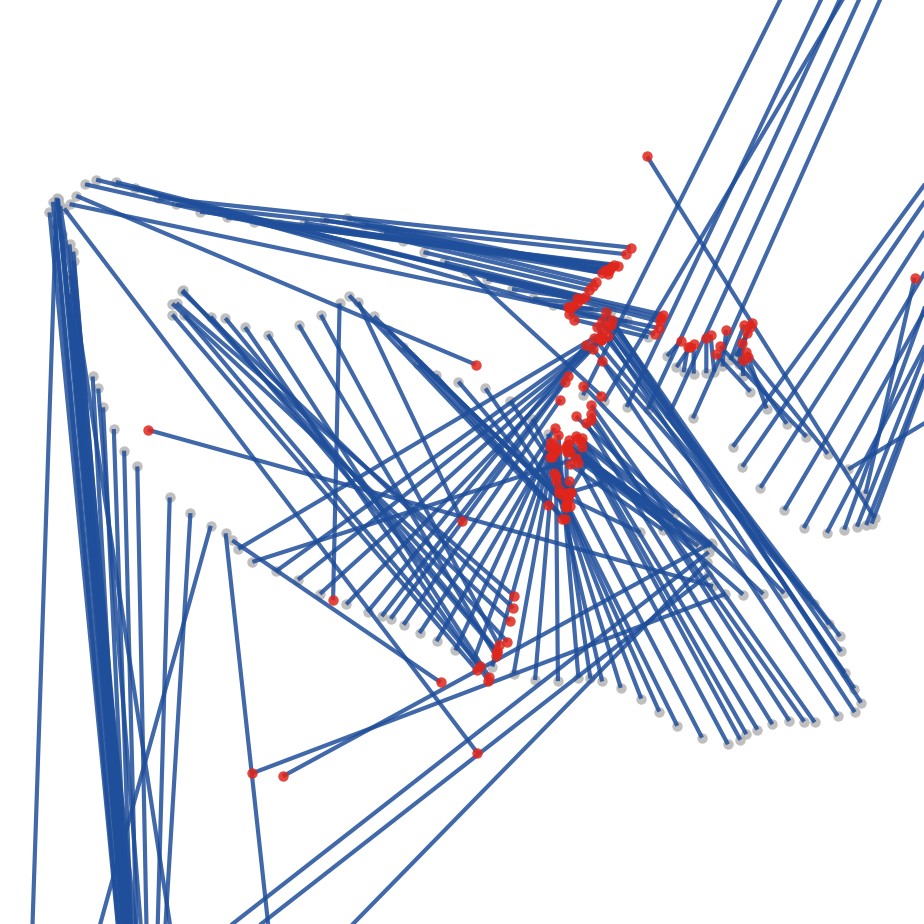}{0.0} & \mapaucbf{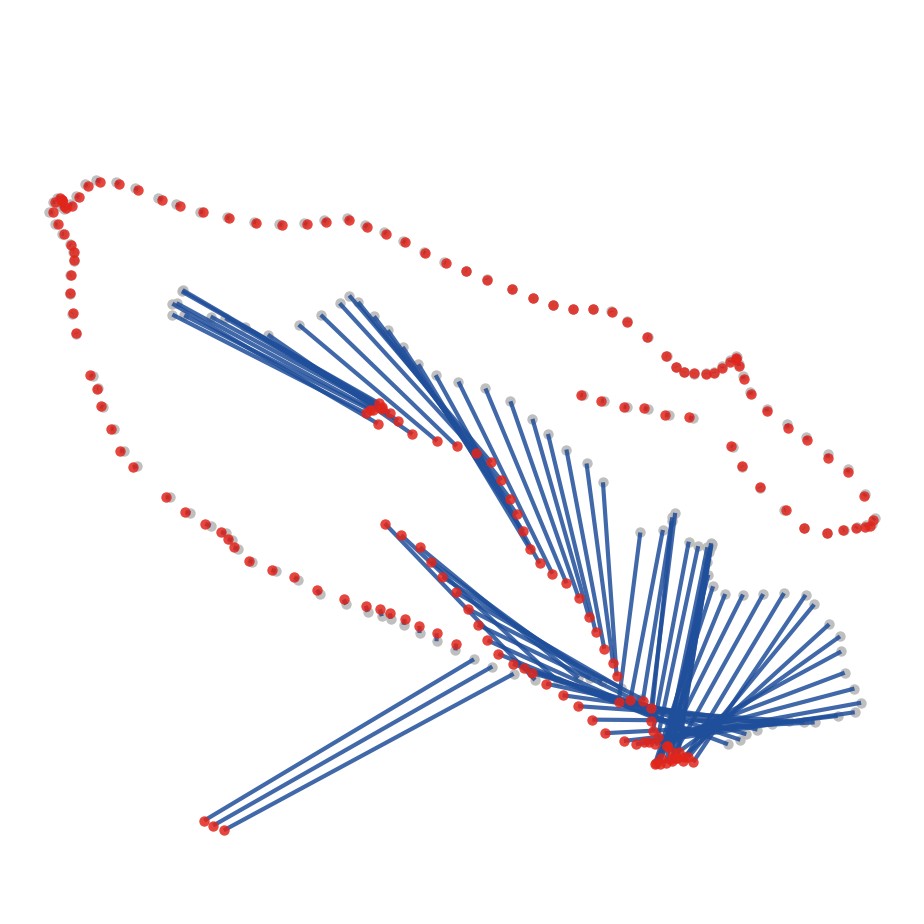}{36.5} & \mapaucbf{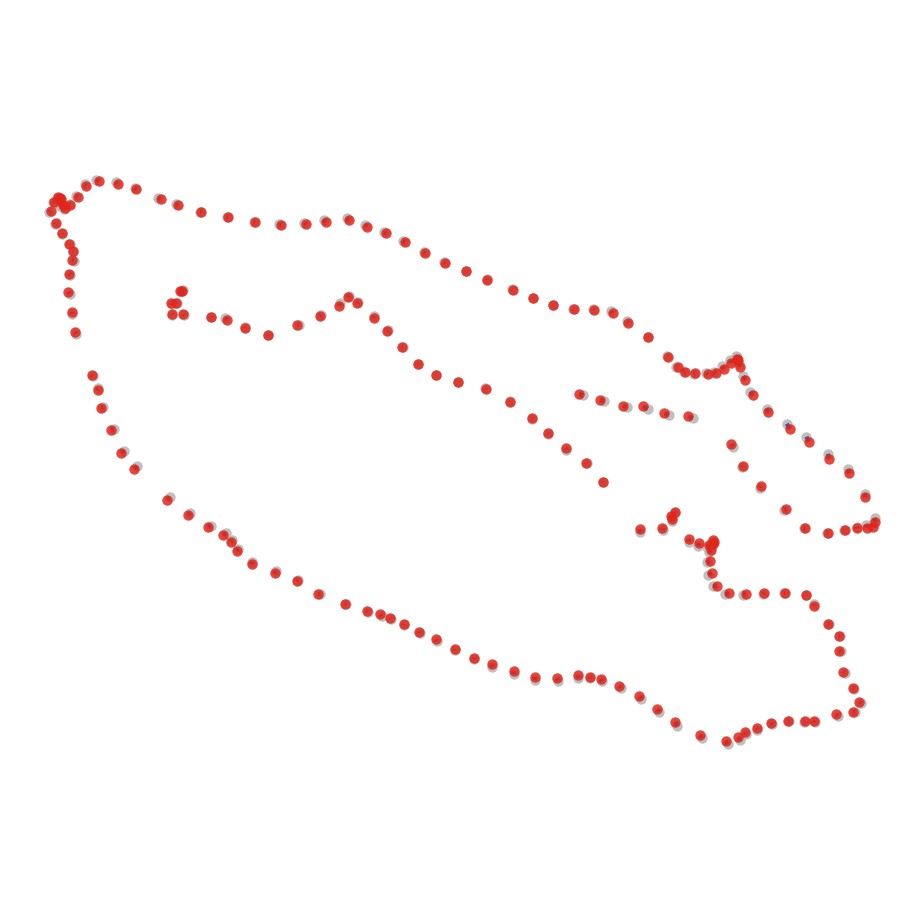}{73.7} & \mapaucbf{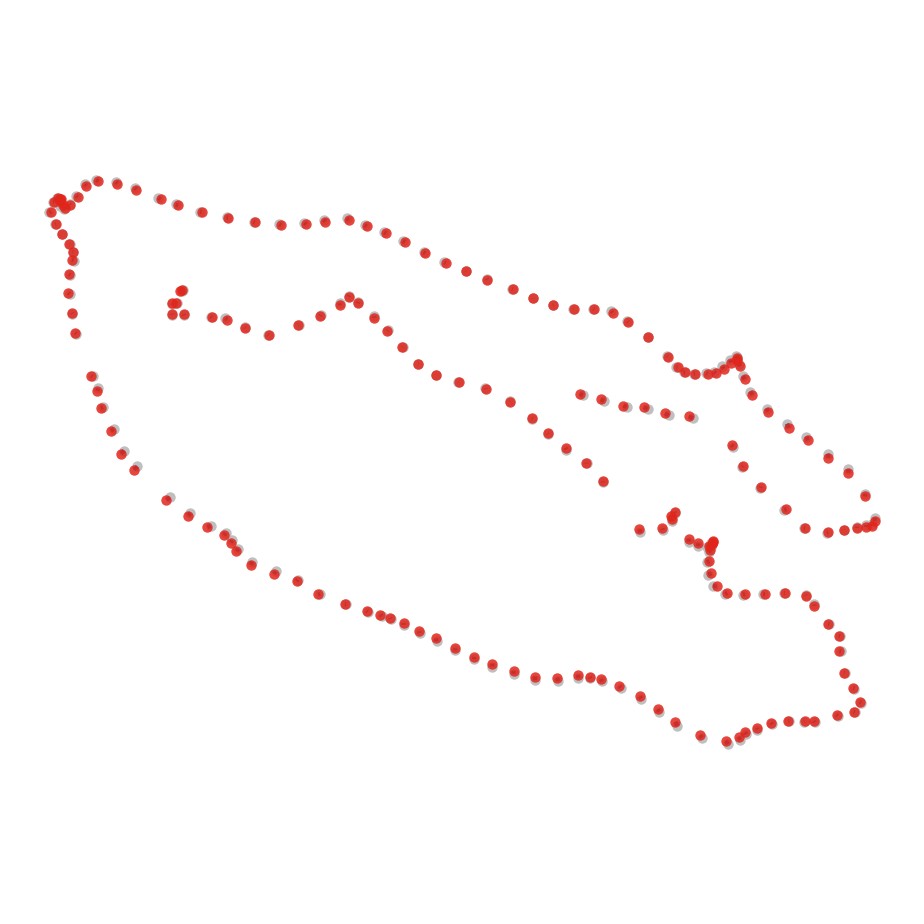}{77.2} \\[4pt]
\end{tabular}
\caption{Trajectory comparison on ETH3D-SLAM sequences: the initialization cameras of each method (upper row of each sequence) and the same cameras after \method (lower row), aligned to the ground truth. $^\ddagger$: our chunking. $^\S$: the method's own chunking.}
\label{fig:eth3d_slam2_map}
\end{figure*}

\ifdefined\aseqW\else\newlength{\aseqW}\fi\setlength{\aseqW}{1.31in}
\begin{figure*}[p]
\def\mappanel#1{{\setlength{\fboxsep}{0pt}\setlength{\fboxrule}{0.3pt}\fcolorbox{black!20}{white}{\includegraphics[width=\dimexpr\aseqW-0.6pt\relax]{fig/aachen_seq_map/#1}}}}
\def\mapauc#1#2{\mappanel{#1}\llap{\raisebox{1.2pt}{\setlength{\fboxsep}{0.8pt}\colorbox{white}{\tiny AUC@10$^\circ$ #2}}\hspace{1.2pt}}}
\def\mapaucbf#1#2{\mappanel{#1}\llap{\raisebox{1.2pt}{\setlength{\fboxsep}{0.8pt}\colorbox{white}{\tiny\bfseries AUC@10$^\circ$ #2}}\hspace{1.2pt}}}
\def\maprow#1{\rotatebox{90}{\scriptsize #1}}
\def\mapname#1#2{\raisebox{0pt}[0pt][0pt]{\parbox{\aseqW}{\centering\includegraphics[width=0.97\aseqW]{fig/aachen_seq_map/#1}\\[1pt]\footnotesize\bfseries #2}}}  
\centering
\setlength{\tabcolsep}{0.8pt}
\renewcommand{\arraystretch}{0.7}
\begin{tabular}{@{}>{\centering\arraybackslash}m{0.9em} >{\centering\arraybackslash}m{\aseqW}@{\hspace{4pt}} *{4}{>{\centering\arraybackslash}m{\aseqW}} @{}}
 & \small\bfseries Ground truth & \small\bfseries DGSfM & \small\bfseries CUT3R & \small\bfseries MERG3R$^\S$ & \small\bfseries DVLT$^\ddagger$ \\[2pt]
\maprow{Initialization} & \mappanel{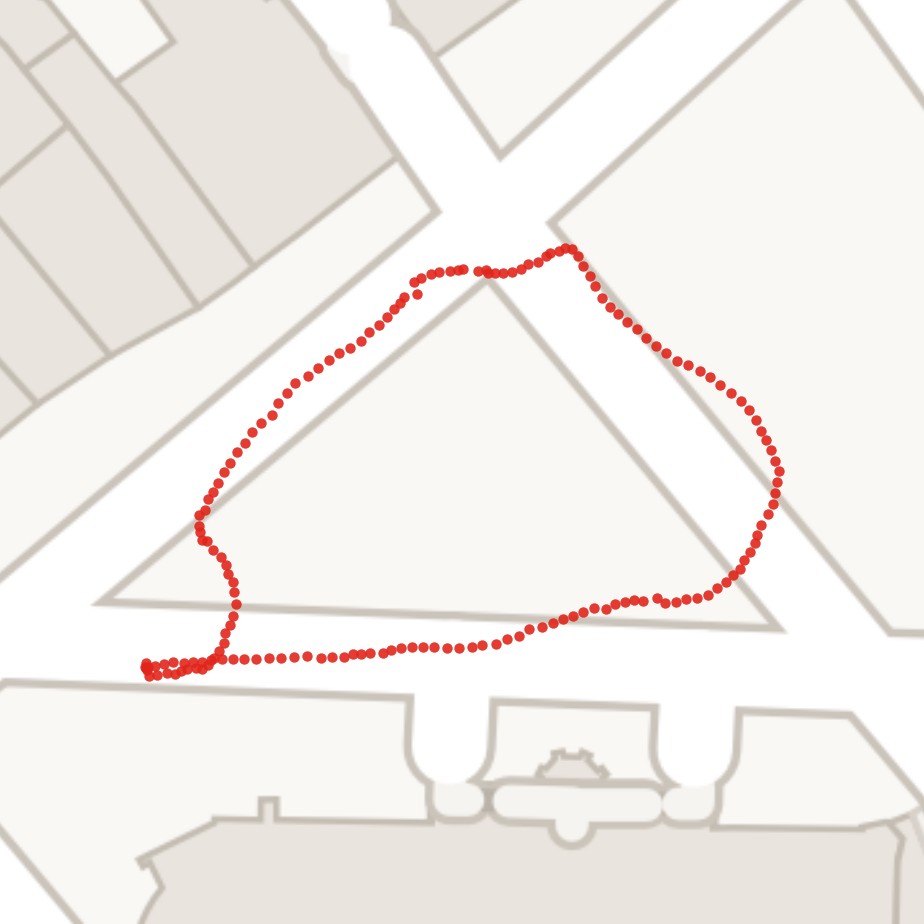} & \mapauc{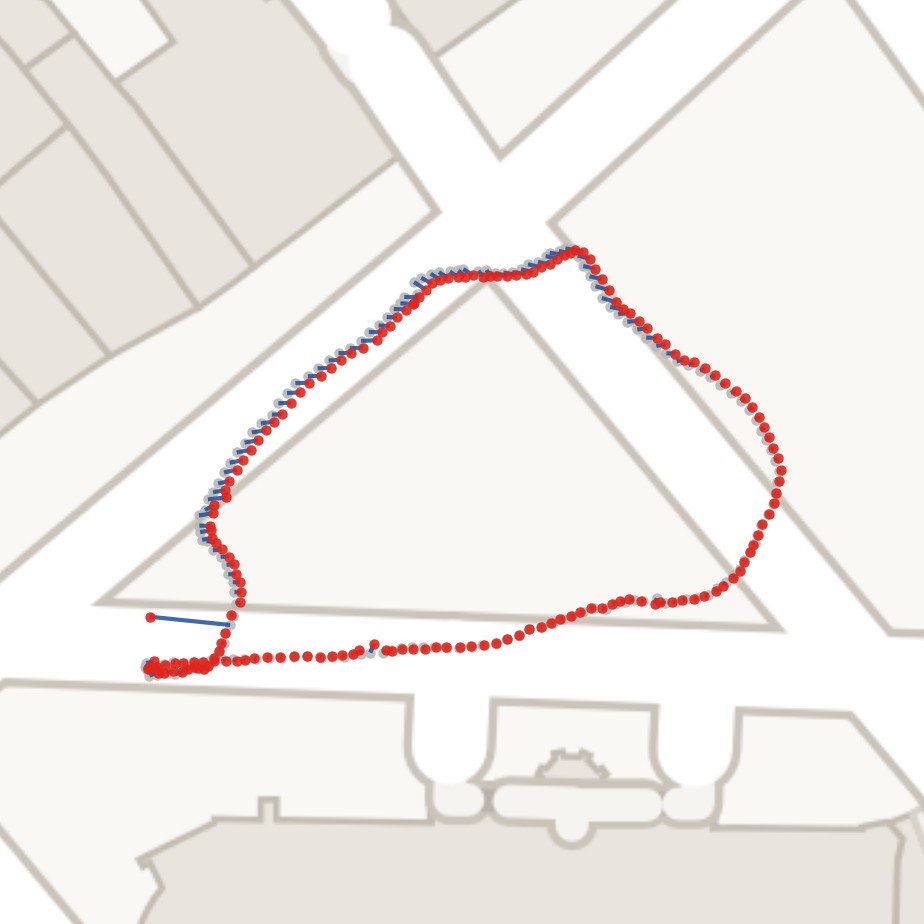}{85.2} & \mapauc{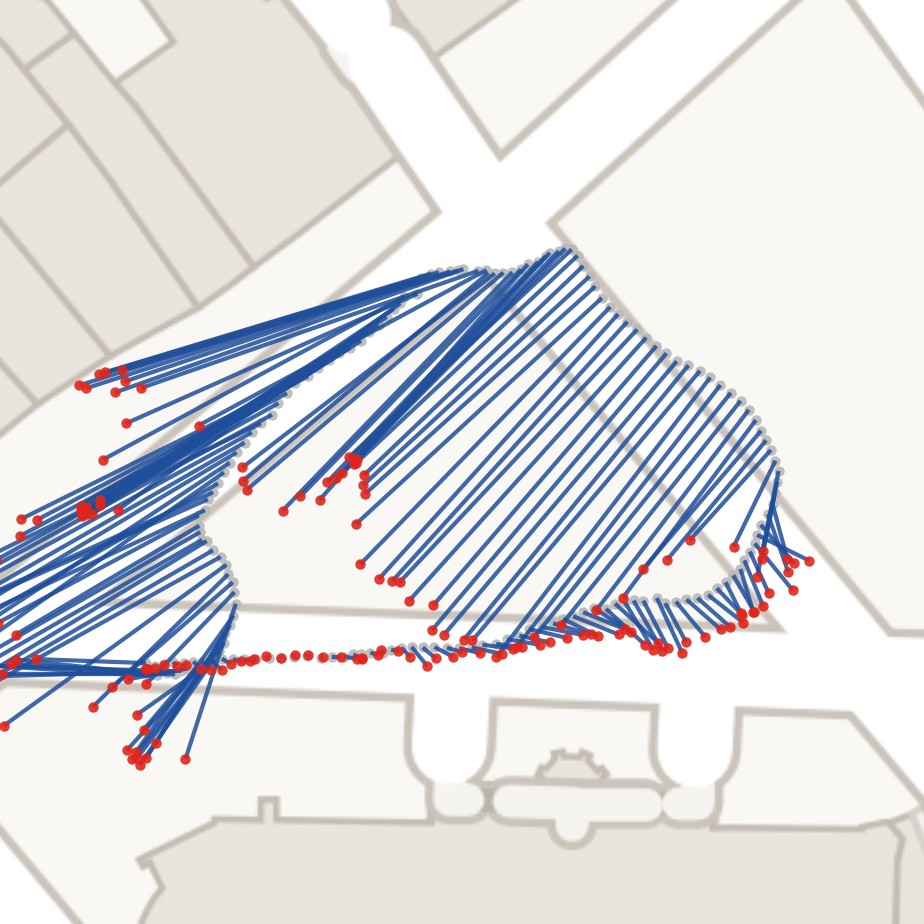}{6.5} & \mapauc{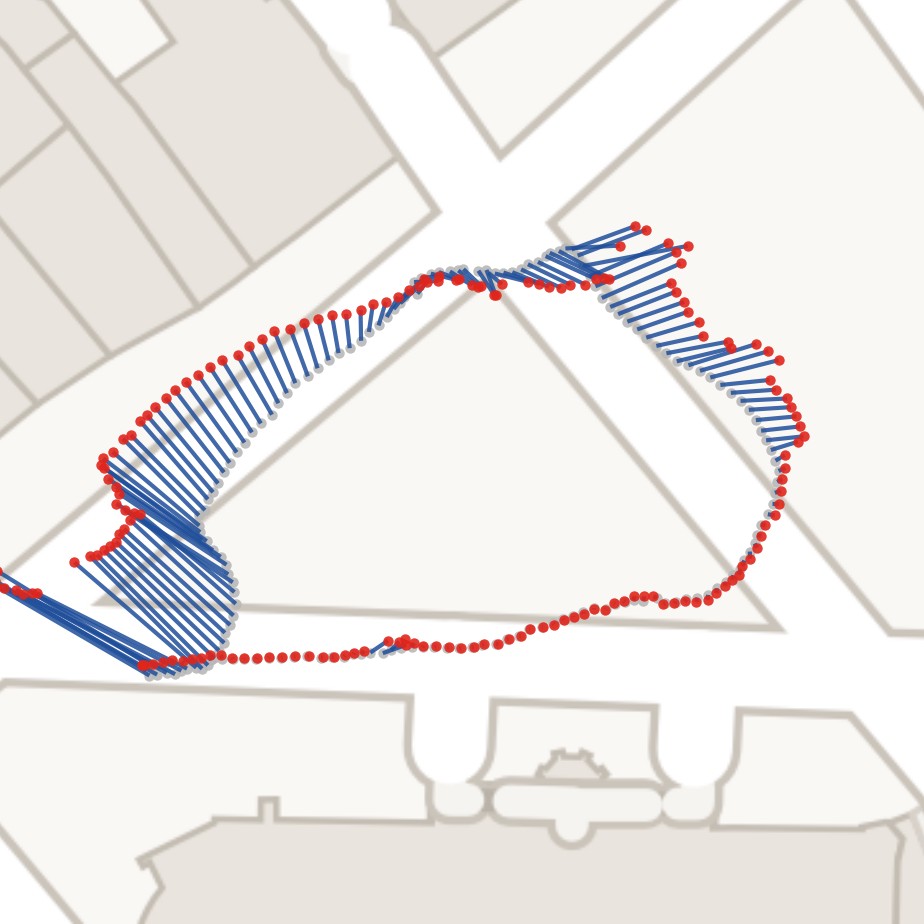}{29.1} & \mapauc{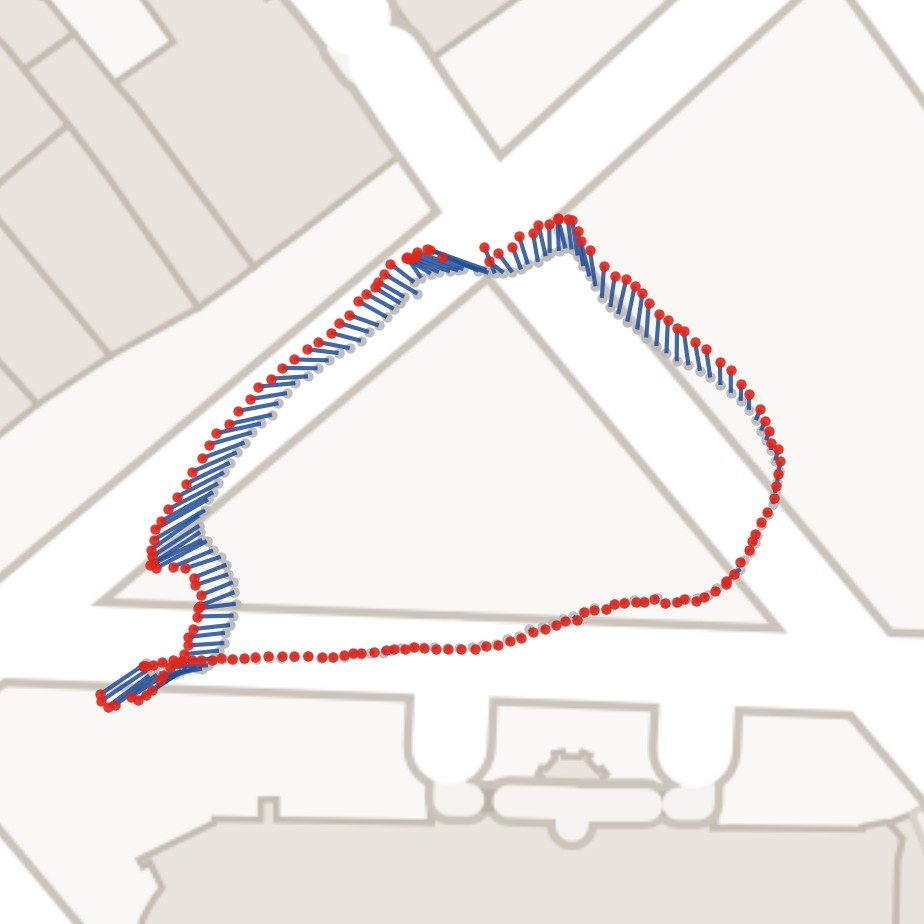}{49.4} \\[0pt]
\maprow{\textbf{+ \method}} & \mapname{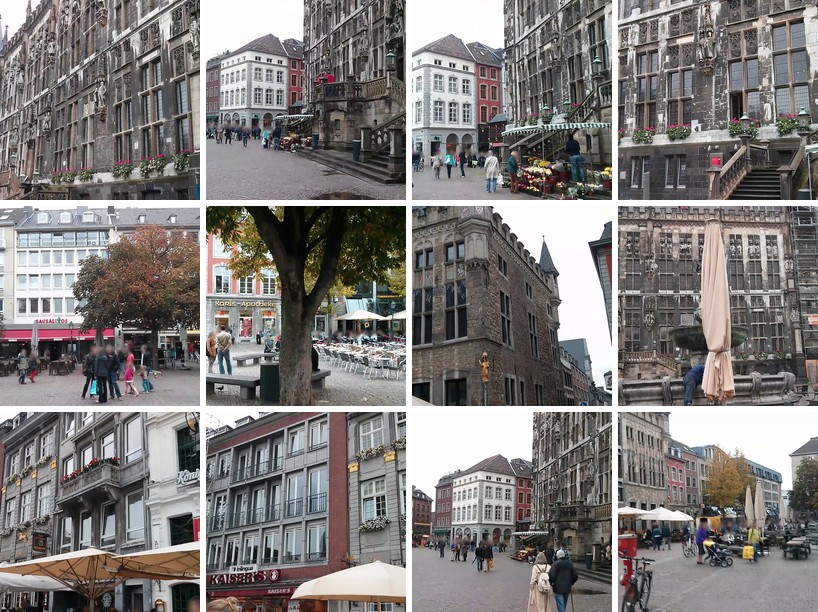}{nexus 1} & \mapaucbf{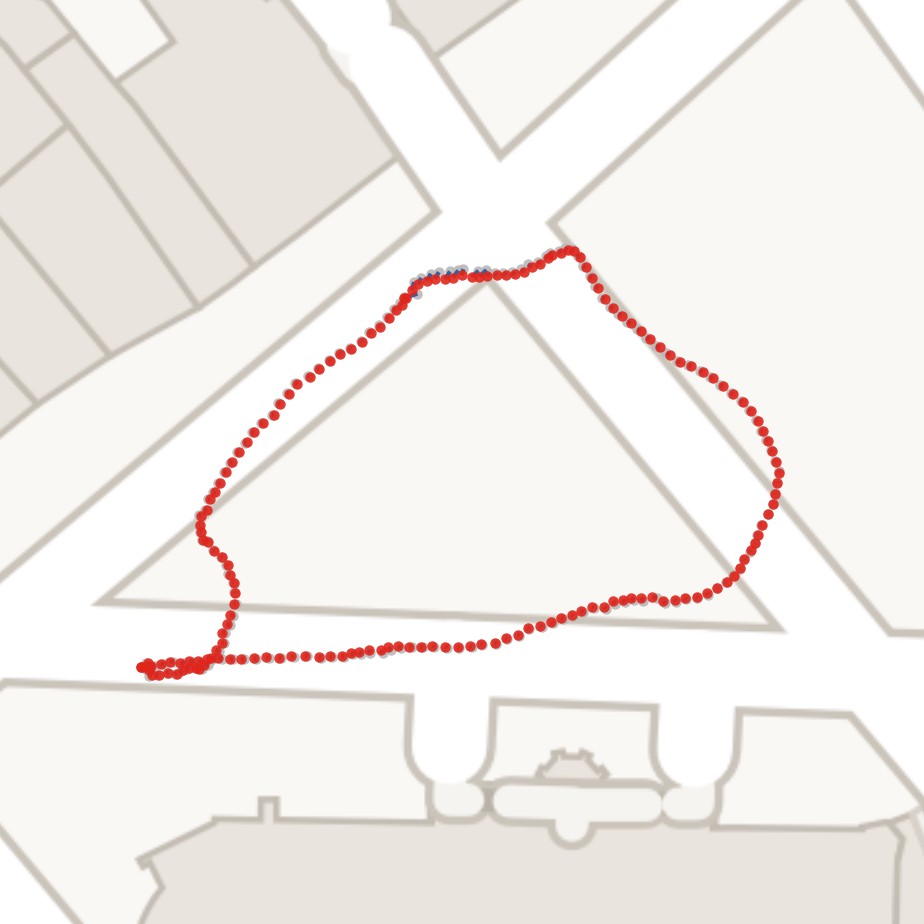}{90.9} & \mapaucbf{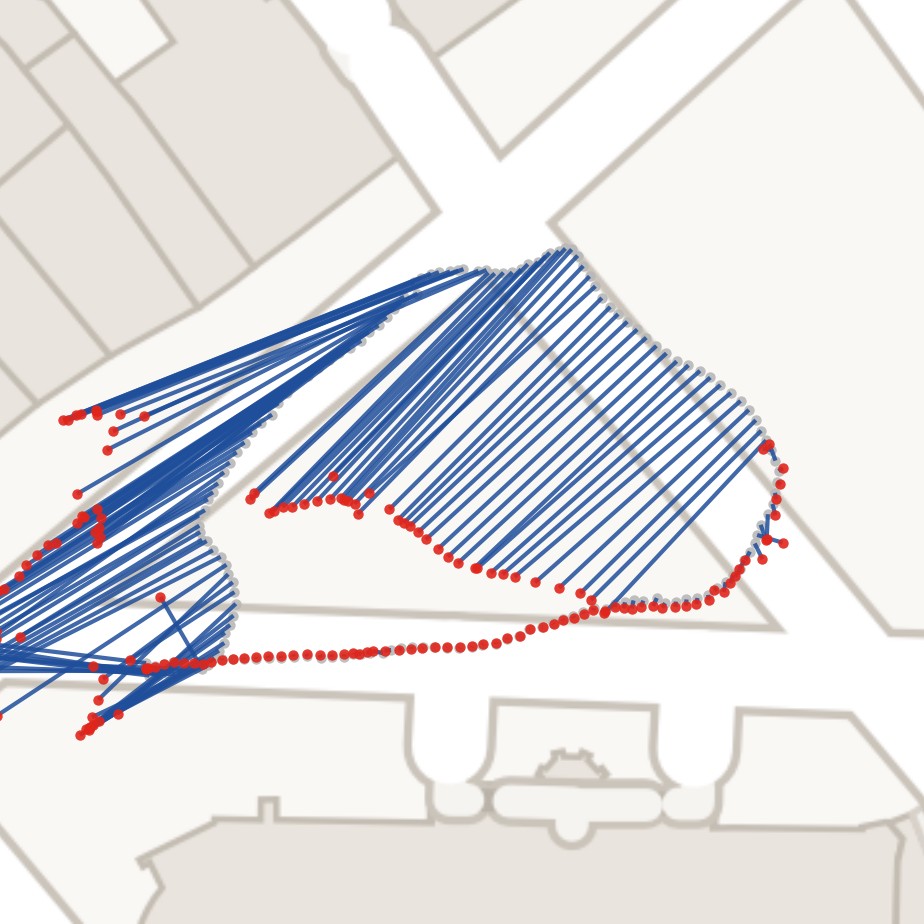}{13.2} & \mapaucbf{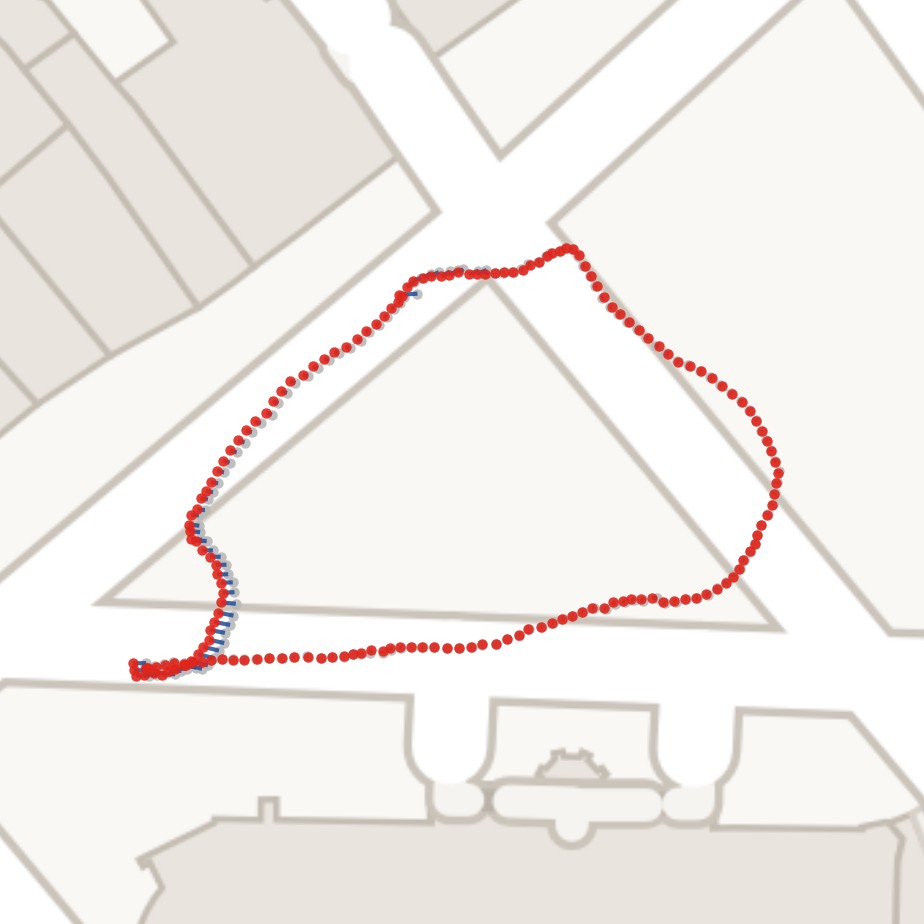}{86.9} & \mapaucbf{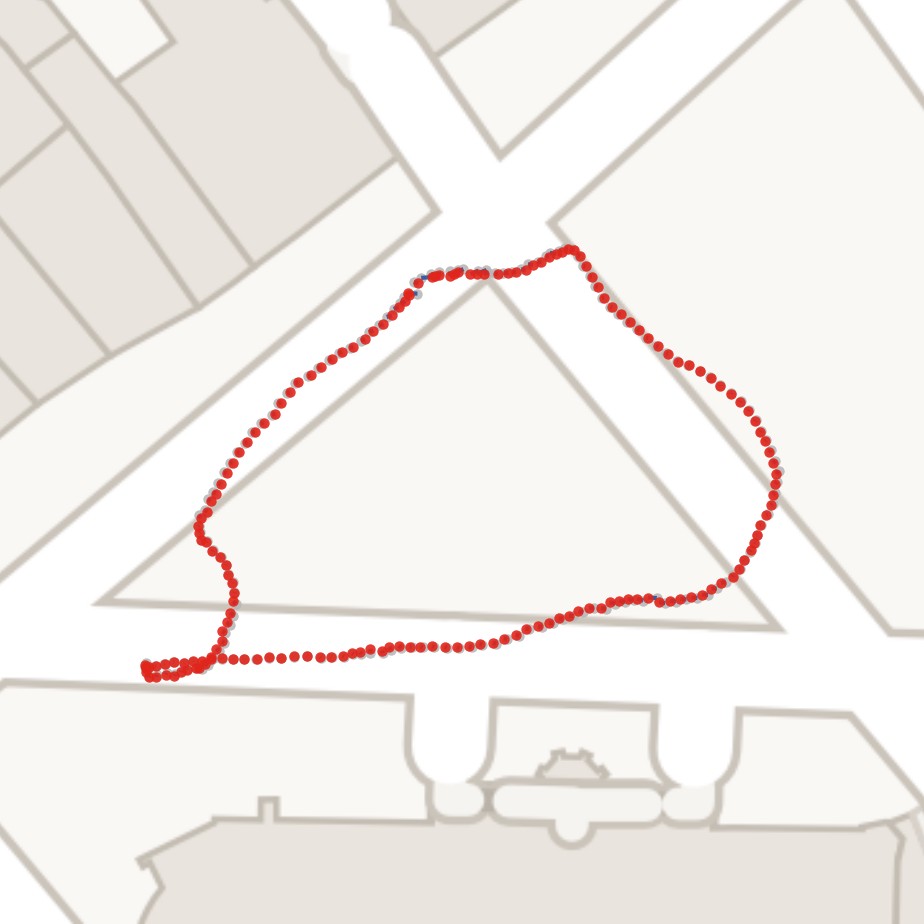}{86.8} \\[4pt]
\maprow{Initialization} & \mappanel{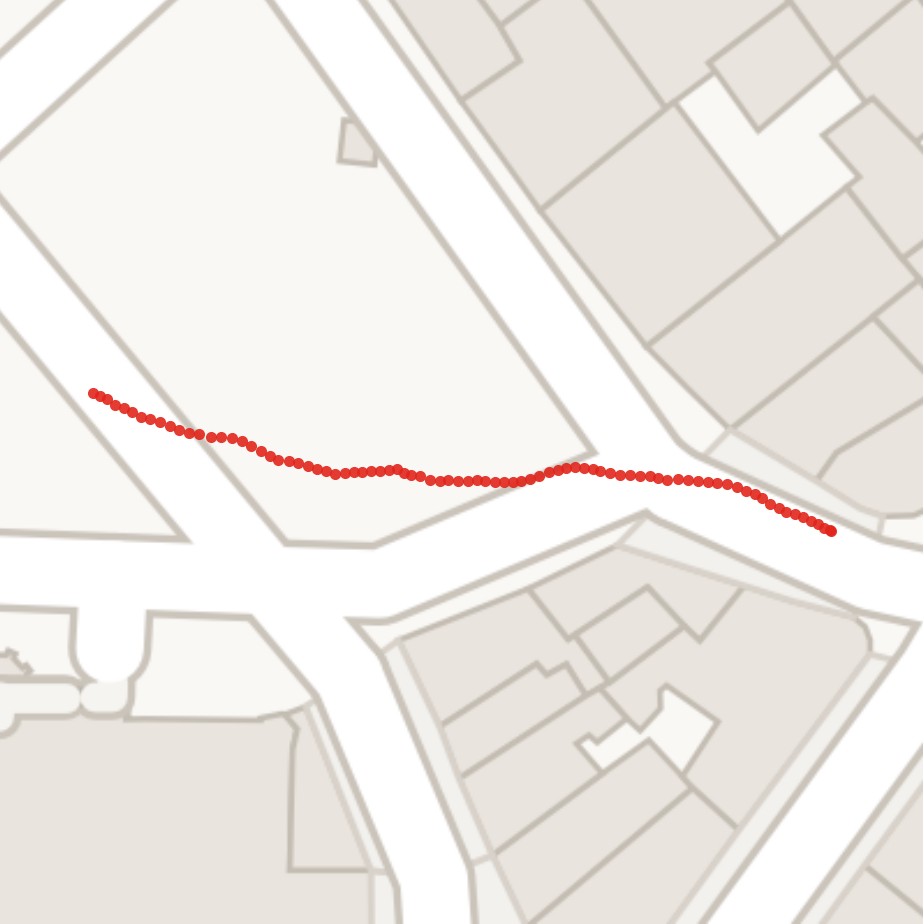} & \mapauc{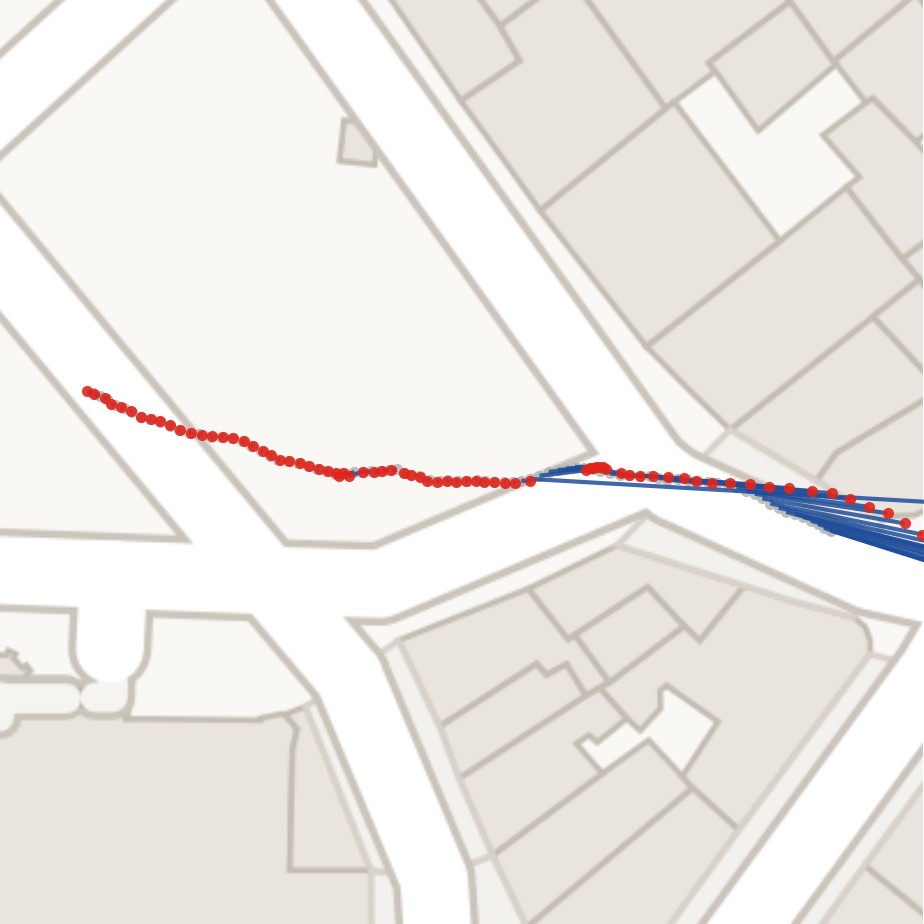}{85.7} & \mapauc{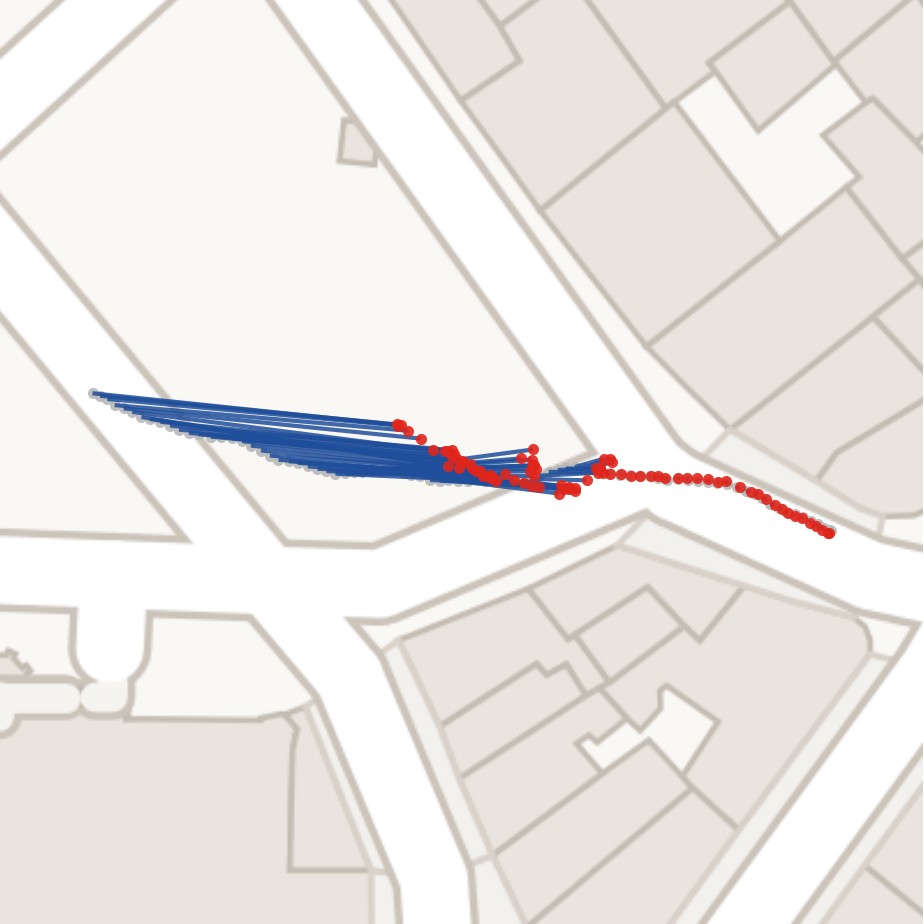}{38.4} & \mapauc{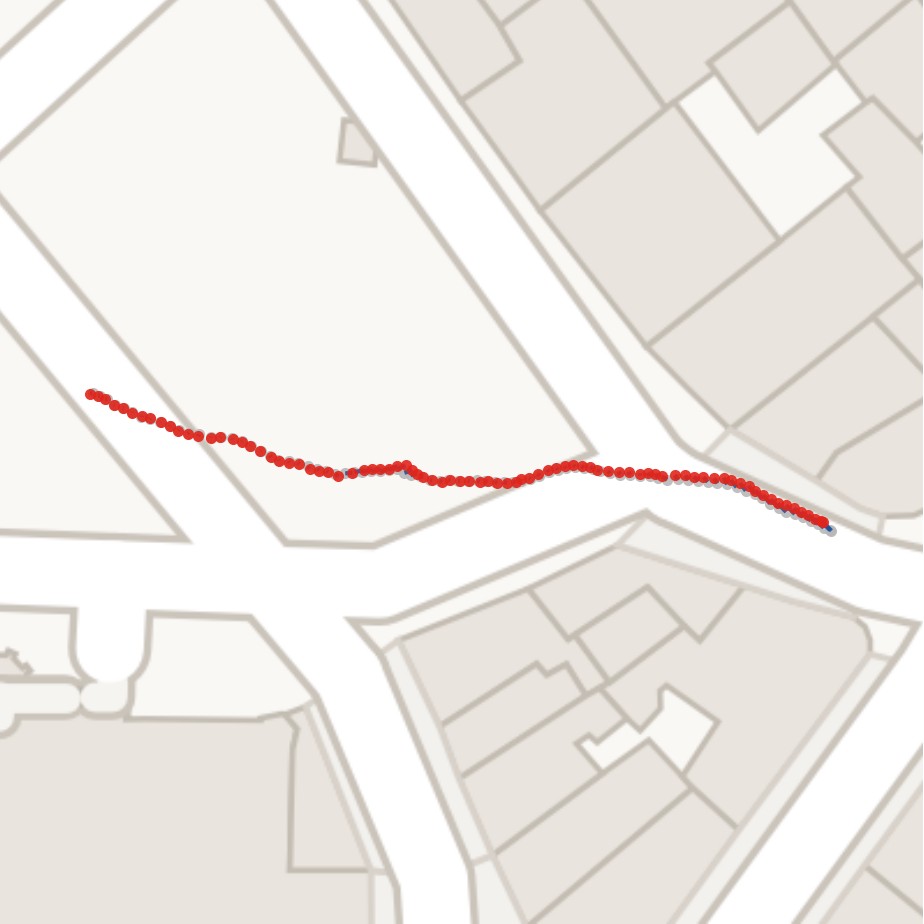}{87.2} & \mapauc{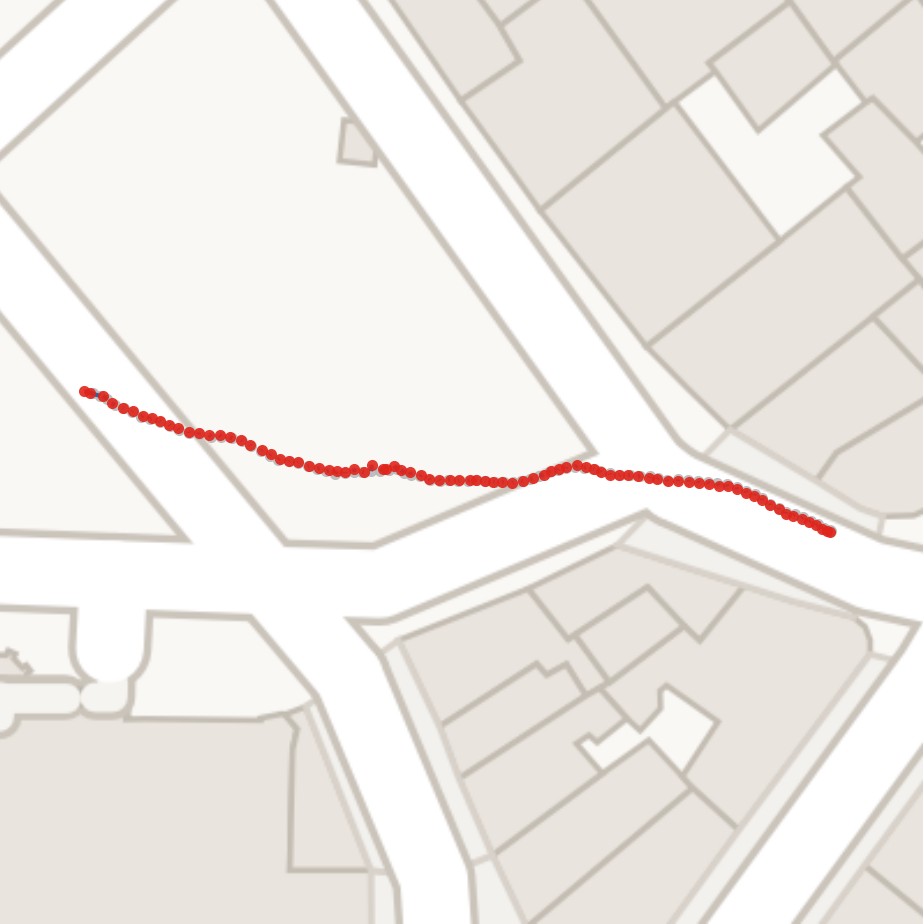}{88.7} \\[0pt]
\maprow{\textbf{+ \method}} & \mapname{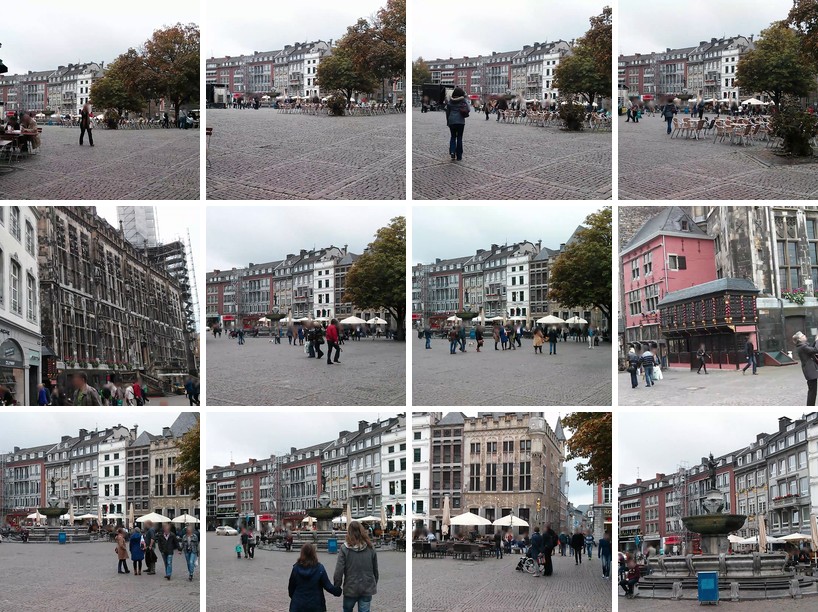}{nexus 6} & \mapaucbf{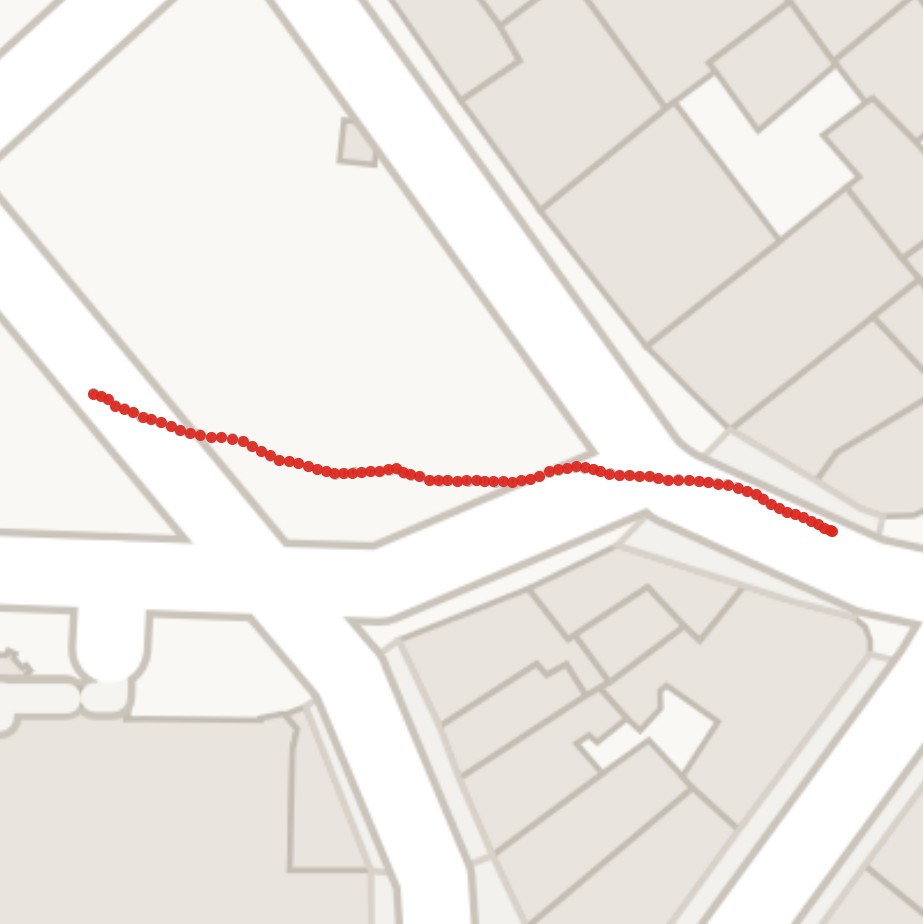}{96.9} & \mapaucbf{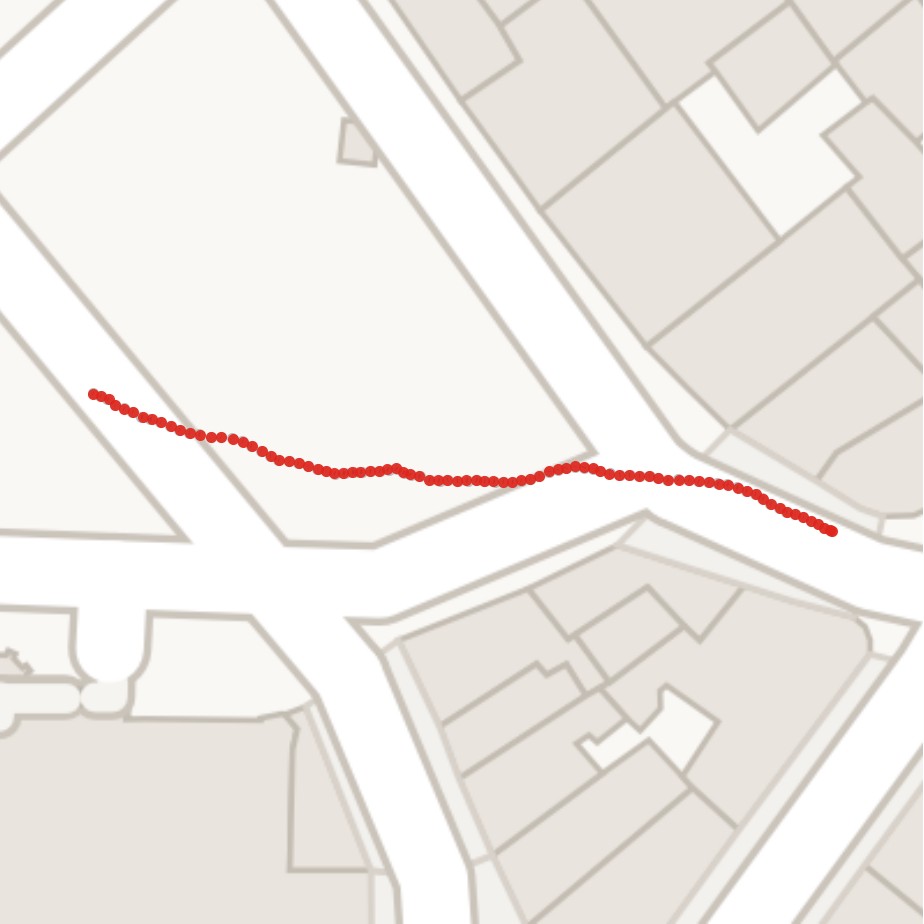}{97.1} & \mapaucbf{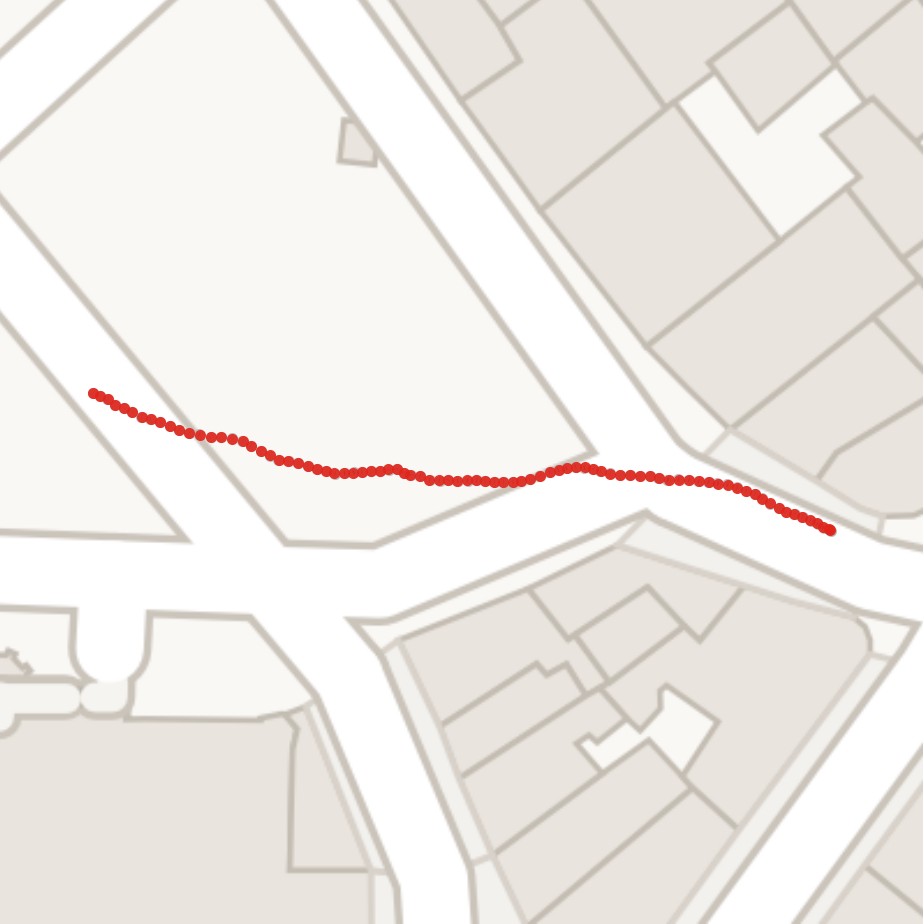}{97.1} & \mapaucbf{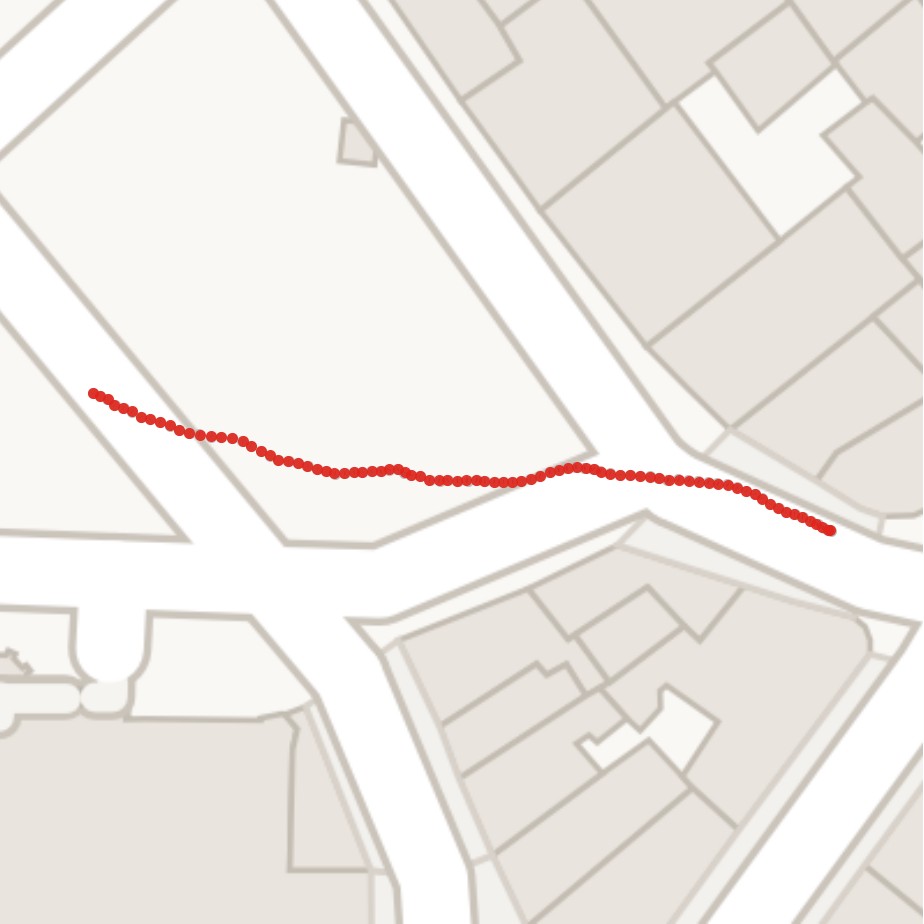}{96.8} \\[4pt]
\maprow{Initialization} & \mappanel{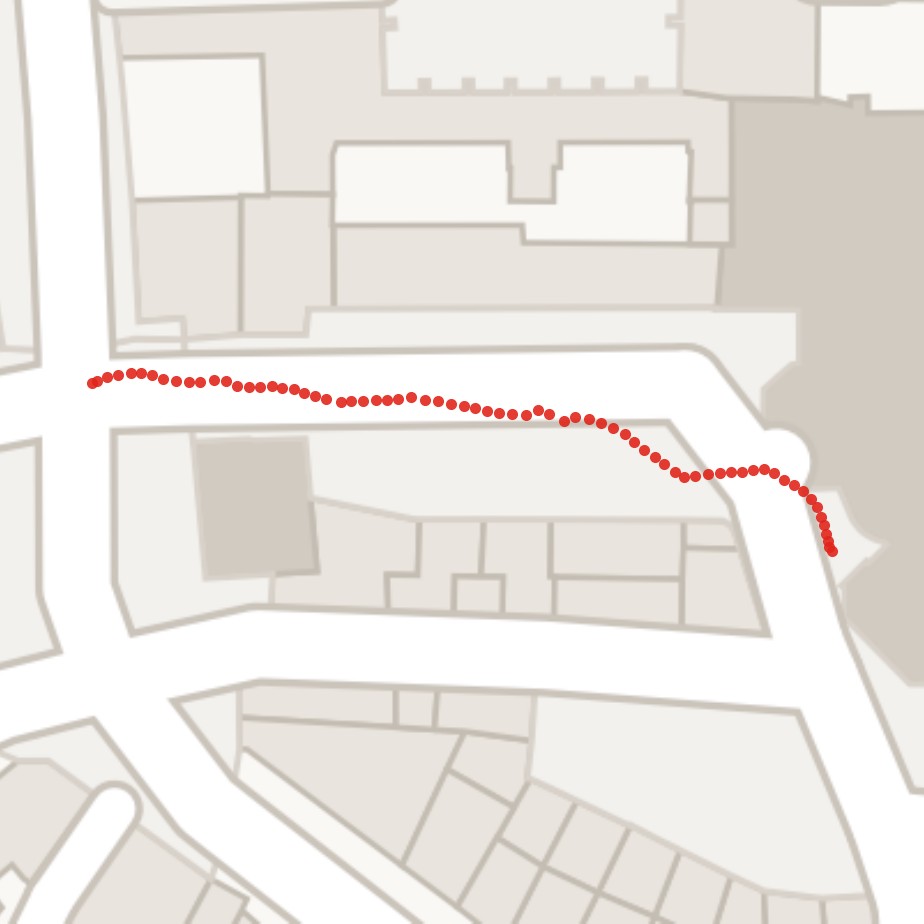} & \mapauc{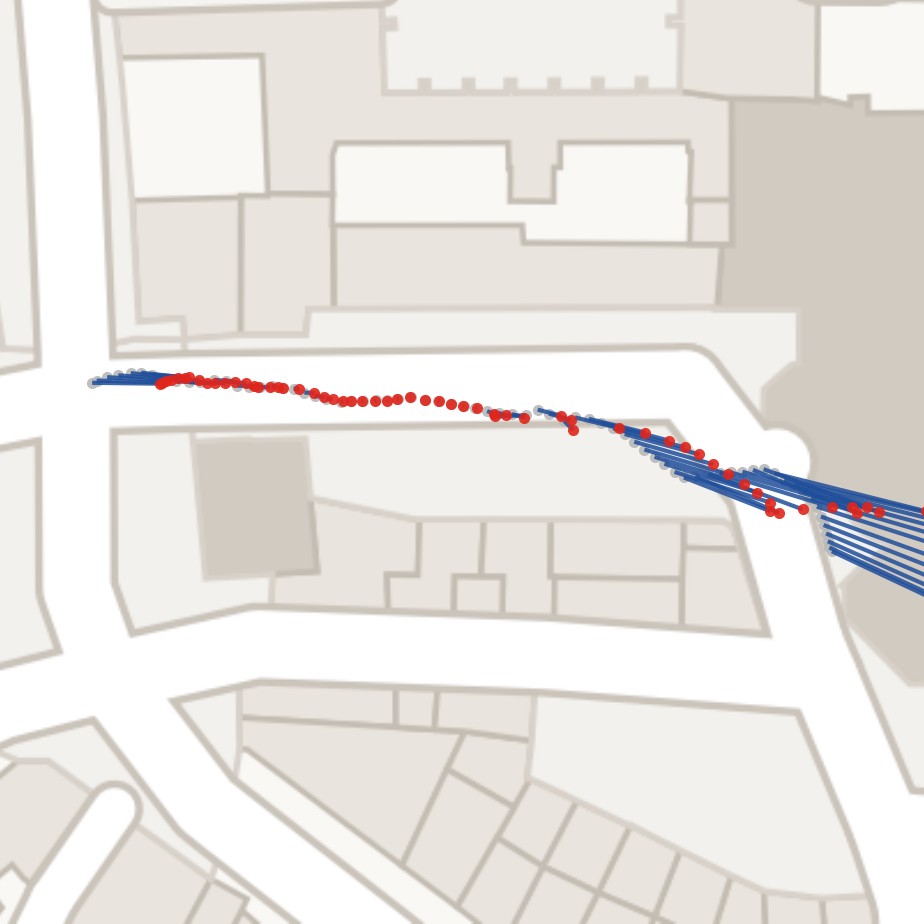}{52.9} & \mapauc{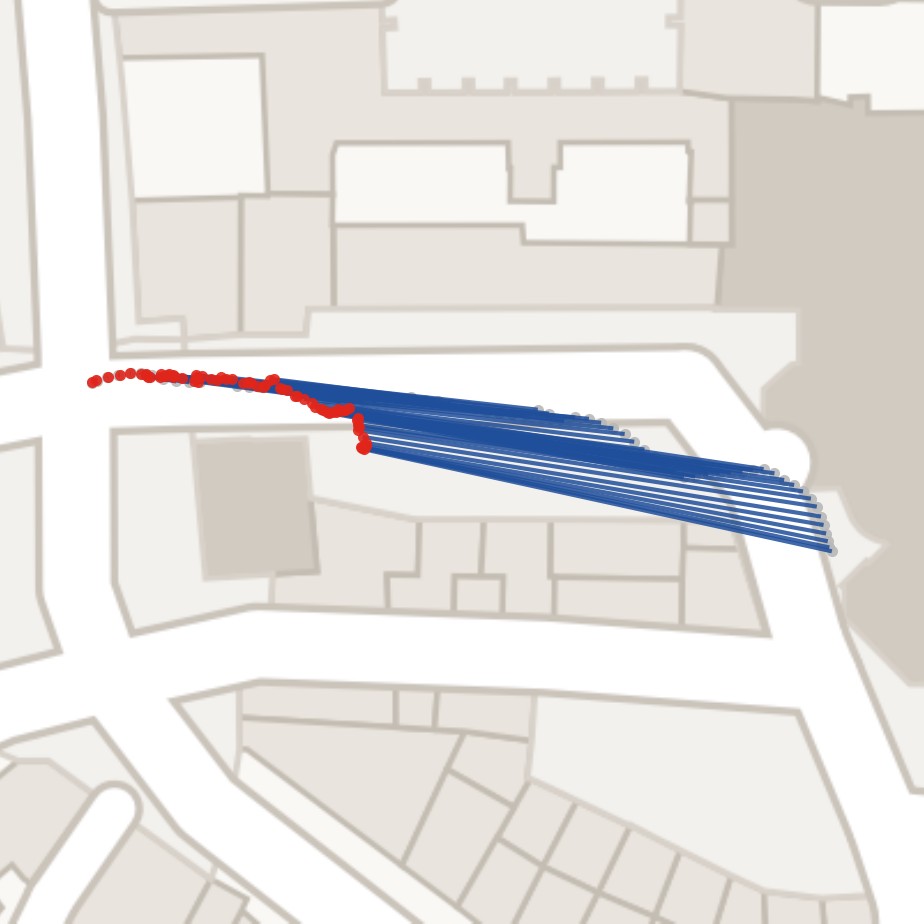}{31.5} & \mapauc{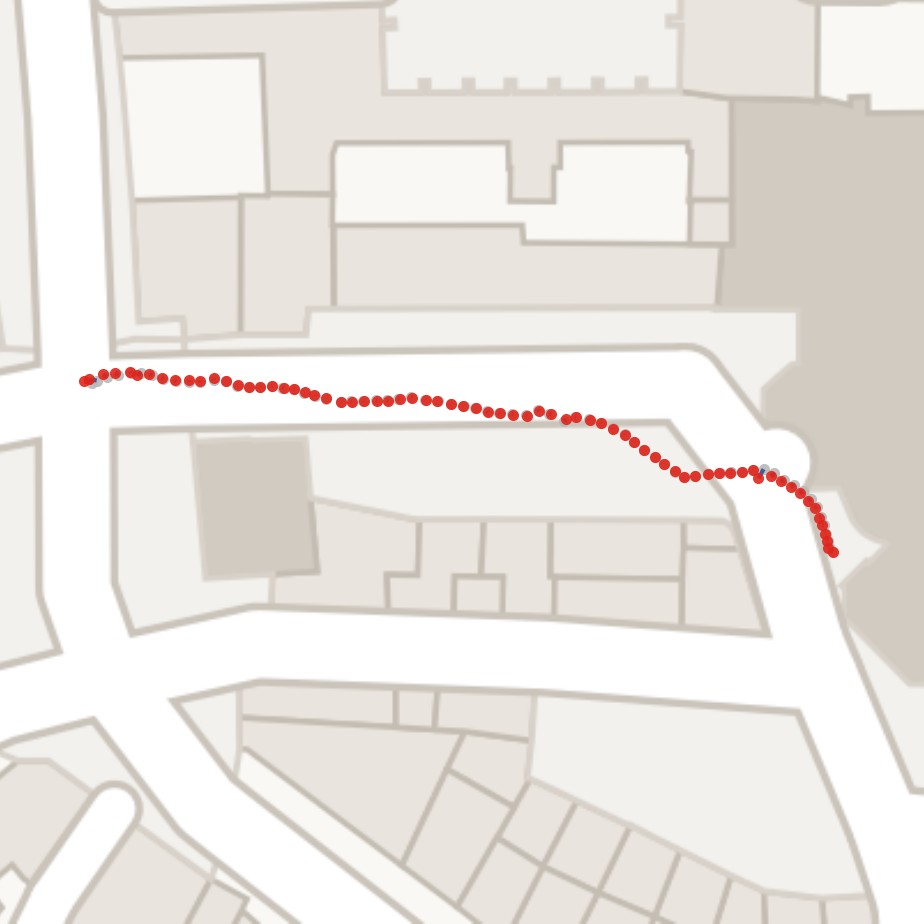}{82.0} & \mapauc{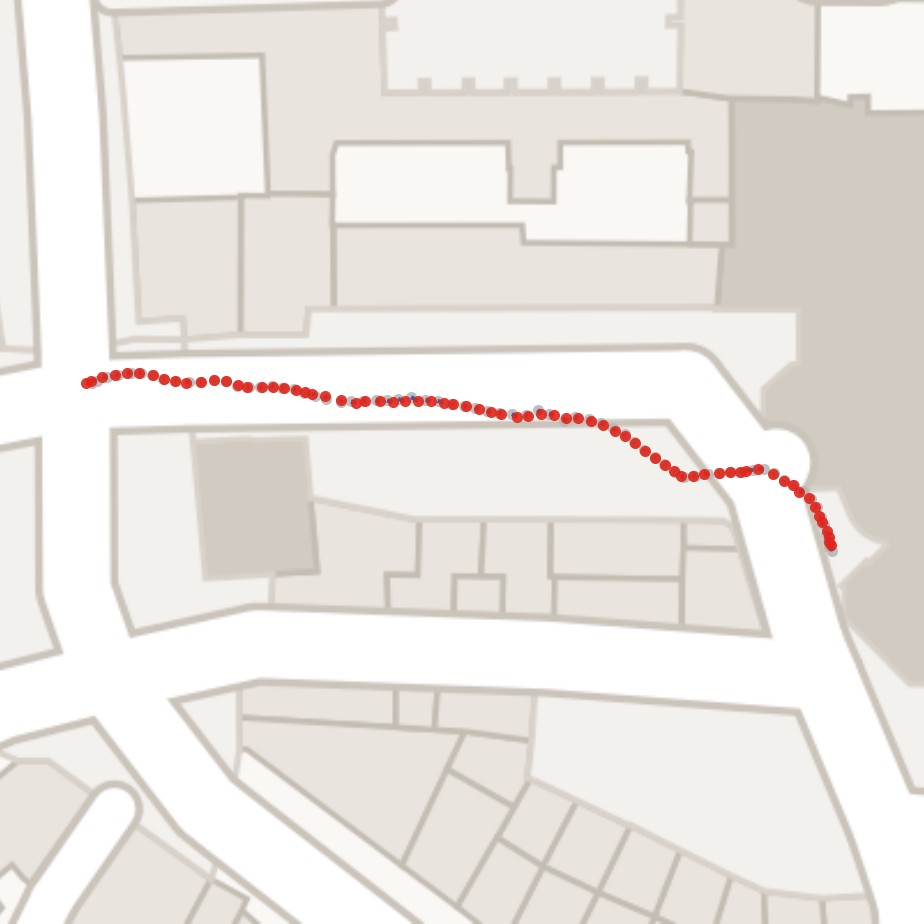}{83.0} \\[0pt]
\maprow{\textbf{+ \method}} & \mapname{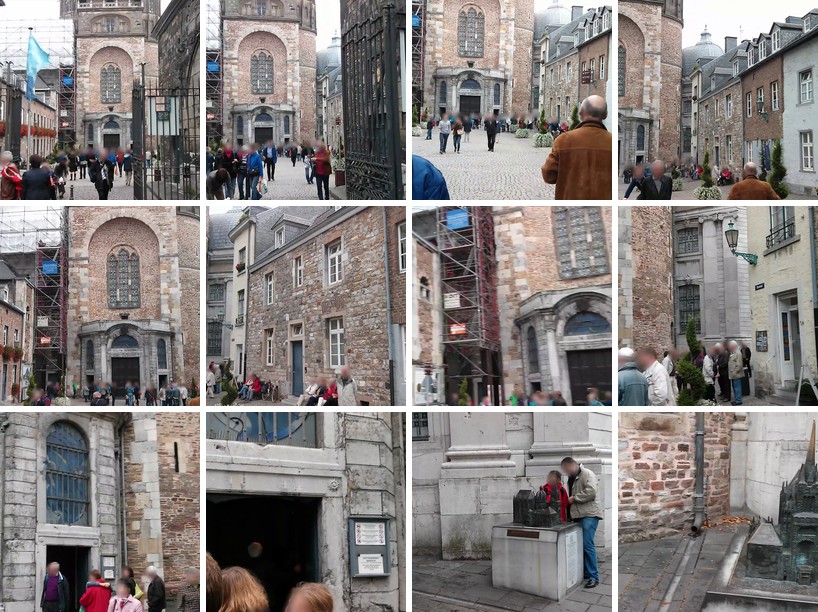}{nexus 8} & \mapaucbf{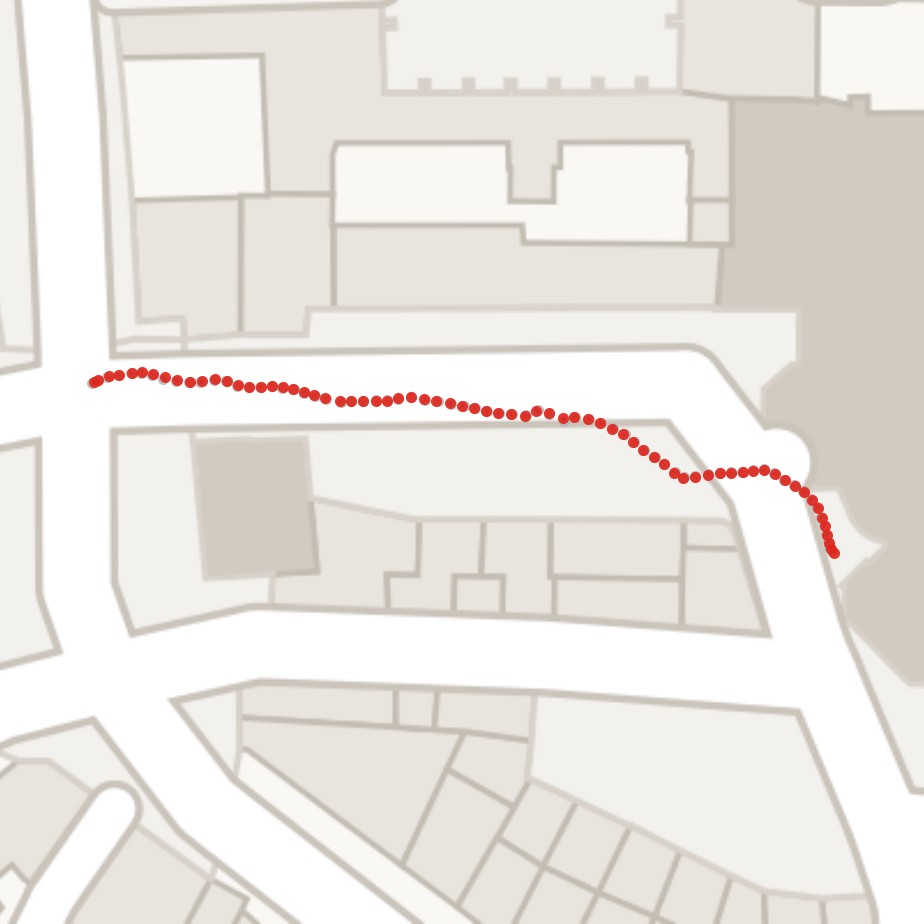}{94.5} & \mapaucbf{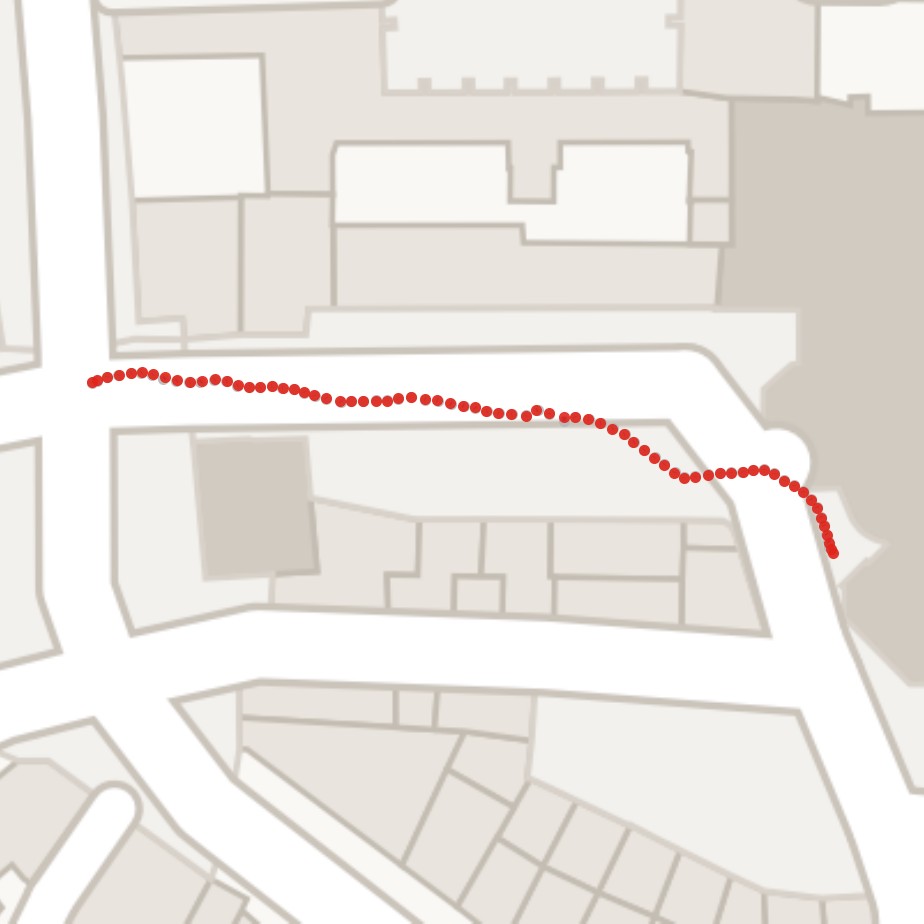}{93.7} & \mapaucbf{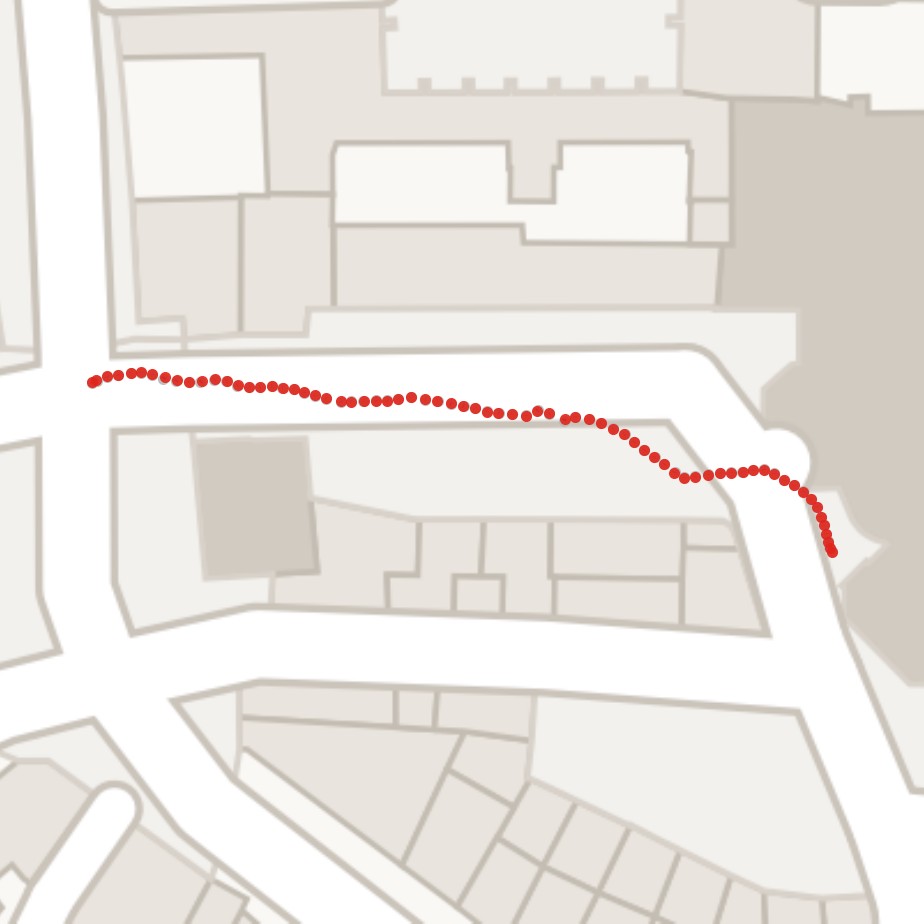}{94.7} & \mapaucbf{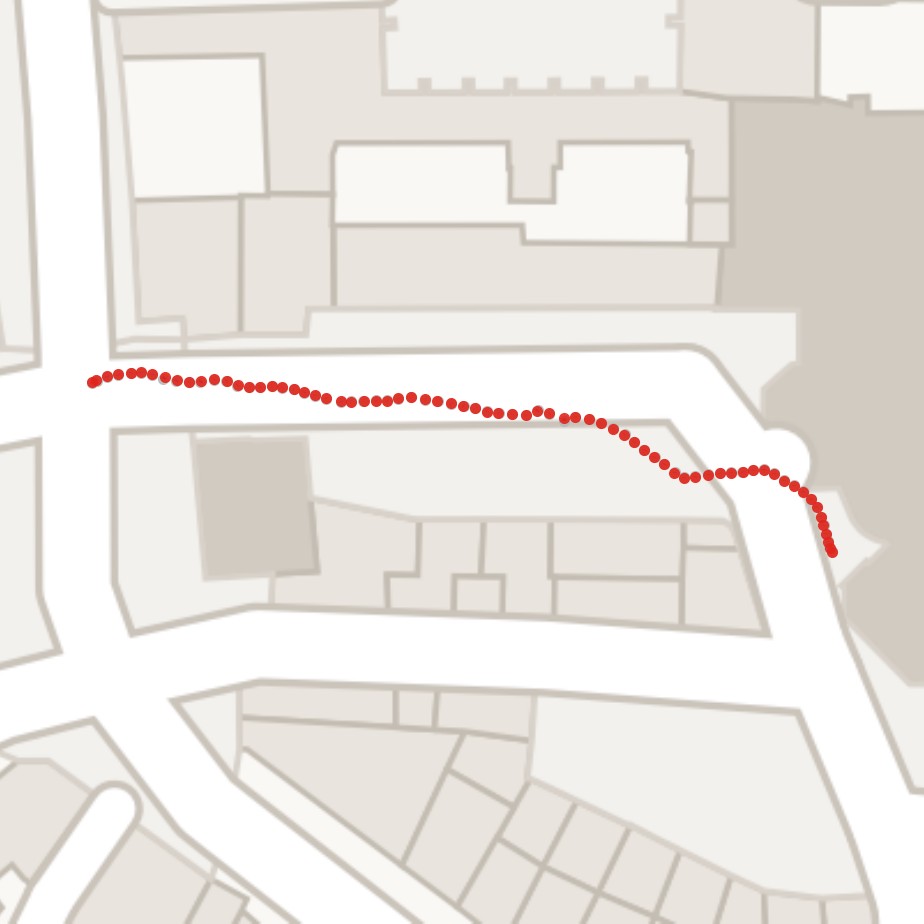}{94.7} \\[4pt]
\end{tabular}
\caption{Trajectory comparison on Aachen sequences on a map of the old town (map data \copyright~OpenStreetMap contributors): the Initialization cameras of each method (upper row of each sequence) and the same cameras after \method (lower row), aligned to the reference model by Sim(3). $^\ddagger$: our chunking. $^\S$: the method's own chunking.}
\label{fig:aachen_seq_map}
\end{figure*}

\ifdefined\mapW\else\newlength{\mapW}\fi\setlength{\mapW}{1.22in}
\begin{figure*}[p]
\def\mappanel#1{{\setlength{\fboxsep}{0pt}\setlength{\fboxrule}{0.3pt}\fcolorbox{black!20}{white}{\includegraphics[width=\dimexpr\mapW-0.6pt\relax]{fig/large_scenes_map/#1}}}}
\def\mapauc#1#2{\mappanel{#1}\llap{\raisebox{1.2pt}{\setlength{\fboxsep}{0.8pt}\colorbox{white}{\tiny AUC@3$^\circ$ #2}}\hspace{1.2pt}}}  
\def\mapaucbf#1#2{\mappanel{#1}\llap{\raisebox{1.2pt}{\setlength{\fboxsep}{0.8pt}\colorbox{white}{\tiny\bfseries AUC@3$^\circ$ #2}}\hspace{1.2pt}}}
\def\maprow#1{\rotatebox{90}{\scriptsize #1}}
\def\mapname#1#2{\raisebox{0pt}[0pt][0pt]{\parbox{\mapW}{\centering\includegraphics[width=0.97\mapW]{fig/large_scenes_map/#1}\\[1pt]\footnotesize\bfseries #2}}}  
\centering
\setlength{\tabcolsep}{0.8pt}
\renewcommand{\arraystretch}{0.7}
\begin{tabular}{@{}>{\centering\arraybackslash}m{0.9em} >{\centering\arraybackslash}m{\mapW}@{\hspace{4pt}} *{4}{>{\centering\arraybackslash}m{\mapW}} @{}}
 & \small\bfseries GT & \small\bfseries COLMAP & \small\bfseries GLUEMAP$^\S$ & \small\bfseries Pi3X$^\ddagger$ & \small\bfseries VGGT-$\Omega^\ddagger$ \\[2pt]
\maprow{Initialization} & \mappanel{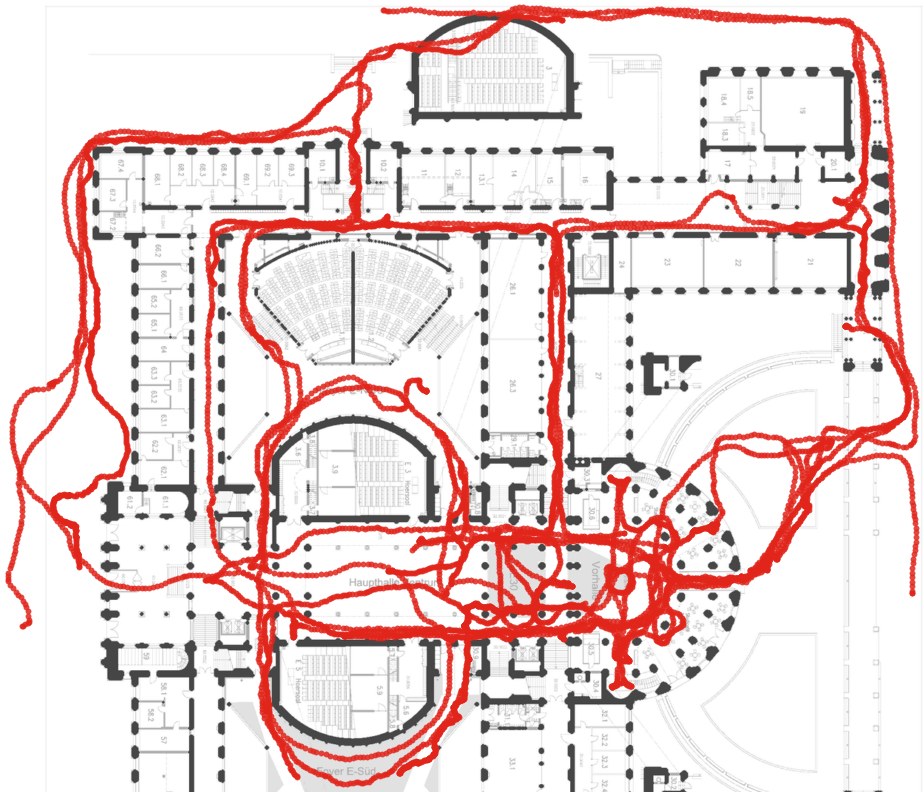} & \mapauc{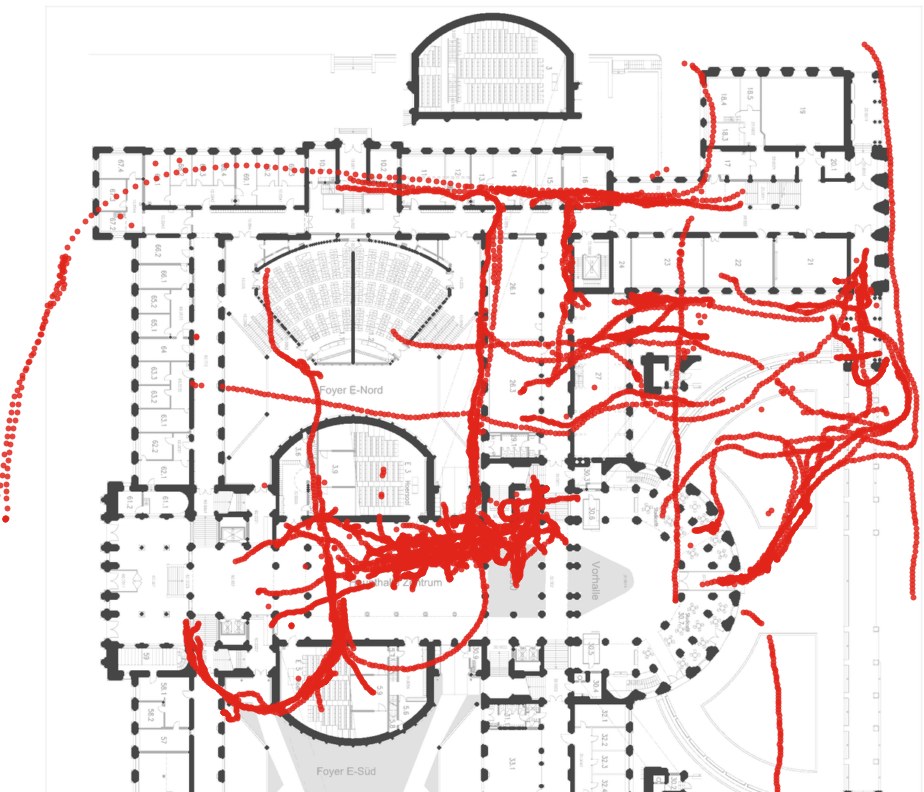}{3.2} & \mapauc{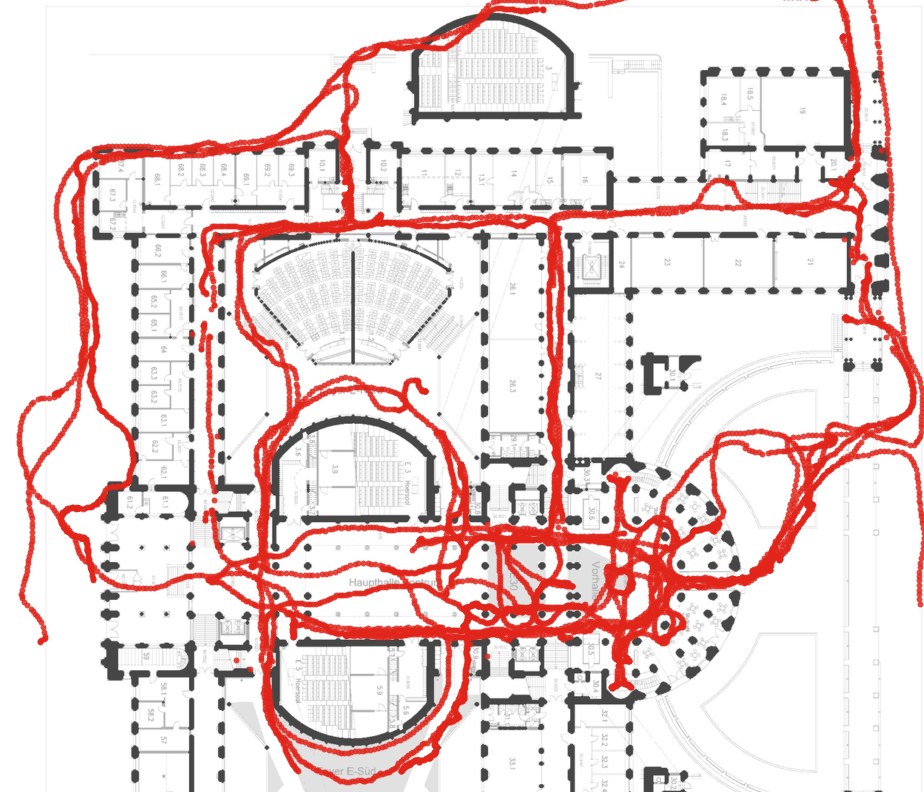}{37.1} & \mapauc{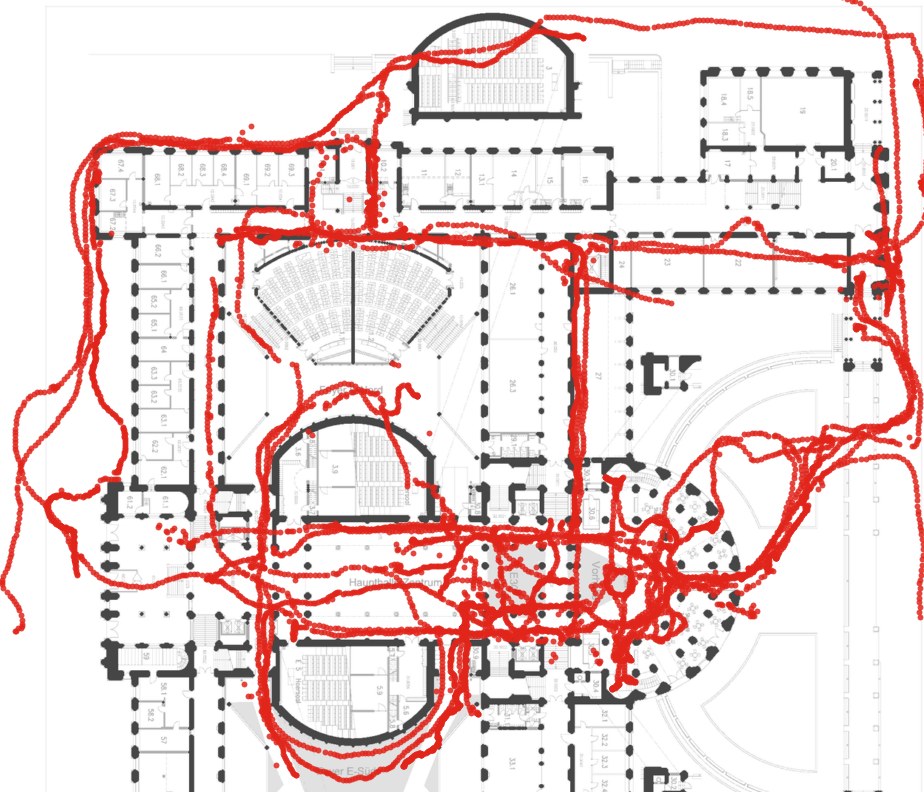}{13.3} & \mapauc{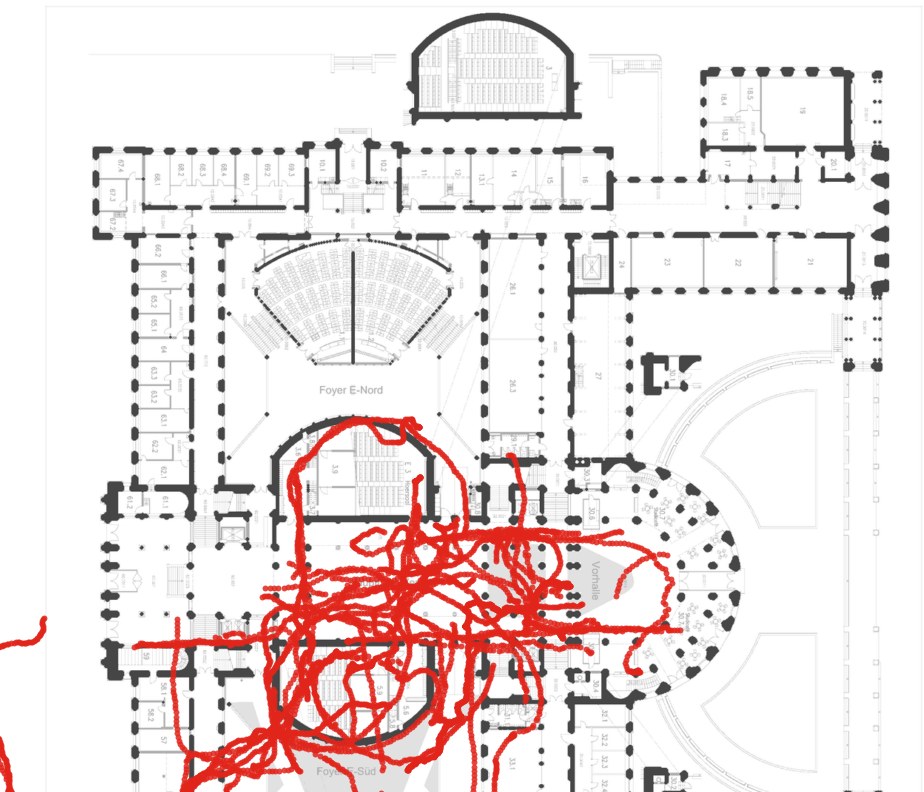}{1.9} \\[0pt]
\maprow{\textbf{+ \method}} & \mapname{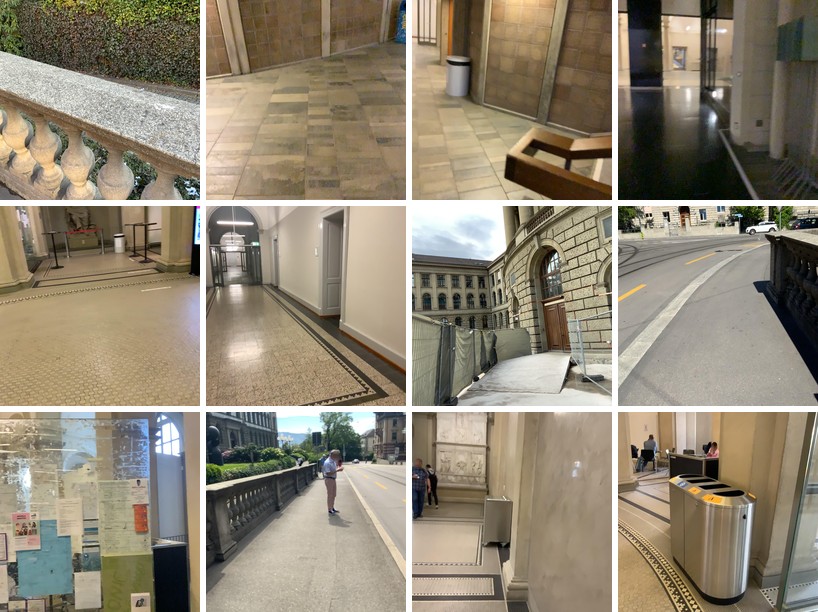}{LaMAR (HGE)} & \mapaucbf{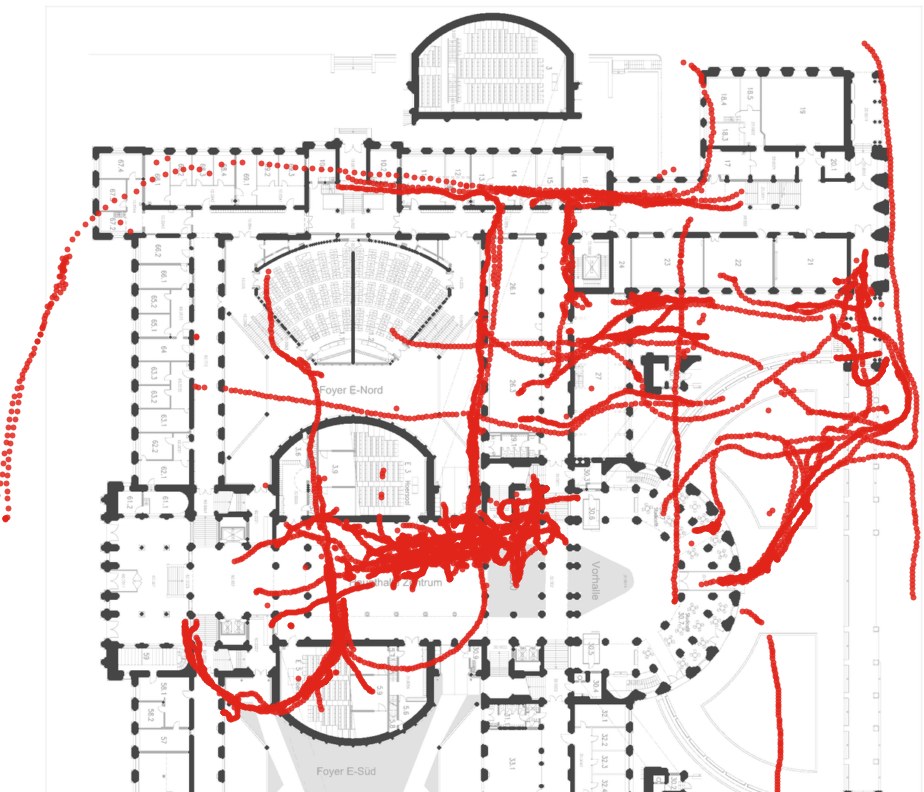}{3.5} & \mapaucbf{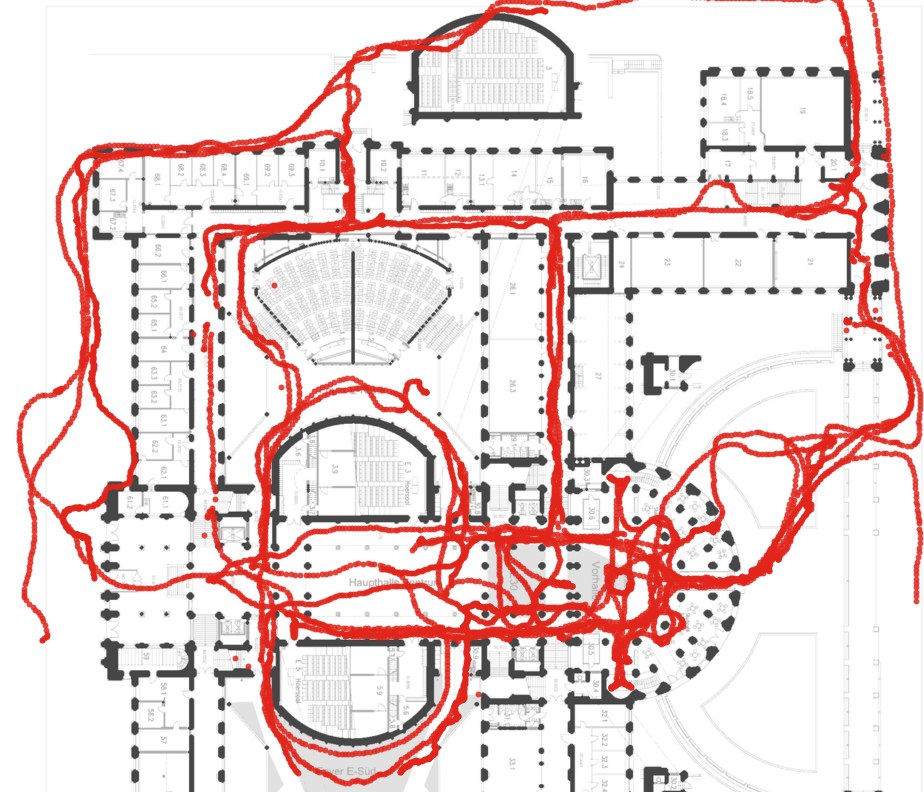}{40.6} & \mapaucbf{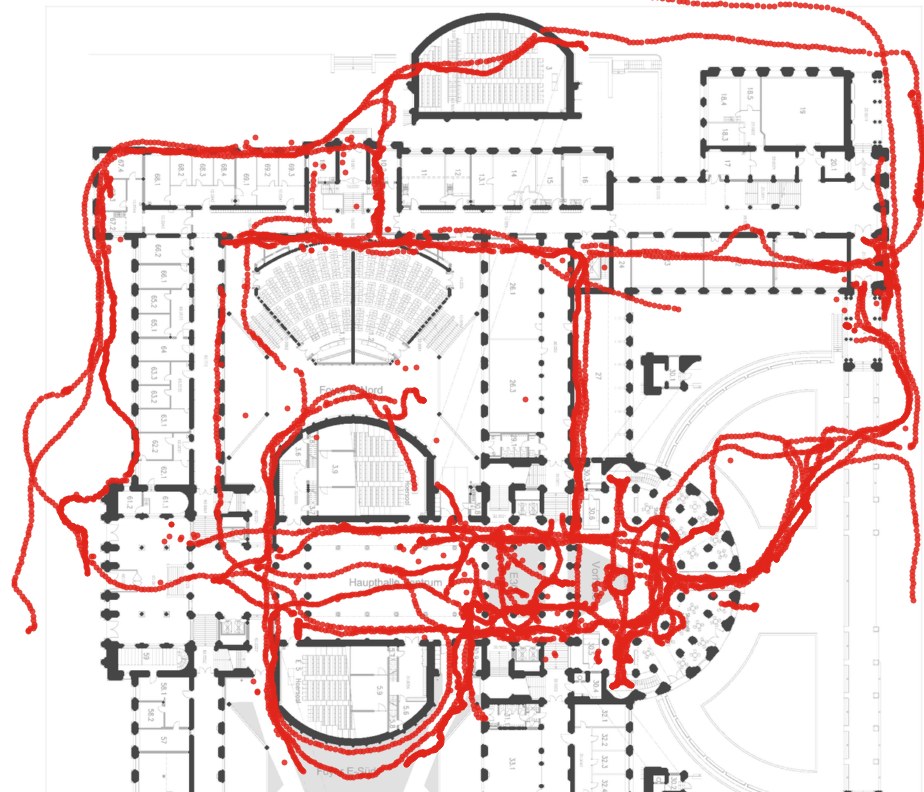}{24.3} & \mapaucbf{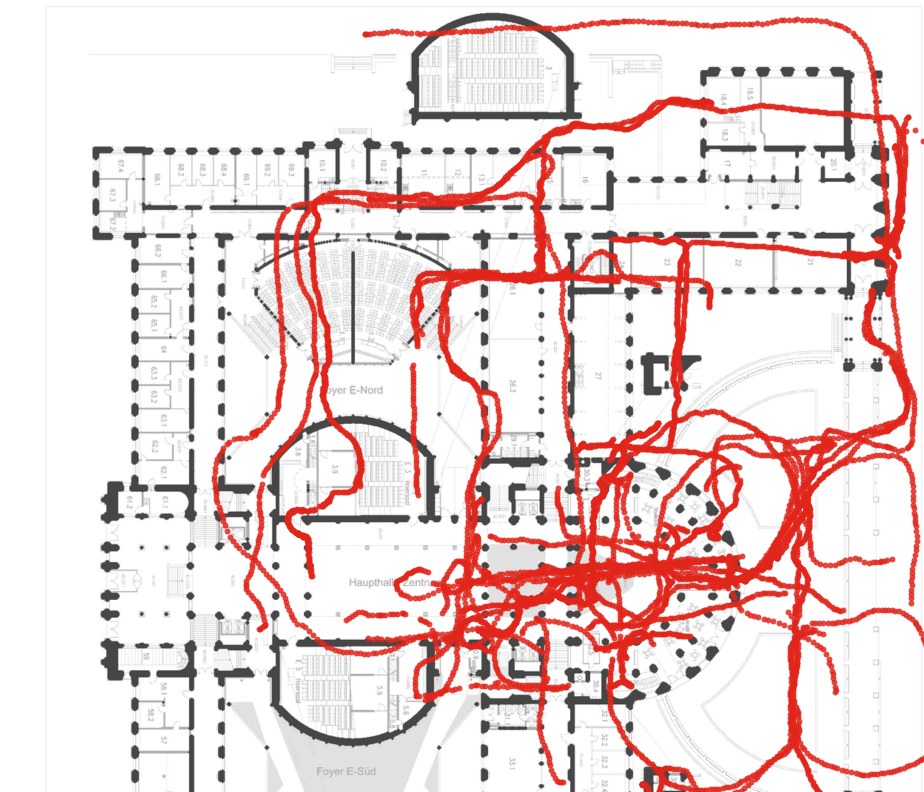}{4.9} \\[4pt]
\maprow{Initialization} & \mappanel{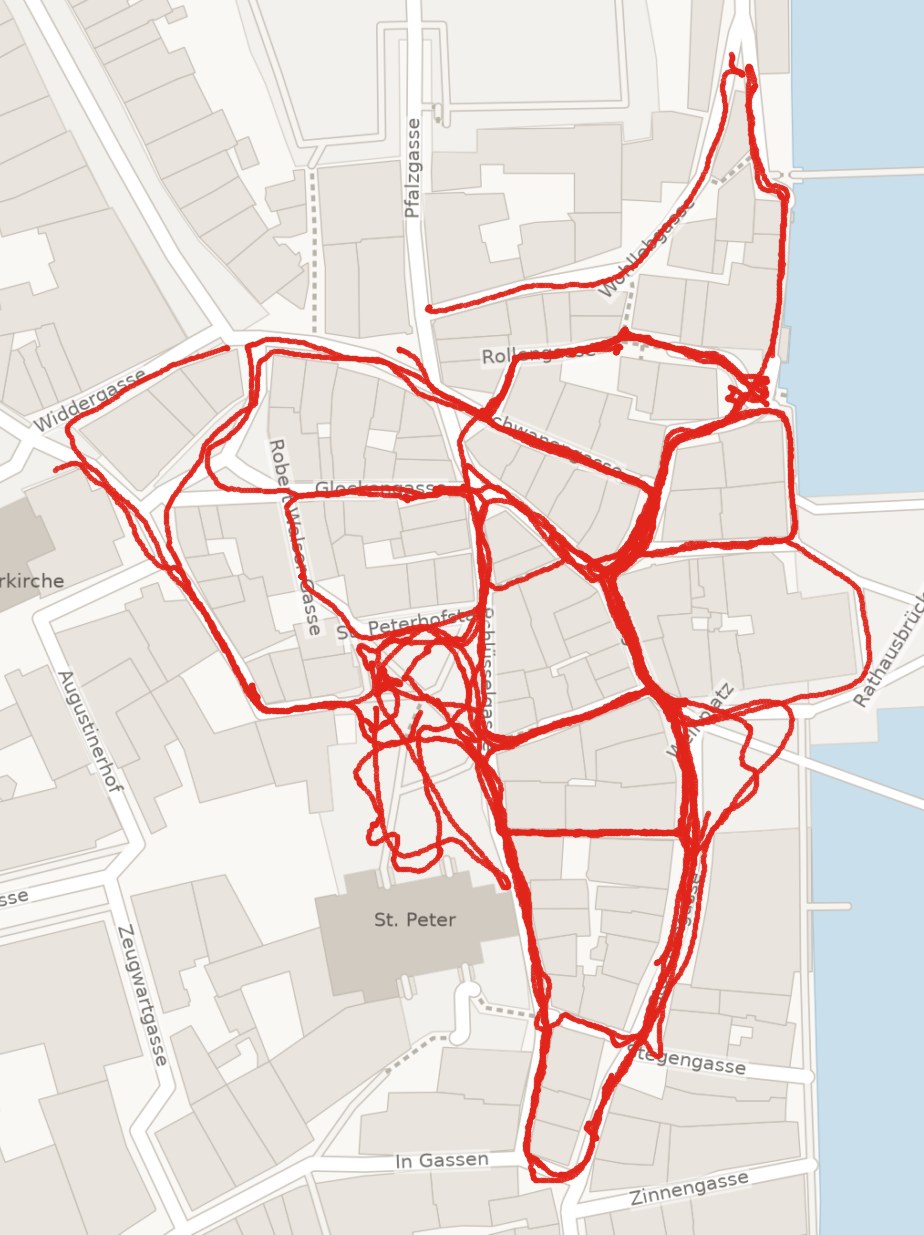} & \mapauc{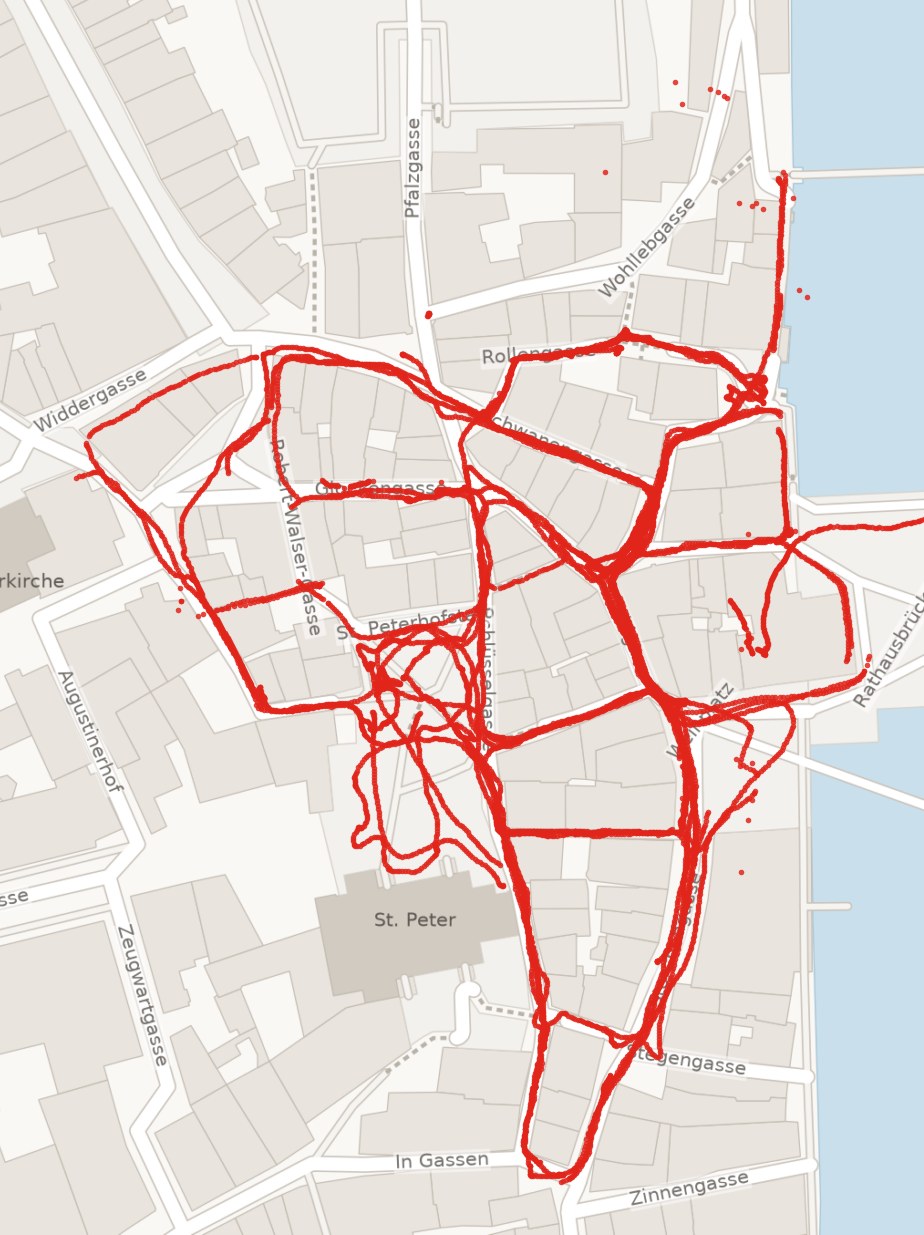}{40.4} & \mapauc{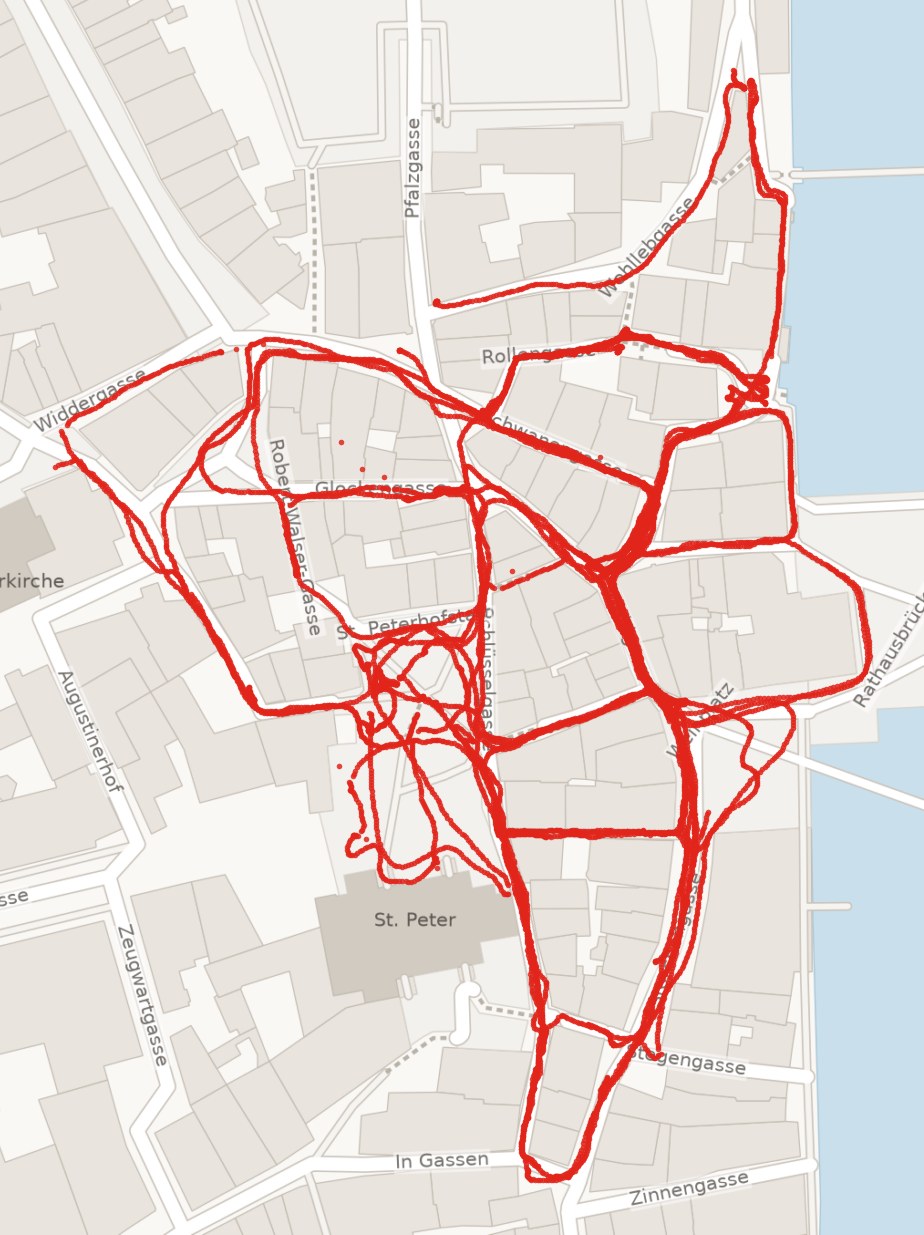}{44.6} & \mapauc{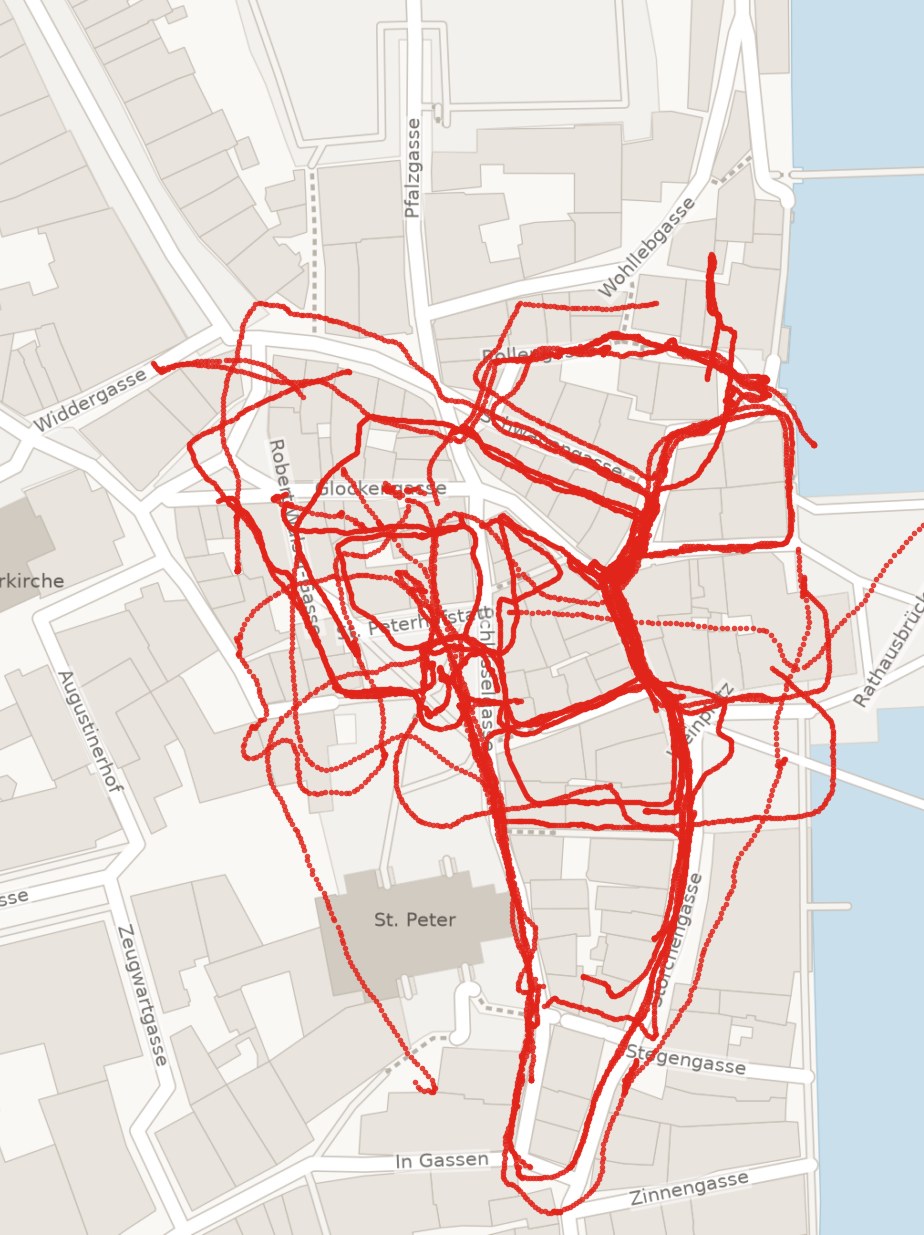}{4.0} & \mapauc{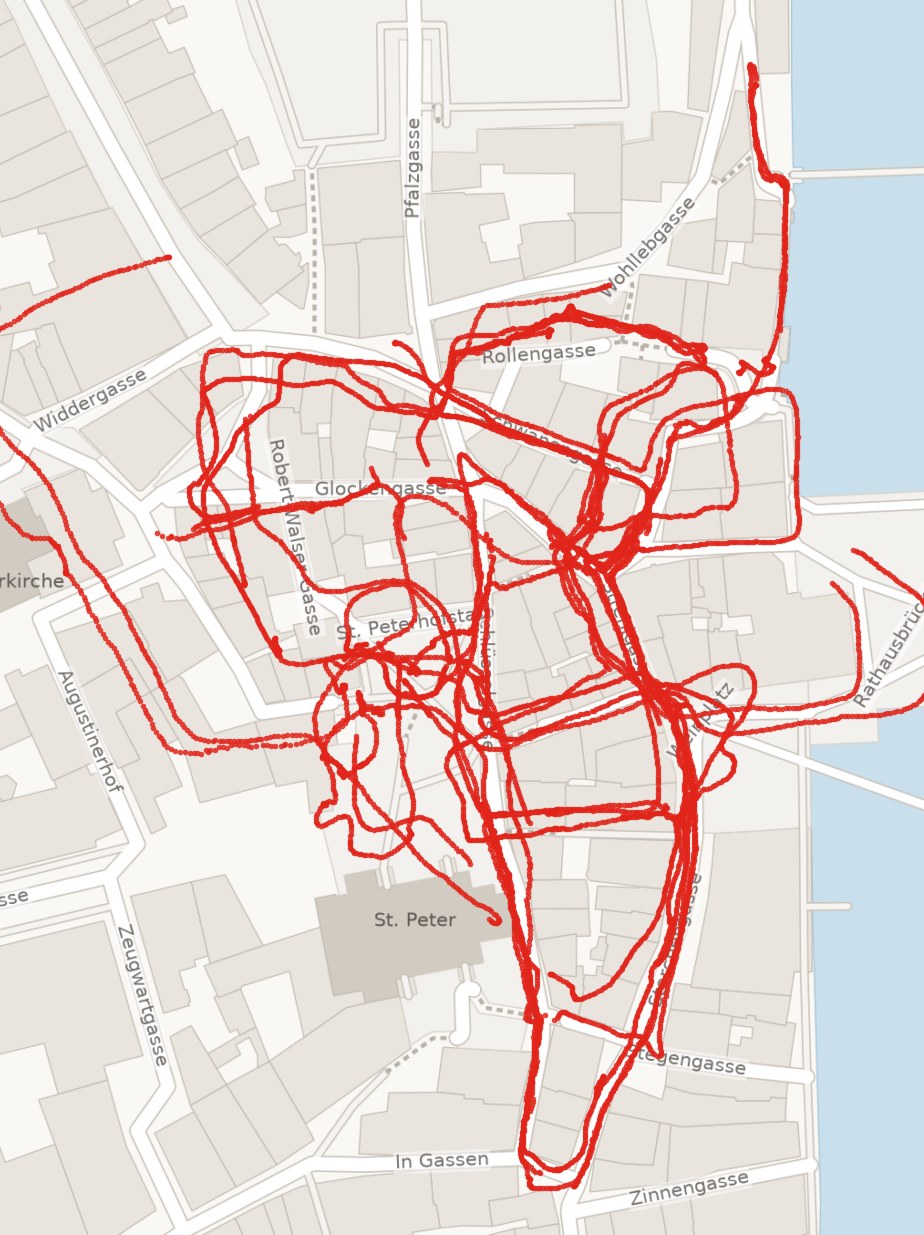}{3.3} \\[0pt]
\maprow{\textbf{+ \method}} & \mapname{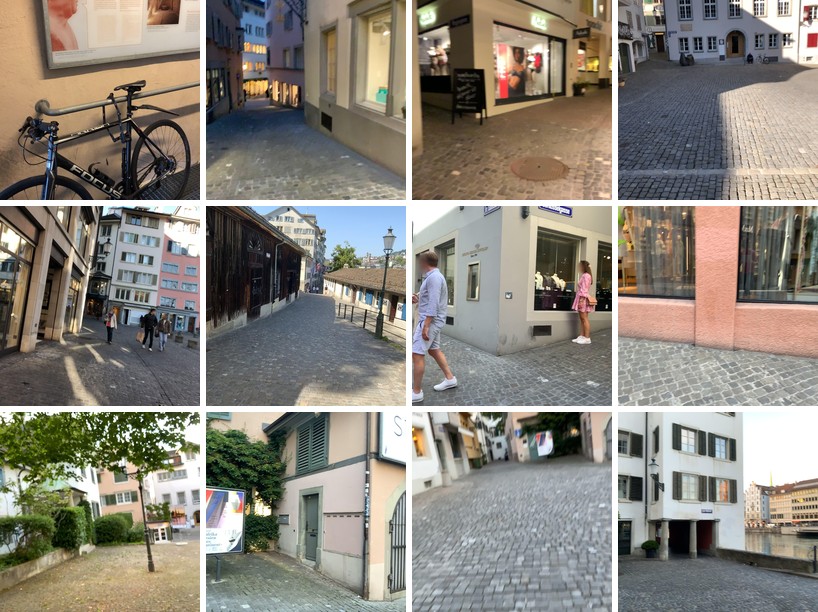}{LaMAR (LIN)} & \mapaucbf{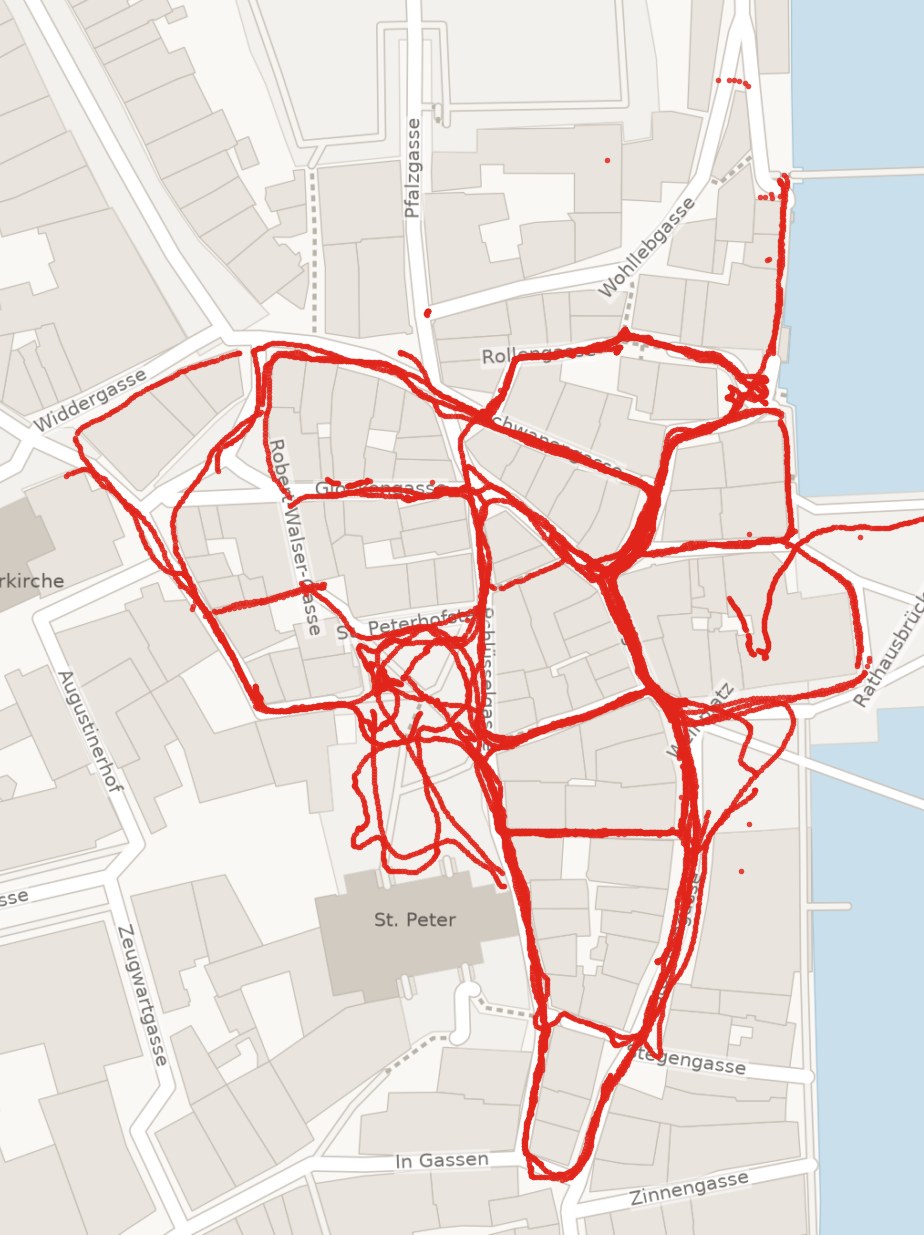}{46.6} & \mapaucbf{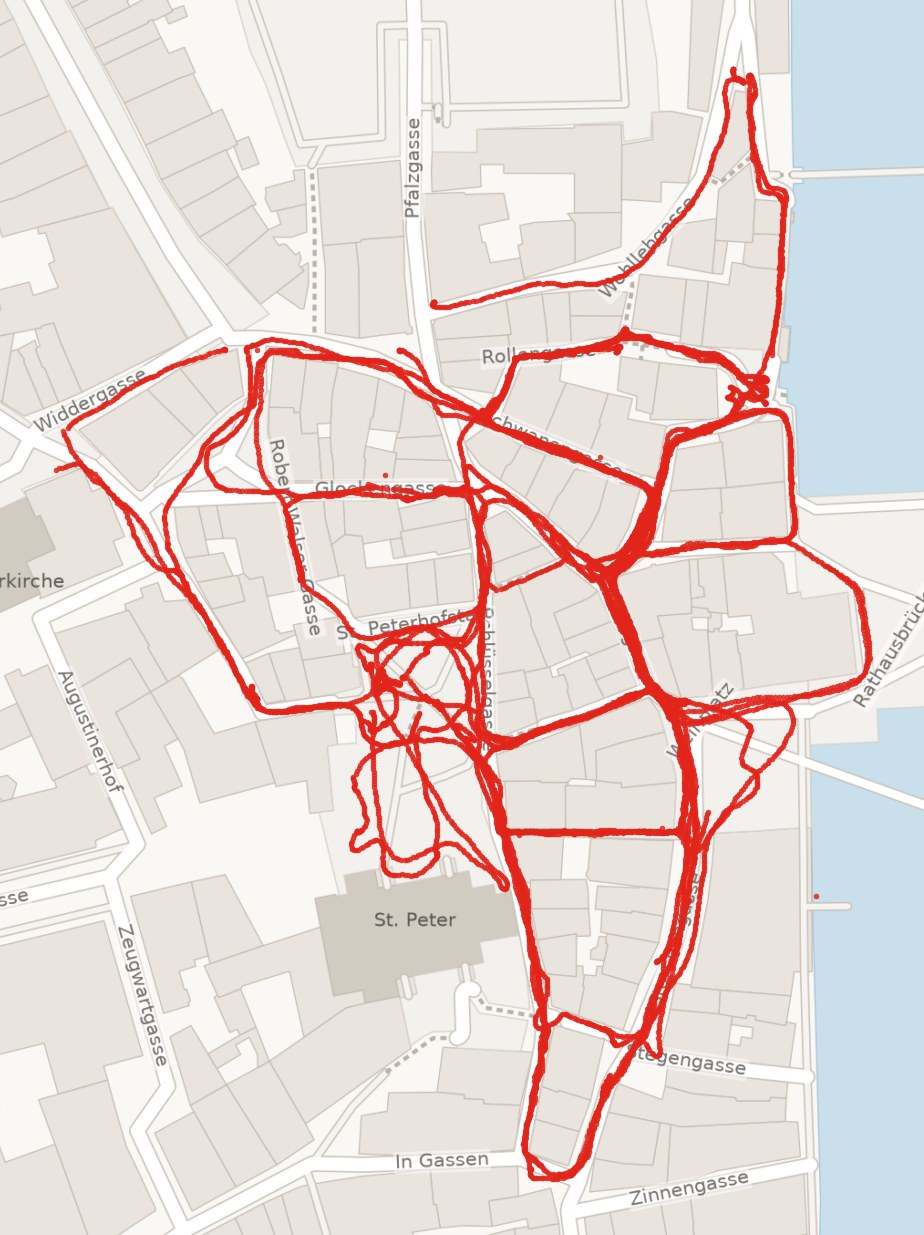}{49.3} & \mapaucbf{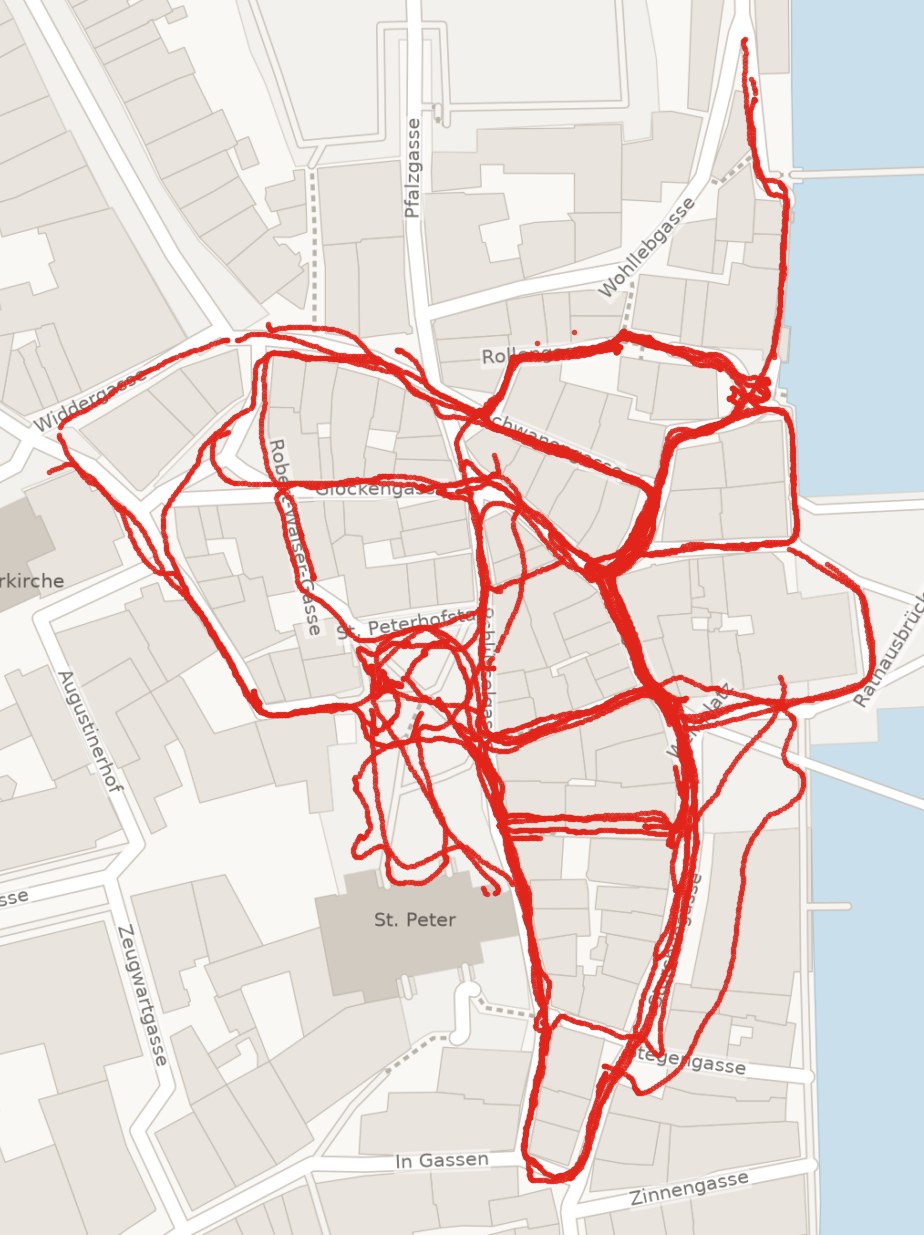}{26.3} & \mapaucbf{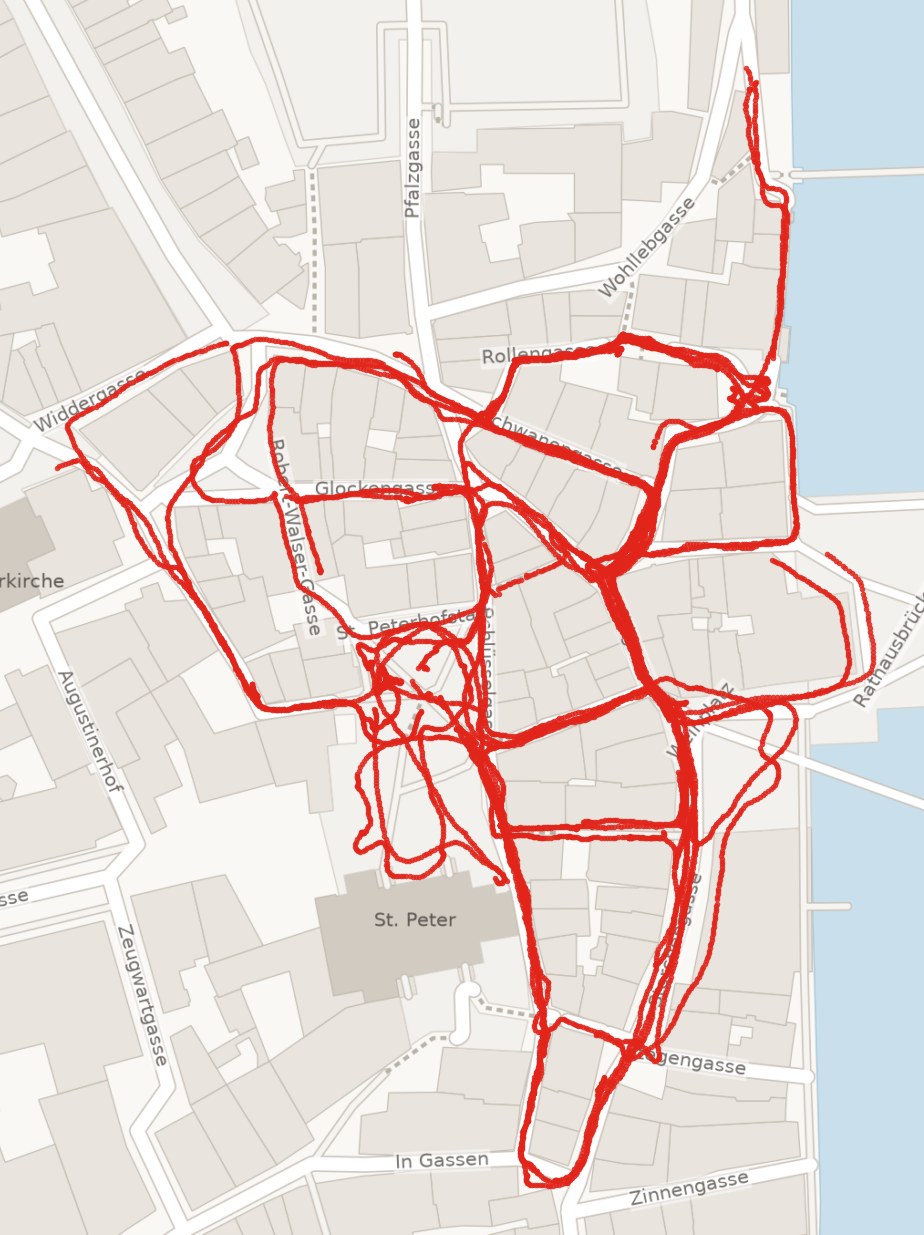}{30.5} \\[4pt]
\maprow{Initialization} & \mappanel{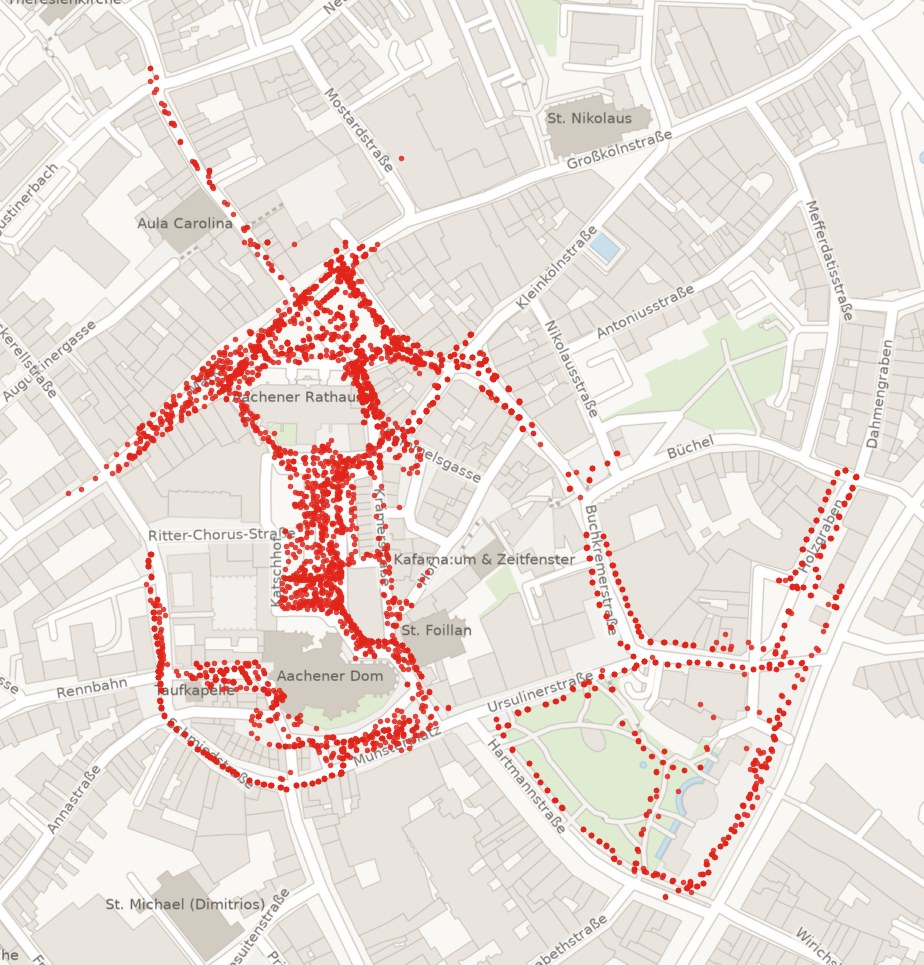} & \mapauc{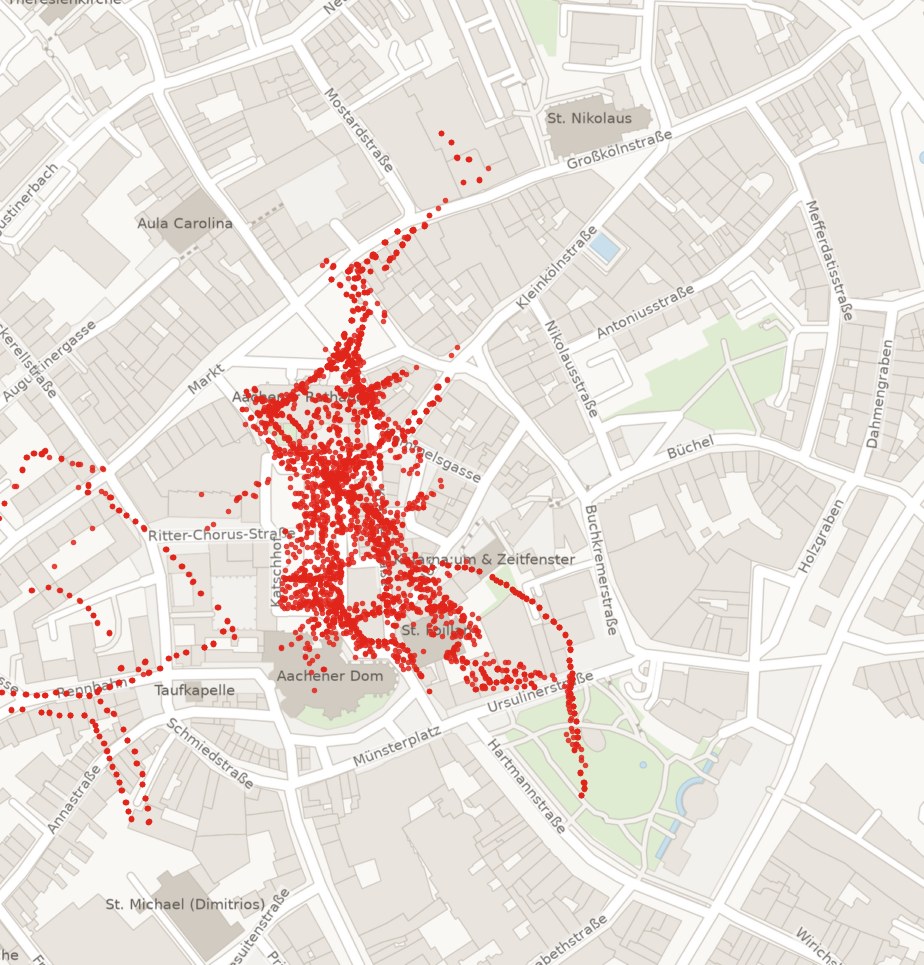}{23.8} & \mapauc{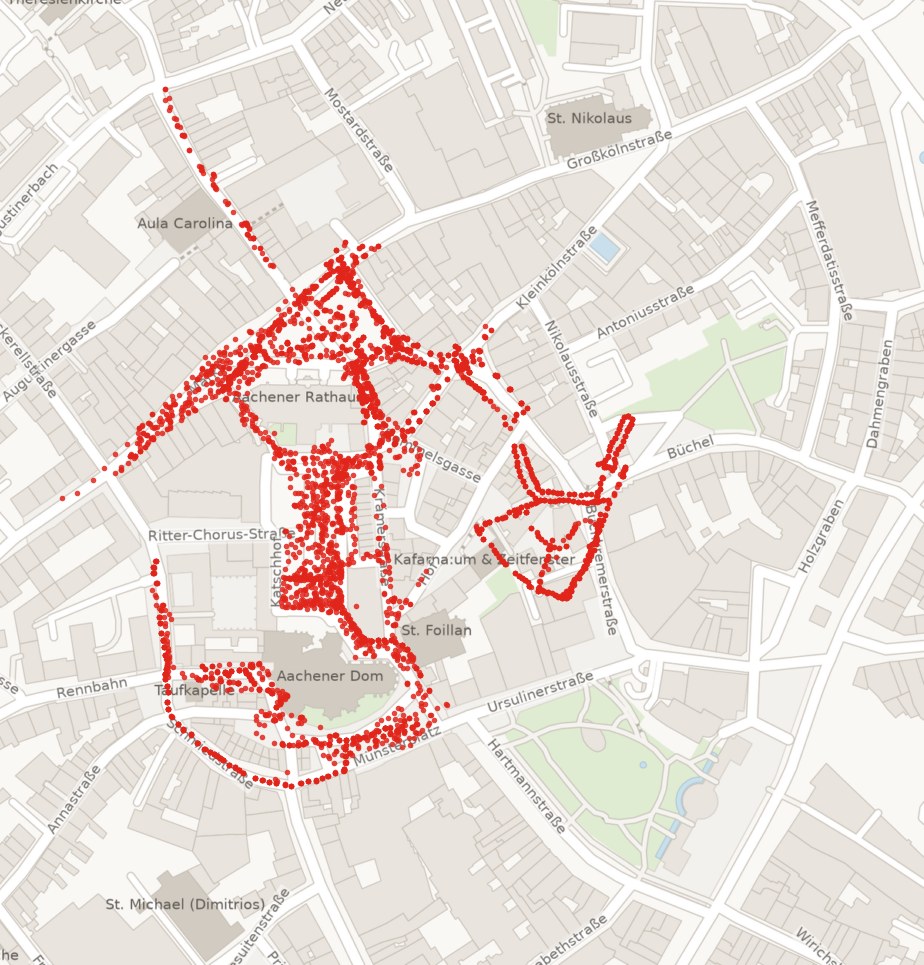}{51.4} & \mapauc{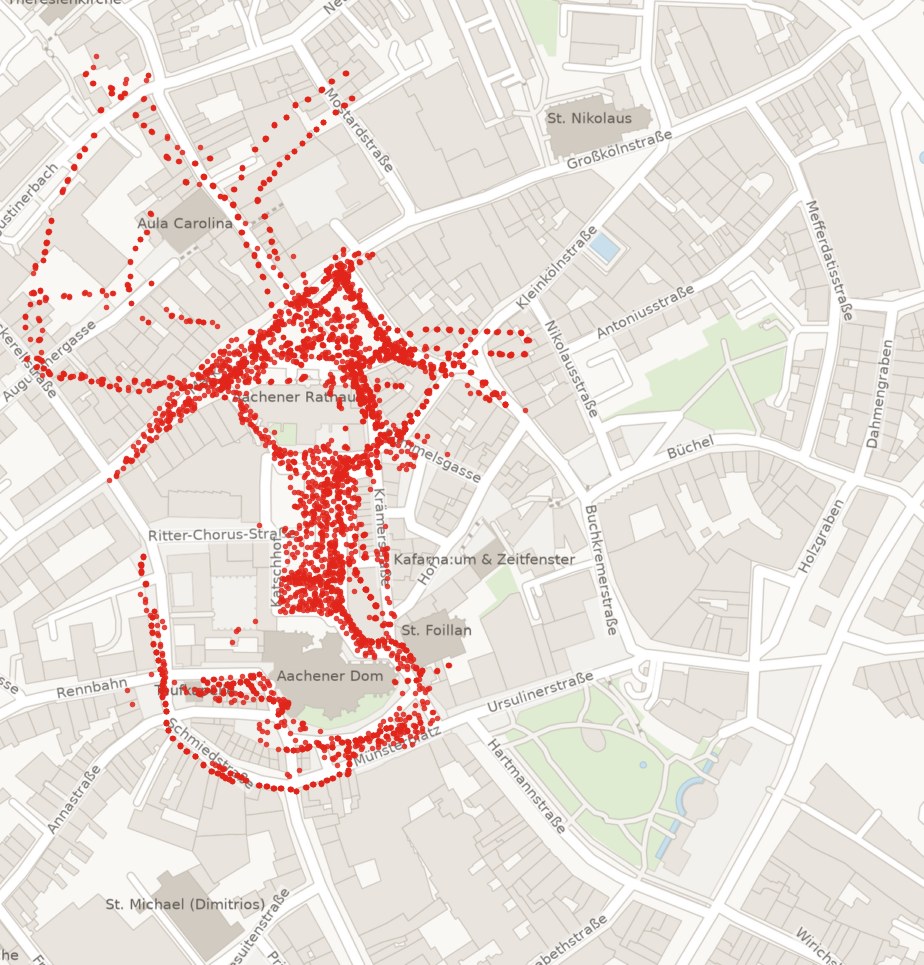}{14.6} & \mapauc{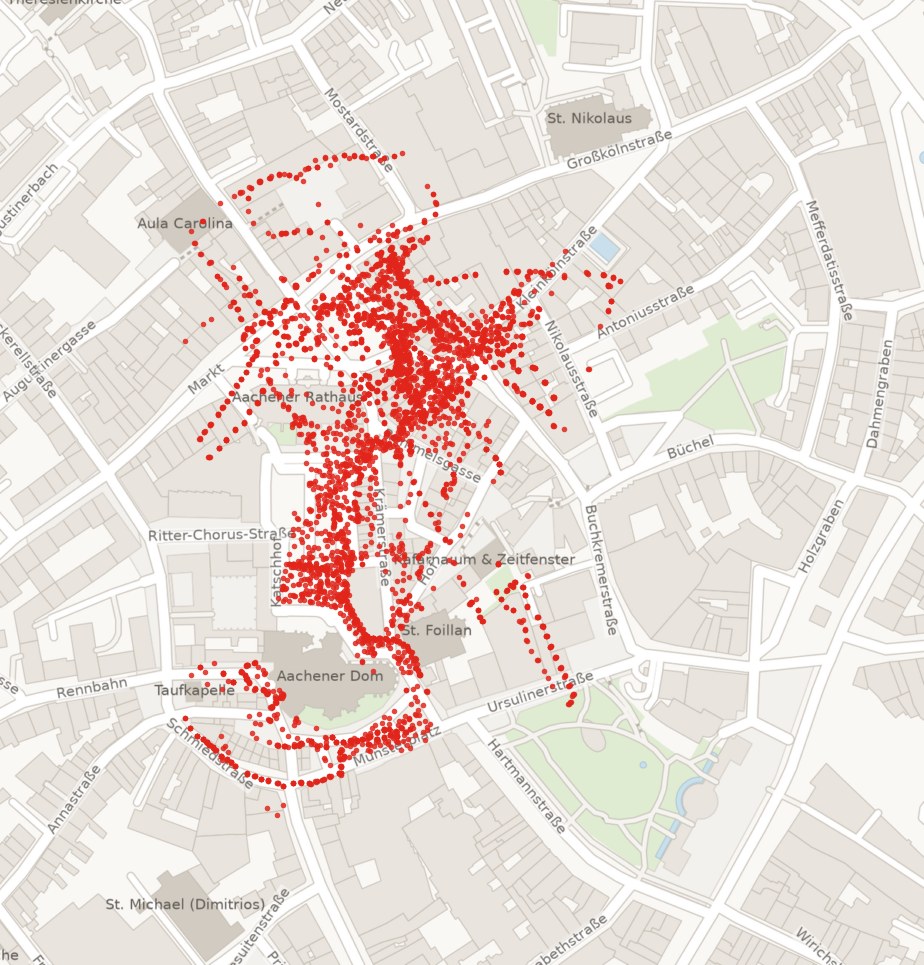}{4.9} \\[0pt]
\maprow{\textbf{+ \method}} & \mapname{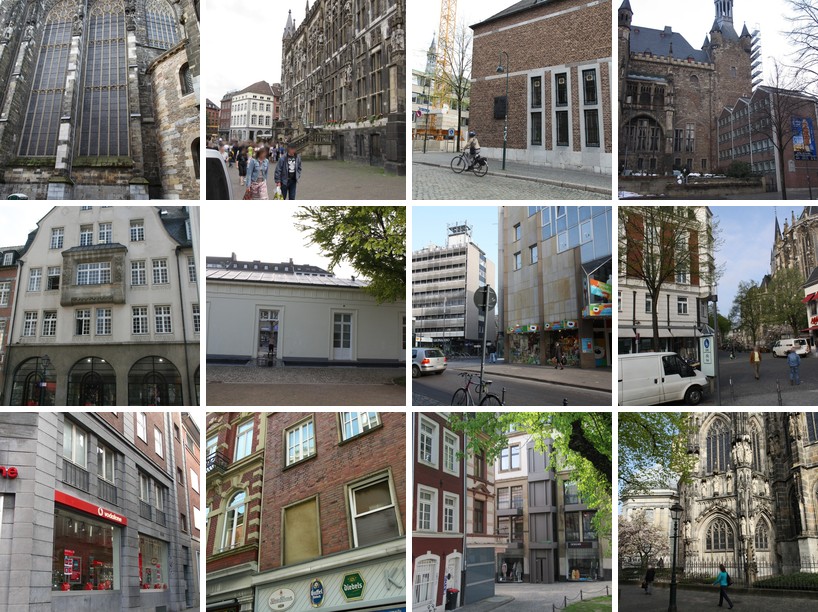}{Aachen DB} & \mapaucbf{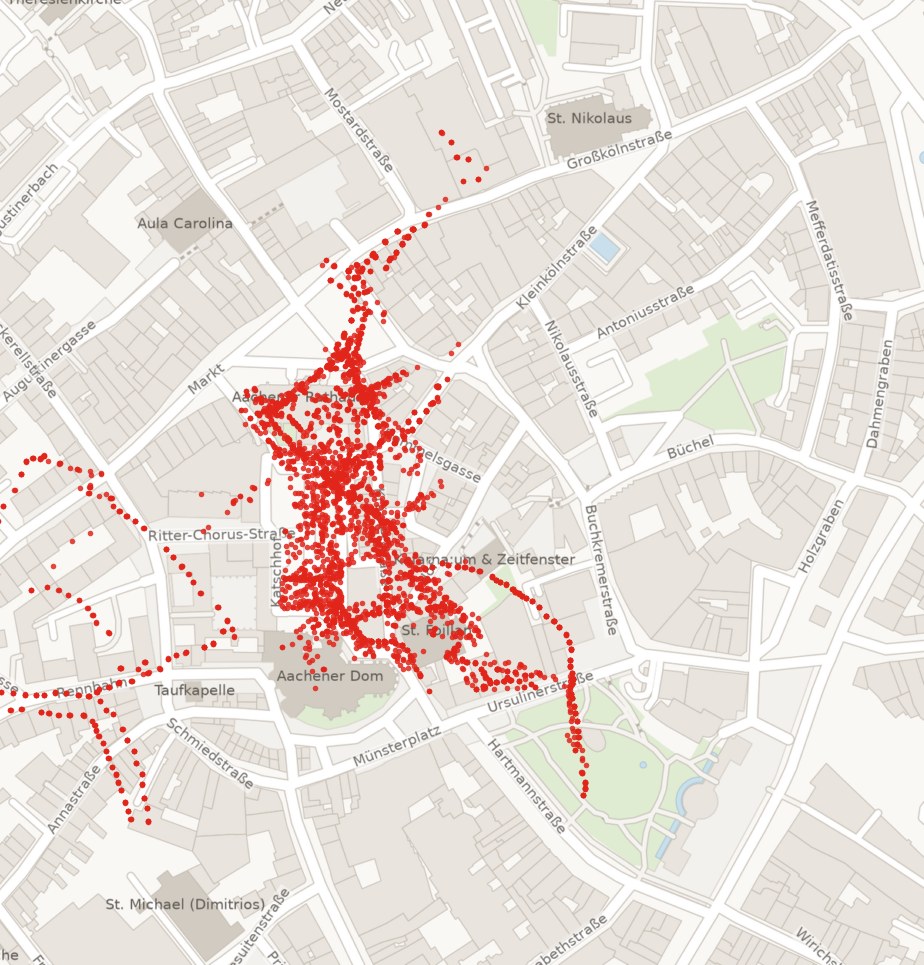}{23.6} & \mapaucbf{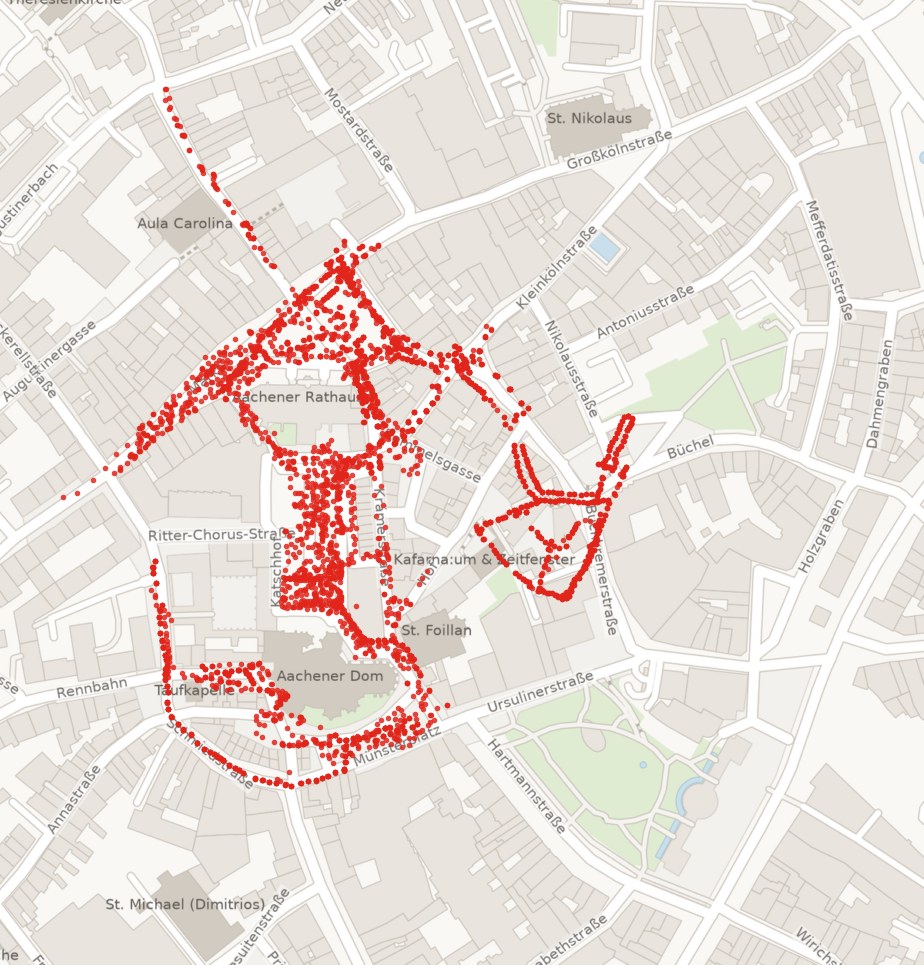}{50.8} & \mapaucbf{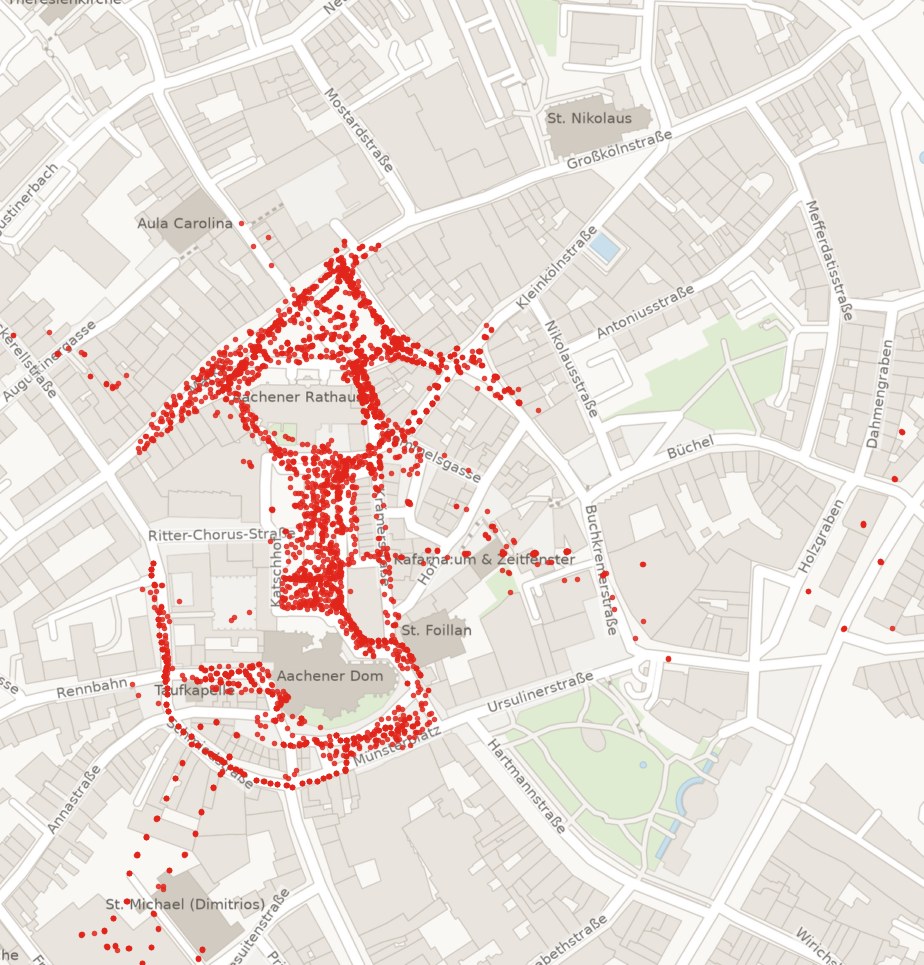}{41.9} & \mapaucbf{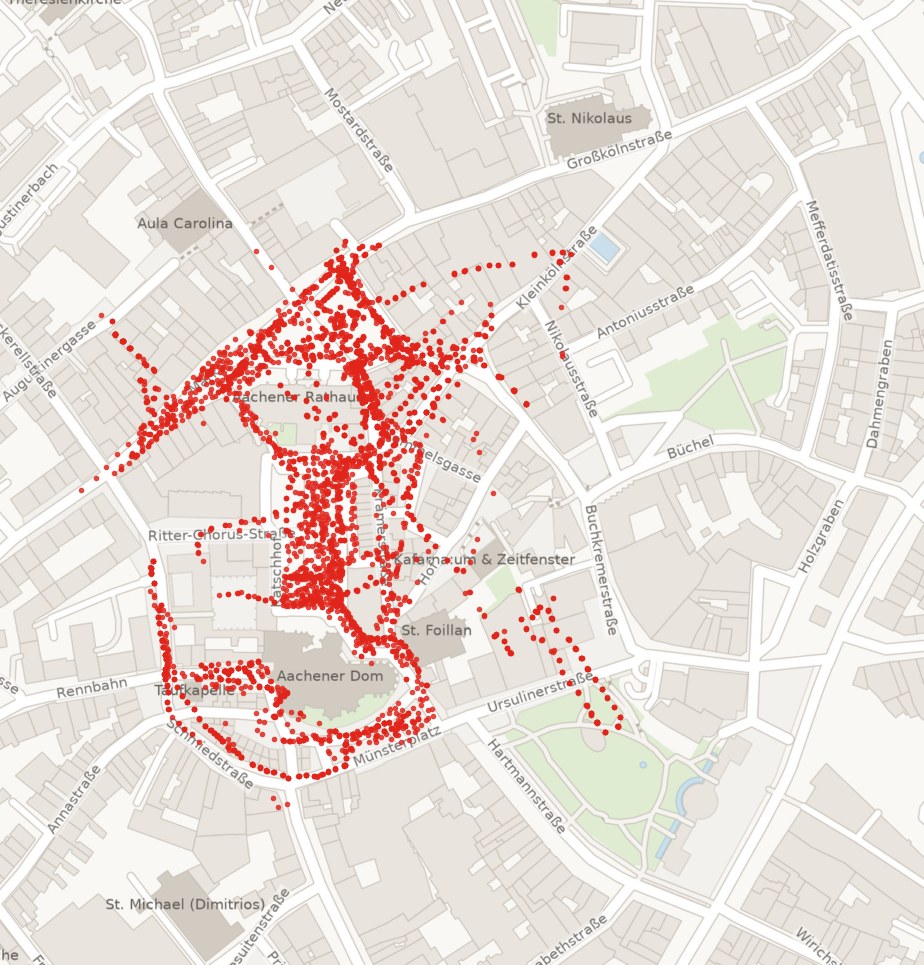}{37.1} \\[4pt]
\end{tabular}
\caption{Camera positions on the large collections. HGE is drawn on the floor plan of the ETH main building (floor E), LIN and Aachen DB on maps drawn from OpenStreetMap data (map data \copyright~OpenStreetMap contributors). $^\ddagger$: our chunking. $^\S$: the method's own chunking.}
\label{fig:large_scenes_map}
\end{figure*}

\clearpage

{
    \small
    \bibliographystyle{ieeenat_fullname}
    \bibliography{main}
}

\end{document}